\documentclass[article]{IEEEtran}
\usepackage{graphics, subfigure, theorem, times, amsfonts, amsmath, amssymb, cite, verbatim, math tools}
\usepackage{tikz}
\usepackage{framed}
\usetikzlibrary{shapes,arrows}
\tikzstyle{phantom vertex} = [ ellipse, 
                               anchor = center, 
                               minimum height = 1*\unit, 
                               minimum width  = 1*\unit,
                               inner sep=0pt,
                               anchor=center]
\tikzstyle{red vertex}   = [black, fill = red!20,   phantom vertex, draw]
\tikzstyle{black vertex} = [black, fill = black!20, phantom vertex, draw]
\tikzstyle{blue vertex}  = [black, fill = blue!20,  phantom vertex, draw]
\tikzstyle{green vertex} = [black, fill = green!20,  phantom vertex, draw]
\tikzstyle{vertex}       = [draw, phantom vertex]

\tikzstyle{point} = [ellipse, inner sep=0pt, draw, fill=white, anchor = center,
                     minimum height = 0.05*\unit, minimum width  = 0.05*\unit]

\usepackage{pgfplots}
\usepackage{color}
\usepackage{needspace}

\newcommand{\myindentedparagraph}[1]{\needspace{1\baselineskip}\medskip \hangindent=11pt \hangafter=0 \noindent{\it #1.}}

\newenvironment{indentedparagraph}[1] 
{\begin{list}{}%
         {\setlength{\leftmargin}{11pt}
          \setlength{\topsep}{10pt}}
         \item[]{\it #1.}}
{\end{list}}

\usepackage{hyperref}
\usepackage{flushend}
\usepackage{multirow}
\usepackage{rotating}
\usepackage{framed}
\input{mysymbol.sty}

\newtheorem{lemma}{\hspace{0pt}\bf Lemma}
\newtheorem{claim}{\hspace{0pt}\bf Claim}
\newtheorem{proposition}{\hspace{0pt}\bf Proposition}

\newtheorem{theorem}{\hspace{0pt}\bf Theorem}
\newtheorem{corollary}{\hspace{0pt}\bf Corollary}

\newtheorem{remark}{\hspace{0pt}\bf Remark}

\newtheorem{definition}{\hspace{0pt}\bf Definition}

\def \mlc {\text{\normalfont mlc}}
\def \sep {\text{\normalfont sep}}

\def\R {\text{\normalfont R}}
\def\C {\text{\normalfont C}}
\def\NR {\text{\normalfont NR}}
\def\SR {\text{\normalfont SR}}
\def\SL {\text{\normalfont SL}}
\def\U {\text{\normalfont U}}
\def\dis {\text{\normalfont dis}}
\def\ciclo {\circlearrowright}

\def\N{\mathbb{N}}

\title{Axiomatic Construction of Hierarchical Clustering in Asymmetric Networks}

\author{\IEEEauthorblockN{Gunnar Carlsson, Facundo M\'emoli, Alejandro Ribeiro, and Santiago Segarra}
\thanks{Work in this paper is supported by NSF CCF-0952867, AFOSR MURI FA9550-10-1-0567, DARPA GRAPHS FA9550-12-1-0416, AFOSR FA9550-09-0-1-0531, AFOSR FA9550-09-1-0643, NSF DMS 0905823, and NSF DMS-0406992. G. Carlsson is with the Department of Mathematics, Stanford University. F. M\'emoli is with the Department of Mathematics and the Department of Computer Science and Engineering, Ohio State University. A. Ribeiro and S. Segarra are with the Department of Electrical and Systems Engineering, University of Pennsylvania. Email: gunnar@math.stanford.edu, memoli@math.osu.edu, aribeiro@seas.upenn.edu, and ssegarra@seas.upenn.edu. Parts of the results in this paper appeared in \cite{Carlssonetal13, Carlssonetal13_2, Carlssonetal13_3, Carlssonetal14}.}}

\begin{document}
\maketitle


%
\begin{abstract}
This paper considers networks where relationships between nodes are represented by directed dissimilarities. The goal is to study methods for the determination of hierarchical clusters, i.e., a family of nested partitions indexed by a connectivity parameter, induced by the given dissimilarity structures. Our construction of hierarchical clustering methods is based on defining admissible methods to be those methods that abide by the axioms of value -- nodes in a network with two nodes are clustered together at the maximum of the two dissimilarities between them -- and transformation -- when dissimilarities are reduced, the network may become more clustered but not less. Several admissible methods are constructed and two particular methods, termed reciprocal and nonreciprocal clustering, are shown to provide upper and lower bounds in the space of admissible methods. Alternative clustering methodologies and axioms are further considered. Allowing the outcome of hierarchical clustering to be asymmetric, so that it matches the asymmetry of the original data, leads to the inception of quasi-clustering methods. The existence of a unique quasi-clustering method is shown. Allowing clustering in a two-node network to proceed at the minimum of the two dissimilarities generates an alternative axiomatic construction. There is a unique clustering method in this case too. The paper also develops algorithms for the computation of hierarchical clusters using matrix powers on a min-max dioid algebra and studies the stability of the methods proposed. We proved that most of the methods introduced in this paper are such that similar  networks yield similar hierarchical clustering results. Algorithms are exemplified through their application to networks describing internal migration within states of the United States (U.S.) and the interrelation between sectors of the U.S. economy. \end{abstract}

%
\section{Introduction} \label{sec_introduction}

The problem of determining clusters in a data set admits different interpretations depending on whether the underlying data is metric, symmetric but not necessarily metric, or asymmetric. Of these three classes of problems, clustering of metric data is the most studied one in terms of both, practice and theoretical foundations. In terms of practice there are literally hundreds of methods, techniques, and heuristics that can be applied to the determination of hierarchical and nonhierarchical clusters in finite metric spaces -- see, e.g., \cite{RuiWunsch05}. Theoretical foundations of clustering methods, while not as well developed as their practical applications \cite{vonlux-david, sober,science_art}, have been evolving over the past decade \cite{ben-david-ackermann,ben-david-reza, CarlssonMemoli10, kleinberg, clust-um,multi-param}. Of particular relevance to our work is the case of hierarchical clustering where, instead of a single partition, we look for a family of partitions indexed by a resolution parameter; see e.g., \cite{lance67general}, \cite[Ch. 4]{clusteringref}, and \cite{ZhaoKarypis05}. In this context, it has been shown in \cite{clust-um} that single linkage \cite[Ch. 4]{clusteringref} is the unique hierarchical clustering method that satisfies three reasonable axioms. These axioms require that the hierarchical clustering of a metric space with two points is the same metric space, that there be no non singleton clusters at resolutions smaller than the smallest distance in the space, and that when distances shrink, the metric space may become more clustered but not less.

When we remove the condition that the data be metric, we move into the realm of clustering in networks, i.e. a set of nodes with pairwise and possibly \emph{directed} dissimilarities represented by edge weights. For the undirected case, the knowledge of theoretical underpinnings is incipient but practice is well developed. Determining clusters in this undirected context is often termed community detection and is formulated in terms of finding cuts such that the edges between different groups have high dissimilarities -- meaning points in different groups are dissimilar from each other -- and the edges within a group have small dissimilarities -- which means that points within the same cluster are similar to each other, \cite{ShiMalik00, GirvanNewman02, GirvanNewman04}. An alternative approach for clustering nodes in graphs is the idea of spectral clustering \cite{Chung97, spectral-clustering, NgEtal02, BachJordan03}. When a graph contains several connected components its Laplacian matrix has multiple eigenvectors associated with the null eigenvalue and the nonzero elements of the corresponding eigenvectors identify the different connected components. The underlying idea of spectral clustering is that different communities should be identified by examining the eigenvectors associated with eigenvalues close to zero. 

\begin{figure}
\centering
\def \thisplotscale {1.5}
\def \unit {\thisplotscale cm}
\tikzstyle {blue vertex here} = [blue vertex, 
                                 minimum width = 0.7*\unit, 
                                 minimum height = 0.7*\unit, 
                                 anchor=center]
{\begin{tikzpicture}[thick, x = 1.2*\unit, y = 0.96*\unit]


	\path[draw, thin] (0,0) + (-0.9, 1.2) node[blue vertex here] (a) {{$a$}}
	                        + ( 0.9, 1.2) node[blue vertex here] (b) {{$b$}}
	                        + (-1.4,-1.2) node[blue vertex here] (c) {{$c$}}
	                        + ( 1.4,-1.2) node[blue vertex here] (d) {{$d$}};
    \path[thin, -stealth] (a) edge [bend left = 15, above] node {{\small $1$}} (b);		                            
    \path[thin, -stealth] (a) edge [bend left = 15, right] node {{\small $6$}} (c);	        
    \path[thin, -stealth] (b) edge [bend left = 15, below] node {{\small $7$}} (a);		                            
    \path[thin, -stealth] (b) edge [bend left = 15, right] node {{\small $3$}} (d);	
    \path[thin, -stealth] (c) edge [bend left = 15, left]  node {{\small $2$}} (a);	        
    \path[thin, -stealth] (c) edge [bend left = 15, above] node {{\small $4$}} (d);	
    \path[thin, -stealth] (d) edge [bend left = 15, left]  node {{\small $5$}} (b);	
    \path[thin, -stealth] (d) edge [bend left = 15, below] node {{\small $2$}} (c);	

    \path[thin, -stealth, black!40] (a) edge [bend left=10, above] node {{\small $10$}} (d);	
    \path[thin, -stealth, black!40] (b) edge [bend left=10, below] node {{\small $10$}} (c);	
    \path[thin, -stealth, black!40] (c) edge [bend left=10, above] node {{\small $10$}} (b);	
    \path[thin, -stealth, black!40] (d) edge [bend left=10, below] node {{\small $10$}} (a);	        

\end{tikzpicture}}
\vspace{-4pt}
\caption{Asymmetric network. Edges denote directed dissimilarities between nodes. Clustering intuition is precarious because there is not a clear proximity notion between nodes. E.g., the pair $a,b$ has the smallest dissimilarity in one direction whereas the pair $c, d$ has the smallest average dissimilarity in both directions. It is not clear which of the two pairs is less dissimilar.}
\vspace{-0.15in}
\label{fig_network_definition}
\end{figure}
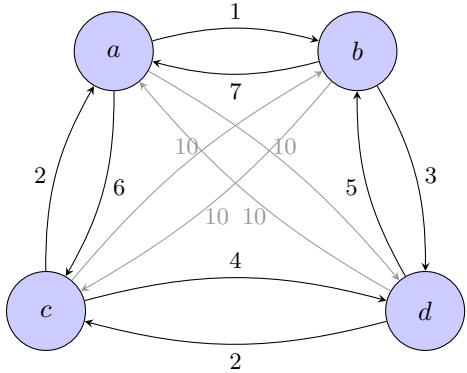

Further relaxing symmetry so that we can allow for asymmetric relationships between nodes \cite{SaitoYadohisa04} reduces the number of available methods that can deal with such data \cite{hubert-min,slater1976hierarchical,boyd-asymmetric,tarjan-improved,slater1984partial,murtagh-multidimensional,PentneyMeila05, MeilaPentney07, ZhouEtal05}. Examples of these methods are the adaptation of spectral clustering to asymmetric graphs by using a random walk perspective \cite{PentneyMeila05} and the use of weighted cuts of minimum aggregate cost \cite{MeilaPentney07}. In spite of these contributions, the rarity of clustering methods for asymmetric networks is expected because the interpretation of clusters as groups of nodes that are closer to each other than to the rest is difficult to generalize when nodes are close in one direction but far apart in the other. E.g., in the network in Fig. \ref{fig_network_definition} nodes $a$ and $b$ are closest to each other in a clockwise sense, but farthest apart in a counterclockwise manner, $c$ and $d$ seem to be closest on average, yet, it seems that all nodes are relatively close as it is possible to loop around the network clockwise without encountering a dissimilarity larger than 3. 

%

%
Although it seems difficult to articulate a general intuition for clustering of asymmetric networks, there are nevertheless some behaviors that we should demand from any reasonable clustering method. Following \cite{kleinberg, CarlssonMemoli10, clust-um,multi-param}, the perspective taken in this paper is to impose these desired behaviors as axioms and proceed to characterize the space of methods that are admissible with respect to them. While different axiomatic constructions are discussed here, the general message is that surprisingly strong structure can be induced by seemingly weak axioms. E.g., by defining the result of clustering networks with two nodes and specifying the behavior of admissible methods when the given dataset shrinks, we encounter that two simple methods are uniformly minimal and maximal across nodes and networks among all those that are admissible (Section \ref{sec_extremal_ultrametrics}). 

Besides axiomatic constructions, this paper also studies stability with respect to perturbations of the original data and establishes that most methods are stable (Section \ref{sec_stability}). We also introduce computationally tractable algorithms to determine the hierarchical clusters that result from applying the different methods that we propose (Section \ref{sec_algorithms}). These algorithms are applied to cluster the network of internal migration between states of the United States (U.S.) and the network of interactions between sectors of the U.S. economy (Section \ref{sec_numerical_experiments}). We also introduce the concept of hierarchical quasi-clustering that generalizes the idea of hierarchical clustering to permit retaining asymmetric influences (Section \ref{sec_full_characterization_asymmetric}) between the clusters. The following sections present a more detailed preview of the results outlined above.

%
\subsection{Fundamental axioms and admissible methods}\label{sec_preview_axioms}
Recall that hierarchical clustering methods produce a resolution dependent clustering of a given network. Throughout this paper we introduce various axioms and properties that represent several desirable features of hierarchical clustering methods. Among these, the axioms of value and transformation underly most of the results presented in this paper. These axioms are stated formally in Sections \ref{sec_axioms} and \ref{sec_dendrograms_and_ultrametrics} but they correspond to the following intuitions:

\begin{indentedparagraph}{(A1) Axiom of Value} For a network with two nodes, the nodes are first clustered together at a resolution level equal to the maximum of the two intervening dissimilarities. 
\end{indentedparagraph}

\begin{indentedparagraph}{(A2) Axiom of Transformation} If we consider a domain network and map it into a target network in a manner such that no pairwise dissimilarity is increased by the mapping, then the resolution level at which two nodes in the target network become part of the same cluster is not larger than the level at which they were clustered together in the original domain network.
\end{indentedparagraph}

\noindent The intuition supporting the Axiom of Transformation is that if some nodes become closer to each other, it may be that new clusters arise, but no cluster can disappear. The intuition supporting the Axiom of Value is that the two nodes in the two-node network form a single cluster at resolutions that allow them to influence each other directly, i.e., resolutions larger than the dissimilarities between them. A hierarchical clustering method satisfying axioms (A1) and (A2)  is said to be {\it admissible.}

Our first theoretical study is the relationship between clustering and mutual influence in networks of arbitrary size (Section \ref{sec_axiomatic_structure}). In particular, we show that the outcome of any admissible hierarchical clustering method is such that a necessary condition for two nodes to cluster together is the existence of chains that allow for direct or indirect influence between the nodes. We can interpret this result as showing that the requirement of direct influence in the two-node network in the Axiom of Value (A1) induces a requirement for, possibly indirect, influence in general networks. This result is termed the Property of Influence and plays an instrumental role in the theoretical developments presented throughout the paper. 

Two hierarchical clustering methods that abide by axioms (A1) and (A2), and that therefore satisfy the Property of Influence, are then derived (Section \ref{sec_reicprocal_and_nonreciprocal}). The first method, \emph{reciprocal clustering}, requires clusters to form through edges exhibiting low dissimilarity in both directions whereas the second method, \emph{nonreciprocal clustering}, allows clusters to form through cycles of small dissimilarity. More specifically, reciprocal clustering defines the cost of an edge as the maximum of the two directed dissimilarities. Nodes are clustered together at a given resolution if there exists a chain linking them such that all links in the chain have a cost smaller than said resolution. In nonreciprocal clustering we consider directed chains and define the cost of a chain as the maximum dissimilarity encountered when traversing it from beginning to end. Nodes are clustered together at a given resolution if it is possible to find directed chains in both directions whose edge costs do not exceed the given resolution. Observe that both of these methods rely on the determination of chains of minimax cost linking any pair of nodes. This fact is instrumental in the derivation of algorithms for the computation of output dendrograms as we discuss in Section \ref{sec_preview_algorithms_stability}.

A fundamental result regarding admissible methods is the proof that any clustering method that satisfies axioms (A1) and (A2) lies between reciprocal and nonreciprocal clustering in a well-defined sense (Section \ref{sec_extremal_ultrametrics}). Specifically, any clustering method that satisfies axioms (A1) and (A2) forms clusters at resolutions larger than the resolutions at which they are formed with nonreciprocal clustering, and smaller than the resolutions at which they are formed with reciprocal clustering. The clustering resolutions vary from method to method, but they are always contained within the specified bounds. When restricted to symmetric networks, reciprocal and nonreciprocal clustering yield equivalent outputs, which coincide with the output of single linkage (Section \ref{secsymmetric_networks}). This observation is consistent with the existence and uniqueness result in \cite{clust-um} since axioms (A1) and (A2) are reduced to two of the axioms considered in \cite{clust-um} when we restrict attention to metric data. The derivations in this paper show that the existence and uniqueness result in \cite{clust-um} is true for all symmetric, not necessarily metric, datasets and that a third axiom considered there is redundant because it is implied by the other two.

We then unveil some of the clustering methods that lie between reciprocal and nonreciprocal clustering and study their properties (Section \ref{sec_intermediate_ultrametrics}). Three families of intermediate clustering methods are introduced. The grafting methods consist of attaching the clustering output structures of the reciprocal and nonreciprocal methods in a way such that admissibility is guaranteed (Section \ref{sec_grafting}). We further present a construction of methods that can be regarded as a convex combination in the space of clustering methods. This operation is shown to preserve admissibility therefore giving rise to a second family of admissible methods (Section \ref{sec_convex_comb}). A third family of admissible clustering methods is defined in the form of semi-reciprocal methods that allow the formation of cyclic influences in a more restrictive sense than nonreciprocal clustering but more permissive than reciprocal clustering (Section \ref{sec_inter_reciprocal}).

In some applications the requirement for bidirectional influence in the Axiom of Value is not justified as unidirectional influence suffices to establish proximity. This alternative value statement leads to the study of alternative axiomatic constructions and their corresponding admissible hierarchical clustering methods (Section \ref{sec_alternative_axioms}). We first propose an Alternative Axiom of Value in which clusters in two-node networks are formed at the minimum of the two dissimilarities: 

\begin{indentedparagraph}{(A1'') Alternative Axiom of Value} For a network with two nodes, the nodes are clustered together at the {\it minimum} of the two dissimilarities between them.
\end{indentedparagraph}

\noindent Under this axiomatic framework we define unilateral clustering as a method in which influence propagates through chains of nodes that are close in at least one direction (Section \ref{sec_unilateral_clustering}). Contrary to the case of admissibility with respect to (A1)-(A2) in which a range of methods exist, unilateral clustering is the unique method that is admissible with respect to (A1'') and (A2). A second alternative is to take an agnostic position and allow nodes in two-node networks to cluster at any resolution between the minimum and the maximum dissimilarity between them. All methods considered in the paper satisfy this agnostic axiom and, not surprisingly, outcomes of methods that satisfy this agnostic axiom are uniformly bounded between unilateral and reciprocal clustering (Section \ref{sec_agnostic_axiom_of_value}).

%
\subsection{Hierarchical quasi-clustering}

Dendrograms are symmetric structures used to represent the outputs of hierarchical clustering methods. Having a symmetric output is, perhaps, a mismatched requirement for the processing of asymmetric data. This mismatch motivates the development of asymmetric structures that generalize the concept of dendrogram (Section \ref{sec_full_characterization_asymmetric}).

Start by observing that a hierarchical clustering method is a map from the space of networks to the space of dendrograms, that a dendrogram is a collection of nested partitions indexed by a resolution parameter and that each partition is induced by an equivalence relation, i.e., a relation satisfying the reflexivity, {\it symmetry}, and transitivity properties. Hence, the symmetry in hierarchical clustering derives from the symmetry property of equivalence relations which we remove in order  to construct the asymmetric equivalent of hierarchical clustering.

To do so we define a quasi-equivalence relation as one that is reflexive and transitive but not necessarily symmetric and define a quasi-partition as the structure induced by a quasi-equivalence relation -- these structures are also known as partial orders \cite{Harzheim05}. Just like regular partitions, quasi-partitions contain disjoint blocks of nodes but also include an influence structure between the blocks derived from the asymmetry in the original network. A quasi-dendrogram is further defined as a nested collection of quasi-partitions, and a hierarchical quasi-clustering method is regarded as a map from the space of networks to the space of quasi-dendrograms (Section \ref{sec_quasi_dendrograms}). 

As in the case of (regular) hierarchical clustering we proceed to study admissibility with respect to asymmetric versions of the axioms of value and transformation (Section \ref{sec_quasi_clustering_axioms}). We show that there is a unique quasi-clustering method admissible with respect to these axioms and that this method is an asymmetric version of the single linkage clustering method (Section \ref{sec_existance_uniqueness_quasi_clustering}). The analysis in this section hinges upon an equivalence between quasi-dendrograms and quasi-ultrametrics (Section \ref{sec_quasi_ultrametrics}) that generalizes the known equivalence between dendrograms and ultrametrics \cite{jardine-sibson}. If we further recall that, for symmetric networks, single linkage is the only hierarchical clustering method that is admissible with respect to the axioms of value and transformation [cf. \cite{clust-um} and Section \ref{secsymmetric_networks}], we conclude that there is a strong parallelism between symmetric networks, equivalence relations, partitions, dendrograms, ultrametrics, and single linkage on the one hand and asymmetric networks, quasi-equivalence relations, quasi-partitions, quasi-dendrograms, quasi-ultrametrics, and directed single linkage on the other. In the same way that dendrograms are particular cases of quasi-dendrograms, every element in the former list is a particular case of the corresponding element in the latter. Moreover, every result relating two elements of the former list can be generalized as relating the two corresponding, more general elements in the latter.

%
\subsection{Algorithms and Stability}\label{sec_preview_algorithms_stability}

Besides the characterization of methods that are admissible with respect to different sets of axioms we also develop algorithms to compute the dendrograms associated with the methods introduced throughout the paper and study their stability with respect to perturbations.

The determination of algorithms for all of the methods introduced is given by the computation of matrix powers in a min-max dioid algebra \cite{GondranMinoux08}. In this dioid algebra we operate in the field of positive reals and define the addition operation between two scalars to be their minimum and the product operation of two scalars to be their maximum (Section \ref{sec_algorithms}). From this definition it follows that the $(i,j)$-th entry of the $n$-th dioid power of a matrix of network dissimilarities represents the minimax cost of a chain linking node $i$ to node $j$ with at most $n$ edges. As we have already mentioned, reciprocal and nonreciprocal clustering require the determination of chains of minimax cost. Similarly, other clustering methods introduced in this paper can be interpreted as minimax chain costs of a previously modified matrix of dissimilarities which can therefore be framed in terms of dioid matrix powers as well. E.g., in unilateral clustering we define the cost of an edge as the minimum of the dissimilarities in both directions and then search for minimax chain costs whereas in semi-reciprocal clustering we limit the length of allowable chains. 

In order to study the stability of clustering methods with respect to perturbations of a network, following \cite{clust-um,metric-structures} we adopt and adapt the Gromov-Hausdorff distance between finite metric spaces \cite[Chapter 7.3]{burago-book} to furnish a notion of distance between asymmetric networks (Section \ref{sub_sec_network_distance}). This distance allows us to compare any two networks, even when they have different node sets. Since dendrograms are equivalent to finite ultrametric spaces which in turn are particular cases of asymmetric networks, we can use the Gromov-Hausdorff distance to quantify the difference between two dendrograms obtained when clustering two different networks. We then say that a clustering method is stable if the clustering outputs of similar networks are close to each other. More precisely, we say that a clustering method is stable if, for any pair of networks, the distance between the output dendrograms can be bounded by the distance between the original networks. In particular, stability of a method guarantees robustness to the presence of noise in the dissimilarity values. Although not every method considered in this paper is stable, we show stability for most of the methods including the reciprocal, nonreciprocal, semi-reciprocal, and unilateral clustering methods (Section \ref{sec_ultrametric_stability}).

%
\subsection{Applications}

Clustering methods are exemplified through their application to two real-world networks: the network of internal migration between states of the U.S. for the year 2011 and the network of interactions between economic sectors of the U.S. economy for the year 2011 (Section \ref{sec_numerical_experiments}). The purpose of these examples is to understand which information can be extracted by performing hierarchical clustering and quasi-clustering analyses based on the different methods proposed. Analyzing migration clusters provides information on population mixing (Section \ref{sec_state_to_state_migration}). Analyzing interactions between economic sectors unveils their relative importances and their differing levels of coupled interactions (Section \ref{sec_economic_sectors}).

The migration network example illustrates the different clustering outputs obtained when we consider the Axiom of Value (A1) or the Alternative Axiom of Value (A1'') as conditions for admissibility. Unilateral clustering, the unique method compatible with (A1''), forms clusters around influential states like California and Texas by merging each of these states with other smaller ones around them (Section \ref{sec_unilateral_migration}). On the other hand, methods compatible with (A1) like reciprocal clustering, tend to first merge states with balanced bidirectional influence such as two different populous states or states sharing urban areas. In this way, reciprocal clustering sees California first merging with Texas for being two very influential states and Washington merging with Oregon for sharing the urban area of Portland (Section \ref{sec_reciprocal_migration}). Moreover, the similarity between the reciprocal and nonreciprocal outcomes (Section \ref{sec_nonreciprocal_migration}) indicates that no other clustering method satisfying axiom (A1) would reveal new information, thus, intermediate clustering methods are not applied. Clustering methods provide information about grouping but obscure information about influence. To study the latter we apply the directed single linkage quasi-clustering method (Section \ref{sec_directed_migration}). Analysis of the output quasi-dendrograms show, e.g., the dominant roles of California and Massachusetts in the population influxes into the West Coast and New England, respectively.

The network of interactions between sectors of the U.S. economy records how much of a sector's output is used as input to another sector of the economy. For this network, reciprocal and nonreciprocal clustering output essentially different dendrograms, indicating the ubiquity of influential cycles between sectors. Reciprocal clustering first merges sectors of bidirectional influence such as professional services with administrative services and the farming sector with the food and beverage sector (Section \ref{sec_reciprocal_io}). Nonreciprocal clustering, on the other hand, captures cycles of influence such as the one between oil and gas extraction, petroleum and coal products, and the construction sector (Section \ref{sec_nonreciprocal_io}). However, nonreciprocal clustering propagates influence through arbitrarily large cycles, a feature which might be undesirable in practice. The observed difference between the reciprocal and the nonreciprocal dendrograms motivates the application of a clustering method with intermediate behavior such as the semi-reciprocal clustering method with parameter 3 (Section \ref{sec_semi_reciprocal_io}). Its cyclic propagation of influence is closer to the real behavior of sectors within the economy and, thus, we obtain a more reasonable clustering output. Finally, the application of the directed single linkage quasi-clustering method reveals the dominant influence of energy, manufacturing, and financial and professional services over the rest of the economy (Section \ref{sec_directed_io}).


%
\section {Preliminaries}\label{sec_preliminaries}
%
We define a network $N_X$ to be a pair $(X, A_X)$ where $X$ is a finite set of points or nodes and $A_X: X \times X \to \reals_+$ is a dissimilarity function. The dissimilarity $A_X(x,x')$ between nodes $x\in X$ and $x'\in X$ is assumed to be non negative for all pairs $(x,x')$ and 0 if and only if $x=x'$. We do not, however, require $A_X$ to be a metric on the finite space $X$: dissimilarity functions $A_X$ need not satisfy the triangle inequality and, more consequential for the problem considered here, they may be asymmetric in that it is possible to have $A_X(x,x')\neq A_X(x',x)$ for some $x \neq x'$. In some discussions it is convenient to introduce a labeling of of the elements in $X$, $X=\{x_1,\ldots,x_n\}$, and reinterpret the dissimilarity function $A_X$ as the possibly asymmetric matrix $A_X \in \reals_+^{n \times n}$ with $(A_X)_{i,j}=A_X(x_i, x_j)$ for all $i, j \in \{1, ... ,n\}$. The diagonal elements $(A_X)_{i,i}=A_X(x_i, x_i)$ are zero. As it doesn't lead to confusion we use $A_X$ to denote both, the dissimilarity function and its matrix representation. We further define $\ccalN$ as the set of all networks $N_X$. Networks in $\ccalN$ can have different node sets $X$ as well as different dissimilarities functions $A_X$.

An example network is shown in Fig. \ref{fig_network_definition}. The set of nodes is $X=\{a, b, c, d\}$ with dissimilarities $A_X$ represented by a weighted directed graph. The dissimilarity from, e.g,  $a$ to $b$ is $A_X(a,b)=1$, which is different from the dissimilarity $A_X(b,a)=7$ from $b$ to $a$. The smallest nontrivial networks contain two nodes $p$ and $q$ and two dissimilarities $\alpha$ and $\beta$ as depicted in Fig. \ref{fig_axioms_value_influence}. The following special networks appear often throughout our paper: consider the dissimilarity function $A_{p,q}$ with $A_{p,q}(p,q)=\alpha$ and $A_{p,q}(q,p)=\beta$ for some $\alpha, \beta >0$ and define the {\it two-node network} $\vec{\Delta}_2(\alpha, \beta)$ with parameters $\alpha$ and $\beta$ as
\begin{equation}\label{eqn_preliminaries_two_node_network}
   \vec{\Delta}_2(\alpha, \beta):= (\{p,q\}, A_{p,q}).
\end{equation} 
By a clustering of the set $X$ we always mean a partition $P_X$ of $X$; i.e., a collection of sets $P_X=\{B_1,\ldots, B_J\}$ which are pairwise disjoint, $B_i\cap B_j =\emptyset$ for $i\neq j$, and are required to cover $X$, $\cup_{i=1}^{J} B_i = X$. The sets $B_1, B_2, \ldots B_J$ are called the \emph{blocks} or \emph{clusters} of $P_X$. We define the \emph{power set} $\ccalP(X)$ of $X$ as the set containing every subset of $X$, thus $B_i \in \ccalP(X)$ for all $i$. An equivalence relation $\sim $ on $X$ is a binary relation such that for all $x, x', x'' \in X$ we have that (1) $x \sim x$, (2) $x \sim x'$ if and only if $x' \sim x$, and (3)  $x \sim x'$  and $x' \sim x''$ imply $x \sim x''$.

A partition $P_X=\{B_1,\ldots, B_J\}$ of $X$ always induces and is induced by an equivalence relation $ \sim_{P_X} $ on $X$ where for all $x, x' \in X$ we have that $x \sim_{P_X} x'$ if and only if $x$ and $x'$ belong to the same block $B_i$ for some $i$. In this paper we focus on hierarchical clustering methods. The output of hierarchical clustering methods is not a single partition $P_X$ but a nested collection $D_X$ of partitions $D_X(\delta)$ of $X$ indexed by a resolution parameter $\delta\geq 0$. In consistency with our previous notation, for a given $D_X$, we say that two nodes $x$ and $x'$ are equivalent at resolution $\delta \geq 0$ and write $x\sim_{D_X(\delta)} x'$ if and only if nodes $x$ and $x'$ are in the same block of $D_X(\delta)$. The nested collection $D_X$ is termed a \emph{dendrogram} and is required to satisfy the following properties (cf. \cite{clust-um}): 

\myindentedparagraph{(D1) Boundary conditions} For $\delta=0$ the partition $D_X(0)$ clusters each $x\in X$ into a separate singleton and for some $\delta_0$ sufficiently large $D_X(\delta_0)$ clusters all elements of $X$ into a single set,
\begin{align}\label{eqn_dendrogram_boundary_conditions}
   & D_X(0)  = \Big\{ \{x\}, \, x\in X\Big\},  \nonumber\\
   & D_X(\delta_0) = \Big\{ X \Big\} \quad \forsome\ \delta_0 > 0.
\end{align}

\myindentedparagraph{(D2) Hierarchy} As $\delta$ increases clusters can be combined but not separated. I.e., for any $\delta_1 < \delta_2$ any pair of points $x,x'$ for which $x\sim_{D_X(\delta_1)} x'$ must be  $x\sim_{D_X(\delta_2)} x'$.

\myindentedparagraph{(D3) Right continuity} For all $\delta \geq 0$, there exists $\epsilon > 0$ such that $D_X(\delta)=D_X(\delta')$ for all $\delta' \in [\delta, \delta+\epsilon]$.

\medskip\noindent The second boundary condition in \eqref{eqn_dendrogram_boundary_conditions} together with (D2) implies that we must have $D_X(\delta) = \big\{ X \big\}$ for all $\delta\geq\delta_0$. We denote  by $[x]_{\delta}$ the equivalence class to which the node $x \in X$ belongs at resolution $\delta$, i.e. $[x]_\delta := \{x' \in X \given x \sim_{D_X(\delta)} x'\}$. From requirement (D1) we must have that $[x]_0=\{x\}$ and $[x]_{\delta_0}=\{X\}$ for all $x \in X$.

\begin{figure*}
\centering
\def \thisplotscale {1.35}
\def \unit {\thisplotscale cm}
\tikzstyle {blue vertex here} = [blue vertex, 
                                 minimum width = 0.4*\unit, 
                                 minimum height = 0.4*\unit, 
                                 anchor=center]

{\small
\begin{tikzpicture}[thick, x = 1.2*\unit, y = 0.96*\unit]

    \coordinate (1 end) at (2.0, 0.8); 
    \coordinate (2 end) at (2.0, 1.6); 
    \coordinate (3 end) at (3.8, 2.4); 
    \coordinate (4 end) at (3.8, 3.2); 
    \coordinate (1 and 2 end) at (5.3, 1.2); 
    \coordinate (3 and 4 end) at (5.3, 2.8); 
    \coordinate (1 2 3 and 4 end) at (6.6, 2.0); 
            
    \path[draw, thick] (0, 0.8) node[left] {$a$} -- (1 end); 
    \path[draw, thick] (0, 1.6) node[left] {$b$} -- (2 end); 
    \path[draw, thick] (0, 2.4) node[left] {$c$} -- (3 end); 
    \path[draw, thick] (0, 3.2) node[left] {$d$} -- (4 end); 

    \path[draw, thick] (1 end) -- (2 end); 
    \path[draw, thick] (3 end) -- (4 end); 

    \path[draw, thick] (1 end) ++ (0,0.4) -- (1 and 2 end); 
    \path[draw, thick] (3 end) ++ (0,0.4) -- (3 and 4 end); 

    \path[draw, thick] (1 and 2 end) -- (3 and 4 end); 

    \path[draw, thick] (1 and 2 end) ++ (0,0.8) -- (1 2 3 and 4 end); 

    \path[draw, thin, draw=black!50] (0,3.6) node [above] {{$0$}}-- ++ (0,-3.2) 
           node [right] {{$\ \{a\}, \{b\}, \{c\}, \{d\}$}} 
           -- ++ (0,-0.3); 
    \path[draw, thin, draw=black!50] (2 end) -- ++ (0,2.0) node [above] {{$2$}};
    \path[draw, thin, draw=black!50] (1 end) -- ++ (0,-0.4) 
           node [right] {{$\ \{a, b\}, \{c\}, \{d\}$}} 
           -- ++ (0,-0.3); 
    \path[draw, thin, draw=black!50] (4 end) -- ++ (0,0.4) node [above] {{$4$}};
    \path[draw, thin, draw=black!50] (3 end) -- ++ (0,-2.0) 
           node [right] {{$\ \{a, b\}, \{c, d\}$}} 
           -- ++ (0,-0.3); 
    \path[draw, thin, draw=black!50] (3 and 4 end) -- ++ (0,0.8) node [above] {{$5$}};
    \path[draw, thin, draw=black!50] (1 and 2 end) -- ++ (0,-0.8) 
           node [right] {{$\ \{a, b, c, d\}$}} 
           -- ++ (0,-0.3); 
    

	\path[draw, thin] (-2.8,3.15) ++ ( 0.0, 0.0) node[blue vertex here] (1) {{$a$}}
	                            ++ ( 1.0, 0.0) node[blue vertex here] (2) {{$b$}}
	                            ++ (-1.3,-2.2) node[blue vertex here] (3) {{$c$}}
	                            ++ ( 1.6, 0.0) node[blue vertex here] (4) {{$d$}};
    \path[thin, -stealth] (1) edge [bend left, above] node {{$2$}} (2);		                            
    \path[thin, -stealth] (2) edge [bend left, right] node {{$5$}} (4);	
    \path[thin, -stealth] (4) edge [bend left, below] node {{$4$}} (3);	
    \path[thin, -stealth] (3) edge [bend left, left]  node {{$5$}} (1);	        
    \path[thin, -stealth] (2) edge [bend left, below] node {{$2$}} (1);		                            
    \path[thin, -stealth] (4) edge [bend left, left]  node {{$5$}} (2);	
    \path[thin, -stealth] (3) edge [bend left, above] node {{$4$}} (4);	
    \path[thin, -stealth] (1) edge [bend left, right] node {{$5$}} (3);	        

\end{tikzpicture}
} 
\caption{Single linkage dendrogram for a symmetric network. Dendrograms are trees representing the outcome of hierarchical clustering algorithms. The single linkage dendrogram as defined by \eqref{eqn_single_linkage} for the network on the left is shown on the right. For resolutions $\delta<2$ each node is in a separate partition, for $2\leq\delta<4$ nodes $a$ and $b$ form the cluster $\{a, b\}$, for $4\leq\delta<5$ we add the cluster $\{c, d\}$, and for $5\leq\delta$ all nodes are part of a single cluster.}
\vspace{-0.1in}
\label{fig_dendrogram_example}
\end{figure*}
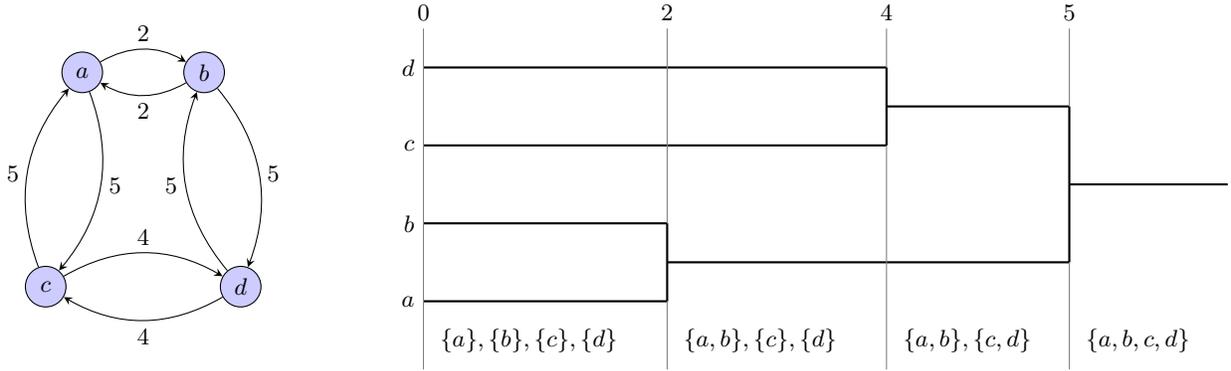

The interpretation of a dendrogram is that of a structure which yields different clusterings at different resolutions. At resolution $\delta=0$ each point is in a cluster of its own. As the resolution parameter $\delta$ increases, nodes start forming clusters. According to condition (D2), nodes become ever more clustered since once they join together in a cluster, they stay together in the same cluster for all larger resolutions. Eventually, the resolutions become coarse enough so that all nodes become members of the same cluster and stay that way as $\delta$ keeps increasing. A dendrogram can be represented as a rooted tree; see Fig. \ref{fig_dendrogram_example}. Its root represents $D_X(\delta_0)$ with all nodes clustered together and the leaves represent $D_X(0)$ with each node separately clustered. Forks in the tree happen at resolutions $\delta$ at which the partitions become finer -- or coarser if we move from leaves to root.

%

%
Denoting by $\ccalD$ the space of all dendrograms we define a hierarchical clustering method as a function 
\begin{equation}\label{eqn_clust_from_networks_to_dendrograms}
\ccalH:\ccalN \to \ccalD,
\end{equation}
from the space of networks $\ccalN$ to the space of dendrograms $\ccalD$ such that the underlying space $X$ is preserved. For the network $N_X=(X,A_X)$ we denote by $D_X=\ccalH(X,A_X)$ the output of clustering method $\ccalH$. 

In the description of hierarchical clustering methods $\ccalH$ in general, and in those derived on this paper in particular, the concepts of \emph{chain}, \emph{chain cost}, and \emph{minimum chain cost} are important. Given a network $(X, A_X)$ and $x, x' \in X$, a chain from $x$ to $x'$ is any \emph{ordered} sequence of nodes in $X$, 
\begin{equation}\label{eqn_definition_chain}
   [x=x_0, x_1, \ldots , x_{l-1}, x_l=x'],
\end{equation}
which starts at $x$ and finishes at $x'$. We will frequently use the notation $C(x,x')$ to denote one such chain. We say that $C(x, x')$ links or connects $x$ to $x'$. Given two chains $C(x, x')=[x=x_0, x_1, ... , x_l=x']$ and $C(x', x'')=[x'=x'_0, x'_1, ... , x'_{l'}=x'']$ such that the end point $x'$ of the first one coincides with the starting point of the second one we define the \emph{concatenated chain} $C(x, x') \uplus C(x',x'')$ as
\begin{align}\label{eqn_definition_concatenation}
   &C(x, x') \uplus C(x',x'')  \nonumber \\ &\qquad\quad
      := [x=x_0,\ldots , x_l=x'=x'_0,\ldots , x'_{l'}=x'']. 
\end{align}
It follows from \eqref{eqn_definition_concatenation} that the concatenation operation $\uplus$ is associative in that $\big[C(x, x') \uplus C(x', x'')\big] \uplus C(x'', x''') = C(x, x') \uplus [C(x', x'') \uplus C(x'', x''')]$. Observe that the chain $C(x, x')=[x=x_0, x_1, \ldots , x_{l-1}, x_l=x']$ and its reverse $[x'=x_l, x_{l-1}, \ldots , x_{1}, x_0=x]$ are different entities even if the intermediate hops are the same.

The \emph{links} of a chain are the edges connecting its consecutive nodes in the direction imposed by the chain. We define the \emph{cost} of a given chain $C(x, x')=[x=x_0,\ldots, x_l=x']$ as
\begin{equation}\label{eqn:cost_of_chain}
\max_{i | x_i\in C(x,x')}A_X(x_i,x_{i+1}),
\end{equation}
i.e., the maximum dissimilarity encountered when traversing its links in order. The directed minimum chain cost $\tdu^*_X(x, x')$ between $x$ and $x'$ is then defined as the minimum cost among all the chains connecting $x$ to $x'$,
\begin{align}\label{eqn_nonreciprocal_chains} 
   \tdu^*_X(x, x') := \min_{C(x,x')} \,\,
                        \max_{i | x_i\in C(x,x')} A_X(x_i,x_{i+1}).
\end{align} 
In asymmetric networks the minimum chain costs $\tdu^*_X(x, x')$ and $\tdu^*_X(x', x)$ are different in general but they are equal on symmetric networks. In this latter case, the costs $\tdu^*_X(x, x') = \tdu^*_X(x', x)$ are instrumental in the definition of the single linkage dendrogram \cite{clust-um}. Indeed, for resolution $\delta$, single linkage makes $x$ and $x'$ part of the same cluster if and only if they can be linked through a chain of cost not exceeding $\delta$. Formally, the equivalence classes at resolution $\delta$ in the single linkage dendrogram $\text{SL}_X$ over a symmetric network $(X, A_X)$ are defined by
\begin{equation}\label{eqn_single_linkage}
   x\sim_{\text{SL}_X(\delta)} x' \iff  
      \tdu^*_X(x, x') = \tdu^*_X(x', x) \leq\delta.
\end{equation}
Fig. \ref{fig_dendrogram_example} shows a finite metric space along with the corresponding single linkage dendrogram. For resolutions 
$\delta<2$ the dendrogram partitions are $D_X(\delta) = \big\{\{a\}, \{b\}, \{c\}, \{d\}\big\}$. For resolutions $2\leq\delta<4$ nodes $a$ and $b$ get clustered together to yield $D_X(\delta) = \big\{ \{a, b\}, \{c\}, \{d\}\big\}$. As we keep increasing the parameter $\delta$, $c$ and $d$ also get clustered together yielding $D_X(\delta) = \big\{\{a, b\}, \{c,d\}\big\}$ for resolutions $4\leq\delta<5$. For $5\leq\delta$ all nodes are part of a single cluster, $D_X(\delta) = \big\{\{a, b,c,d\}\big\}$ because we can build chains between any pair of nodes incurring maximum cost smaller than or equal to $\delta$. 

We further define a \emph{loop} as a chain of the form $C(x,x)$ for some $x \in X$ such that $C(x, x)$ contains at least one node other than $x$. Since a loop is a particular case of a chain, the cost of a loop is given by \eqref{eqn:cost_of_chain}. Furthermore, consistently with \eqref{eqn_nonreciprocal_chains}, we define the \emph{minimum loop cost} $\mlc(X,A_X)$ of a network $(X, A_X)$ as the minimum across all possible loops of each individual loop cost,
\begin{equation}\label{eqn_def_mlc}
    \mlc(X,A_X):=\min_x \, \min_{C(x,x)} \,\,  \max_{i | x_i\in C(x,x)}A_X(x_i,x_{i+1}),
\end{equation}
where, we recall, $C(x,x)$ contains at least one node different from $x$. Another relevant property of a network $(X, A_X)$ is the \emph{separation} of the network $\sep(X,A_X)$ which we define as its minimum positive dissimilarity, 
\begin{equation}\label{eqn_def_separation_network}
   \sep(X,A_X) := \min_{x \neq x'} A_X(x, x').
\end{equation}
Notice that from \eqref{eqn_def_mlc} and \eqref{eqn_def_separation_network} we must have 
\begin{equation}\label{eqn:sep_less_mlc}
\sep(X,A_X) \leq \mlc(X,A_X).
\end{equation}
Further observe that in the particular case of networks with symmetric dissimilarities the two quantities coincide, i.e., $\sep(X,A_X)=\mlc(X,A_X)$, when $A_X(x,x')=A_X(x',x)$ for all $x,x' \in X$. For example, the network in Fig. \ref{fig_network_definition} has separation equal to $1$ and minimum loop cost equal to $3$.

When one restricts attention to networks $(X,A_X)$ having dissimilarities $A_X$ that conform to the definition of a finite metric space -- i.e., dissimilarities $A_X$ are symmetric and satisfy the triangle inequality -- it has been shown \cite{clust-um} that single linkage is the unique hierarchical clustering method satisfying axioms (A1)-(A2) in Section \ref{sec_axioms} plus a third axiom stating that clusters cannot form at resolutions smaller than the minimum distance between different points of the space. In the case of asymmetric networks the space of admissible methods is richer, as we demonstrate throughout this paper.

%
\section {Axioms of value and transformation}\label{sec_axioms}

To study hierarchical clustering methods on asymmetric networks we start from intuitive notions that we translate into the axioms of value and transformation discussed in this section. 

%

The Axiom of Value is obtained from considering the two-node network $\vec{\Delta}_2(\alpha, \beta)$ defined in \eqref{eqn_preliminaries_two_node_network} and depicted in Fig. \ref{fig_axioms_value_influence}. We say that node $x$ is able to influence node $x'$ at resolution $\delta$ if the dissimilarity from $x$ to $x'$ is not greater than $\delta$. In two-node networks, our intuition dictates that a cluster is formed if nodes $p$ and $q$ are able to influence each other. This implies that the output dendrogram should be such that $p$ and $q$ are part of the same cluster at resolutions $\delta\geq\max(\alpha,\beta)$ that allow direct mutual influence. Conversely, we expect nodes $p$ and $q$ to be in separate clusters at resolutions $0 \leq \delta<\max(\alpha,\beta)$ that do {\it not} allow for mutual influence. At resolutions $\delta<\min(\alpha,\beta)$ there is no influence between the nodes and at resolutions $\min(\alpha,\beta) \leq \delta<\max(\alpha,\beta)$ there is unilateral influence from one node over the other. In either of the latter two cases the nodes are different in nature. If we think of dissimilarities as, e.g., trust, it means one node is trustworthy whereas the other is not. If we think of the network as a Markov chain, at resolutions $0 \leq \delta<\max(\alpha,\beta)$ the states are different singleton equivalence classes -- one of the states would be transient and the other one absorbent. Given that, according to \eqref{eqn_clust_from_networks_to_dendrograms}, a hierarchical clustering method is a map $\ccalH$ from networks to dendrograms, we formalize this intuition as the following requirement on the set of admissible maps:

\myindentedparagraph{(A1) Axiom of Value} The dendrogram $D_{p, q}=\ccalH(\vec{\Delta}_2(\alpha, \beta))$ produced by $\ccalH$ applied to the network $\vec{\Delta}_2(\alpha, \beta)$ is such that $D_{p, q}(\delta)=\big\{\{p\},\{q\}\big\}$ for $0 \leq \delta<\max(\alpha,\beta)$ and $D_{p, q}(\delta)=\big\{\{p,q\}\big\}$ otherwise; see Fig. \ref{fig_axioms_value_influence}.

\medskip\noindent Clustering nodes $p$ and $q$ together at resolution $\delta=\max(\alpha,\beta)$ is somewhat arbitrary, as any monotone increasing function of $\max(\alpha, \beta)$ would be admissible. As a value claim, however, it means that the clustering resolution parameter $\delta$ is expressed in the same units as the elements of the dissimilarity function.

%
\begin{figure}
  \centering
  \centerline{\def \thisplotscale {0.9}
\def \unit {\thisplotscale cm}
\def \xdendogram{{1, 2}}
\def \ydendogram{{1, 2}}

{\small
\begin{tikzpicture}[shorten >=2, scale = \thisplotscale]

    \node [blue vertex] at (-4.5,1) (p) {$p$};
    \node [blue vertex] at (-2,1) (q) {$q$};
    \path [-stealth](p) edge [bend left, above] node {$\alpha$} (q);	
    \path [-stealth] (q) edge [bend left, below] node {$\beta$}  (p);	
    
    \draw [-stealth] (-0.5,0) -- (5.3,0) node [below, at end] {$\delta$};
    \draw [-stealth] (0,-0.5) -- (0,2.9);
    
    \draw[thick] (0,0.7) -- ++(2.6,0) -- ++(0,1.2) -- +(-2.6,0) ++(0,-0.6) -- +(2.1,0);
    \draw[dashed](2.6,0.7) -- ++(0,-1.1) node [right, at end] {$\max(\alpha,\beta)$};
    \node [left] at (0,0.7) {$p$};
    \node [left] at (0,1.9) {$q$};
    
    \node at (-4,2.5) {$\vec{\Delta}_2(\alpha, \beta)$};
    \node at (-0.7,2.5) {$D_{p,q}$};
 
\end{tikzpicture}
}


    
}
\caption{Axiom of Value. Nodes in a two-node network cluster at the minimum resolution at which both can influence each other.}
\vspace{-0.1in}
\label{fig_axioms_value_influence}
\end{figure}
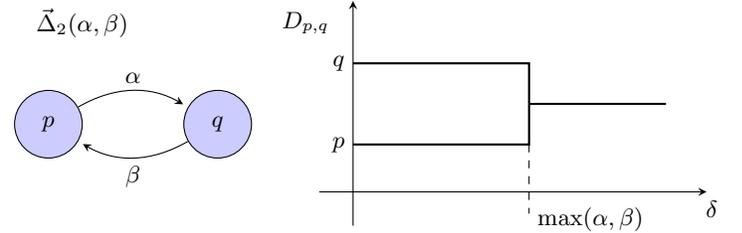

%
\begin{figure*}
\centering
\def \thisplotscale {0.65}
\def \unit {\thisplotscale cm}

{\small
\begin{tikzpicture}[-stealth, shorten >=2, x = 1.1*\unit, y=1*\unit]

    \node [blue vertex] at (   0,  3.4) (1) {$x_1$};
    \node [blue vertex] at ( 2.5, -0.6) (2) {$x_2$};    
    \node [blue vertex] at (-2.5, -0.6) (3) {$x_3$};

    \path (1) edge [bend left=20, right] node {{$1$}} (2);	
    \path (2) edge [bend left=20, below] node {{$2$}} (3);
    \path (3) edge [bend left=20, left] node  {{$2$}} (1);    	

    \path (2) edge [bend left=20, right] node {{$2$}} (1);	
    \path (3) edge [bend left=20, below] node {{$3$}} (2);
    \path (1) edge [bend left=20, left]  node {{$3$}} (3);    	
    
    \node [blue vertex] at (8.0,2.8) (1p) {$y_1$};
    \node [blue vertex] at (9.7,0.0) (2p) {$y_2$};    
    \node [blue vertex] at (6.3,0.0) (3p) {$y_3$};

    \path (1p) edge [bend left=20, right] node {{ $1/2$}} (2p);	
    \path (2p) edge [bend left=20, below] node {{$1/2$}} (3p);
    \path (3p) edge [bend left=20, left]  node {{$1/2$}} (1p);    	

    \path (2p) edge [bend left=20, right] node {{$1$}} (1p);	
    \path (3p) edge [bend left=20, below] node {{$1$}} (2p);
    \path (1p) edge [bend left=20, left]  node {{$1$}} (3p);    	
    
    \path (1) edge [bend left, above, red, very thick, pos=0.6] node {$\bbphi$} (1p);	
    \path (2) edge [bend left, above, red, very thick, pos=0.05] node {$\bbphi$} (2p);	    
    \path (3) edge [bend left, above, red, very thick, pos=0.75] node {$\bbphi$} (3p);	    

   
    \draw [-stealth] (12.5,2) -- (18.3,2) node [below, at end] {$\delta$};
    \draw [-stealth] (13,2) -- (13,4.9);
    
    \draw[thick, -] (13,2.4) -- ++(2.5,0) -- ++(0,0.8) -- ++(-2.5,0) ++(0,0.8) -- ++(3.5,0) -- ++(0,-1.2) -- ++(-1,0) -- ++(1,0)--++ (0, 0.6)--++(1.5, 0);
    \node [left] at (13,2.4) {$x_1$};
    \node [left] at (13,3.2) {$x_2$};
    \node [left] at (13,4) {$x_3$};
    \node [left] at (12.5,4.5) {$D_{X}$};
    
    
    \draw [-stealth] (12.5,-1) -- (18.3,-1) node [below, at end] {$\delta$};
    \draw [-stealth] (13,-1) -- (13,1.9);
    
    \draw[thick, -] (13,-0.6) -- ++(2,0) -- ++(0,0.8) -- ++(-2,0) ++(0,0.8) -- ++(2,0) -- ++(0,-0.8) -- ++(2.1,0);
    \draw[dashed, thick, red, -](16,4.5) -- ++(0,-6) node [right, at end] {$\delta'$};
    \node [left] at (13,-0.6) {$y_1$};
    \node [left] at (13,0.2) {$y_2$};
    \node [left] at (13,1) {$y_3$};
    \node [left] at (12.5,1.5) {$D_{Y}$};
     \node at (-2.5,3.5) {$N_{X}$};
      \node at (9, 3.5) {$N_{Y}$};

\end{tikzpicture}
}
\caption{Axiom of Transformation. If the network $N_X$ can be mapped to the network $N_Y$ using a dissimilarity reducing map $\phi$, then for every resolution $\delta$ nodes clustered together in $D_X(\delta)$ must also be clustered in $D_Y(\delta)$. E.g., since points $x_1$ and $x_2$ are clustered together at resolution $\delta'$, their image through $\phi$, i.e. $y_1=\phi(x_1)$ and $y_2=\phi(x_2)$, must also be clustered together at this resolution because the map $\phi$ is dissimilarity reducing.}
\vspace{-0.1in}
\label{fig_axiom_of_transformation}
\end{figure*}
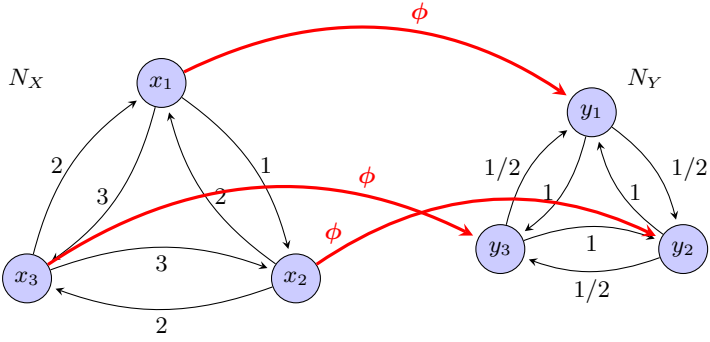

%
The second restriction on the space of allowable methods $\ccalH$ formalizes our expectations for the behavior of $\ccalH$ when confronted with a transformation of the underlying space $X$ and the dissimilarity function $A_X$; see Fig. \ref{fig_axiom_of_transformation}. Consider networks $N_X=(X,A_X)$ and $N_Y=(Y,A_Y)$ and denote by $D_X=\ccalH(X,A_X)$ and $D_Y=\ccalH(Y,A_Y)$ the corresponding dendrogram outputs. If we map all the nodes of the network $N_X=(X,A_X)$ into nodes of the network $N_Y=(Y,A_Y)$ in such a way that no pairwise dissimilarity is increased we expect the latter network to be more clustered than the former at any given resolution.  Intuitively, nodes in $N_Y$ are more capable of influencing each other, thus, clusters should be formed more easily. In terms of the respective dendrograms we expect that nodes co-clustered at resolution $\delta$ in $D_X$ are mapped to nodes that are also co-clustered at this resolution in $D_Y$. In order to formalize this notion, we introduce the following concept: given two networks $N_X=(X,A_X)$ and $N_Y=(Y,A_Y)$, map $\phi:X\to Y$ is called \emph{dissimilarity-reducing} map if it holds that $A_X(x,x')\geq A_Y(\phi(x),\phi(x'))$ for all $x,x'\in X$.

The Axiom of Transformation that we introduce next is a formal statement of the intuition described above :

\myindentedparagraph{(A2) Axiom of Transformation} Consider any two networks $N_X=(X,A_X)$ and $N_Y=(Y,A_Y)$ and any dissimilarity-reducing map $\phi:X\to Y$. Then, the method $\ccalH$ satisfies the axiom of transformation if the output dendrograms $D_X=\ccalH(X,A_X)$ and $D_Y=\ccalH(Y,A_Y)$ are such that $x\sim_{D_X(\delta)}x'$ for some $\delta\geq 0$ implies that $\phi(x)\sim_{D_Y(\delta)}\phi(x')$.

\medskip\noindent We say that a hierarchical clustering method $\ccalH$ is \emph{admissible} with respect to (A1) and (A2), or \emph{admissible} for short, if it satisfies Axioms (A1) and (A2). Axiom (A1) states that units of the resolution parameter $\delta$ are the same units of the elements of the dissimilarity function. Axiom (A2) states that if we reduce dissimilarities, clusters may be combined but cannot be separated. 

These axioms are an adaptation of the axioms proposed in \cite{clust-um,CarlssonMemoli10,multi-param} for the case of finite metric spaces.

%
\subsection {Dendrograms as ultrametrics}\label{sec_dendrograms_and_ultrametrics}

%
Dendrograms are convenient graphical representations but otherwise cumbersome to handle. A mathematically more convenient representation is obtained when one identifies dendrograms with finite \emph{ultrametric} spaces. An ultrametric defined on the space $X$ is a metric function $u_X: X \times X \to \reals_+$ that satisfies a stronger triangle inequality as we formally define next.

%
\begin{definition}\label{def_ultrametric} Given a node set $X$ an ultrametric $u_X$ is a nonnegative function $u_X: X \times X \to \reals_+$ satisfying the following properties:
\begin{mylist}
\item[{\it (i) Identity.}] The ultrametric $u_X(x, x')=0$ if and only if $x=x'$ for all $x, x' \in X$.
\item[{\it (ii) Symmetry.}] For all pairs of points $x,x'\in X$ it holds that $u_X(x, x')= u_X(x', x)$.
\item[{\it (iii) Strong triangle inequality.}] Given points $x,x',x''\in X$ the ultrametrics $u_X(x,x'')$, $u_X(x,x')$, and $u_X(x',x'')$ satisfy the strong triangle inequality
\begin{equation}\label{eqn_strong_triangle_inequality}
    u_X(x,x'') \leq \max \Big(u_X(x,x'),  u_X(x',x'') \Big).
\end{equation} \end{mylist}\vspace{-10pt} \end{definition}

%
\noindent Since \eqref{eqn_strong_triangle_inequality} implies the usual triangle inequality $u_X(x,x'') \leq u_X(x,x') + u_X(x',x'')$ for all $x, x', x'' \in X$, ultrametric spaces are particular cases of metric spaces.

Our interest in ultrametrics stems from the fact that it is possible to establish a structure preserving bijective mapping between dendrograms and ultrametrics as proved by the following construction and theorem; see also Fig. \ref{fig_dendrograms_as_ultrametrics}.

%
\begin{figure}
\centering\vspace{5pt}
\def \thisplotscale {0.8}
\def \unit {\thisplotscale cm}

{\small
\begin{tikzpicture}[thick, x = 1.4*\unit, y = 1.0*\unit]

    \coordinate (1 end) at (2.0, 1.0); 
    \coordinate (2 end) at (2.0, 2.0); 
    \coordinate (3 end) at (3.8, 3.0); 
    \coordinate (4 end) at (3.8, 4.0); 
    \coordinate (1 and 2 end) at (5.3, 1.5); 
    \coordinate (3 and 4 end) at (5.3, 3.5); 
    \coordinate (1 2 3 and 4 end) at (6.6, 2.5); 
            
    \path[draw, thick] (0, 1) node[left] {$a$} -- (1 end); 
    \path[draw, thick] (0, 2) node[left] {$b$} -- (2 end); 
    \path[draw, thick] (0, 3) node[left] {$c$} -- (3 end); 
    \path[draw, thick] (0, 4) node[left] {$d$} -- (4 end); 

    \path[draw, thick] (1 end) -- (2 end); 
    \path[draw, thick] (3 end) -- (4 end); 

    \path[draw, thick] (1 end) ++ (0,0.5) -- (1 and 2 end); 
    \path[draw, thick] (3 end) ++ (0,0.5) -- (3 and 4 end); 

    \path[draw, thick] (1 and 2 end) -- (3 and 4 end); 

    \path[draw, thick] (1 and 2 end) ++ (0,1) -- (1 2 3 and 4 end); 

    \path[draw, thin] (2 end) -- ++ (0,2.5) 
                                ++ (-0.7,0) node [above] {{$u_X(a,b)=2$}};
    \path[draw, thin] (4 end) -- ++ (0,0.5) 
                                ++ (-0.5,0) node [above] {{$u_X(c,d)=4$}};
    \path[draw, thin] (3 and 4 end) -- ++ (0,1.0) 
                                ++ (0.4,0) node [above] {{$u_X(a/b,c/d)=5$}};
           
\end{tikzpicture}
}
\caption{Equivalence of dendrograms and ultrametrics. Given a dendrogram $D_X$ define distance $u_X(x,x') := \min \big\{ \delta \geq 0 , x\sim_{D_X(\delta)} x' \big\}$. This distance is an ultrametric because it satisfies the strong triangle inequality \eqref{eqn_strong_triangle_inequality} and is symmetric.}
\vspace{-0.1in}
\label{fig_dendrograms_as_ultrametrics}
\end{figure} 

Consider the map $\Psi:\mathcal{D} \rightarrow \mathcal{U}$ from the space of dendrograms to the space of networks endowed with ultrametrics, defined as follows:

For a given dendrogram $D_X$ over the finite set $X$ write $\Psi(D_X) = (X, u_X)$, where we define $u_X(x,x')$ for all $x, x' \in X$ as the smallest resolution at which $x$ and $x'$ are clustered together
\begin{equation}\label{eqn_theo_dendrograms_as_ultrametrics_10}
   u_X(x,x') := \min \Big\{ \delta\geq 0,  x\sim_{D_X(\delta)} x' \Big\}.
\end{equation}
We also consider the map $\Upsilon:\mathcal{U} \rightarrow \mathcal{D}$ constructed as follows: for a given ultrametric $u_X$ on the finite set $X$ and each $\delta \geq 0$ define the relation $\sim_{u_X(\delta)}$ on $X$ as
\begin{equation}\label{eqn_theo_dendrograms_as_ultrametrics_20}
   x \sim_{u_X(\delta)} x' \iff u_X(x,x')\leq \delta.
\end{equation}
Further define $D_X(\delta) :=\big\{X \mod \sim_{u_X(\delta)}\big\}$ and $\Upsilon(X, u_X):= D_X$.

%

\begin{theorem}\label{theo_dendrograms_as_ultrametrics}
The maps $\Psi:\mathcal{D} \rightarrow \mathcal{U}$ and $\Upsilon: \mathcal{U} \rightarrow \mathcal{D}$ are both well defined. Furthermore, $\Psi\circ\Upsilon$ is the identity on $\mathcal{U}$ and $\Upsilon\circ\Psi$ is the identity on $\mathcal{D}$.
\end{theorem}

%
The proof of this result can be found in \cite{clust-um}, yet, for the reader's convenience, we present the proof here. 

\begin{myproofnoname}
First notice that the technical condition (D3) of dendrograms ensures that the minimum in \eqref{eqn_theo_dendrograms_as_ultrametrics_10} exists rendering a well-defined function $u_X$. Thus, to show that $\Psi$ is a well-defined map, we must prove that $u_X$ is an ultrametric. In order to do this, we have to show the symmetry, identity, non negativity and strong triangle inequality properties. Non negativity follows from the non negativity of the resolution parameter $\delta$ in \eqref{eqn_theo_dendrograms_as_ultrametrics_10}. Symmetry $u_X(x,x')=u_X(x',x)$ for all $x, x' \in X$ follows from the symmetry property of the equivalence relation $\sim_{D_X(\delta)}$. The identity property $u_X(x,x')=0 \Leftrightarrow x=x'$ follows from reflexivity of the equivalence relation $\sim_{D_X(\delta)}$ and the boundary condition (D1) on dendrograms. To see that $u_X$ satisfies the strong triangle inequality in \eqref{eqn_strong_triangle_inequality} consider points $x$, $x'$, and $x'' \in X$ such that the lowest resolution for which $x\sim_{D_X(\delta)} x''$ is $\delta_1$ and the smallest resolution for which $x'\sim_{D_X(\delta)} x''$ is $\delta_2$. According to \eqref{eqn_theo_dendrograms_as_ultrametrics_10} we then have
\begin{alignat}{3}\label{eqn_theo_pf_dendrograms_as_ultrametrics_10}
   &u_X(x, x'') &&   = \delta_1 
                && := \min \Big\{ \delta \geq 0, x\sim_{D_X(\delta)} x'' \Big\}, \nonumber\\
   &u_X(x',x'') &&   = \delta_2 
                && := \min \Big\{ \delta \geq 0, x'\sim_{D_X(\delta)} x'' \Big\}.
\end{alignat}
Denote by $\delta_0:=\max(\delta_1,\delta_2)$. Because the dendrogram is a nested set of partitions [cf. (D2)] it must be $x\sim_{D_X(\delta_0)} x''$ and $x'\sim_{D_X(\delta_0)} x''$. Furthermore, being $\sim_{D_X(\delta_0)}$ an equivalence relation it satisfies transitivity from where it follows that $x\sim_{D_X(\delta_0)} x'$. Using \eqref{eqn_theo_dendrograms_as_ultrametrics_10} for $x$, $x'$ we conclude that
\begin{equation}\label{eqn_theo_pf_dendrograms_as_ultrametrics_20}
   u_X(x,x') := \min \Big\{ \delta \geq 0, x\sim_{D_X(\delta)} x' \Big\} \leq \delta_0.
\end{equation}
But now observe that by definition $\delta_0:=\max(\delta_1,\delta_2)$. Substitute this expression in \eqref{eqn_theo_pf_dendrograms_as_ultrametrics_20} and compare with \eqref{eqn_theo_pf_dendrograms_as_ultrametrics_10} to write
\begin{equation}\label{eqn_theo_pf_dendrograms_as_ultrametrics_30}
   u_X(x,x') \leq \max(\delta_1,\delta_2) = \max \Big(u_X(x,x''), u_X(x',x'')\Big).
\end{equation}
Thus, $u_X$ satisfies the strong triangle inequality and is therefore an ultrametric, proving that the map $\Psi$ is well-defined.

For the converse result, we need to show that $\Upsilon$ is a well-defined map. In order to do so, we first need to show that the relation $\sim_{u_X(\delta)}$ as defined in \eqref{eqn_theo_dendrograms_as_ultrametrics_20} is an equivalence relation. Symmetry and reflexivity are implied by the symmetry and identity properties of the ultrametric $u_X$, respectively. To see that $\sim_{u_X(\delta)}$ is also transitive consider points $x$, $x'$, and $x'' \in X$ such that $x \sim_{u_X(\delta)} x''$ and $x' \sim_{u_X(\delta)} x''$. Consequently, it follows from  \eqref{eqn_theo_dendrograms_as_ultrametrics_20} that 
\begin{align}\label{eqn_theo_pf_dendrograms_as_ultrametrics_40}
   u_X(x, x'') \leq \delta, \qquad u_X(x',x'') \leq \delta.
\end{align}
Further note that being $u_X$ an ultrametric it satisfies the strong triangle inequality in \eqref{eqn_strong_triangle_inequality}. Combining this with \eqref{eqn_theo_pf_dendrograms_as_ultrametrics_40} yields
\begin{align}\label{eqn_theo_pf_dendrograms_as_ultrametrics_50}
    u_X(x,x') \leq \max \Big( u_X(x,x'),  u_X(x',x'') \Big) \leq \delta,
\end{align}
from where it follows that $x \sim_{u_X(\delta)} x'$ [cf. \eqref{eqn_theo_dendrograms_as_ultrametrics_20}]. Thus, $\sim_{u_X(\delta)}$ is an equivalence relation which, as such, induces a partition $D_X(\delta) := \{X \mod \sim_{u_X(\delta)}\}$ of the set $X$ for every $\delta \geq 0$. Now, we need to show that $D_X$ is a well-defined dendrogram, i.e., we need to show that the partitions $D_X(\delta)$ for $\delta \geq 0$ satisfy (D1)-(D3). The boundary conditions (D1) are satisfied from the identity property of $u_X$ and the fact that the maximum value of $u_X$ in the finite set $X$ must be upper bounded by some $\delta_0$. To see that partitions are nested in the sense of condition (D2) notice that for $\delta_1<\delta_2$, the condition $u_X(x,x')\leq\delta_1$ implies $u_X(x,x')\leq\delta_2$. This latter inequality substituted in the definition in \eqref{eqn_theo_dendrograms_as_ultrametrics_20} leads to the conclusion that $x \sim_{u_X(\delta_1)} x'$ implies $x \sim_{u_X(\delta_2)} x'$ for $\delta_1\leq \delta_2$ as in condition (D2). Finally, to see that the technical condition (D3) is satisfied, for each $\delta \geq 0$ such that $D_X(\delta) \neq \{ X \}$ we may define $\epsilon(\delta)$ as any positive real satisfying
\begin{equation}\label{eqn_epsilon_d_3_ultrametric_dendrogram}
0 < \epsilon(\delta) < \displaystyle \min_{\substack{x, x' \in X \\ u_X(x, x') > \delta}} u_X(x, x')- \delta,
\end{equation} 
where the finiteness of $X$ ensures that $\epsilon(\delta)$ is well-defined.
Hence, \eqref{eqn_theo_dendrograms_as_ultrametrics_20} guarantees that the equivalence relation $\sim_{u_X(\delta)}$ is the same as $\sim_{u_X(t)}$ for $t \in [\delta, \delta + \epsilon(\delta)]$. Consequently, the partition $D_X(t)$ induced by the equivalence relation is the same for this range of resolutions, proving (D3) for these resolutions. For resolutions $\delta$ such that $D_X(\delta) = \{ X \}$, (D3) is trivially satisfied since the dendrogram remains unchanged for all larger resolutions, proving that $\Upsilon$ is well-defined.

In order to conclude the proof, we need to show that $\Psi \circ \Upsilon$ and $\Upsilon \circ \Psi$ are the identities on $\mathcal{U}$ and $\mathcal{D}$, respectively. To see why the former is true, pick any ultrametric network $(X, u_X)$ and consider an arbitrary pair of nodes $x, x' \in X$ such that $u_X(x, x')=\delta_0$. Also, consider the ultrametric network $\Psi \circ \Upsilon (X, u_X):=(X, u^*_X)$. From \eqref{eqn_theo_dendrograms_as_ultrametrics_20}, in the dendrogram $\Upsilon (X, u_X)$ the nodes $x$ and $x'$ are not merged for resolutions $\delta < \delta_0$ and at resolution $\delta=\delta_0$ both nodes merge into one single cluster. When we apply $\Psi$ to the resulting dendrogram, from \eqref{eqn_theo_dendrograms_as_ultrametrics_10} we obtain $u^*_X(x, x')=\delta_0$. Since $x, x' \in X$ were chosen arbitrarily, we have that $u_X = u^*_X$, showing that $\Psi \circ \Upsilon$ is the identity on $\mathcal{U}$. A similar argument shows that $\Upsilon \circ \Psi$ is the identity on $\mathcal{D}$.
\end{myproofnoname}

%
\noindent Given the equivalence between dendrograms and ultrametrics established by 
Theorem \ref{theo_dendrograms_as_ultrametrics} 
we can regard hierarchical clustering methods $\ccalH$ as inducing ultrametrics in node spaces $X$ based on dissimilarity functions $A_X$. However, ultrametrics are particular cases of dissimilarity functions. Thus, we can reinterpret the method $\ccalH$ as a map
\begin{equation}\label{eqn_clustering_from_networks_to_ultrametrics}
\ccalH:\ccalN\to\ccalU 
\end{equation}
mapping the space of networks to the space $\ccalU$ of networks endowed with ultrametrics.  For all $x, x' \in X$, the ultrametric value $u_X(x,x')$ induced by $\ccalH$ is the minimum resolution at which $x$ and $x'$ are co-clustered by $\ccalH$. Observe that the outcome of a hierarchical clustering method defines an ultrametric in the space $X$ even when the original data does not correspond to a metric, as is the case of asymmetric networks. At any rate, a simple observation with important consequences \cite{clust-um} for the study of the stability of methods, is that $\ccalU \subset \ccalN$.

We say that two methods $\ccalH_1$ and $\ccalH_2$ are \emph{equivalent}, and we write $\ccalH_1 \equiv \ccalH_2$, if and only if
\begin{equation}\label{eqn:method_equivalence}
\ccalH_1(N) = \ccalH_2(N),
\end{equation}
for all $N \in \ccalN$.

A further consequence of the equivalence provided by Theorem \ref{theo_dendrograms_as_ultrametrics} is that we can now rewrite axioms (A1)-(A2) in a manner that refers to properties of the output ultrametrics. We then say that a hierarchical clustering method $\ccalH$ is \emph{admissible} if and only if it satisfies the following two axioms:

\myindentedparagraph{(A1) Axiom of Value} The ultrametric output $(\{p,q\}, u_{p,q})=\ccalH(\vec{\Delta}_2(\alpha, \beta))$ produced by $\ccalH$ applied to the two-node network $\vec{\Delta}_2(\alpha, \beta)$ satisfies 
\begin{equation}\label{eqn_two_node_network_ultrametric}
   u_{p,q}(p,q) = \max(\alpha,\beta).
\end{equation} 

\myindentedparagraph{(A2) Axiom of Transformation} Consider two networks $N_X=(X,A_X)$ and $N_Y=(Y,A_Y)$ and a dissimilarity-reducing map $\phi:X\to Y$, i.e. a map $\phi$ such that for all $x,x' \in X$ it holds that $A_X(x,x')\geq A_Y(\phi(x),\phi(x'))$. Then, for all $x, x' \in X$, the output ultrametrics $(X,u_X)=\ccalH(X,A_X)$ and $(Y,u_Y)=\ccalH(Y,A_Y)$ satisfy 
\begin{equation}\label{eqn_dissimilarity_reducing_ultrametric}
    u_X(x,x') \geq u_Y(\phi(x),\phi(x')).
\end{equation} 

\noindent The axioms in Section \ref{sec_axioms} restrict admissible methods $\ccalH$ by placing conditions on the dendrograms that the methods may produce. The axioms here do the same by imposing conditions on the ultrametrics produced by the methods. Axiom (A1) implies that the units of the dissimilarity function $A_X$ and the ultrametric $u_X$ are the same. Axiom (A2) implies that not increasing any dissimilarity in the network cannot result in an increase of the output ultrametric between some pair of nodes. Despite the somewhat different interpretations, by virtue of Theorem \ref{theo_dendrograms_as_ultrametrics}, the requirements here imposed on the output ultrametrics are equivalent to the requirements imposed on the output dendrograms in the axiom statements introduced earlier in Section \ref{sec_axioms}. 

For the particular case of symmetric networks $(X, A_X)$ we defined the single linkage dendrogram $\text{SL}_X$ through the equivalence relations in \eqref{eqn_single_linkage}. According to Theorem \ref{theo_dendrograms_as_ultrametrics} this dendrogram is equivalent to an ultrametric space that we denote by $(X, u^{\SL}_X)$. Comparing \eqref{eqn_single_linkage} with \eqref{eqn_theo_dendrograms_as_ultrametrics_20} we conclude, as is well known \cite{clust-um}, that the single linkage ultrametric $u^{\SL}_X$ in symmetric networks is given by 
\begin{align}\label{eqn_single_linkage_ultrametric}
   u^{\SL}_X(x,x') &\ =\   \tdu^*_X(x, x') = \tdu^*_X(x', x) \\\nonumber
                 &\ = \  \min_{C(x,x')} \,\,
                         \max_{i | x_i\in C(x,x')} A_X(x_i,x_{i+1}),
\end{align}
where we also used \eqref{eqn_nonreciprocal_chains} to write the last equality. We read \eqref{eqn_single_linkage_ultrametric} as saying that the single linkage ultrametric $u^{\SL}_X(x,x')$ between $x$ and $x'$ is the minimum chain cost $\tdu^*_X(x, x') = \tdu^*_X(x', x)$ among all chains linking $x$ to $x'$.

%
\begin{figure}
\centering\vspace{5pt}
\def \thisplotscale {0.66}
\def \unit {\thisplotscale cm}

{\small
\begin{tikzpicture}[scale = \thisplotscale]

    \path   (-4.1, 3.5)   node [blue vertex] (1) {$a$}   
          ++( 2.3,-3.5) node [blue vertex] (2) {$b$}    
          ++(-4.6, 0)   node [blue vertex] (3) {$c$};

    \path (1) edge [bend left=20, red, right, shorten >=2, -stealth] node {$1/2$} (2);	
    \path (2) edge [bend left=20, red, below, shorten >=2, -stealth] node {$1$} (3);
    \path (3) edge [bend left=20, red, left, shorten >=2,  -stealth] node {$1$} (1);    	

    \path (2) edge [bend left=20, right, shorten >=2, -stealth] node {$2$} (1);	
    \path (3) edge [bend left=20, below, shorten >=2, -stealth] node {$3$} (2);
    \path (1) edge [bend left=20, left, shorten >=2, -stealth]  node {$3$} (3);

    \path [draw, -stealth]   (0,-1) ++ (-0.5,   0) -- ++ (5.8,0) node [below, at end] {$\delta$};
    \path [draw, -stealth]   (0,-1) ++ (   0,-0.5) -- ++ (0,6.0);

    \path [draw, thick]      (0,-1) ++ (0, 0.7) node [left] {$a$} -- ++(3.0,0) --
                                    ++ (0, 0.7) -- ++ (-3.0,0) node [left] {$b$};
    \path [draw, thick]      (0,-1) ++ (0, 2.1) node [left] {$c$} -- ++(4.0,0) --
                                    ++ (0,-1.05) -- ++ (-1,0);
    \path [draw, thick]      (0,-1) ++ (0, 1.575) ++ (4.0,0) -- ++(1,0);

    \path [draw, thick]      (0,1.3) ++ (0, 0.7) node [left] {$a$} -- ++(1.5,0) --
                                     ++ (0, 0.7) -- ++ (-1.5,0) node [left] {$b$};
    \path [draw, thick]      (0,1.3) ++ (0, 2.1) node [left] {$c$} -- ++(3.3,0) --
                                     ++ (0,-1.05) -- ++ (-1.8,0);
    \path [draw, thick]      (0,1.3) ++ (0, 1.575) ++ (3.3,0) -- ++(1.7,0);

    \path [draw, dashed]      (0,-1) ++ (2, -0.5) node [right] {$\delta=1$} -- ++(0,6);

\end{tikzpicture}
}
\caption{Property of Influence. No clusters can be formed at resolutions for which it is impossible to form influence loops. Here, the loop of minimum cost is formed by circling the network clockwise where the maximum cost encountered is $A_X(b,c)=A_X(c,a)=1$. The top dendrogram is an invalid outcome because it has $a$ and $b$ clustering together at resolution $\delta<1$. The bottom dendrogram satisfies the Property of Influence (P1), [cf \eqref{eqn_mlc_lowerbounds_ultrametric}].}
\vspace{-0.1in}
\label{fig_axiom_of_influence}
\end{figure}
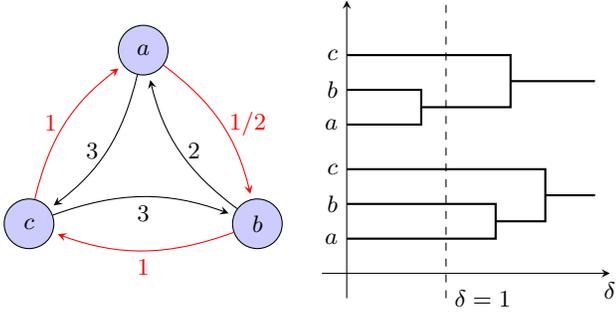 

%
\section{Influence modalities} \label{sec_axiomatic_structure}

The axiom of value states that in order for two nodes to belong to the same cluster they have to be able to exercise mutual influence on each other. When we consider a network with more than two nodes the concept of mutual influence is more difficult because it is possible to have direct influence as well as indirect chains of influence through other nodes. In this section we introduce two intuitive notions of mutual influence in networks of arbitrary size and show that they can be derived from the axioms of value and transformation. Besides their intrinsic value, these influence modalities are important for later developments in this paper; see, e.g. the proof of Theorem \ref{theo_extremal_ultrametrics}.

Consider first the intuitive notion that for two nodes to be part of a cluster there has to be a way for each of them to exercise influence on the other, either directly or indirectly. To formalize this idea, recall the concept of minimum loop cost \eqref{eqn_def_mlc} which we exemplify in Fig. \ref{fig_axiom_of_influence}. For this network, the loops $[a,b,a]$ and $[b,a,b]$ have maximum cost $2$ corresponding to the link $(b,a)$ in both cases. All other two-node loops have cost $3$. All of the counterclockwise loops, e.g., $[a,c,b,a]$, have cost $3$ and any of the clockwise loops have cost $1$. Thus, the minimum loop cost of this network is $\mlc(X,A_X)=1$. 

For resolutions $0 \leq \delta<\mlc(X,A_X)$ it is impossible to find chains of mutual influence with maximum cost smaller than $\delta$ between any pair of points. Indeed, suppose we can link $x$ to $x'$ with a chain of maximum cost smaller than $\delta$, and also link $x'$ to $x$ with a chain having the same property. Then, we can form a loop with cost smaller than $\delta$ by concatenating these two chains. Thus, the intuitive notion that clusters cannot form at resolutions for which it is impossible to observe mutual influence can be translated into the requirement that no clusters can be formed at resolutions $0 \leq \delta<\mlc(X,A_X)$. In terms of ultrametrics, this implies that it must be $u_X(x,x') \geq \mlc(X,A_X)$ for any pair of different nodes $x, x' \in X$ as we formally state next:

\myindentedparagraph{(P1) Property of Influence} For any network $N_X=(X,A_X)$ the output ultrametric $(X, u_X)=\ccalH(X,A_X)$ corresponding to the application of hierarchical clustering method $\ccalH$ is such that the ultrametric $u_X(x,x')$ between any two distinct points $x$ and $x'$ cannot be smaller than the minimum loop cost $\mlc(X,A_X)$ [cf. \eqref{eqn_def_mlc}] of the network
\begin{equation}\label{eqn_mlc_lowerbounds_ultrametric}
    u_X(x,x') \geq \mlc(X,A_X)   \quad\quad \forall \quad x \neq x'.
\end{equation}

\noindent Since for the network in Fig. \ref{fig_axiom_of_influence} the minimum loop cost is $\mlc(X,A_X)=1$, then the Property of Influence implies that $u_X(x,x')\geq\mlc(X,A_X)=1$ for any pair of nodes $x \neq x'$. Equivalently, the output dendrogram is such that for resolutions $\delta<\mlc(X,A_X)=1$ each node is in its own block. Observe that (P1) does not imply that a cluster with more than one node {\it is} formed at resolution $\delta=\mlc(X,A_X)$ but states that achieving this minimum resolution is a necessary condition for the formation of clusters.

%
\begin{figure}
\centering

\def \thisplotscale {0.6}
\def \unit {\thisplotscale cm}

{\small
\begin{tikzpicture}[-stealth, shorten >=2 ,scale = \thisplotscale, font=\footnotesize]

	\path                node [blue vertex]    (1)   {$1$} 
              ++ (4,  0) node [blue vertex]    (2)   {$2$}
	          ++ (4,  0) node [phantom vertex] (3)   {$\ldots$}
	          ++ (0.6,0) node [phantom vertex] (4)   {$\ldots$} 
	          ++ (3.2,0) node [blue vertex]    (n)   {$n$}; 

	\path (1) edge [bend left, above, draw=black!80, near start] node {$\alpha$} (2);	
	\path (1) edge [bend left, above, draw=black!80]             node {}         (3);	
	\path (1) edge [bend left, above, draw=black!80]             node {}         (n);	
	\path (2) edge [bend left, above, draw=black!80, near start] node {$\alpha$} (3);	
	\path (2) edge [bend left, above, draw=black!80]             node {}         (n);	

	\path (n) edge [bend left, below, draw=black!80, near start] node {$\beta$} (3);	
	\path (n) edge [bend left, below, draw=black!80]             node {}        (2);	
	\path (n) edge [bend left, below, draw=black!80]             node {}        (1);	
	\path (2) edge [bend left, below, draw=black!80, near start] node {$\beta$} (1);	

\end{tikzpicture}
} 
\vspace{-0.1in}
\caption{Canonical network $\vec{\Delta}_n(\alpha,\beta)$ for Extended Axiom of Value. Edges from a node to another node identified with a higher number have weight $\alpha$, whereas edges going to nodes identified with lower numbers have weight $\beta.$ All admissible methods $\ccalH$ cluster the $n$ nodes together at resolution $\max(\alpha, \beta)$.}
\vspace{-0.15in}
\label{fig_extended_axiom_of_value}
\end{figure}
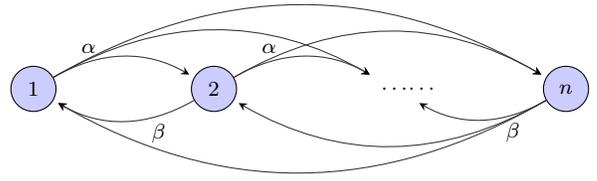

%
A second intuitive statement about influence in networks of arbitrary size comes in the form of the \emph{Extended Axiom of Value}. To introduce this concept define a family of canonical asymmetric networks
\begin{equation}\label{eqn:canonical_asymmetric_networks}
\vec{\Delta}_n(\alpha,\beta):=(\{1,\ldots,n\}, A_{n,\alpha,\beta}),
\end{equation}
with $n\in\N$ and $\alpha, \beta > 0$, where the underlying set $\{1,\ldots,n\}$ is the set of the first $n$ natural numbers and the dissimilarity value $A_{n,\alpha,\beta}(i,j)$ between points $i$ and $j$ depends on whether $i>j$ or not; see Fig. \ref{fig_extended_axiom_of_value}. For points $i>j$ we let $A_{n,\alpha,\beta}(i,j)=\alpha$ whereas for points $i<j$ we have $A_{n,\alpha,\beta}(i,j)=\beta$. Or, in matrix form,
\begin{equation}\label{eqn_extended_axiom_of_value_canonical_matrix}
   A_{n,\alpha,\beta}
      := \begin{pmatrix}
             0      & \alpha & \alpha & \alpha & \cdots & \alpha \\ 
             \beta  & 0      & \alpha & \alpha & \cdots & \alpha \\ 
             \beta  & \beta  & 0      & \alpha & \cdots & \alpha \\ 
             \vdots & \vdots & \vdots & \ddots & \vdots & \vdots \\
             \beta  & \beta  & \beta  & \cdots & 0      & \alpha \\
             \beta  & \beta  & \beta  & \cdots & \beta  & 0      
         \end{pmatrix}.
\end{equation}
In the network $\vec{\Delta}_n(\alpha,\beta)$ all pairs of nodes have dissimilarities $\alpha$ in one direction and $\beta$ in the other direction. This symmetry entails that all nodes should cluster together at the same resolution, and the requirement of mutual influence along with consistency with the Axiom of Value entails that this resolution should be $\max(\alpha,\beta)$. Before formalizing this definition notice that having clustering outcomes that depend on the ordering of the nodes in the space $\{1,\ldots,n\}$ is not desirable. Thus, we consider a permutation $\Pi= \{\pi_1, \pi_2, \ldots, \pi_n\}$ of $\{1,2, \ldots, n\}$ and the action $\Pi(A)$ of $\Pi$ on a $n \times n$ matrix $A$, which we define by $(\Pi(A))_{i,j} = A_{\pi_i,\pi_j}$ for all $i$ and $j$. Define now the network $\vec{\Delta}_n(\alpha,\beta,\Pi) : =(\{1,\ldots,n\},\Pi(A_{n,\alpha,\beta}))$ with underlying set $\{1,\ldots,n\}$ and dissimilarity matrix given $\Pi(A_{n,\alpha,\beta})$. With this definition we can now formally introduce the Extended Axiom of Value as follows: 

\myindentedparagraph{(A1') Extended Axiom of Value} Consider the network $\vec{\Delta}_n(\alpha,\beta,\Pi)=(\{1,\ldots,n\},\Pi(A_{n,\alpha,\beta}))$. Then, for all indices $n\in\mbN$, constants $\alpha,\beta>0$, and permutations $\Pi$ of $\{1,\ldots,n\}$, the outcome $(\{1,\ldots,n\},u)= \mathcal{H}\big(\vec{\Delta}_n(\alpha,\beta,\Pi)\big)$ of hierarchical clustering method $\ccalH$ applied to the network $\vec{\Delta}_n(\alpha,\beta,\Pi)$ satisfies
\begin{equation}
   u(i,j) = \max(\alpha,\beta),
\end{equation}  
for all pairs of nodes $i \neq j$.

\medskip\noindent Observe that the Axiom of Value (A1) is subsumed into the Extended Axiom of Value for $n=2$. Further note that the minimum loop cost of the canonical network $\vec{\Delta}_n(\alpha,\beta)$ is $\mlc \big(\vec{\Delta}_n(\alpha,\beta)\big)=\max(\alpha,\beta)$ because forming a loop requires traversing a link while moving right and a link while moving left at least once in Fig. \ref{fig_extended_axiom_of_value}. Since a permutation of indices does not alter the minimum loop cost of the network we have that
\begin{align}\label{eqn_mlc_extended_value} 
   \mlc \big(\vec{\Delta}_n(\alpha,\beta,\Pi)\big)= \mlc\big(\vec{\Delta}_n(\alpha,\beta)\big) = \max(\alpha, \beta).
\end{align} 
By the Property of Influence (P1) it follows from \eqref{eqn_mlc_extended_value} and \eqref{eqn_mlc_lowerbounds_ultrametric} that for the network $\vec{\Delta}_n(\alpha,\beta,\Pi)$ we must have $ u(i,j) \geq \mlc(\vec{\Delta}_n(\alpha,\beta))=\max(\alpha, \beta)$ for $i \neq j$. By the Extended Axiom of Value (A1') we have $u(i,j) = \max(\alpha,\beta)$ for $i \neq j$, which means that (A1') and (P1) are compatible requirements. We can then conceive of two alternative axiomatic formulations where admissible methods are required to abide by the Axiom of Transformation (A2), the Property of Influence (P1), and either the (regular) Axiom of Value (A1) or the Extended Axiom of Value (A1') -- Axiom (A1) and (P1) are compatible because (A1) is a particular case of (A1') which we already argued is compatible with (P1). We will see in the following section that these two alternative axiomatic formulations are equivalent to each other in the sense that a clustering method satisfies one set of axioms if and only if it satisfies the other. We further show that (P1) and (A1') are implied by (A1) and (A2). As a consequence, it follows that both alternative axiomatic formulations are equivalent to simply requiring validity of axioms (A1) and (A2). 

%
\subsection{Equivalent Axiomatic Formulations}

We begin by stating the equivalence between admissibility with respect to (A1)-(A2) and (A1')-(A2). A theorem stating that methods admissible with respect to (A1') and (A2) satisfy the Property of Influence (P1) is presented next to conclude that (A1)-(A2) imply (P1) as a consequence. 

%
\begin{theorem}\label{theo_extended_value}
Assume the hierarchical clustering method $\ccalH$ satisfies the Axiom of Transformation (A2). Then, $\ccalH$ satisfies the Axiom of Value (A1) if and only if it satisfies the Extended Axiom of Value (A1').
\end{theorem}

In proving Theorem \ref{theo_extended_value}, we make use of the following lemma which proves that, given a network, if the directed minimum chain cost $\tdu^*_X(x, x') \geq \delta$ between $x, x' \in X$ is at least $\delta$ it is possible to find a network partition separating $x$ and $x'$ such that the dissimilarities between points in different partitions are bounded below by $\delta$.

%
\begin{lemma}\label{lemma_axiom_redundancy}
Let $N=(X, A_X)$ be any network and $\delta$ any positive constant. Suppose that $x,x' \in X$ are such that their associated minimum chain cost [cf. \eqref{eqn_nonreciprocal_chains}] satisfies
\begin{align}\label{eqn_lem_axiom_redundancy_01} 
   \tdu^*_X(x, x') \geq \delta.
\end{align} 
Then, there exists a partition $P_\delta(x,x')=\{B_\delta(x), B_\delta(x')\}$ of the node space $X$ into blocks $B_\delta(x)$ and $B_\delta(x')$ with $x \in B_\delta(x)$ and $x' \in B_\delta(x')$ such that for all points $b \in B_\delta(x)$ and $b' \in B_\delta(x')$
\begin{align}\label{eqn_lem_axiom_redundancy_02} 
   A_X(b, b') \geq \delta.
\end{align} \end{lemma}

%
\begin{myproofnoname}
We prove this result by contradiction. If a partition $P_\delta(x,x')=\{B_\delta(x), B_\delta(x')\}$ with $x \in B_\delta(x)$ and $x' \in B_\delta(x)$ satisfying \eqref{eqn_lem_axiom_redundancy_02} does not exist for all pairs of points $x,x'\in X$ satisfying \eqref{eqn_lem_axiom_redundancy_01}, then there is at least one pair of nodes $x,x'\in X$ satisfying \eqref{eqn_lem_axiom_redundancy_01} such that for {\it all} partitions of $X$ into two blocks $P=\{B, B'\}$ with $x \in B$ and $x' \in B'$ we can find at least a pair of elements $b_P \in B$ and $b'_P \in B'$ for which
\begin{equation}\label{eqn_lem_axiom_redundancy_03} 
   A_X(b_P, b'_P) < \delta.
\end{equation}
Begin by considering the partition $P_1=\{B_1, B'_1\}$ where $B_1 = \{ x \}$ and $B'_1 = X \backslash \{x\}$. Since \eqref{eqn_lem_axiom_redundancy_03} is true for all partitions having $x\in B$ and $x'\in B'$ and $x$ is the unique element of $B_1$, there must exist a node $b'_{P_1} \in B'_1$ such that  
\begin{equation}\label{eqn_lem_axiom_redundancy_04} 
   A_X(x, b'_{P_1}) <  \delta. 
\end{equation}
Hence, the chain $C(x, b'_{P_1})= [x, b'_{P_1}]$ composed of these two nodes has cost smaller than $\delta$. Moreover, since $\tdu^*_X(x, b'_{P_1})$ represents the minimum cost among all chains $C(x, b'_{P_1})$ linking $x$ to $b'_{P_1}$, we can assert that
\begin{equation}\label{eqn_lem_axiom_redundancy_04_1} 
   \tdu^*_X(x, b'_{P_1}) \leq  A_X(x, b'_{P_1}) <  \delta.
\end{equation}
Consider now the partition $P_2=\{B_2, B'_2\}$ where $B_2= \{ x, b'_{P_1} \}$ and $B'_2=X \backslash B_2$. From \eqref{eqn_lem_axiom_redundancy_03}, there must exist a node $b'_{P_2} \in B'_2$ that satisfies at least one of the two following conditions 
\begin{align}
   &A_X(x, b'_{P_2}) <  \delta,  \label{eqn_lem_axiom_redundancy_05} \\
   &A_X(b'_{P_1}, b'_{P_2}) <  \delta. \label{eqn_lem_axiom_redundancy_06}
\end{align}
If \eqref{eqn_lem_axiom_redundancy_05} is true, the chain $C(x, b'_{P_2})=[x, b'_{P_2}]$ has cost smaller than $\delta$. If \eqref{eqn_lem_axiom_redundancy_06} is true, we combine the dissimilarity bound  with the one in \eqref{eqn_lem_axiom_redundancy_04} to conclude that the chain $C(x, b'_{P_2})=[x, b'_{P_1}, b'_{P_2}]$ has cost smaller than $\delta$. In either case we conclude that there exists a chain $C(x, b'_{P_2})$ linking $x$ to $b'_{P_2}$ whose cost is smaller than $\delta$. Therefore, the minimum chain cost must satisfy
\begin{equation}\label{eqn_lem_axiom_redundancy_04_2} 
\tdu^*_X(x, b'_{P_2}) <  \delta.
\end{equation}
Repeat the process by considering the partition $P_3$ with $B_3= \{ x, b'_{P_1}, b'_{P_2} \}$ and $B'_3=X\backslash B_3$. As we did in arguing \eqref{eqn_lem_axiom_redundancy_05}-\eqref{eqn_lem_axiom_redundancy_06} it must follow from \eqref{eqn_lem_axiom_redundancy_03} that there exists a point $b'_{P_3}$ such that at least one of the dissimilarities $A_X(x, b'_{P_3})$, $A_X(b'_{P_1}, b'_{P_3})$, or $A_X( b'_{P_2}, b'_{P_3})$ is smaller than $\delta$. This observation implies that at least one of the chains $[x, b'_{P_3}]$, $[x, b'_{P_1}, b'_{P_3}]$, $[x, b'_{P_2}, b'_{P_3}]$, or $[x, b'_{P_1}, b'_{P_2}, b'_{P_3}]$ has cost smaller than $\delta$ from where it follows
\begin{equation}\label{eqn_lem_axiom_redundancy_04_3} 
   \tdu^*_X(x, b'_{P_3}) <  \delta.
\end{equation}
This recursive construction can be repeated $n-1$ times to obtain partitions $P_1, P_2, ... , P_{n-1}$ and corresponding nodes $b'_{P_1}, b'_{P_2}, \dots, b'_{P_{n-1}}$ such that the minimum chain cost satisfies
\begin{equation}\label{eqn_lem_axiom_redundancy_04_4} 
   \tdu^*_X(x, b'_{P_i}) <  \delta, \qquad \text{for all\ } i.
\end{equation}
Observe that nodes $b'_{P_i}$ are distinct by construction and distinct from $x$. Since there are $n$ nodes in the network it must be that $x'=b'_{P_k}$ for some $i \in \{1, \ldots , n-1\}$. It follows from \eqref{eqn_lem_axiom_redundancy_04_4} that
\begin{equation}\label{eqn_lem_axiom_redundancy_04_5} 
\tdu^*_X(x, x') <  \delta.
\end{equation}
This is a contradiction because $x,x'\in X$ were assumed to satisfy \eqref{eqn_lem_axiom_redundancy_01}. Thus, the assumption that \eqref{eqn_lem_axiom_redundancy_03} is true for {\it all} partitions is incorrect. Hence, the claim that there is a partition $P_\delta(x,x')=\{B_\delta(x), B_\delta(x')\}$ satisfying \eqref{eqn_lem_axiom_redundancy_02} must be true. \end{myproofnoname}

%

\begin{myproof}[of Theorem \ref{theo_extended_value}]
To prove that (A1)-(A2) imply (A1')-(A2) let $\ccalH$ be a method that satisfies (A1) and (A2) and denote by $(\{1, 2, \ldots, n\},u_{n, \alpha, \beta})= \ccalH(\vec{\Delta}_n(\alpha,\beta,\Pi))$ the output ultrametric resulting of applying $\ccalH$ to the network $\vec{\Delta}_n(\alpha,\beta,\Pi)$ considered in the Extended Axiom of Value (A1'). We want to prove that (A1') is satisfied which means that we have to show that for all indices $n\in\mbN$, constants $\alpha,\beta>0$, permutations $\Pi$ of $\{1,\ldots,n\}$, and points $i\neq j$, we have $u_{n, \alpha, \beta}(i,j)=\max(\alpha,\beta)$. We will do so by showing both
\begin{align}
   &u_{n, \alpha, \beta}(i,j)\ \leq\ \max(\alpha,\beta)  \label{eqn_lem_axiom_redundancy_001}, \\
   &u_{n, \alpha, \beta}(i,j)\ \geq\ \max(\alpha,\beta)  \label{eqn_lem_axiom_redundancy_002},
\end{align}
for all $n\in\mbN$, $\alpha,\beta>0$, $\Pi$, and $i\neq j$.

To prove \eqref{eqn_lem_axiom_redundancy_001} define a symmetric two-node network $\vec{\Delta}_2(\max(\alpha, \beta), \max(\alpha, \beta))=(\{p,q\}, A_{p,q})$ where $A_{p,q}(p,q)=A_{p,q}(q,p)=\max(\alpha, \beta)$ and denote by $\big(\{p, q\}, u_{p,q}\big)=\ccalH(\vec{\Delta}_2(\max(\alpha, \beta), \max(\alpha, \beta)))$ the outcome of method $\ccalH$ when applied to $\vec{\Delta}_2(\max(\alpha, \beta), \max(\alpha, \beta))$. Since the method $\ccalH$ abides by (A1),
\begin{align}\label{eqn_lem_axiom_redundancy_110}
   u_{p,q}(p,q)
      = \max\big(\max(\alpha, \beta),\max(\alpha, \beta)\big)
      = \max(\alpha, \beta).
\end{align}
Consider now the map $\phi_{i,j}:\{p,q\} \to \{1,\ldots,n\}$ from the two-node network $\vec{\Delta}_2(\max(\alpha, \beta), \max(\alpha, \beta))$ to the permuted canonical network $\vec{\Delta}_n(\alpha,\beta,\Pi)$ where $\phi_{i,j}(p)=i$ and $\phi_{i,j}(q)=j$. Since dissimilarities in $\vec{\Delta}_n(\alpha,\beta,\Pi)$ are either $\alpha$ or $\beta$ and the dissimilarities in $\vec{\Delta}_2(\max(\alpha, \beta), \max(\alpha, \beta))$ are $\max(\alpha,\beta)$ it follows that the map $\phi_{i,j}$ is dissimilarity reducing regardless of the particular values of $i$ and $j$. Since the method $\ccalH$ was assumed to satisfy (A2) as well, we must have 
\begin{align}\label{eqn_lem_axiom_redundancy_120}
   u_{p,q} (p,q) 
      \geq  u_{n, \alpha, \beta}\big(\phi_{i,j}(p),\phi_{i,j}(q)\big) 
      =     u_{n, \alpha, \beta}(i,j) .
\end{align}
The inequality in \eqref{eqn_lem_axiom_redundancy_001} follows form substituting \eqref{eqn_lem_axiom_redundancy_110} into \eqref{eqn_lem_axiom_redundancy_120}.

In order to show inequality \eqref{eqn_lem_axiom_redundancy_002}, pick two arbitrary distinct nodes $i, j \in \{1,\ldots,n\}$ in the node set of $\vec{\Delta}_n(\alpha,\beta,\Pi)$. Denote by $C(i,j)$ and $C(j,i)$ two minimizing chains in the definition  \eqref{eqn_nonreciprocal_chains} of the directed minimum chain costs $\tdu^*_{n, \alpha, \beta}(i, j)$ and $\tdu^*_{n, \alpha, \beta}(j, i)$ respectively. Observe that at least one of the following two inequalities must be true
\begin{align}
   \tdu^*_{n, \alpha, \beta}(i, j) \geq \max(\alpha, \beta), \label{eqn_inequality_max_i_j} \\
   \tdu^*_{n, \alpha, \beta}(j, i) \geq \max(\alpha, \beta). \label{eqn_inequality_max_j_i}
\end{align}
Indeed, if both \eqref{eqn_inequality_max_i_j} and \eqref{eqn_inequality_max_j_i} were false, the concatenation of $C(i,j)$ and $C(j,i)$ would form a loop $C(i,i)=C(i,j) \uplus C(j,i)$ of cost strictly less than $\max(\alpha, \beta)$. This cannot be true because $\max(\alpha, \beta)$ is the minimum loop cost of the network  $\vec{\Delta}_n(\alpha,\beta,\Pi)$ as we already showed in \eqref{eqn_mlc_extended_value}.

Without loss of generality assume \eqref{eqn_inequality_max_i_j} is true and consider $\delta = \max(\alpha, \beta)$. By Lemma \ref{lemma_axiom_redundancy} we are therefore guaranteed to find a partition of the node set $\{1,\ldots,n\}$ into two blocks $B_\delta(i)$ and $B_\delta(j)$  with $i \in B_\delta(i)$ and $j \in B_\delta(j)$ such that for all $b \in B_\delta(i)$ and $b' \in B_\delta(j)$ it holds that 
\begin{equation}\label{eqn_b_bprime_max}
   \Pi(A_{n,\alpha,\beta})(b, b') \geq \delta = \max(\alpha, \beta).
\end{equation}
Define a two-node network $\vec{\Delta}_2(\max(\alpha, \beta), \min(\alpha, \beta))=(\{r, s\}, A_{r,s})$ where $A_{r,s}(r,s)=\max(\alpha, \beta)$ and $A_{r,s}(s,r)=\min(\alpha, \beta)$ and denote by $(\{r,s\},u_{r,s})=\ccalH(\vec{\Delta}_2(\max(\alpha, \beta), \min(\alpha, \beta)))$. Since the method $\ccalH$ satisfies (A1) we must have
\begin{equation}\label{eqn_max_max_min_a_b}
   u_{r,s}(r, s)
      = \max \big( \max(\alpha, \beta), \min(\alpha, \beta) \big) 
      = \max(\alpha, \beta).
\end{equation}
Consider the map $\phi'_{i,j} : \{1,\ldots,n\} \to \{r, s\}$ such that $\phi'_{i,j}(b)=r$ for all $b \in B_\delta(i)$ and $\phi'_{i,j}(b')=s$ for all $b' \in B_\delta(j)$. The map $\phi'_{i,j}$ is dissimilarity reducing because
\begin{equation}\label{eqn_phi_dissim_reducing_i_j}
      \Pi(A_{n,\alpha,\beta})(k,l) \geq A_{r,s}(\phi'_{i,j}(k), \phi'_{i,j}(l)),
\end{equation}
for all $k, l \in \{1,\ldots,n\}$. To see the validity of \eqref{eqn_phi_dissim_reducing_i_j} consider three different possible cases. If $k$ and $l$ belong both to the same block, i.e., either $k,l \in B_\delta(i)$ or $k,l \in B_\delta(j)$, then $\phi'_{i,j}(k)=\phi'_{i,j}(l)$ and $A_{r,s}(\phi'_{i,j}(k), \phi'_{i,j}(l))=0$ which cannot exceed the nonnegative $\Pi(A_{n,\alpha,\beta})(k,l)$. If $k \in B_\delta(j)$ and $l \in B_\delta(i)$ it holds that $A_{r,s}(\phi'_{i,j}(k), \phi'_{i,j}(l)) = A_{r,s}(s, r)= \min(\alpha, \beta)$ which cannot exceed $\Pi(A_{n,\alpha,\beta})(k,l)$ which is either equal to $\alpha$ or $\beta$. If $k \in B_\delta(i)$ and $l \in B_\delta(j)$, then we have $A_{r,s}(\phi'_{i,j}(k), \phi'_{i,j}(l)) = A_{r,s}(r, s)= \max(\alpha, \beta)$ but we also have $\Pi(A_{n,\alpha,\beta})(k,l)= \max(\alpha, \beta)$ as it follows by taking $b=k$ and $b'=l$ in \eqref{eqn_b_bprime_max}.

Since $\ccalH$ satisfies the Axiom of Transformation (A2) and the map $\phi'_{i,j}$ is dissimilarity reducing we must have
\begin{equation}\label{eqn_phi_dissim_reducing_i_j_2}
   u_{n, \alpha, \beta}(i,j) \geq u_{r,s}\big(\phi'_{i,j}(i), \phi'_{i,j}(j)\big) = u_{r,s}(r, s).
\end{equation}
Substituting \eqref{eqn_max_max_min_a_b} in \eqref{eqn_phi_dissim_reducing_i_j_2} we obtain the inequality \eqref{eqn_lem_axiom_redundancy_002}. Combining this result with the validity of \eqref{eqn_lem_axiom_redundancy_001}, it follows that $u_{n, \alpha, \beta}(i,j)= \max(\alpha,\beta)$ for all $n\in\mbN$, $\alpha,\beta>0$, $\Pi$, and $i\neq j$. Thus, admissibility with respect to (A1)-(A2) implies admissibility with respect to (A1')-(A2). That admissibility with respect to (A1')-(A2) implies admissibility with respect to (A1)-(A2) is immediate because (A1) is a particular case of (A1'). Hence, if a method satisfies axioms (A1') and (A2) it must satisfy (A1) and (A2). The stated equivalence between admissibility with respect to (A1)-(A2) and (A1')-(A2) follows.
\end{myproof}

%
The Axiom of Extended Value (A1') is stronger than the (regular) Axiom of Value (A1). However, Theorem \ref{theo_extended_value} shows that when considered together with the Axiom of Transformation (A2), both axioms of value are equivalent in the restrictions they impose in the set of admissible clustering methods $\ccalH$. In the following theorem we show that the Property of Influence (P1) can be derived from axioms (A1') and (A2).

%
\begin{theorem}\label{theo_influence_redundancy}
If a clustering method $\ccalH$ satisfies the axioms of extended value (A1') and transformation (A2) then it satisfies the Property of Influence (P1).
\end{theorem}

The following lemma is instrumental towards the proof of Theorem \ref{theo_influence_redundancy}.

%
\begin{lemma}\label{lemma_influence_redundancy}
Let $N=(X, A_X)$ be an arbitrary network with $n$ nodes and $\vec{\Delta}_n(\alpha,\beta)=(\{1,\ldots,n\}, A_{n,\alpha,\beta})$ be the canonical network in \eqref{eqn_extended_axiom_of_value_canonical_matrix} with $0 < \alpha \leq \sep(X, A_X)$ [cf. \eqref{eqn_def_separation_network}] and $\beta=\mlc(X, A_X)$ [cf. \eqref{eqn_def_mlc}]. Then, there exists a bijective map $\phi:X\to\{1,\ldots,n\}$ such that
\begin{align}\label{eqn_dissim_m_3}
   A_X(x, x') \geq A_{n, \alpha,\beta}(\phi(x), \phi(x')),
\end{align}
for all $x, x' \in X$.
 \end{lemma}

%
\begin{myproofnoname} To construct the map $\phi$ consider the function $P:X \to \mathcal{P}(X)$ from the node set $X$ to its power set $\mathcal{P}(X)$ such that
\begin{equation}\label{eqn_function_predecessor}
   P(x) :=\{ x' \in X \, | \, x' \neq x \,\, , \,\, A_X(x', x)<\beta\},
\end{equation} 
for all $x \in X$.
Having $r \in P(s)$ for some $r,s \in X$ implies that $A_X(r, s) < \beta=\mlc(X,A_X)$. An important observation is that we must have a node $x\in X$ whose $P$-image is empty. Otherwise, pick a node $x_n\in X$ and construct the chain $[x_0, x_1, \ldots , x_n]$ where the $i$th element of the chain $x_{i-1}$ is in the $P$-image of $x_i$. From the definition of the map $P$ it follows that all dissimilarities along this chain satisfy $A_X(x_{i-1},x_{i})<\beta=\mlc(X,A_X)$. But since the chain $[x_0, x_1, \ldots , x_n]$ contains $n+1$ elements, at least one node must be repeated. Hence, we have found a loop for which all dissimilarities are bounded above by $\beta=\mlc(X,A_X)$, which is impossible because it contradicts the definition of the minimum loop cost in \eqref{eqn_def_mlc}.  We can then find a node $x_{i_1}$ for which $P(x_{i_1})=\emptyset$. Fix $\phi(x_{i_1})=1$. 

Select now a node $x_{i_2}\neq x_{i_1}$ whose $P$-image is either $\{x_{i_1}\}$ or $\emptyset$, which we write jointly as $P(x_{i_2}) \subseteq \{x_{i_1}\}$. Such a node must exist, otherwise, pick a node $x_{n-1}\in X\backslash \{x_{i_1}\}$ and construct the chain $[x_0, x_1, \ldots , x_{n-1}]$ where $x_{i-1} \in P(x_i) \backslash \{x_{i_1}\}$, i.e. $x_{i-1}$ is in the $P$-image of $x_i$ and $x^{i-1} \neq \{x_{i_1}\}$. Since the chain $[x_0, x_1, ... , x_{n-1}]$ contains $n$ elements from the set $X\backslash \{x_{i_1}\}$ of cardinality $n-1$, at least one node must be repeated. Hence, we have found a loop where all dissimilarities between consecutive nodes satisfy $A_X(x^{i-1},x^{i})<\beta=\mlc(X,A_X)$, contradicting the definition of minimum loop cost. We can then find a node $x_{i_2} \neq x_{i_1}$ for which $P(x_{i_2}) \subseteq \{x_{i_1}\}$. Fix $\phi(x_{i_2})=2$.

Repeat this process $k$ times so that at step $k$ we have $\phi(x_{i_k})=k$ for a node $x_{i_k} \not\in \{x_{i_1},x_{i_2},\ldots x_{i_{k-1}}\}$ whose P-image is a subset of the nodes already picked, that is
\begin{equation}
   P(x_{i_k}) \subseteq \{x_{i_1},x_{i_2},\ldots x_{i_{k-1}}\}.
\end{equation}
This node must exist, otherwise, we could start with a node $x_{n-k+1} \in X \backslash \{x_{i_1},x_{i_2},\ldots x_{i_{k-1}}\}$ and construct a chain $[x_0, x_1, \ldots , x_{n-k+1}]$ where $x_{i-1} \in P(x_i) \backslash \{x_{i_1},x_{i_2},\ldots x_{i_{k-1}}\}$ and arrive to the same contradiction as for the case $k=2$.

Since all the nodes $x_{i_k}$ are different, the map $\phi$ with $\phi(x_{i_k})=k$ is bijective. By construction, $\phi$ is such that for all $l>k$, $x_{i_l}\notin P(x_{i_k})$. From \eqref{eqn_function_predecessor}, this implies that the dissimilarity from $x_{i_l}$ to $x_{i_k}$ must satisfy
\begin{equation}\label{eqn_dissim_m_1}
   A_X(x_{i_l},x_{i_k}) \geq \beta, \qquad \text{for all\ } l>k.
\end{equation} 
Moreover, from the definition of the canonical matrix $A_{n, \alpha,\beta}$ in \eqref{eqn_extended_axiom_of_value_canonical_matrix} we have that for $l>k$
\begin{equation}\label{eqn_dissim_m_2}
   A_{n, \alpha,\beta}(\phi(x_{i_l}), \phi(x_{i_k})) 
      = A_{n, \alpha,\beta}(l,k)    
      = \beta.
\end{equation}
By comparing \eqref{eqn_dissim_m_2} with \eqref{eqn_dissim_m_1} we conclude that \eqref{eqn_dissim_m_3} is true for all points with $\phi(x)>\phi(x')$. When $\phi(x)<\phi(x')$, we have $A_{n, \alpha,\beta}(\phi(x), \phi(x'))=\alpha$ which was assumed to be bounded above by the separation of the network $(X, A_X)$, thus, $A_{n, \alpha,\beta}(\phi(x), \phi(x'))$ is not greater than any positive dissimilarity in the range of $A_X$.
\end{myproofnoname}

%
\begin{myproof}[of Theorem \ref{theo_influence_redundancy}]
Consider a given arbitrary network $N=(X, A_X)$ with $X=\{x_1, x_2, ... , x_n\}$ and denote by $(X, u_X)=\ccalH(X,A_X)$ the output of applying the clustering method $\ccalH$ to the network $N$. The method $\ccalH$ is known to satisfy (A1') and (A2) and we want to show that it satisfies (P1) for which we need to show that $u_X(x,x')\geq\mlc(X,A_X)$ for all $x\neq x'$ [cf. \eqref{eqn_mlc_lowerbounds_ultrametric}]. 

Consider the canonical network $\vec{\Delta}_n(\alpha,\beta)=(\{1,\ldots,n\}, A_{n,\alpha,\beta})$ in \eqref{eqn_extended_axiom_of_value_canonical_matrix} with $\beta=\mlc(X, A_X)$ being the minimum loop cost of the network $N$ [cf. \eqref{eqn_def_mlc}] and $\alpha>0$ a constant not exceeding the separation of the network \eqref{eqn_def_separation_network}. Thus, we have $\alpha\leq\sep(X,A_X)\leq\mlc(X,A_X)=\beta$. Note that networks $N$ and $\vec{\Delta}_n(\alpha,\beta)$ have equal number of nodes.

%
Denote by $(\{1,\ldots,n\}, u_{\alpha,\beta}) = \ccalH(\vec{\Delta}_n(\alpha,\beta))$ the ultrametric space obtained when we apply the clustering method $\ccalH$ to the network $\vec{\Delta}_n(\alpha,\beta)$. Since $\ccalH$ satisfies the Extended Axiom of Value (A1'), then for all indices $i, j \in \{1,\ldots,n\}$ with $i \neq j$ we have
\begin{equation}\label{eqn_ultram_epsilon_m_1}
   u_{\alpha,\beta}(i,j) = \max(\alpha,\beta) = \beta = \mlc(X,A_X) .
\end{equation}
Further, focus on the bijective dissimilarity reducing map considered in Lemma \ref{lemma_influence_redundancy} and notice that since the method $\ccalH$ satisfies the Axiom of Transformation (A2) it follows that for all $x, x' \in X$ 
\begin{equation}\label{eqn_ultram_epsilon_m_2}
u_X(x,x') \geq u_{\alpha,\beta}(\phi(x), \phi(x')).
\end{equation}
Since the equality in \eqref{eqn_ultram_epsilon_m_1} is true for all $i\neq j$ and since all points $x\neq x'$ are mapped to points $\phi(x)\neq\phi(x')$ because $\phi$ is bijective, \eqref{eqn_ultram_epsilon_m_2} implies
\begin{equation}\label{eqn_ultram_epsilon_m_3}
   u_X(x,x') \geq \beta = \mlc(X, A_X),
\end{equation}
for all distinct $x, x' \in X$. This is the definition of the Property of Influence (P1). 
\end{myproof}

%
The fact that (P1) is implied by (A1') and (A2) as claimed by Theorem \ref{theo_influence_redundancy} implies that adding (P1) as a third axiom on top of these two is moot. Since we have already established in Theorem \ref{theo_extended_value} that (A1) and (A2) yield the same space of admissible methods as (A1') and (A2) we can conclude as a corollary of theorems \ref{theo_extended_value} and \ref{theo_influence_redundancy} that (P1) is also satisfied by all methods $\ccalH$ that satisfy (A1) and (A2).

%
\begin{corollary}\label{cor_axiom_redundancy}
If a given clustering method $\ccalH$ satisfies the axioms of value (A1) and transformation (A2), then it also satisfies the Property of Influence (P1).
\end{corollary}

%
\begin{myproofnoname}
If a clustering method $\ccalH$ satisfies (A1) and (A2), then by Theorem \ref{theo_extended_value} it must satisfy (A1') and (A2). If the latter is true, by Theorem \ref{theo_influence_redundancy} method $\ccalH$ must satisfy property (P1).
\end{myproofnoname}

%
\begin{figure*}
\centering

\def \thisplotscale {0.9}
\def \unit {\thisplotscale cm}

{
\begin{tikzpicture}[-stealth, shorten >=2 ,scale = \thisplotscale]

	\node[blue vertex] (x) {$x$} ;

	\path (x) ++ (4,0)   node [blue vertex]    (x1)   {$x_1$} 
	          ++ (4,0)   node [phantom vertex] (x2)   {$\ldots$}
	          ++ (0.5,0) node [phantom vertex] (xlm1) {$\ldots$} 
	          ++ (4,0)   node [blue vertex]    (xl)   {$x_{l-1}$}
	          ++ (4,0)    node [blue vertex]   (xp)   {$x'$}; 

	\path (x)    edge [bend left, above] node {$A_X(x,x_1)$}       (x1);	
	\path (x1)   edge [bend left, above] node {$A_X(x_1,x_2)$}     (x2);	
	\path (xlm1) edge [bend left, above] node {$A_X(x_{l-2},x_{l-1})$} (xl);	
	\path (xl)   edge [bend left, above] node {$A_X(x_{l-1},x')$}      (xp);	

	\path (x1)   edge [bend left, below] node {$A_X(x_1,x)$}       (x);
	\path (x2)   edge [bend left, below] node {$A_X(x_2,x_1)$}     (x1);
	\path (xl)   edge [bend left, below] node {$A_X(x_{l-1},x_{l-2})$} (xlm1);
	\path (xp)   edge [bend left, below] node {$A_X(x',x_{l-1})$}      (xl);

\end{tikzpicture}
} 
\caption{Reciprocal clustering. Nodes $x$ and $x'$ are clustered together at resolution $\delta$ if they can be joined with a (reciprocal) chain whose maximum dissimilarity is smaller than or equal to $\delta$ in both directions [cf. \eqref{eqn_reciprocal_clustering}]. Of all methods that satisfy the axioms of value and transformation, reciprocal clustering yields the largest ultrametric between any pair of nodes.}
\vspace{-0.1in}
\label{fig_reciprocal_path}
\end{figure*}
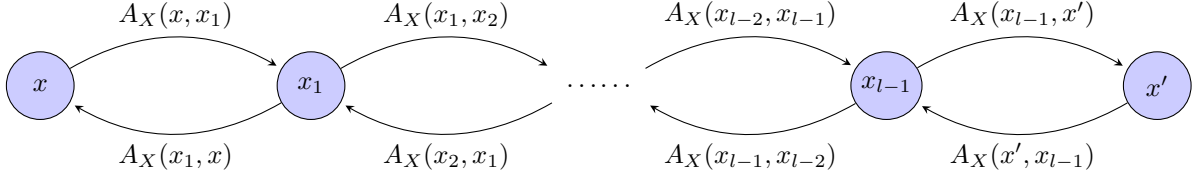

%
In the discussion leading to the introduction of the Axiom of Value (A1) in Section \ref{sec_axioms} we argued that the intuitive notion of a cluster dictates that it must be possible for co-clustered nodes to influence each other. In the discussion leading to the definition of the Property of Influence (P1) at the beginning of this section we argued that in networks with more than two nodes the natural extension is that co-clustered nodes must be able to influence each other either directly or through their indirect influence on other intermediate nodes. The Property of Influence is a codification of this intuition because it states the impossibility of cluster formation at resolutions where influence loops cannot be formed. While (P1) and (A1) seem quite different and seemingly independent, we have shown in this section that if a method satisfies axioms (A1) and (A2) it must satisfy (P1). Therefore, requiring direct influence on a two-node network as in (A1) restricts the mechanisms for indirect influence propagation so that clusters cannot be formed at resolutions that do not allow for mutual, possibly indirect, influence as stated in (P1). In that sense the restriction of indirect influence propagation in (P1) is not just {\it intuitively} reasonable but {\it formally} implied by the more straightforward restrictions on direct influence in (A1) and dissimilarity reducing maps in (A2).


%
\section {Reciprocal and nonreciprocal clustering}\label{sec_reicprocal_and_nonreciprocal}

%

Pick any network $N_X=(X,A_X)\in\ccalN$. One particular clustering method satisfying axioms (A1)-(A2) can be constructed by considering the \emph{symmetric} dissimilarity
\begin{equation}\label{eqn_reciprocal_clustering_sym_matrix} 
 \bbarA_X(x, x') := \max(A_X(x,x'), A_X(x',x)),
\end{equation}
for all $x,x' \in X$. This effectively reduces the problem to clustering of symmetric data, a scenario in which the single linkage method in \eqref{eqn_single_linkage} is known to satisfy axioms analogous to (A1)-(A2), \cite{clust-um}. Drawing upon this connection we define the \emph{reciprocal} clustering method $\ccalH^{\R}$ with output $(X,u^{\R}_X)=\ccalH^{\R}(X,A_X)$ as the one for which the ultrametric $u^{\R}_X(x,x')$ between points $x$ and $x'$ is given by
\begin{align}\label{eqn_reciprocal_clustering} 
    u^{\R}_X(x,x')
    &:= \min_{C(x,x')} \, \max_{i | x_i\in C(x,x')}
              \bbarA_X(x_i,x_{i+1}).
\end{align} 
%
%
%
An illustration of the definition in \eqref{eqn_reciprocal_clustering} is shown in Fig. \ref{fig_reciprocal_path}. We search for chains $C(x,x')$ linking nodes $x$ and $x'$. For a given chain we walk from $x$ to $x'$  and for every link, connecting say $x_i$ with $x_{i+1}$, we determine the maximum dissimilarity in both directions, i.e. the value of $\bbarA_X(x_i, x_{i+1})$. We then determine the maximum across all the links in the chain. The reciprocal ultrametric $u^{\R}_X(x,x')$ between points $x$ and $x'$ is the minimum of this value across all possible chains. Recalling the equivalence of dendrograms and ultrametrics provided by Theorem \ref{theo_dendrograms_as_ultrametrics}, we know that $\R_X$, the dendrogram produced by reciprocal clustering, clusters $x$ and $x'$ together for resolutions $\delta\geq u^{\R}_X(x,x')$. Combining the latter observation with \eqref{eqn_reciprocal_clustering}, we can write the reciprocal clustering equivalence classes as
\begin{equation}\label{eqn_reciprocal_clustering_dendrogram} 
   x\sim_{\R_X(\delta)} x' \iff  
       \min_{C(x,x')} \, \max_{i | x_i\in C(x,x')}\bbarA_X(x_i,x_{i+1})\leq\delta.
\end{equation}
Comparing \eqref{eqn_reciprocal_clustering_dendrogram} with the definition of single linkage in \eqref{eqn_single_linkage} with $\tdu^*_X(x, x')$ as defined in \eqref{eqn_nonreciprocal_chains}, we see that reciprocal clustering is equivalent to single linkage for the symmetrized network $N=(X,\bbarA_X)$ where dissimilarities between nodes are symmetrized to the maximum value of each directed dissimilarity.

For the method $\ccalH^{\R}$ specified in \eqref{eqn_reciprocal_clustering} to be a properly defined hierarchical clustering method, we need to establish that $u^{\R}_X$ is a valid ultrametric. One way of seeing that this is true is to observe that  $u^{\R}_X$ arises from applying single linkage hierarchical clustering to the symmetric dissimilarity $\bbarA_X$, which is known to output valid ultrametrics.

%
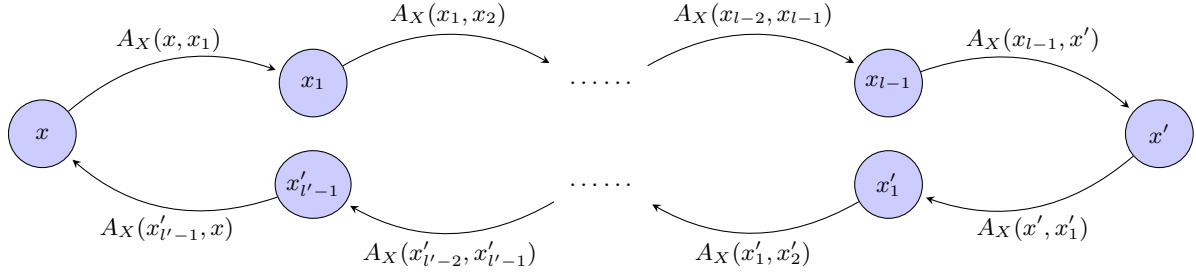
\begin{figure*}
\centering

\def \thisplotscale {0.9}
\def \unit {\thisplotscale cm}

{\small
\begin{tikzpicture}[-stealth, shorten >=2 ,scale = \thisplotscale]
	
	\node[blue vertex] (x) {$x$} ;

	\path (x) ++ (4,0.75)   node [blue vertex]    (x1)   {$x_1$} 
	          ++ (4,0)   node [phantom vertex] (x2)   {$\ldots$}
	          ++ (0.5,0) node [phantom vertex] (xlm1) {$\ldots$} 
	          ++ (4,0)   node [blue vertex]    (xl)   {$x_{l-1}$}
	          ++ (4,-0.75)  node [blue vertex]    (xp)   {$x'$}; 

	\path (x) ++ (4,-0.75)  node [blue vertex]    (xlp)   {$x'_{l'-1}$} 
	          ++ (4,0)   node [phantom vertex] (xlm1p) {$\ldots$}
	          ++ (0.5,0) node [phantom vertex] (x2p)   {$\ldots$} 
	          ++ (4,0)   node [blue vertex]    (x1p)   {$x'_{1}$}; 

	\path (x)    edge [bend left, above] node {$A_X(x,x_1)$}       (x1);	
	\path (x1)   edge [bend left, above] node {$A_X(x_1,x_2)$}     (x2);	
	\path (xlm1) edge [bend left, above] node {$A_X(x_{l-2},x_{l-1})$} (xl);	
	\path (xl)   edge [bend left, above] node {$A_X(x_{l-1},x')$}      (xp);	

	\path (xp)    edge [bend left, below] node {$A_X(x',x'_1)$}       (x1p);
	\path (x1p)   edge [bend left, below] node {$A_X(x'_1,x'_2)$}     (x2p);
	\path (xlm1p) edge [bend left, below] node {$A_X(x'_{l'-2},x'_{l'-1})$} (xlp);
	\path (xlp)   edge [bend left, below] node {$A_X(x'_{l'-1},x)$}       (x);

\end{tikzpicture}
} 
\caption{Nonreciprocal clustering. Nodes $x$ and $x'$ are co-clustered at resolution $\delta$ if they can be joined in both directions with possibly different (nonreciprocal) chains of maximum dissimilarity not greater than $\delta$ [cf. \eqref{eqn_nonreciprocal_clustering}]. Of all methods abiding to the axioms of value and transformation, nonreciprocal clustering yields the smallest ultrametric between any pair of nodes.}
\vspace{-0.1in}
\label{fig_nonreciprocal_path}
\end{figure*}

Nevertheless, here we directly verify that $u^{\R}_X$ as defined by \eqref{eqn_reciprocal_clustering} is indeed an ultrametric on the space $X$. It is clear that $u^{\R}_X(x,x')=0$ only if $x=x'$ and that $u^{\R}_X(x,x')=u^{\R}_X(x',x)$ because the definition is symmetric on $x$ and $x'$. To verify that the strong triangle inequality in \eqref{eqn_strong_triangle_inequality} holds, let $C^*(x,x')$ and $C^*(x',x'')$ be chains that achieve the minimum in \eqref{eqn_reciprocal_clustering} for $u^{\R}_X(x,x')$ and $u^{\R}_X(x',x'')$, respectively. The maximum cost in the concatenated chain $C(x,x'')=C^*(x,x')\uplus C^*(x',x'')$ does not exceed the maximum cost in each individual chain. Thus, while the maximum cost may be smaller on a different chain, the chain $C(x,x'')$ suffices to bound $u^{\R}_X(x,x'') \leq \max \big( u^{\R}_X(x,x'), u^{\R}_X(x',x'')\big)$ as in \eqref{eqn_strong_triangle_inequality}. It is also possible to prove that $\ccalH^{\R}$ satisfies axioms (A1)-(A2) as in the following proposition. 

%
\begin{proposition}\label{prop_reciprocal_axioms}
The reciprocal clustering method $\ccalH^{\R}$ is valid and admissible. I.e., $u_X^{\R}$ defined by \eqref{eqn_reciprocal_clustering} is an ultrametric for all networks $N_X=(X, A_X)$ and $\ccalH^{\R}$ satisfies axioms (A1)-(A2).
\end{proposition}

%
\begin{myproofnoname}
That $u_X^{\R}$ conforms to the definition of an ultrametric was proved in the paragraph preceding this proposition. To see that the Axiom of Value (A1) is satisfied pick an arbitrary two-node network $\vec{\Delta}_2(\alpha, \beta)$ as defined in Section \ref{sec_preliminaries} and denote by $(\{p,q\}, u^{\R}_{p,q})=\ccalH^\R(\vec{\Delta}_2(\alpha, \beta))$ the output of applying the reciprocal clustering method to $\vec{\Delta}_2(\alpha, \beta)$. Since every possible chain from $p$ to $q$ must contain $p$ and $q$ as consecutive nodes, applying the definition in \eqref{eqn_reciprocal_clustering} yields
\begin{equation}\label{eqn_theo_reciprocal_axioms_pf_10}
    u^{\R}_{p,q}(p,q) = \max \Big(A_{p,q}(p,q), A_{p,q}(q,p)\Big) 
                 = \max(\alpha,\beta).
\end{equation} 
Axiom (A1) is thereby satisfied. 

To show fulfillment of axiom (A2), consider two networks $(X, A_X)$ and $(Y, A_Y)$ and a dissimilarity reducing map $\phi:X \to Y$. Let $(X, u^\R_X)=\ccalH^\R(X, A_X)$ and $(Y, u^\R_Y)=\ccalH^\R(Y, A_Y)$ be the outputs of applying the reciprocal clustering method to networks $(X, A_X)$ and $(Y, A_Y)$. For an arbitrary pair of nodes $x, x' \in X$, denote by $C^*_X(x,x')=[x=x_0,\ldots, x_l=x']$ a chain that achieves the minimum reciprocal cost in \eqref{eqn_reciprocal_clustering} so as to write
\begin{equation}\label{eqn_theo_reciprocal_axioms_pf_40}
    u^{\R}_X(x,x') = \max_{i | x_i\in C^*_X(x,x')} \, \bbarA_X(x_i,x_{i+1}).
\end{equation}
Consider the transformed chain $C_Y(\phi(x),\phi(x'))=[\phi(x)=\phi(x_0),\ldots, \phi(x_l)=\phi(x')]$ in the space $Y$. Since the transformation $\phi$ does not increase dissimilarities we have that for all links in this chain $A_Y(\phi(x_i),\phi(x_{i+1}))\leq A_X(x_i,x_{i+1})$ and $A_Y(\phi(x_{i+1}),\phi(x_i))\leq A_X(x_{i+1},x_i)$. Combining this observation with \eqref{eqn_theo_reciprocal_axioms_pf_40} we obtain,
\begin{align}\label{eqn_theo_reciprocal_axioms_pf_50}
    \max_{\phi(x_i)\in C_Y(\phi(x),\phi(x'))} \bbarA_Y(\phi(x_i),\phi(x_{i+1})) \leq u^{\R}_X(x,x').
\end{align}
Further note that $C_Y(\phi(x),\phi(x'))$ is a particular chain joining $\phi(x)$ and $\phi(x')$ whereas the reciprocal ultrametric is the minimum across all such chains. Therefore,
\begin{equation}\label{eqn_theo_reciprocal_axioms_pf_60}
    u^{\R}_Y(\phi(x),\phi(x')) \leq \max_{\phi(x_i)\in C_Y(\phi(x),\phi(x'))} \bbarA_Y(\phi(x_i),\phi(x_{i+1})).
   \end{equation}
Substituting \eqref{eqn_theo_reciprocal_axioms_pf_50} in \eqref{eqn_theo_reciprocal_axioms_pf_60}, it follows that $u^{\R}_Y(\phi(x),\phi(x'))\leq u^{\R}_X(x,x')$. This is the requirement in \eqref{eqn_dissimilarity_reducing_ultrametric} for dissimilarity reducing transformations in the statement of Axiom (A2).\end{myproofnoname}

%
In reciprocal clustering, nodes $x$ and $x'$ belong to the same cluster at a resolution $\delta$ whenever we can go back and forth from $x$ to $x'$ at a maximum cost $\delta$ through the same chain. In nonreciprocal clustering we relax the restriction about the chain being the same in both directions and cluster nodes $x$ and $x'$ together if there are chains, possibly different, linking $x$ to $x'$ and $x'$ to $x$. To state this definition in terms of ultrametrics consider a given network $N=(X,A_X)$ and recall the definition of the unidirectional minimum chain cost $\tdu^*_X$ in \eqref{eqn_nonreciprocal_chains}. We define the \emph{nonreciprocal} clustering method $\ccalH^{\NR}$ with output $(X,u^{\NR}_X)=\ccalH^{\NR}(X,A_X)$ as the one for which the ultrametric $u^{\NR}_X(x,x')$ between points $x$ and $x'$ is given by the maximum of the unidirectional minimum chain costs $\tdu^*_X(x, x')$ and $\tdu^*_X(x', x)$ in each direction,
\begin{equation}\label{eqn_nonreciprocal_clustering} 
    u^{\NR}_X(x,x') := \max \Big( \tdu^*_X(x,x'),\ \tdu^*_X(x',x )\Big).
\end{equation} 
An illustration of the definition in \eqref{eqn_nonreciprocal_clustering} is shown in Fig. \ref{fig_nonreciprocal_path}. We consider forward chains $C(x,x')$ going from $x$ to $x'$ and backward chains $C(x',x)$ going from $x'$ to $x$. For each of these chains we determine the maximum dissimilarity across all the links in the chain. We then search independently for the best forward chain $C(x,x')$ and the best backward chain $C(x',x)$ that minimize the respective maximum dissimilarities across all possible chains. The nonreciprocal ultrametric $u^{\NR}_X(x,x')$ between points $x$ and $x'$ is the maximum of these two minimum values.

As it is the case with reciprocal clustering we can verify that $u_X^{\NR}$ is a properly defined ultrametric and that, as a consequence, the nonreciprocal clustering method $\ccalH^{\NR}$ is properly defined. Identity and symmetry are immediate. For the strong triangle inequality consider chains $C^*(x,x')$ and $C^*(x',x'')$ that achieve the minimum costs in $\tdu^*_X(x,x')$  and $\tdu^*_X(x',x'')$ as well as the chains $C^*(x'',x')$ and $C^*(x',x)$ that achieve the minimum costs in $\tdu^*_X(x'',x')$ and $\tdu^*_X(x',x)$. The concatenation of these chains permits concluding that $u^{\NR}_X(x,x'')\leq \max \big( u^{\NR}_X(x,x'), u^{\NR}_X(x',x'')\big)$, which is the strong triangle inequality in \eqref{eqn_strong_triangle_inequality}. The method $\ccalH^{\NR}$ also satisfies axioms (A1)-(A2) as the following proposition shows.

%
\begin{figure}
\centering

\def \thisplotscale {0.7}
\def \unit {\thisplotscale cm}

{\small
\begin{tikzpicture}[scale = \thisplotscale]

    \path   (-4.1, 3.5)   node [blue vertex] (1) {$a$}   
          ++( 2.3,-3.5) node [blue vertex] (2) {$b$}    
          ++(-4.6, 0)   node [blue vertex] (3) {$c$};

    \path (1) edge [bend left=20, right, shorten >=2, -stealth] node {$1/2$} (2);	
    \path (2) edge [bend left=20, below, shorten >=2, -stealth] node {$1/2$} (3);
    \path (3) edge [bend left=20, red, left, shorten >=2,  -stealth] node {$1$} (1);    	

    \path (2) edge [bend left=20, blue, right, shorten >=2, -stealth] node {$2$} (1);	
    \path (3) edge [bend left=20, mygreen, below, shorten >=2, -stealth] node {$3$} (2);
    \path (1) edge [bend left=20, left, shorten >=2, -stealth]  node {$4$} (3);

    \path [draw, -stealth]   (0,-1) ++ (-0.5,   0) -- ++ (6.0,0) node [below, at end] {$\delta$};
    \path [draw, -stealth]   (0,-1) ++ (   0,-0.5) -- ++ (0,6.0);

    \path [draw, thick, draw=blue]     (0,-1) ++ (0, 0.7) node [left] {$a$} -- ++(3.0,0) --
                                          ++ (0, 0.7) -- ++ (-3.0,0) node [left] {$b$};
    \path [draw, thick, draw=mygreen]  (0,-1) ++ (0, 2.1) node [left] {$c$} -- ++(4.5,0) --
                                          ++ (0,-1.05) -- ++ (-1.5,0);
    \path [draw, thick]                (0,-1) ++ (0, 1.575) ++ (4.5,0) -- ++(0.9,0);

    \path [draw, thick, draw=red] (0,1.3) ++ (0, 0.7) node [left] {$a$} -- ++(1.5,0) --
                                     ++ (0, 0.7) -- ++ (-1.5,0) node [left] {$b$};
    \path [draw, thick, draw=red] (0,1.3) ++ (0, 2.1) node [left] {$c$} -- ++(1.5,0) --
                                     ++ (0,-0.7);
    \path [draw, thick]           (0,1.3) ++ (0, 1.4) ++ (1.5,0) -- ++(3.9,0);


    \path [draw=black!50] (0,-1) ++ (3,   -0.5) node [left] {$2$} -- ++(0,0.5) -- ++(0,0.7);
    \path [draw=black!50] (0,-1) ++ (4.5, -0.5) node [left] {$3$} -- ++(0,0.5) -- ++(0,1.05);
    \path [draw=black!50] (0,-1) ++ (1.5, -0.5) node [left] {$1$} -- ++(0,0.5) -- ++(0,3.0);
    
    \node at (5,1.5) {$\R_X$};
        \node at (5,3.5) {$\NR_X$};

\end{tikzpicture}
}
\vspace{-0.15in}
\caption{Reciprocal and nonreciprocal dendrograms. An example network with its corresponding reciprocal (bottom) and nonreciprocal (top) dendrograms is shown. The optimal reciprocal chain linking $a$ and $b$ is $[a,b]$ the optimal chain linking $b$ and $c$ is $[b,c]$ and the optimal chain linking $a$ and $c$ is $[a,b,c]$. The optimal nonreciprocal chains linking $a$ and $b$ are $[a,b]$ and $[b,c,a]$. Of these two the cost of $[b,c,a]$ is larger.}
\vspace{-0.1in}
\label{fig_reciprocal_nonreciprocal_dendrograms}
\end{figure}
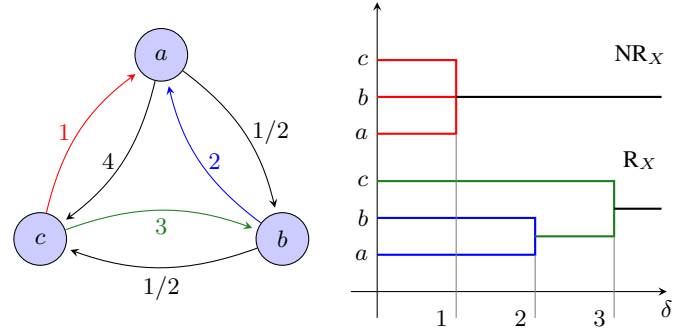

%
\begin{proposition}\label{prop_nonreciprocal_axioms}
The nonreciprocal clustering method $\ccalH^{\NR}$ is valid and admissible. I.e., $u_X^{\NR}$ defined by \eqref{eqn_nonreciprocal_clustering} is an ultrametric for all networks $N=(X, A_X)$ and $\ccalH^{\NR}$ satisfies axioms (A1)-(A2).
\end{proposition}

%
\begin{myproofnoname} See Appendix \ref{appendix_sec_reicprocal_and_nonreciprocal}.
\end{myproofnoname}

We denote by $\NR_X$ the dendrogram output by the nonreciprocal method $\ccalH^{\NR}$, equivalent to $u^{\NR}_X$ by Theorem \ref{theo_dendrograms_as_ultrametrics}.

The reciprocal and nonreciprocal dendrograms for an example network are shown in Fig. \ref{fig_reciprocal_nonreciprocal_dendrograms}. Notice that these dendrograms \emph{different}. In the reciprocal dendrogram nodes $a$ and $b$ cluster together at resolution $\delta=2$ due to their direct connections $A_X(a,b)=1/2$ and $A_X(b,a)=2$. Node $c$ joins this cluster at resolution $\delta=3$ because it links bidirectionally with $b$ through the direct chain $[b,c]$ whose maximum cost is $A_X(c,b)=3$.  The optimal reciprocal chain linking $a$ and $c$ is $[a,b,c]$ whose maximum cost is also $A_X(c,b)=3$. In the nonreciprocal dendrogram we can link nodes with different chains in each direction. As a consequence, $a$ and $b$ cluster together at resolution $\delta=1$ because the directed cost of the chain $[a,b]$ is $A_X(a,b)=1/2$ and the directed cost of the chain $[b,c,a]$ is $A_X(c,a)=1$. Similar chains demonstrate that $a$ and $c$ as well as $b$ and $c$ also cluster together at resolution $\delta=1$.


%
\section {Extremal ultrametrics}\label{sec_extremal_ultrametrics}

%
Given that we have constructed two admissible methods satisfying axioms (A1)-(A2), the question whether these two constructions are the only possible ones arises and, if not, whether they are special in some sense. We will see in Section \ref{sec_intermediate_ultrametrics} that there are constructions other than reciprocal and nonreciprocal clustering that satisfy axioms (A1)-(A2). However, we prove in this section that reciprocal and nonreciprocal clustering are a peculiar pair in that all possible admissible clustering methods are contained between them in a well-defined sense. To explain this sense properly, observe that since reciprocal chains [cf. Fig. \ref{fig_reciprocal_path}] are particular cases of nonreciprocal chains [cf. Fig. \ref{fig_nonreciprocal_path}] we must have that for all pairs of nodes $x,x'$
\begin{equation}\label{eqn_nonreciprocal_smaller_than_reciprocal} 
    u^{\NR}_X(x,x') \leq  u^{\R}_X(x,x').
\end{equation} 
I.e., nonreciprocal ultrametrics do not exceed reciprocal ultrametrics. An important characterization is that any method $\ccalH$ satisfying axioms (A1)-(A2) yields ultrametrics that lie between $u_X^{\NR}$ and $u_X^{\R}$ as we formally state in the following generalization of Theorem 18 in \cite{clust-um}.

%
\begin{theorem}\label{theo_extremal_ultrametrics}
Consider an admissible clustering method $\ccalH$ satisfying axioms (A1)-(A2). For an arbitrary given network $N=(X,A_X)$ denote by $(X,u_X)=\ccalH(N)$ the output of $\ccalH$ applied to $N$. Then, for all pairs of nodes $x,x'$
\begin{equation}\label{eqn_theo_extremal_ultrametrics} 
    u^{\NR}_X(x,x') \leq  u_X(x,x') \leq  u^{\R}_X(x,x'),
\end{equation} 
where $u^{\NR}_X(x,x')$ and $u^{\R}_X(x,x')$ denote the nonreciprocal and reciprocal ultrametrics as defined by \eqref{eqn_nonreciprocal_clustering} and \eqref{eqn_reciprocal_clustering}, respectively.
\end{theorem}

%
\begin{myproof}[of  ${\bf u^{\NR}_X(x,x') \leq  u_X(x,x')}$] Recall that validity of (A1)-(A2) implies validity of (P1) by Corollary \ref{cor_axiom_redundancy}. To show the first inequality in \eqref{eqn_theo_extremal_ultrametrics}, consider the nonreciprocal clustering equivalence relation $\sim_{\NR_X(\delta)}$ at resolution $\delta$ according to which $x \sim_{\NR_X(\delta)} x'$ if and only if $x$ and $x'$ belong to the same nonreciprocal cluster at resolution $\delta$. Notice that this is true if and only if $u^{\NR}_X(x,x')\leq\delta$. Further consider the space $Z := X \mod \sim_{\NR_X(\delta)}$ of corresponding equivalence classes and the map $\phi_{\delta}:X\to Z$ that maps each point of $X$ to its equivalence class. Notice that $x$ and $x'$ are mapped to the same point $z$ if they belong to the same cluster at resolution $\delta$, which allows us to write
\begin{equation}\label{eqn_theo_extremal_ultrametrics_pf_010}
    \phi_\delta(x) = \phi_\delta(x') \iff  u^{\NR}_X(x,x')\leq\delta.
\end{equation} 
We define the network $N_Z:=(Z,A_Z)$ by endowing $Z$ with the dissimilarity $A_Z$ derived from the dissimilarity $A_X$ as
\begin{equation}\label{eqn_theo_extremal_ultrametrics_pf_020}
    A_Z(z,z') := \min_{x\in\phi_\delta^{-1}(z), x'\in\phi_\delta^{-1}(z')} A_X(x,x').
\end{equation} 
The dissimilarity $A_Z(z,z')$ compares all the dissimilarities $A_X(x,x')$ between a member of the equivalence class $z$ and a member of the equivalence class $z'$ and sets $A_Z(z,z')$ to the value corresponding to the least dissimilar pair; see Fig. \ref{fig_proof_theo_extremal_ultrametrics}. Notice that according to construction, the map $\phi_\delta$ is dissimilarity reducing 
\begin{equation}\label{eqn_theo_extremal_ultrametrics_pf_025}
    A_X(x,x') \geq A_Z(\phi_\delta(x),\phi_\delta(x')),
\end{equation} 
because we either have $A_Z(\phi_\delta(x),\phi_\delta(x'))=0$ if $x$ and $x'$ are co-clustered at resolution $\delta$, or $A_X(x,x') \geq \min_{x\in\phi_\delta^{-1}(z), x'\in\phi_\delta^{-1}(z')} A_X(x,x') = A_Z(\phi_\delta(x),\phi_\delta(x'))$ if they are mapped to different equivalent classes. 

Consider now an arbitrary method $\ccalH$ satisfying axioms (A1)-(A2) and denote by $(Z,u_Z) = \ccalH(N_Z)$ the outcome of $\ccalH$ when applied to $N_Z$. To apply Property (P1) to this outcome we determine the minimum loop cost of $N_Z$ in the following claim.

%
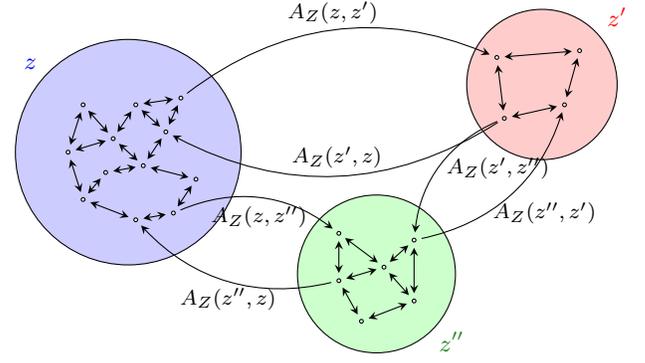
\begin{figure}
\centering
\def \thisplotscale {1}
\def \unit {\thisplotscale cm}

{\footnotesize
\begin{tikzpicture}[-stealth, shorten >=2,  shorten <=2, x = \unit, y=0.9*\unit]

    \node [blue vertex, minimum size = 3*\unit] at (0,0.1) (z) {};
    \path (z) ++ (-1.3,1.3) node {\blue{\small $z$}};
    \node [point] at ( 0.2,-0.1) (1) {}; \node [point] at ( 0.7, 0.9) (2) {};
    \node [point] at (-0.6,-0.6) (3) {}; \node [point] at ( 0.6,-0.8) (4) {};
    \node [point] at (-0.8, 0.1) (5) {}; \node [point] at ( 0.9,-0.3) (6) {};
    \node [point] at (-0.6, 0.8) (7) {}; \node [point] at ( 0.1, 0.8) (8) {};
    \node [point] at ( 0.1,-0.9) (9) {}; \node [point] at ( 0.5, 0.4) (10) {}; 
    \node [point] at (-0.2, 0.3) (11) {}; \node [point] at (-0.3,-0.2) (12) {};
    \path [stealth-stealth] (3) edge (5)   (3) edge (12)  (3) edge (9)   (5) edge (11);
    \path [stealth-stealth] (1) edge (12)  (1) edge (11)  (1) edge (10)  (1) edge (6);
    \path [stealth-stealth] (4) edge (9)   (6) edge (4)   (2) edge (10)  (7) edge (11);
    \path [stealth-stealth] (8) edge (11)  (8) edge (10)  (8) edge (2)   (5) edge (7);

    \path (z) ++ (5.5,1.0) node [red vertex, minimum size = 2*\unit] (zp) {}; 
    \path (zp) ++ (1,1) node {\red{\small $z'$}};    
    \path (zp) ++ (-0.6,0.4) node [point] (1p) {};  \path (zp) ++ (0.3,-0.3) node [point] (2p) {};
    \path (zp) ++ (0.5,0.5) node [point] (3p) {};  \path (zp) ++ (-0.5,-0.5) node [point] (4p) {};
    \path [stealth-stealth] (1p) edge (3p) (3p) edge (2p) (2p) edge (4p) (4p) edge (1p);

    \path (z) ++ (3.3,-1.8) node [green vertex, minimum size = 2.1*\unit] (zpp) {}; 
    \path (zpp) ++ (1,-1) node {\green{\small $z''$}};    
    \path (zpp) ++ (-0.5,0.6) node [point] (1pp) {};  \path (zpp) ++ (0.5,-0.4) node [point] (2pp) {};
    \path (zpp) ++ (0.5,0.5) node [point] (3pp) {};  \path (zpp) ++ (-0.2,-0.7) node [point] (4pp) {};
    \path (zpp) ++ (0.1,0.1) node [point] (5pp) {};  \path (zpp) ++ (-0.5,-0.1) node [point] (6pp) {};
    \path [stealth-stealth] (5pp) edge (3pp) (5pp) edge (1pp) (5pp) edge (2pp) (5pp) edge (6pp);
    \path [stealth-stealth] (1pp) edge (6pp) (6pp) edge (4pp) (4pp) edge (2pp) (2pp) edge (3pp);    
     
    \path (2) edge [bend left, above] node {$A_Z(z,z')$} (1p);
    \path (4p) edge [bend left, above] node {$A_Z(z',z)$} (10);    
    \path (4p) edge [bend right, pos=0.52, right] node {$A_Z(z',z'')$} (3pp);    
    \path (3pp) edge [bend right, pos=0.35, right] node {$A_Z(z'',z')$} (2p);   
    \path (4) edge [bend left, below] node {$A_Z(z,z'')$} (1pp);             
    \path (6pp) edge [bend left, below] node {$A_Z(z'',z)$} (9);             
\end{tikzpicture}
} 
\caption{\small Network of equivalence classes for a given resolution. Each shaded subset of nodes represent an equivalence class. The Axiom of Transformation permits relating the clustering of nodes in the original network and the clustering of nodes in the network of equivalence classes.}
\label{fig_proof_theo_extremal_ultrametrics}
\end{figure}

%
\begin{claim}\label{lemma_mlc_greater_delta}
The minimum loop cost of the network $N_Z$ is 
\begin{equation}\label{eqn_theo_extremal_ultrametrics_pf_028}
    \mlc(N_Z) > \delta.
\end{equation} \end{claim}

%
\begin{myproofnoname} According to the definition in \eqref{eqn_theo_extremal_ultrametrics_pf_020}, if $z \neq z'$ it must be that either $A_Z(z,z')>\delta$ or $A_Z(z',z)>\delta$ for otherwise $z$ and $z'$ would be the same equivalent class. Indeed, if both $A_Z(z,z')\leq\delta$ and $A_Z(z',z)\leq\delta$ we can build chains $C(x, x')$ and $C(x',x)$ with maximum cost smaller than $\delta$ for any $x\in\phi_\delta^{-1}(z)$ and $x'\in\phi_\delta^{-1}(z')$. For the chain $C(x, x')$ denote by $x_o\in\phi_\delta^{-1}(z)$ and $x_i'\in\phi_\delta^{-1}(z')$ the points achieving the minimum in \eqref{eqn_theo_extremal_ultrametrics_pf_020} so that $A_Z(z,z') = A_X(x_o,x_i')$. Since $x$ and $x_o$ are in the same equivalence class there is a chain $C(x, x_o)$ of maximum cost smaller than $\delta$. Likewise, since $x'_i$ and $x'$ are in the same class there is a chain $C(x_i', x')$ that joins them at maximum cost smaller than $\delta$. Therefore, the concatenated chain 
\begin{equation}\label{eqn_theo_extremal_ultrametrics_pf_030}
    C(x,x') = C(x,x_o) \uplus  [x_o,x_i'] \uplus C(x_i',x') ,
\end{equation}
has maximum cost smaller than $\delta$. The construction of the chain $C(x', x)$ is analogous. However, the existence of these two chains implies that $x$ and $x'$ are clustered together at resolution $\delta$ [cf \eqref{eqn_nonreciprocal_clustering}] contradicting the assumption that $z$ and $z'$ are different equivalent classes.

To prove that the minimum loop cost of $N_Z$ is $\mlc(Z,A_Z) > \delta$ assume that \eqref{eqn_theo_extremal_ultrametrics_pf_028} is not true and denote by $[z, z', \ldots, z^{(l)}, z]$ a loop of cost smaller than $\delta$. For any $x\in\phi_\delta^{-1}(z)$ and $x'\in\phi_\delta^{-1}(z')$ we can join $x$ to $x'$ using the chain $C(x, x')$ in \eqref{eqn_theo_extremal_ultrametrics_pf_030}. To join $x'$ and $x$ denote by $x_o^{(k)}$ and $x_i^{(k+1)}$ the points for which $A_Z(z^{(k)},z^{(k+1)}) = A_X(x_o^{(k)},x_i^{(k+1)})$ as in \eqref{eqn_theo_extremal_ultrametrics_pf_020}. We can then join $x'_o$ and $x^{(l)}_o$ with the concatenated chain
\begin{equation}\label{eqn_theo_extremal_ultrametrics_pf_050}
    C(x'_o, x^{(l)}_o) =
         \biguplus_{k=1}^{l-1} \left[x_o^{(k)},x_i^{(k+1)}\right] 
             \uplus C \left(x_i^{(k+1)},x_o^{(k+1)}\right).
\end{equation}
The maximum cost in traversing this chain is smaller than $\delta$ because the maximum cost in $C (x_i^{(k+1)},x_o^{(k+1)})$ is smaller than $\delta$ since both nodes belong to the same class $z^{(k+1)}$, and because $A_X(x_o^{(k)},x_i^{(k+1)})\leq\delta$ by assumption. We can now join $x'$ to $x$ with the concatenated chain
\begin{equation}\label{eqn_theo_extremal_ultrametrics_pf_060}
    C(x',x) =  C(x',x_o') \uplus  C(x'_o, x^{(l)}_o) \uplus  [x^{(l)}_o,x_i] \uplus  C(x_i,x),
\end{equation}
whose maximum cost is smaller than $\delta$. Using the chains \eqref{eqn_theo_extremal_ultrametrics_pf_030} and \eqref{eqn_theo_extremal_ultrametrics_pf_060} it follows that $u^{\NR}_X(x,x')\leq\delta$ contradicting the assumption that $x$ and $x'$ belong to different equivalent classes. Therefore, the assumption that \eqref{eqn_theo_extremal_ultrametrics_pf_028} is false cannot hold. The opposite must be true. \end{myproofnoname}

%
Continuing with the main proof, recall that $(Z,u_Z) = \ccalH(Z,A_Z)$. Since the minimum loop cost of $Z$ satisfies \eqref{eqn_theo_extremal_ultrametrics_pf_028} it follows from Property (P1) that for all pairs  $z,z'$, 
\begin{equation}\label{eqn_theo_extremal_ultrametrics_pf_070}
    u_Z(z,z') > \delta .
\end{equation}
Further note that according to \eqref{eqn_theo_extremal_ultrametrics_pf_025} and Axiom (A2) we must have $u_X(x,x') \geq u_Z(z,z')$. This fact, combined with \eqref{eqn_theo_extremal_ultrametrics_pf_070} allows us to conclude that when $x$ and $x'$ map to different equivalence classes
\begin{equation}\label{eqn_theo_extremal_ultrametrics_pf_080}
    u_X(x,x') \geq u_Z(z,z') >\delta.
\end{equation}
Notice that according to \eqref{eqn_theo_extremal_ultrametrics_pf_010}, $x$ and $x'$ mapping to different equivalence classes is equivalent to $ u^{\NR}_X(x,x')>\delta$. Consequently, we can claim that $u^{\NR}_X(x,x')>\delta$ implies $u_X(x,x')>\delta$, or, in set notation that
\begin{equation}\label{eqn_theo_extremal_ultrametrics_pf_090}
    \{(x,x') : u^{\NR}_X(x,x')>\delta\} \subseteq \{(x,x') : u_X(x,x')>\delta\}.
\end{equation}
Because \eqref{eqn_theo_extremal_ultrametrics_pf_090} is true for arbitrary $\delta>0$ it implies that $u^{\NR}_X(x,x') \leq  u_X(x,x')$ for all $x, x' \in X$ as in the first inequality in \eqref{eqn_theo_extremal_ultrametrics}. \end{myproof}

%

\begin{myproof}[of ${\bf u_X(x,x') \leq  u^{\R}_X(x,x')}$] To prove the second inequality in \eqref{eqn_theo_extremal_ultrametrics} consider points $x$ and $x'$ with reciprocal ultrametric $u^{\R}_X(x,x') = \delta$. Let $C(x,x')=[x=x_0,\ldots, x_l=x']$ be a chain achieving the minimum in \eqref{eqn_reciprocal_clustering} so that we can write
\begin{equation}\label{eqn_theo_extremal_ultrametrics_pf_100}
   \delta =  u^{\R}_X(x,x') = \max_i \,  \max \Big(A_X(x_i,x_{i+1}),  A_X(x_{i+1},x_i)\Big) .
\end{equation}
Turn attention to the symmetric two-node network $\vec{\Delta}_2(\delta,\delta)= (\{p,q\}, A_{p,q})$ with $A_{p,q}(p,q)=A_{p,q}(q,p)=\delta$. Denote the output of clustering method $\ccalH$ applied to network $\vec{\Delta}_2(\delta,\delta)$ as $(\{p,q\}, u_{p,q}) = \ccalH(\vec{\Delta}_2(\delta,\delta))$. Notice that according to Axiom (A1) we have $u_{p,q}(p,q) = \max(\delta, \delta)=\delta$.

Focus now on transformations $\phi_i:\{p,q\}\to X$ given by $\phi_i(p)=x_i$, $\phi_i(q)=x_{i+1}$ so as to map $p$ and $q$ to subsequent points in the chain $C(x,x')$ used in \eqref{eqn_theo_extremal_ultrametrics_pf_100}. Since it follows from \eqref{eqn_theo_extremal_ultrametrics_pf_100} that $A_X(x_i,x_{i+1})\leq\delta$  and $A_X(x_{i+1},x_i) \leq \delta$ for all $i$, it is just a simple matter of notation to observe that
\begin{align}\label{eqn_theo_extremal_ultrametrics_pf_110}
   A_X(\phi_i(p),\phi_i(q)) \leq A_{p,q}(p,q) = \delta, \nonumber\\
   A_X(\phi_i(q),\phi_i(p)) \leq A_{p,q}(q,p) = \delta.
\end{align}
Since according to \eqref{eqn_theo_extremal_ultrametrics_pf_110} transformations $\phi_i$ are dissimilarity-reducing, it follows from Axiom (A2) that 
\begin{align}\label{eqn_theo_extremal_ultrametrics_pf_120}
   u_X(\phi_i(p),\phi_i(q)) \leq u_{p,q}(p,q) = \delta.
\end{align}
Substituting the equivalences $\phi_i(p)=x_i$, $\phi_i(q)=x_{i+1}$ and recalling that \eqref{eqn_theo_extremal_ultrametrics_pf_120} is true for all $i$ we can equivalently write
\begin{align}\label{eqn_theo_extremal_ultrametrics_pf_130}
   u_X(x_i,x_{i+1}) \leq \delta, \quad \forall\ i .
\end{align}
To complete the proof we use the fact that since $u_X$ is an ultrametric and $C(x,x')=[x=x_0,\ldots, x_l=x']$ is a chain joining $x$ and $x'$ the strong triangle inequality dictates that
\begin{align}\label{eqn_theo_extremal_ultrametrics_pf_140}
   u_X(x,x') \leq \max_i u_X(x_i,x_{i+1}) \leq \delta,
\end{align}
where we used \eqref{eqn_theo_extremal_ultrametrics_pf_130} in the second inequality. The proof of the second inequality in \eqref{eqn_theo_extremal_ultrametrics} follows by substituting $\delta =  u^{\R}_X(x,x')$ [cf. \eqref{eqn_theo_extremal_ultrametrics_pf_100}] into \eqref{eqn_theo_extremal_ultrametrics_pf_140}. \end{myproof}

%
According to Theorem \ref{theo_extremal_ultrametrics}, nonreciprocal clustering applied to a given network $N=(X,A_X)$ yields a uniformly minimal ultrametric among those output by all clustering methods satisfying axioms (A1)-(A2). Reciprocal clustering yields a uniformly maximal ultrametric. Any other clustering method abiding by (A1)-(A2) yields an ultrametric such that the value $u_X(x,x')$ for any two points $x, x' \in X$ lies between the values $u^{\NR}_X(x,x')$ and $u^{\R}_X(x,x')$ assigned by nonreciprocal and reciprocal clustering. In terms of dendrograms, \eqref{eqn_theo_extremal_ultrametrics} implies that among all possible clustering methods, the smallest possible resolution at which nodes are clustered together is the one corresponding to nonreciprocal clustering. The highest possible resolution is the one that corresponds to reciprocal clustering.

%
\subsection{Hierarchical clustering on symmetric networks}\label{secsymmetric_networks}

Restrict attention to the subspace $\ccalM \subset \ccalN$ of symmetric networks, that is $N=(X,A_X)\in\mathcal{M}$ if and only if $A_X(x,x')=A_X(x',x)$ for all $x, x' \in X$. When restricted to the space $\ccalM$ reciprocal and nonreciprocal clustering are equivalent methods because, for any pair of points, minimizing nonreciprocal chains are always reciprocal -- more precisely there may be multiple minimizing nonreciprocal chains but at least one of them is reciprocal. To see this formally first fix $x,x'\in X$ and observe that in symmetric networks the symmetrization in \eqref{eqn_reciprocal_clustering_sym_matrix} is unnecessary because  $\bbarA_X(x_i,x_{i+1}) = A_X(x_i,x_{i+1}) = A_X(x_{i+1},x_i)$ and the definition of reciprocal clustering in \eqref{eqn_reciprocal_clustering}  reduces to
\begin{align}\label{eqn_reciprocal_symetric_networks} 
    u^{\R}_X(x,x')
       & = \min_{C(x,x')} \, \max_{i | x_i\in C(x,x')} A_X(x_i,x_{i+1}) \nonumber\\
       & = \min_{C(x',x)} \, \max_{i | x_i\in C(x',x)} A_X(x_i,x_{i+1}).
\end{align} 
Further note that the costs of any given chain $C(x,x')=[x=x_0, x_1, \ldots , x_{l-1}, x_l=x']$ and its reciprocal $C(x',x)=[x'=x_l, x_{l-1}, \ldots , x_{1}, x_0=x]$ are the same. It follows that directed minimum chain costs $\tdu^*_X(x, x')=\tdu^*_X(x', x)$ are equal and according to \eqref{eqn_nonreciprocal_clustering} equal to the nonreciprocal ultrametric 
\begin{align}\label{eqn_nonreciprocal_symetric_networks}
   u^{\NR}_X(x, x')= \tdu^*_X(x, x') = \tdu^*_X(x', x) = u^{\R}_X(x,x').
\end{align} 
To write the last equality in \eqref{eqn_nonreciprocal_symetric_networks} we used the definitions   of $\tdu^*_X(x, x')$ and $\tdu^*_X(x', x)$ in \eqref{eqn_nonreciprocal_chains} which are correspondingly equivalent to the first and second equality in \eqref{eqn_reciprocal_symetric_networks}.

By further comparison of the ultrametric definition of single linkage in \eqref{eqn_single_linkage_ultrametric} with \eqref{eqn_nonreciprocal_symetric_networks} the equivalence of reciprocal, nonreciprocal, and single linkage clustering in symmetric networks follows
\begin{align}\label{eqn_nonreciprocal_sl_reciprocal}
   u^{\NR}_X(x, x')= u^{\SL}_X(x,x') = u^{\R}_X(x,x').
\end{align} 
The equivalence in \eqref{eqn_nonreciprocal_symetric_networks} along with Theorem \ref{theo_extremal_ultrametrics} demonstrates that when considering the application of hierarchical clustering methods $\ccalH:\ccalM\to\ccalU$ to symmetric networks, there exist a unique method satisfying (A1)-(A2). The equivalence in \eqref{eqn_nonreciprocal_sl_reciprocal} shows that this method is single linkage. Before stating this result formally let us define the symmetric version of the Axiom of Value:

\myindentedparagraph{(B1) Symmetric Axiom of Value} Consider a symmetric two-node network $\vec{\Delta}_2(\alpha, \alpha)=(\{p, q\},A_{p, q})$ with $A_{p, q}(p,q)=A_{p, q}(q,p)=\alpha$. The ultrametric output $(\{p, q\},u_{p, q})=\ccalH(\vec{\Delta}_2(\alpha, \alpha))$  produced by $\ccalH$ satisfies 
\begin{equation}\label{eqn_two_node_network_ultrametric_symmetric}
   u_{p, q}(p,q) = \alpha.
\end{equation} 

\medskip \noindent Since there is only one dissimilarity in a symmetric network with two nodes, (B1) states that they cluster together at the resolution that connects them to each other. We can now invoke Theorem \ref{theo_extremal_ultrametrics} and \eqref{eqn_nonreciprocal_sl_reciprocal} to prove that single linkage is the unique hierarchical clustering method in symmetric networks that is admissible with respect to (B1) and (A2). 

%
\begin{corollary}\label{cor_single_linkage} Let $\ccalH:\ccalM\to\ccalU$ be a hierarchical clustering method for symmetric networks $N=(X,A_X)\in\ccalM$, that is $A_X(x,x')=A_X(x',x)$ for all $x, x' \in X$, and $\ccalH^{\SL}$ be the single linkage method with output ultrametrics as defined in \eqref{eqn_single_linkage_ultrametric}. If $\ccalH$ satisfies axioms (B1) and (A2) then $\ccalH\equiv\ccalH^{\SL}$.
\end{corollary}

%
\begin{myproofnoname} When restricted to symmetric networks (B1) and (A1) are equivalent statements. Thus, $\ccalH$ satisfies the hypotheses of Theorem \ref{theo_extremal_ultrametrics} and as a consequence \eqref{eqn_theo_extremal_ultrametrics} is true for any pair of points $x,x'$ of any network $N\in\ccalM$. But by \eqref{eqn_nonreciprocal_sl_reciprocal} nonreciprocal, single linkage, and reciprocal ultrametrics coincide. Thus, we can reduce \eqref{eqn_theo_extremal_ultrametrics} to
\begin{equation}\label{eqn_theo_extremal_ultrametrics_single_linkage} 
    u^{\text{SL}}_X(x,x') \leq  u_X(x,x') \leq  u^{\text{SL}}_X(x,x').
\end{equation} 
It then must be $u^{\text{SL}}_X(x,x')=u_X(x,x')$ for any pair of points $x,x'$ of any network $N\in\ccalM$. This means $\ccalH\equiv\ccalH^{\SL}$. \end{myproofnoname}

%
The uniqueness result claimed by Corollary \ref{cor_single_linkage} strengthens the uniqueness result in \cite[Theorem 18]{clust-um}. To explain the differences consider the symmetric version of the Property of Influence. In a symmetric network there is always a loop of minimum cost of the form $[x,x',x]$ for some pair of points $x,x'$. Indeed, say that $C^*(x^*,x^*)$ is one of the loops achieving the minimum cost in \eqref{eqn_def_mlc} and let $A_X(x,x')=\mlc(X,A_X)$ be the maximum dissimilarity in this loop. Then, the cost of the loop $[x,x',x]$ is $A_X(x,x')=A_X(x',x)=\mlc(X,A_X)$ which means that either the loop $C^*(x^*,x^*)$ was already of the form $[x,x',x]$ or that the cost of the loop $[x,x',x]$ is the same as $C^*(x^*,x^*)$. In any event, there is a loop of minimum cost of the form $[x,x',x]$ which implies that in symmetric networks we must have
\begin{equation}\label{eqn_mlc_equals_sep_symetric}
    \mlc(X,A_X) = \min_{x \neq x'} A_X(x, x') = \sep(X,A_X),
\end{equation}
where we recalled the definition of the separation of a network stated in \eqref{eqn_def_separation_network} to write the second equality. With this observation we can now introduce the symmetric version of the Property of Influence (P1):

\myindentedparagraph{(Q1) Symmetric Property of Influence} For any symmetric network $N_X=(X,A_X)$ the output $(X, u_X)=\ccalH(X,A_X)$ corresponding to the application of hierarchical clustering method $\ccalH$ is such that the ultrametric $u_X(x,x')$ between any two distinct points $x$ and $x'$ cannot be smaller than the separation of the network [cf. \eqref{eqn_mlc_equals_sep_symetric}],
\begin{equation}\label{eqn_separation_lowerbounds_ultrametric}
    u_X(x,x') \geq \sep(X,A_X).
\end{equation}

\medskip\noindent In \cite{clust-um} admissibility is defined with respect to (B1), (A2), and (Q1), which corresponds to conditions (I), (II), and (III) of \cite[Theorem 18]{clust-um}. Corollary \ref{cor_single_linkage} shows that Property (Q1) is redundant when given Axioms (B1) and (A2) -- respectively, Condition (III) of \cite[Theorem 18]{clust-um} is redundant when given conditions (I) and (II) of  \cite[Theorem 18]{clust-um}. Corollary \ref{cor_single_linkage} also shows that single linkage is the unique admissible method for all symmetric, not necessarily metric, networks.


%
\section{Intermediate Clustering Methods}\label{sec_intermediate_ultrametrics}
Reciprocal and nonreciprocal clustering bound the range of clustering methods satisfying axioms (A1)-(A2) in the sense specified by Theorem \ref{theo_extremal_ultrametrics}. Since methods $\ccalH^{\R}$ and $\ccalH^{\NR}$ are in general different (e.g. recall the example in Fig. \ref{fig_reciprocal_nonreciprocal_dendrograms}) a question of great interest is whether one can identify methods which are \emph{intermediate} to $\ccalH^{\R}$ and $\ccalH^{\NR}$.

In this section we study three types of intermediate methods. In Section \ref{sec_grafting} we introduce grafting methods, which are built by exchanging branches between dendrograms generated by different admissible methods. In Section \ref{sec_convex_comb}, we compute a form of convex combination of dendrograms generated by admissible methods to obtain new admissible methods. In Section \ref{sec_inter_reciprocal}, we present the semi-reciprocal family which requires part of the influence to be reciprocal and allows the rest to propagate through loops. These latter methods arise as natural intermediate ultrametrics in an algorithmic sense, as further discussed in Section \ref{sec_algorithms}.

\subsection{Grafting and related constructions}\label{sec_grafting}
\begin{figure}
\centering

\def \thisplotscale {0.9}
\def \unit {\thisplotscale cm}

{\small
\begin{tikzpicture}[scale = \thisplotscale, x=1.0*\unit, y = 0.8*\unit]

    \path (-0.5,2.9) ++ (0, 0.0) node [blue vertex, scale=0.8] (1) {\large{$a$}}
                     +  (4, 0.0) node [blue vertex, scale=0.8] (2) {\large{$b$}}
                     +  (4,-3.3) node [blue vertex, scale=0.8] (3) {\large{$c$}}    
                     +  (0,-3.3) node [blue vertex, scale=0.8] (4) {\large{$d$}};

    \path (1) edge [bend left, above, -stealth, shorten >=2] node {$1$} (2);	
    \path (2) edge [bend left, right, -stealth, shorten >=2] node {$1$} (3);
    \path (3) edge [bend left, below, -stealth, shorten >=2] node {$1$} (4);    
    \path (4) edge [bend left, left,  -stealth, shorten >=2] node {$1$} (1);  	

    \path (2) edge [bend left, below, -stealth, shorten >=2] node {$3$} (1);	
    \path (3) edge [bend left, left,  -stealth, shorten >=2] node {$5$} (2);
    \path (4) edge [bend left, above, -stealth, shorten >=2] node {$2$} (3);   
    \path (1) edge [bend left, right, -stealth, shorten >=2] node {$5$} (4);

    \path [draw, -stealth] (-2,-9.5) 
           + (-0.5,   0) -- + (7,0.0) node [below, at end] {$\delta$};
    \path [draw, -stealth] (-2,-9.5)  
           + (   0,-0.5) -- + ( 0.0,7.5);
    \path [thin, draw=black!30]  (-2,-9.5) ++ (0,-0.2) 
                    ++ (1,0) node [below] {1} -- + (0,7.5)
                    ++ (1,0) node [below] {2} -- + (0,7.5)
                    ++ (1,0) node [below] {3} -- + (0,7.5)
                    ++ (1,0) 
                    ++ (1,0) node [below] {5} -- + (0,7.5)
                    ++ (1,0) node [below] {6} -- + (0,7.5);

    \path [draw, thick, draw=red] (-2,-9.5) ++ (0,5.5)
                      node [left] {$d$} -- +  (2, 0) 
           ++ (0,0.5) node [left] {$c$} -- ++ (2, 0) -- ++ (0,-0.5);
    \path [draw, thick, draw=blue](-2,-9.5) ++ (0,5.5)
           ++ (0,1.0) node [left] {$b$} -- +  (3, 0) 
           ++ (0,0.5) node [left] {$a$} -- ++ (3, 0) -- ++ (0,-0.5);
    \path [draw, thick, draw=black](-2,-9.5) ++ (0,5.5)
           ++ (0,0.25) ++ (2,0) -- +  (3, 0) 
           ++ (0,1.0)  ++ (1,0) -- ++ (2, 0) -- ++ (0,-1.0);
    \path [draw, thick, draw=black](-2,-9.5) ++ (0,5.5)
           ++ (0,0.75) ++ (5,0) -- +  (1.5, 0);
    \path (-2,-9.5) ++ (0,5.5) ++ (6.0,1.2) node [right] {$\ccalH^\R$};

    \path [draw, thick, draw=red] (-2,-9.5) ++ (0,3.0)
                      node [left] {$d$} -- +  (1, 0) 
           ++ (0,0.5) node [left] {$c$} -- ++ (1, 0) -- ++ (0,-0.5);
    \path [draw, thick, draw=blue](-2,-9.5) ++ (0,3.0)
           ++ (0,1.0) node [left] {$b$} -- +  (1, 0) 
           ++ (0,0.5) node [left] {$a$} -- ++ (1, 0) -- ++ (0,-0.5);
    \path [draw, thick, draw=black](-2,-9.5) ++ (0,3.0)
           ++ (0,1.0)  ++ (1,0) -- ++ (0,-0.5);
    \path [draw, thick, draw=black](-2,-9.5) ++ (0,3.0)
           ++ (0,0.75) ++ (1,0) -- +  (5.5, 0);
    \path (-2,-9.5) ++ (0,3.0) ++ (6.0,1.2) node [right] {$\ccalH^{\NR}$};

    \path [draw, thick, draw=red] (-2,-9.5) ++ (0,0.5)
                      node [left] {$d$} -- +  (1, 0) 
           ++ (0,0.5) node [left] {$c$} -- ++ (1, 0) -- ++ (0,-0.5);
    \path [draw, thick, draw=blue](-2,-9.5) ++ (0,0.5)
           ++ (0,1.0) node [left] {$b$} -- +  (1, 0) 
           ++ (0,0.5) node [left] {$a$} -- ++ (1, 0) -- ++ (0,-0.5);
    \path [draw, thick, draw=black](-2,-9.5) ++ (0,0.5)
           ++ (0,0.25) ++ (1,0) -- +  (4, 0) 
           ++ (0,1.0)  ++ (0,0) -- ++ (4, 0) -- ++ (0,-1.0);
    \path [draw, thick, draw=black](-2,-9.5) ++ (0,0.5)
           ++ (0,0.75) ++ (5,0) -- +  (1.5, 0);
    \path (-2,-9.5) ++ (0,0.5) ++ (6.0,1.2) node [right] {$\ccalH^{\R/\NR}$};
           
    \path [thick, draw=mygreen]  (-2,-9.5) ++ (0,-0.2) 
                    ++ (4,0) node [below] {$\beta=4$} -- + (0,7.5);

\end{tikzpicture}}
\vspace{-0.1in}
\caption{Dendrogram grafting. Reciprocal ($\ccalH^{\R}$) and nonreciprocal ($\ccalH^{\NR}$) dendrograms for the given network are shown -- edges not drawn have dissimilarities greater than 5. Grafting according to \eqref{def_mu_beta_1} with $\beta=4$ is performed to construct the dendrogram corresponding to the method $\ccalH^{\R/\NR}(4)$. Branches of the reciprocal dendrogram are cut at resolution $\beta=4$ and replaced by corresponding branches of the nonreciprocal dendrogram.}
\vspace{-0.1in}
\label{fig_beta_ultrametrics_2}
\end{figure}
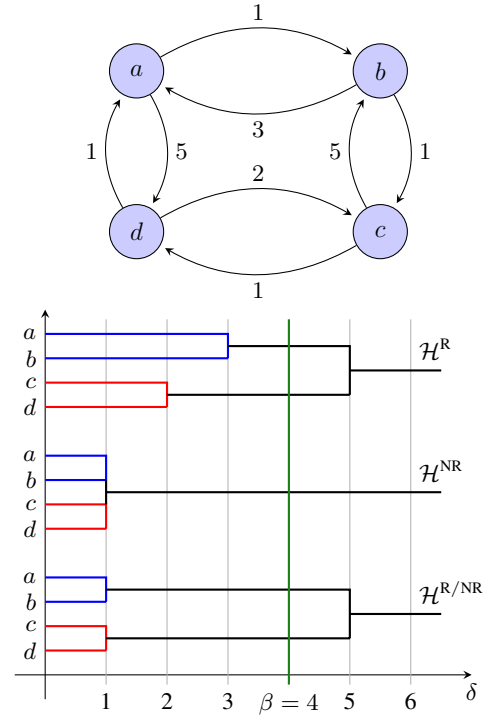

A family of admissible methods can be constructed by grafting branches of the nonreciprocal dendrogram into corresponding branches of the reciprocal dendrogram; see Fig. \ref{fig_beta_ultrametrics_2}. To be precise, consider a given positive constant $\beta>0$. For any given network $N=(X,A_X)$ compute the reciprocal and nonreciprocal dendrograms and cut all branches of the reciprocal dendrogram at resolution $\beta$. For each of these branches define the corresponding branch in the nonreciprocal tree as the one whose leaves are the same. Replacing the previously cut branches of the reciprocal tree by the corresponding branches of the nonreciprocal tree yields the $\ccalH^{\R/\NR}(\beta)$ method. Grafting is equivalent to providing the following piecewise definition of the output ultrametric; for $x,x'\in X$ let
\begin{equation}\label{def_mu_beta_1}
   u^{\R/\NR}_X(x,x';\beta) :=
   \begin{cases}
      u^{\NR}_X(x,x'), & \text{if } u^{\R}_X(x,x') \leq \beta, \\
      u^{\R}_X(x,x'),  & \text{if } u^{\R}_X(x,x') >    \beta .
   \end{cases}
\end{equation}
For pairs $x,x'$ having large reciprocal ultrametric $u^{\R}_X(x,x')>\beta$ we keep the reciprocal ultrametric value $u^{\R/\NR}_X(x,x';\beta)=u^{\R}_X(x,x')$. For pairs $x,x'$ with small reciprocal ultrametric $u^{\R}_X(x,x')\leq\beta$ we replace the reciprocal by the nonreciprocal ultrametric and make $u^{\R/\NR}_X(x,x';\beta)=u^{\NR}_X(x,x')$. 
 
To show that \eqref{def_mu_beta_1} is an admissible method we need to show that it defines an ultrametric on the space $X$ and that the method satisfies axioms (A1) and (A2). This is asserted in the following proposition. 

\begin{proposition}\label{prop_beta_1}
The hierarchical clustering method $\ccalH^{\R/\NR}(\beta)$ is valid and admissible. I.e., $u_X^{\R/\NR}(\beta)$ defined by \eqref{def_mu_beta_1} is an ultrametric for all networks $N=(X, A_X)$ and $\ccalH^{\R/\NR}(\beta)$ satisfies axioms (A1)-(A2).
\end{proposition}

\begin{myproofnoname} See Appendix \ref{appendix_sec_intermediate_ultrametrics}. \end{myproofnoname}

Since $u^{\R/\NR}_X(x,x';\beta)$ coincides with either $u^{\NR}_X(x,x')$ or $u^{\R}_X(x,x')$ for all $x, x' \in X$, it satisfies Theorem \ref{theo_extremal_ultrametrics} as it should be the case for the output ultrametric of any admissible method. 

An example construction of $u^{\R/\NR}_X(x,x';\beta)$ for a particular network and $\beta=4$ is illustrated in Fig. \ref{fig_beta_ultrametrics_2}. The nonreciprocal ultrametric \eqref{eqn_nonreciprocal_clustering} is $u^{\NR}_X(x, x') = 1$ for all $x \neq x'$ due to the outmost clockwise loop visiting all nodes at cost 1. This is represented in the nonreciprocal $\ccalH^{\NR}$ dendrogram in Fig. \ref{fig_beta_ultrametrics_2}. For the reciprocal ultrametric \eqref{eqn_reciprocal_clustering} nodes $c$ and $d$ merge at resolution $u^\R_X(c, d)=2$, nodes $a$ and $b$ at resolution $u^\R_X(a,b)=3$, and they all join together at resolution $\delta=5$ . This is represented in the reciprocal $\ccalH^{\R}$ dendrogram in Fig. \ref{fig_beta_ultrametrics_2}. To determine $u^{\R/\NR}_X(x,x';4)$ use the piecewise definition in \eqref{def_mu_beta_1}. Since the reciprocal ultrametrics $u^\R_X(c, d)=2\leq4$, and $u^\R_X(a,b)=3\leq4$ are smaller than $\beta=4$ we set the grafted outcomes to the nonreciprocal ultrametric values to obtain $u^{\R/\NR}_X(c, d)=u^{\NR}_X(c, d)=1$, and $u^{\R/\NR}_X(a,b)=u^{\NR}_X(a,b)=1$. Since the remaining ultrametric distances are $u^{\R}_X(x,x') = 5$ which exceed $\beta=4$ we set $u^{\R/\NR}_X(x,x';4) = u^{\R}_X(x,x') = 5$. This yields the $\ccalH^{\R/\NR}$ dendrogram in Fig. \ref{fig_beta_ultrametrics_2} which we interpret as cutting branches from $\ccalH^{\R}$ that we replace by corresponding branches of $\ccalH^{\NR}$.

In the method $\ccalH^{\R/\NR}(\beta)$ we use the reciprocal ultrametric as a decision variable in the piecewise definition \eqref{def_mu_beta_1} and use nonreciprocal ultrametrics for nodes having small reciprocal ultrametrics. There are three other possible grafting combinations $\ccalH^{\R/\R}(\beta)$, $\ccalH^{\NR/\R}(\beta)$ and $\ccalH^{\NR/\NR}(\beta)$ depending on which ultrametric is used as decision variable to swap branches and which of the two ultrametrics is used for nodes having small values of the decision ultrametric. In the method $\ccalH^{\R/\R}(\beta)$, we use reciprocal ultrametrics as decision variables and as the choice for small values of reciprocal ultrametrics,
\begin{equation}\label{def_mu_beta_4}
   u^{\R/\R}_X(x,x';\beta) :=
      \begin{cases}
         u^{\R}_X(x,x'), & \text{if } u^{\R}_X(x,x')\leq \beta, \\
         u^{\NR}_X(x,x'), & \text{if } u^{\R}_X(x,x')>\beta.
      \end{cases}
\end{equation}
In the same manner in which \eqref{def_mu_beta_1} represents cutting the reciprocal dendrogram at a resolution and grafting branches of the nonreciprocal dendrogram for resolutions lower than the cut, the definition in \eqref{def_mu_beta_4} entails cutting the reciprocal dendrogram at a given resolution and grafting branches of the nonreciprocal tree for resolutions higher than the cut. The method $\ccalH^{\R/\R}(\beta)$ as defined in \eqref{def_mu_beta_4} is not valid, however, because for some networks $N=(X,A_X)$ the function $u^{\R/\R}_X(\beta)$ is not an ultrametric as it violates the strong triangle inequality in \eqref{eqn_strong_triangle_inequality}. As a counterexample consider again the network in Fig. \ref{fig_beta_ultrametrics_2}. Applying the definition in \eqref{def_mu_beta_4} we make $u^{\R/\R}_X(a,b;4) = u^{\R}_X(a,b)$ because $u^{\R/\R}_X(a,b;4) \leq 4$ and we make $u^{\R/\R}_X(a,c;4) = u^{\NR}_X(a,c) = 1$ and $u^{\R/\R}_X(c,b;4) = u^{\NR}_X(c,b) = 1$ because both $u^{\R}_X(a,c;4) > 4$ and $u^{\R}_X(c,b;4) > 4$. However, this implies that $u^{\R/\R}_X(a,b;4) > \max(u^{\R/\R}_X(a,c;4), u^{\R/\R}_X(c,b;4))$ violating the strong triangle inequality \eqref{eqn_strong_triangle_inequality} and proving that the definition in \eqref{def_mu_beta_4} is not a valid output of a hierarchical clustering method.

In $\ccalH^{\NR/\NR}(\beta)$ we use nonreciprocal ultrametrics as decision variables and as the choice for small values of nonreciprocal ultrametrics. In $\ccalH^{\NR/\R}(\beta)$ nonreciprocal ultrametrics are used as decision variables and reciprocal ultrametrics are used for small values of nonreciprocal ultrametrics. Both of these methods are invalid as they can be seen to also violate the strong triangle inequality for some networks.

A second valid grafting alternative can be obtained as a modification of $\ccalH^{\R/\R}(\beta)$ in which reciprocal ultrametrics are kept for pairs having small reciprocal ultrametrics, nonreciprocal ultrametrics are used for pairs having large reciprocal ultrametrics, but all nonreciprocal ultrametrics smaller than $\beta$ are saturated to this value. Denoting the method by $\ccalH^{\R/\R_{\max}}(\beta)$ the output ultrametrics are thereby given as
\begin{equation}\label{def_mu_beta_5}
   u^{\R/\R_{\max}}_X\!(x,x';\beta) :=
   \begin{cases}
      u^{\R}_X(x,x'),                        &\!\!\! \text{if } u^{\R}_X(x,x')\leq \beta, \\
      \max \big(\beta, u^{\NR}_X(x,x')\big), &\!\!\! \text{if } u^{\R}_X(x,x')>\beta.
   \end{cases}
\end{equation}
This alternative definition outputs a valid ultrametric and the method $\ccalH^{\R/\R_{\max}}(\beta)$ satisfies axioms (A1)-(A2) as we claim in the following proposition.

\begin{proposition}\label{prop_beta_5}
The hierarchical clustering method $\ccalH^{\R/\R_{\max}}(\beta)$ is valid and admissible. I.e., $u_X^{\R/\R_{\max}}(\beta)$ defined by \eqref{def_mu_beta_5} is an ultrametric for all networks $N=(X, A_X)$ and $\ccalH^{\R/\R_{\max}}(\beta)$ satisfies axioms (A1)-(A2).
\end{proposition}

\begin{myproofnoname} See Appendix \ref{appendix_sec_intermediate_ultrametrics}. \end{myproofnoname}

\begin{remark}\normalfont
Intuitively, the grafting combination $\ccalH^{\R/\NR}(\beta)$ allows nonreciprocal propagation of influence for resolutions smaller than $\beta$ while requiring reciprocal propagation for higher resolutions. This is of interest if we want tight clusters of small dissimilarity to be formed through loops of influence while looser clusters of higher dissimilarity are required to form through links of bidirectional influence. Conversely, the clustering method $\ccalH^{\R/\R_{\max}}(\beta)$ requires reciprocal influence within tight clusters of resolution smaller than $\beta$ but allows nonreciprocal influence in clusters of higher resolutions. This latter behavior is desirable in, e.g., trust propagation in social interactions, where we want tight clusters to be formed through links of mutual trust but allow looser clusters to be formed through unidirectional trust loops.
\end{remark}


\subsection{Convex combinations}\label{sec_convex_comb}

Another completely different family of intermediate admissible methods can be constructed from the result of performing a convex combination of methods known to satisfy axioms (A1) and (A2). Indeed, consider two admissible clustering methods $\ccalH^1$ and $\ccalH^2$ and a given parameter $0 \leq \theta \leq 1$. For an arbitrary network $N=(X, A_X)$ denote by $(X,u^1_X) = \ccalH^1(N)$ and $(X,u^2_X)=\ccalH^2(N)$ the respective outcomes of methods $\ccalH^1$ and $\ccalH^2$. Construct then the dissimilarity function $A_X^{12}(\theta)$ as the convex combination of ultrametrics $u^1_X$ and $u^2_X$: for all $x,x'\in X$
\begin{equation}\label{eqn_def_sym_net_conv_comb}
   A^{12}_X(x,x';\theta) := \theta\, u^1_X(x, x') + (1-\theta) \, u^2_X(x, x').
\end{equation}
Although it can be shown that $A_X^{12}(\theta)$ is a well-defined dissimilarity function, it is not an ultrametric in general because it may violate the strong triangle inequality. Thus, we can recover the ultrametric structure by applying any admissible clustering method $\ccalH$ to the network $N^{12}_\theta=(X,A^{12}_X)$ to obtain $(X, u_X)=\ccalH(N^{12}_\theta)$. Notice however that the network $N^{12}_\theta$ is symmetric because the ultrametrics $u_X^1$ and $u_X^2$ are symmetric by definition. Also recall that, according to Corollary \ref{cor_single_linkage}, single linkage is the unique admissible clustering method for symmetric networks. Thus, we define the convex combination method $\ccalH^{12}_\theta:\ccalN\to\ccalU$ as the one where the output $(X,u^{12}_{X}(\theta))=\ccalH^{12}_\theta(N)$ corresponding to network $N=(X,A_X)$ is given by
\begin{align}\label{eqn_def_u_1_2_bar}
   u^{12}_{X}(x, x';\theta) 
      := \min_{C(x,x')} \, \max_{i | x_i\in C(x,x')} A^{12}_X(x_i,x_{i+1};\theta),
\end{align}
for all $x, x' \in X$ and $A^{12}_X$ as given in \eqref{eqn_def_sym_net_conv_comb}. The operation in \eqref{eqn_def_u_1_2_bar} is equivalent to the definition of single linkage applied to the symmetric network $N^{12}_\theta$. We show that \eqref{eqn_def_u_1_2_bar} defines a valid ultrametric and that $\ccalH^{12}_\theta$ fulfills axioms (A1) and (A2) as stated in the following proposition.

%
\begin{proposition}\label{prop_convex_combination}
Given two admissible hierarchical clustering methods $\ccalH^1$ and $\ccalH^2$, the convex combination method $\ccalH^{12}_\theta$ is valid and admissible. I.e., $u_X^{12}(\theta)$ defined by \eqref{eqn_def_u_1_2_bar} is an ultrametric for all networks $N=(X, A_X)$ and $\ccalH^{12}_\theta$ satisfies axioms (A1)-(A2).
\end{proposition}

%
\begin{myproofnoname} See Appendix \ref{appendix_sec_intermediate_ultrametrics}. \end{myproofnoname}

%

The construction in \eqref{eqn_def_u_1_2_bar} can be generalized to produce intermediate clustering methods generated by convex combinations of any number (i.e. not necessarily two) of admissible methods (such as reciprocal, nonreciprocal, members of the grafting family of Section \ref{sec_grafting}, members of the semi-reciprocal family to be introduced in Section \ref{sec_inter_reciprocal}, etc). These convex combinations can be seen to satisfy axioms (A1) and (A2) through recursive application of Proposition \ref{prop_convex_combination}.

\vspace{-0.1in}
\begin{remark}\normalfont
Since \eqref{eqn_def_u_1_2_bar} is equivalent to single linkage applied to the symmetric network $N^{12}_\theta$, it follows \cite{clust-um,CarlssonMemoli10} that the ultrametric $u_X^{12}(\theta)$ in \eqref{eqn_def_u_1_2_bar} is the largest ultrametric uniformly bounded by $A^{12}_X(\theta)$, i.e., the largest ultrametric for which $u^{12}_X(x,x';\theta)\leq A^{12}_X(x,x';\theta)$ for all pairs $x,x'$. We can then think of \eqref{eqn_def_u_1_2_bar} as an operation ensuring a valid ultrametric definition while deviating as little as possible from $A^{12}_X(\theta)$, thus, retaining as much information as possible in the convex combination of $u^1_X$ and $u^2_X$.
\end{remark}

%
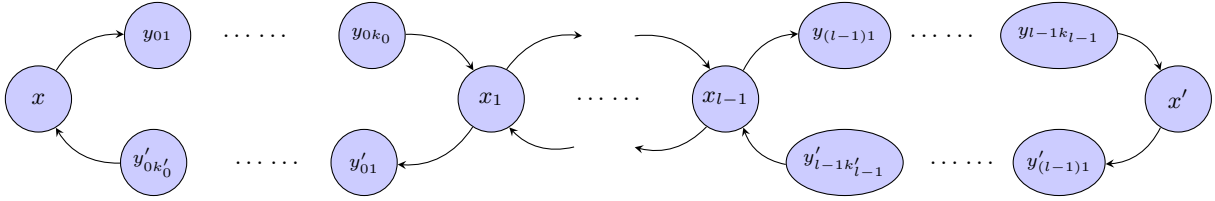
\begin{figure*}
\centering

\def \thisplotscale {2.2}
\def \unit {\thisplotscale cm}
\tikzstyle {blue vertex here} = [blue vertex, 
                                 minimum width = 0.4*\unit, 
                                 minimum height = 0.4*\unit, 
                                 anchor=center]
\tikzstyle {blue vertex here 2} = [blue vertex, 
                                 minimum width = 0.4*\unit, 
                                 minimum height = 0.4*\unit,
                                 font=\scriptsize, 
                                 anchor=center]
{\small \begin{tikzpicture}[thick, x = 1.2*\unit, y = 0.96*\unit]


	\path[draw, thin] (8.0,3.4) ++ ( -4, -0.4) node[blue vertex here] (1) {{$x$}}
	                            ++ ( 0.6, 0.4) node[blue vertex here 2] (2) {{$y_{01}$}}
	                            ++ (0.4,0)   node [phantom vertex] ()   {$\ldots$}
	                            ++ (0.18,0)   node [phantom vertex] ()   {$\ldots$}
	                            ++ (0.5,0) node[blue vertex here 2] (3) {{$y_{0k_0}$}}
	                            ++ (0.6, -0.4) node[blue vertex here] (4) {{$x_1$}}
	                            ++ (0.5, 0.4) node (p1) {}
	                            ++ (0, -0.4) node () {$\ldots$}
	                            ++ (0.18, 0) node () {$\ldots$}

	                            ++ (0, 0.4) node (p2) {}

	                            ++ ( 0.5, -0.4) node[blue vertex here] (5) {{$x_{l-1}$}}
	                            ++ ( 0.6, 0.4) node[blue vertex here 2] (9) {{$y_{(l-1)1}$}}
	                            ++ (0.4,0)   node [phantom vertex] ()   {$\ldots$}
	                            ++ (0.18,0)   node [phantom vertex] ()   {$\ldots$}
	                            ++ (0.5,0) node[blue vertex here 2] (10) {{$y_{l-1k_{l-1}}$}}
	                            ++ (0.6, -0.4) node[blue vertex here] (11) {{$x'$}}
                                     ++ (-0.6, -0.4) node[blue vertex here 2] (12) {{$y'_{(l-1)1}$}}
                                     ++ (-0.4,0)   node [phantom vertex] ()   {$\ldots$}
	                            ++ (-0.18,0)   node [phantom vertex] ()   {$\ldots$}
	                            ++ (-0.5, 0) node[blue vertex here 2] (13) {{$y'_{l-1k'_{l-1}}$}}
	                             ++ (-1.1, 0.1) node (p3) {}
	                             ++ (-0.22, 0) node (p4) {}


	                            ++ ( -1.1, -0.1) node[blue vertex here 2] (7) {{$y'_{01}$}}
	                             ++ (-0.4,0)   node [phantom vertex] ()   {$\ldots$}
	                            ++ (-0.18,0)   node [phantom vertex] ()   {$\ldots$} 
	                            ++ ( -0.48, 0) node[blue vertex here 2] (8) {{$y'_{0k'_0}$}};
    \path[thin, -stealth] (1) edge [bend left, above] node {} (2);		                            
    \path[thin, -stealth] (3) edge [bend left, below] node {} (4);
    \path[thin, -stealth] (4) edge [bend left, below] node {} (p1);	
    \path[thin, -stealth] (p2) edge [bend left, below] node {} (5);
    \path[thin, -stealth] (4) edge [bend left, above] node {} (7);	
    \path[thin, -stealth] (8) edge [bend left, right] node {} (1);	
    \path[thin, -stealth] (5) edge [bend left, right] node {} (9);	
    \path[thin, -stealth] (10) edge [bend left, right] node {} (11);
    \path[thin, -stealth] (5) edge [bend left, right] node {} (p3);	    
       \path[thin, -stealth] (p4) edge [bend left, right] node {} (4);   
       \path[thin, -stealth] (11) edge [bend left, right] node {} (12); 
       \path[thin, -stealth] (13) edge [bend left, right] node {} (5);  
\end{tikzpicture}} 
\caption{Semi-reciprocal chains. The main chain joining $x$ and $x'$ is formed by $[x, x_1, ... , x_{l-1}, x']$. Between two consecutive nodes of the main chain $x_i$ and $x_{i+1}$, we have a secondary chain in each direction $[x_i, y_{i1}, ... , y_{ik_i}, x_{i+1}]$ and $[x_{i+1}, y'_{i1}, ... , y'_{ik'_i}, x_i]$. For $u^{\SR(t)}_X(x,x')$, the maximum allowed node length of secondary chains is $t$, i.e., $k_i, k'_i \leq t-2$ for all $i$.}
\vspace{-0.1in}
\label{fig_inter_reciprocal_example}
\end{figure*}


\subsection{Semi-reciprocal ultrametrics}\label{sec_inter_reciprocal}

In reciprocal clustering we require influence to propagate through bidirectional chains; see Fig. \ref{fig_reciprocal_path}. We could reinterpret bidirectional propagation as allowing loops of node length two in both directions. E.g., the bidirectional chain between $x$ and $x_1$ in Fig. \ref{fig_reciprocal_path} can be interpreted as a loop between $x$ and $x_1$ composed by two chains $[x, x_1]$ and $[x_1, x]$ of node length two. \emph{Semi-reciprocal} clustering is a generalization of this concept where loops consisting of at most $t$ nodes in each direction are allowed. Given $t \in \naturals$ such that $t \geq 2$, we use the notation $C_t(x,x')$ to denote any chain $[x=x_0,x_1,\ldots,x_l=x']$ joining $x$ to $x'$ where $l \leq t-1$. That is, $C_t(x,x')$ is a chain starting at $x$ and finishing at $x'$ with at most $t$ nodes, where $x$ and $x'$ need not be different nodes. Recall that the notation $C(x,x')$ represents a chain linking $x$ with $x'$ where no maximum is imposed on the number of nodes in the chain. Given an arbitrary network $N=(X, A_X)$, define as $A^{\SR(t)}_X(x, x')$ the minimum cost incurred when traveling from node $x$ to node $x'$ using a chain of at most $t$ nodes. I.e.,
\begin{equation}\label{eqn_inter_cost}
A^{\SR(t)}_X(x, x'):=\min_{C_t(x, x')} \,\,\,  \max_{i | x_i\in C_t(x, x')} A_X(x_i, x_{i+1}).
\end{equation}
We define the family of semi-reciprocal clustering methods $\ccalH^{\SR(t)}$ with output $(X,u^{\SR(t)}_X)=\ccalH^{\SR(t)}(X,A_X)$ as the one for which the ultrametric $u^{\SR(t)}_X(x,x')$ between points $x$ and $x'$ is 
\begin{align}\label{eqn_inter_reciprocal_clustering} 
    u^{\SR(t)}_X(x,x') := \min_{C(x,x')} \,\,\, \max_{i | x_i\in C(x,x')} \overline{A^{\SR(t)}_X}(x_i, x_{i+1})
\end{align}     
where the function $\overline{A^{\SR(t)}_X}(x_i, x_{i+1})$ is defined as
\begin{align}\label{eqn_inter_reciprocal_clustering_auxiliary}     
    \overline{A^{\SR(t)}_X}(x_i, x_{i+1}) := \max \big(A^{\SR(t)}_X(x_i, x_{i+1}), A^{\SR(t)}_X(x_{i+1}, x_i)\big).
\end{align} 

The chain $C(x, x')$ of unconstrained length in \eqref{eqn_inter_reciprocal_clustering} is called the \emph{main chain}, represented by $[x=x_0, x_1, ... , x_{l-1},x_{l}= x']$ in Fig. \ref{fig_inter_reciprocal_example}. Between consecutive nodes $x_i$ and $x_{i+1}$ of the main chain, we build loops consisting of secondary chains in each direction, represented in Fig. \ref{fig_inter_reciprocal_example} by $[x_i, y_{i1}, ... , y_{ik_i}, x_{i+1}]$ and $[x_{i+1}, y'_{i1}, ... , y'_{ik'_i}, x_{i}]$ for all $i$. 
For the computation of $u^{\SR(t)}_X(x,x')$, the maximum allowed length of secondary chains is equal to $t$ nodes, i.e., $k_i, k'_i \leq t-2$ for all $i$. In particular, for $t=2$ we recover the reciprocal chain depicted in Fig. \ref{fig_reciprocal_path}.

We can reinterpret \eqref{eqn_inter_reciprocal_clustering} as the application of reciprocal clustering [cf. \eqref{eqn_reciprocal_clustering}] to a network with dissimilarities $A^{\SR(t)}_X$ as in \eqref{eqn_inter_cost}, i.e., a network with dissimilarities given by the optimal choice of secondary chains. Semi-reciprocal clustering methods are valid and satisfy axioms (A1)-(A2) as shown in the following proposition.


\begin{proposition}\label{inter_reciprocal_axioms}
The semi-reciprocal clustering method $\ccalH^{\SR(t)}$ is valid and admissible for all integers $t \geq 2$. I.e., $u_X^{\SR(t)}$ defined by \eqref{eqn_inter_reciprocal_clustering} is an ultrametric for all networks $N=(X, A_X)$ and $\ccalH^{\SR(t)}$ satisfies axioms (A1)-(A2).
\end{proposition}
\begin{myproofnoname}
See Appendix \ref{appendix_sec_intermediate_ultrametrics}.
\end{myproofnoname}

The semi-reciprocal family is a countable family of clustering methods parameterized by integer $t$ representing the allowed maximum node length of secondary chains. Reciprocal and nonreciprocal ultrametrics are equivalent to semi-reciprocal ultrametrics for specific values of $t$. For $t=2$ we have $u^{\SR(2)}_X = u^{\R}_X$ meaning that we recover reciprocal clustering. To see this formally, note that $A^{\SR(2)}_X(x, x')=A_X(x,x')$ [cf. \eqref{eqn_inter_cost}] since the only chain of length two joining $x$ and $x'$ is $[x, x']$. Hence, for $t=2$,  \eqref{eqn_inter_reciprocal_clustering} reduces to
\begin{equation}\label{eqn_inter_reciprocal_clustering_1} 
    u^{\SR(2)}_X(x,x')= \min_{C(x,x')} \,\,\, \max_{i | x_i\in C(x,x')} \bbarA_X(x_i,x_{i+1}),
\end{equation} 
which is the definition of the reciprocal ultrametric [cf. \eqref{eqn_reciprocal_clustering}]. Nonreciprocal ultrametrics can be obtained as $u^{\SR(t)}_X = u^{\NR}_X$ for any parameter $t$ exceeding the number of nodes in the network analyzed. To see this, notice that minimizing over $C(x, x')$ is equivalent to minimizing over $C_t(x, x')$ for all $t \geq n$, since we are looking for minimizing chains in a network with non negative dissimilarities. Therefore, visiting the same node twice is not an optimal choice. This implies that $C_n(x,x')$ contains all possible minimizing chains between $x$ and $x'$. I.e., all chains of interest have at most $n$ nodes. Hence, by inspecting \eqref{eqn_inter_cost}, $A^{\SR(t)}_X(x, x')= \tdu^*_X(x, x')$ [cf. \eqref{eqn_nonreciprocal_chains}] for all $t \geq n$. Furthermore, when $t \geq n$, the best main chain that can be picked is formed only by nodes $x$ and $x'$ because, in this way, no additional meeting point is enforced between the chains going from $x$ to $x'$ and vice versa. As a consequence, definition \eqref{eqn_inter_reciprocal_clustering} reduces to
\begin{equation}\label{eqn_inter_reciprocal_clustering_2} 
    u^{\SR(t)}_X(x,x')= \max \Big(\tdu^*_X(x, x'), \tdu^*_X(x', x) \Big),
\end{equation} 
for all $x, x' \in X$ and for all $t \geq n$. The right hand side of \eqref{eqn_inter_reciprocal_clustering_2} is the definition of the nonreciprocal ultrametric [cf. \eqref{eqn_nonreciprocal_clustering}].

%
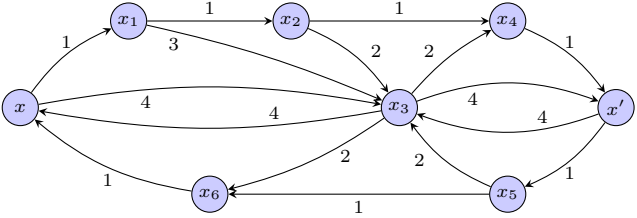
\begin{figure}
\centering
\def \thisplotscale {1.2}
\def \unit {\thisplotscale cm}
\tikzstyle {blue vertex here} = [blue vertex, 
                                 minimum width = 0.4*\unit, 
                                 minimum height = 0.4*\unit, 
                                 anchor=center]
{\small \begin{tikzpicture}[thick, x = 1.2*\unit, y = 0.96*\unit]
         
    \scriptsize

	\path[draw, thin] (8.0,3.4) ++ ( -6, -1) node[blue vertex here] (1) {{$x$}}
	                            ++ ( 1.0, 1) node[blue vertex here] (2) {{$x_1$}}
	                            ++ (1.5,0) node[blue vertex here] (3) {{$x_2$}}
	                            ++ ( 1, -1) node[blue vertex here] (4) {{$x_3$}}
	                            ++ ( 1, 1) node[blue vertex here] (5) {{$x_4$}}
	                            ++ ( 1, -1) node[blue vertex here] (6) {{$x'$}}
	                            ++ ( -1, -1) node[blue vertex here] (7) {{$x_5$}}
	                            ++ ( -2.75, 0) node[blue vertex here] (8) {{$x_6$}};
    \path[thin, -stealth] (1) edge [bend left=15, above] node {$1$} (2);		                            
    \path[thin, -stealth] (2) edge [right, above] node {$1$} (3);
    \path[thin, -stealth] (3) edge [above] node {$1$} (5);	
    \path[thin, -stealth] (4) edge [bend left =10, above left, pos=0.4]  node {$2$} (5);	        
    \path[thin, -stealth] (5) edge [bend left=15, above] node {$1$} (6);		                            
    \path[thin, -stealth] (6) edge [bend left=15, below]  node {$1$} (7);	
    \path[thin, -stealth] (7) edge [below] node {$1$} (8);	
    \path[thin, -stealth] (8) edge [bend left = 15, below] node {$1$} (1);
    \path[thin, -stealth] (2) edge [bend left = 5, below, pos=0.1] node {$3$} (4);
    \path[thin, -stealth] (3) edge [bend left = 15, above right, pos=0.65] node {$2$} (4);
     \path[thin, -stealth] (7) edge [bend left = 15, below left, pos=0.7] node {$2$} (4);	       
     \path[thin, -stealth] (4) edge [bend left = 20, below, pos=0.3] node {$4$} (6);
     \path[thin, -stealth] (6) edge [bend left = 20, above, pos=0.3] node {$4$} (4); 
     \path[thin, -stealth] (1) edge [bend left = 10, below, pos=0.3] node {$4$} (4); 
     \path[thin, -stealth] (4) edge [bend left = 10, above, pos=0.3] node {$4$} (1); 
     \path[thin, -stealth] (4) edge [bend left = 10, below, pos=0.25] node {$2$} (8); 
\end{tikzpicture}} 
\caption{Semi-reciprocal example. Computation of semi-reciprocal ultrametrics between nodes $x$ and $x'$ for different values of parameter $t$. $u_X^{\SR(2)}(x, x')=4$, $u_X^{\SR(3)}(x, x') = 3$, $u_X^{\SR(4)}(x, x') = 2$ and $u_X^{\SR(t)}(x, x')=1$ for all $t \geq 5$; see text for details.}
\vspace{-0.1in}
\label{fig_inter_reciprocal_example_2}
\end{figure}

For the network in Fig. \ref{fig_inter_reciprocal_example_2}, we calculate the semi-reciprocal ultrametrics between $x$ and $x'$ for different values of $t$. The edges which are not delineated are assigned dissimilarity values greater than the ones depicted in the figure. Since the only bidirectional chain between $x$ and $x'$ uses $x_3$ as the intermediate node, we conclude that $u_X^{\R}(x,x')=u_X^{\SR(2)}(x,x')=4$. Furthermore, by constructing a path through the outermost clockwise cycle in the network, we conclude that $u_X^{\NR}(x, x')=1$. Since the longest secondary chain in the minimizing chain for the nonreciprocal case, $[x, x_1, x_2, x_4, x']$, has node length 5, we may conclude that $u_X^{\SR(t)}(x, x')=1$ for all $t \geq 5$. For intermediate values of $t$, if e.g., we fix $t=3$, the minimizing chain is given by the main chain $[x, x_3, x']$ and the secondary chains $[x, x_1, x_3]$, $[x_3, x_4, x']$, $[x', x_5, x_3]$ and $[x_3, x_6, x]$ joining consecutive nodes in the main chain in both directions. The maximum cost among all dissimilarities in this path is $A_X(x_1, x_3)=3$. Hence, $u^{\SR(3)}_X(x, x')=3$. The minimizing chain for $t=4$ is similar to the minimizing one for $t=3$ but replacing the secondary chain $[x, x_1, x_3]$ by $[x, x_1, x_2, x_3]$. In this way, we obtain $u^{\SR(4)}_X(x, x')=2$.

\begin{remark}\normalfont 
Intuitively, when propagating influence through a network, reciprocal clustering requires bidirectional influence whereas nonreciprocal clustering allows arbitrarily large unidirectional cycles. In many applications, such as trust propagation in social networks, it is reasonable to look for an intermediate situation where influence can propagate through cycles but of limited length. Semi-reciprocal ultrametrics represent this intermediate situation where the parameter $t$ represents the maximum length of chains through which influence can propagate in a nonreciprocal manner.
\end{remark}


%
\section{Algorithms}\label{sec_algorithms}

In this section, given a network $N=(X,A_X)$ with $|X|=n$, we interpret $A_X$ as an $n\times n$ matrix of dissimilarities.  Ultrametrics over $X$ will de denoted $u_X$ and will also be regarded as $n\times n$ symmetric matrices. Given a square matrix $A$, its transpose will be denoted by $A^T$.

 By \eqref{eqn_reciprocal_clustering}, reciprocal clustering searches for chains that minimize the maximum dissimilarity in the symmetric matrix \begin{equation}\label{eq_max_sym}\bbarA_X:=\max(A_X, A_X^T),\end{equation}
where the $\max$ is applied element-wise. This is equivalent to finding chains in $\bbarA_X$ that have minimum cost in a $\ell_\infty$ sense. Likewise, nonreciprocal clustering searches for directed chains of minimum $\ell_\infty$-sense cost in $A_X$ to construct the matrix $\tdu^*_X$ [cf. \eqref{eqn_nonreciprocal_chains}] and selects the maximum of the directed costs by performing the operation $u^{\NR}_X = \max(\tdu^*_X,\tdu^{*T}_X)$ [cf. \eqref{eqn_nonreciprocal_clustering}]. These operations can be performed algorithmically using matrix powers in the dioid algebra $\mathfrak{A}:= (\reals^+\cup\{+\infty\},\min,\max)$ \cite{GondranMinoux08}.

In the dioid algebra  $\mathfrak{A}$ the regular sum is replaced by the minimization operator and the regular product by maximization. Using $\oplus$ and $\otimes$ to denote sum and product, respectively, on this dioid algebra we have $a\oplus b := \min(a,b)$ and $a\otimes b := \max(a,b)$ for all $a, b \in \reals^+\cup\{+\infty\}$.  Henceforth, for a natural number $n$, $[1,n]$ will denote the set $\{1,2,\ldots,n\}.$ In the algebra $\mathfrak{A}$, the matrix product $A\otimes B$ of two real valued matrices of compatible sizes is therefore given by the matrix with entries
\begin{equation}\label{def_star}
   \big[A \otimes B\big]_{ij}  
       \ :=\ \bigoplus_{k=1}^n \big(A_{ik} \otimes B_{kj} \big) 
       \ =\ \min_{k\in[1,n]} \, \max \big(A_{ik},B_{kj} \big).
\end{equation}

For integers $k\geq 2$ dioid matrix powers $A_X^{(k)}:=A_X\otimes A_X^{(k-1)}$ with $A_X^{(1)}:=A_X$ of a dissimilarity matrix are related to ultrametric matrices $u_X$. We delve into this relationship in the next section.

\subsection{Dioid powers and ultrametrics}
Notice that the elements of the dioid power $u_X^{(2)}$ of a given ultrametric matrix $u_X$ are given by
\begin{equation}\label{def_diod_algebra_ultrametric}
   \big[u_X^{(2)}\big]_{ij}  
       = \min_{k\in[1,n]} \, \max \big([u_{X}]_{ik},[u_{X}]_{kj} \big).
\end{equation}
Since $u_X$ satisfies the strong triangle inequality we have that $[u_X]_{ij}\leq\max \big([u_{X}]_{ik},[u_{X}]_{kj}\big)$ for all $k\in[1,n]$. And for $k=j$ in particular we further have that $\max \big([u_X]_{ik},[u_X]_{kj})=\max \big([u_X]_{ij},[u_X]_{jj})=\max\big([u_X]_{ij},0)=[u_X]_{ij}$. Combining these two observations it follows that the result of the minimization in \eqref{def_diod_algebra_ultrametric} is $\big[u_X^{(2)}\big]_{ij} =  \big[u_X\big]_{i,j}$ since none of its arguments is smaller that $[u_{X}]_{ij}$ and one of them is exactly $[u_{X}]_{ij}$. This being valid for all $i,j$ implies 
\begin{equation}\label{def_diod_algebra_ultrametric_result}
   u_X^{(2)} =  u_X
\end{equation}
Furthermore, a matrix having the property in \eqref{def_diod_algebra_ultrametric_result} is such that $\big[u_X\big]_{ij} = \big[u_X^{(2)}\big]_{ij} = \min_{k\in[1,n]} \, \max \big([u_{X}]_{ik},[u_{X}]_{kj} \big) \leq \max\big([u_{X}]_{ik},[u_{X}]_{kj} \big)$, which is just a restatement of the strong triangle inequality. Therefore, a nonnegative matrix $u_X$ represents a finite ultrametric if and only if \eqref{def_diod_algebra_ultrametric_result} is true, has null diagonal elements $\big[u_X\big]_{ii}=0$ and positive off-diagonal elements, and is symmetric, $u_X=u_X^T$. We then expect dioid powers and max-symmetrization operations \eqref{eq_max_sym} to play a role in the construction of ultrametrics.

This is indeed the case. From the definition in \eqref{def_star} it follows that for a given dissimilarity matrix $A_X$ the $i,j$ entry $[A_X^{(2)}]_{ij}$ of the dioid power $A_X^{(2)}$ represents the minimum $\ell_\infty$-sense cost of a chain linking $i$ to $j$ in at most $2$ hops. Proceeding recursively we can show that the $l$th dioid power $A_X^{(l)}$ is such that its $i,j$ entry $[A_X^{(l)}]_{ij}$ represents the minimum $\ell_\infty$-sense cost of a chain containing at most $l$ hops. 

The \emph{quasi-inverse} of a matrix in a dioid algebra is a useful concept that simplifies the proofs within this section. In any dioid algebra, we call quasi-inverse of $A$, denoted $A^*$, the limit, when it exists, of the sequence of matrices \cite[Ch.4, Def. 3.1.2]{GondranMinoux08}
\begin{equation}\label{eqn_def_quasi_inverse}
A^* := \lim_{k\to \infty } I \oplus A \oplus A^{(2)} \oplus ... \oplus A^{(k)},
\end{equation}
where $I$ has zeros in the diagonal and $+ \infty$ in the off diagonal elements. The utility of the quasi-inverse resides in the fact that, given a dissimilarity matrix $A_X$, then \cite[Ch.6, Sec 6.1]{GondranMinoux08}
\begin{equation}\label{eqn_quasi_inverse_nonrecip}
[A_X^*]_{ij} = \min_{C(x_i,x_j)} \,\,\, \max_{k | x_k \in C(x_i,x_j)} \,\, A_X(x_k,x_{k+1}),
\end{equation}
where $A_X^*$ is the quasi-inverse of $A_X$ in the dioid $\mathfrak{A}$ as defined in \eqref{eqn_def_quasi_inverse}. I.e., the elements of the quasi-inverse $A_X^*$ correspond to the directed minimum chain costs of the associated network $(X, A_X)$ as defined in \eqref{eqn_nonreciprocal_chains}.

\subsection{Algorithms for reciprocal, nonreciprocal, and semi-reciprocal clustering}
Since, as already discussed in Section \ref{sec_inter_reciprocal}, given $N=(X,A_X)\in\ccalN$, we can restrict candidate minimizing chains to those with at most $|X|-1$ hops, the following result follows.

%
\begin{theorem}\label{theo_algo_recip_nonrecip}
For given network $N=(X, A_X)$ with $n$ nodes the reciprocal ultrametric $u^{\R}_X$ defined in \eqref{eqn_reciprocal_clustering} can be computed as
\begin{align}
   u^{\R}_X = \Big(\max\left( {A}_X, A_X^T \right)\Big)^{(n-1)}, \label{eqn_algo_recip} 
\end{align}
where the operation $(\cdot)^{(n-1)}$ denotes the $(n-1)$st matrix power in the dioid algebra $\mathfrak{A}$ with matrix product as defined in \eqref{def_star}. The nonreciprocal ultrametric $u^{\NR}_X$ defined in \eqref{eqn_nonreciprocal_clustering} can be computed as
\begin{align}
   u^{\NR}_X=\max\left({A}_X^{(n-1)},\left(A_X^T\right)^{(n-1)}\right). \label{eqn_algo_nonrecip}
\end{align}\end{theorem}

%
\begin{myproofnoname} 
By comparing \eqref{eqn_quasi_inverse_nonrecip} with \eqref{eqn_nonreciprocal_chains}, we can see that 
\begin{equation}\label{eqn_quasi_inverse_tdu} 
A_X^*=\tdu^*_X.
\end{equation}
It is just a matter of notation when comparing \eqref{eqn_quasi_inverse_tdu} with \eqref{eqn_nonreciprocal_clustering} to realize that
\begin{align}
u^{\NR}_X = \max \big( A_X^*, (A_X^*)^T \big) \label{eqn_algo_nonrecip_2}.
\end{align}
Similarly, if we consider the quasi inverse of the symmetrized matrix $\bbarA_X:=\max(A_X, A_X^T)$, expression \eqref{eqn_quasi_inverse_nonrecip} becomes
\begin{equation}\label{eqn_quasi_inverse_recip}
[\bar{A}_X^*]_{ij} = \min_{C(x_i,x_j)} \,\,\, \max_{k | x_k \in C(x_i,x_j)} \,\, \bar{A}_X(x_k,x_{k+1}).
\end{equation}
From comparing \eqref{eqn_quasi_inverse_recip} and \eqref{eqn_reciprocal_clustering} it is immediate that
\begin{equation}\label{eqn_algo_recip_2}
u^{\R}_X= \bar{A}_X^* = \big(\max(A_X, A_X^T)\big)^*.
\end{equation}
If we show that $A^*_X=A^{(n-1)}_X$, then \eqref{eqn_algo_recip_2} and \eqref{eqn_algo_nonrecip_2} imply equations \eqref{eqn_algo_recip} and \eqref{eqn_algo_nonrecip} respectively, completing the proof. 

Notice in particular that when $\mathfrak{A}= (\reals^+\cup\{+\infty\},\min,\max)$, the $\min$ or $\oplus$ operation is idempotent, i.e. $a \oplus a = a$ for all $a$. In this case, it can be shown that \cite[Ch.4, Prop. 3.1.1]{GondranMinoux08}
\begin{equation}\label{eqn_quasi_inverse_develop_0}
I \oplus A_X \oplus A_X^{(2)} \oplus ... \oplus A_X^{(k)}= (I \oplus A_X)^{(k)},
\end{equation}
for all $k \geq 1$. Moreover, since diagonal elements are null in both matrices in the right hand side of \eqref{eqn_quasi_inverse_develop_0} and the off diagonal elements in $I$ are $+\infty$, it is immediate that $I \oplus A_X = A_X$. Consequently, \eqref{eqn_quasi_inverse_develop_0} becomes
\begin{equation}\label{eqn_quasi_inverse_develop}
I \oplus A_X \oplus A_X^{(2)} \oplus ... \oplus A_X^{(k)}= A_X^{(k)}.
\end{equation}
Taking the limit to infinity in both sides of equality \eqref{eqn_quasi_inverse_develop} and invoking the definition of the quasi-inverse \eqref{eqn_def_quasi_inverse}, we obtain
\begin{equation}\label{eqn_quasi_inverse_develop_2}
A_X^*= \lim_{k \to \infty} A_X^{(k)}.
\end{equation}
Finally, it can be shown \cite[Ch. 4, Sec. 3.3, Theo. 1]{GondranMinoux08} that $A_X^{(n-1)}=A_X^{(n)}$, proving that the limit in \eqref{eqn_quasi_inverse_develop_2} exists and, more importantly, that $A_X^*=A_X^{(n-1)}$, as desired.
\end{myproofnoname}

%
For the reciprocal ultrametric we symmetrize dissimilarities with a maximization operation and take the $(n-1)$st power of the resulting matrix on the dioid algebra $\mathfrak{A}$. For the nonreciprocal ultrametric we revert the order of these two operations. We
first consider matrix powers ${A}_X^{(n-1)}$ and $\left(A_X^T\right)^{(n-1)}$ of the dissimilarity matrix and its transpose which we then symmetrize with a maximization operator. Besides emphasizing the relationship between reciprocal and nonreciprocal clustering, Theorem \ref{theo_algo_recip_nonrecip} suggests the existence of intermediate methods in which we raise dissimilarity matrices $A_X$ and $A_X^T$ to some power, perform a symmetrization, and then continue matrix multiplications. These procedures yield methods that are not only valid but coincide with the family of semi-reciprocal ultrametrics introduced in Section \ref{sec_inter_reciprocal} as the following proposition asserts.

%
\begin{proposition}\label{prop_general_algo} For a given network $N=(X, A_X)$ with $n$ nodes, the $t$-th semi-reciprocal ultrametric $u_X^{\SR(t)}$ in \eqref{eqn_inter_reciprocal_clustering} for every natural $t \geq 2$ can be computed as
\begin{equation}\label{eqn_algo_semi_reciprocal_2}
   u_X^{\SR(t)} = \left(\max\left({A}_X^{(t-1)},\left(A_X^T\right)^{(t-1)}\right)\right)^{(n-1)},
\end{equation}
where $(\cdot)^{(t-1)}$ and $(\cdot)^{(n-1)}$ denote matrix powers in the dioid algebra $\mathfrak{A}$ with matrix product as defined in \eqref{def_star}.\end{proposition}

%
\begin{myproofnoname} See Appendix \ref{appendix_sec_algorithms}. \end{myproofnoname}

%
The result in \eqref{eqn_algo_semi_reciprocal_2} is intuitively clear. The powers ${A}_X^{(t-1)}$ and $\left(A_X^T\right)^{(t-1)}$ represent the minimum $\ell_\infty$-sense cost among directed chains of at most $t-1$ links. In the terminology of Section \ref{sec_inter_reciprocal} these are the costs of optimal secondary chains containing at most $t$ nodes. Therefore, the maximization $\max\big({A}_X^{(t-1)},\left(A_X^T\right)^{(t-1)}\big)$ computes the cost of joining two  nodes with secondary chains of at most $t$ nodes in each direction. This is the definition of $\overline{A^{\SR(t)}_X}$ in \eqref{eqn_inter_reciprocal_clustering}. Applying the dioid power $(n-1)$ to this new matrix is equivalent to looking for minimizing chains in the network with costs given by the secondary chains. Thus, the outermost dioid power computes the costs of the optimal main chains that achieve the ultrametric values in \eqref{eqn_inter_reciprocal_clustering}.

Observe that we recover \eqref{eqn_algo_recip} by making $t=2$ in \eqref{eqn_algo_semi_reciprocal_2} and that we recover \eqref{eqn_algo_nonrecip} when $t=n$. For this latter case note that when $t=n$ in \eqref{eqn_algo_semi_reciprocal_2}, comparison with \eqref{eqn_algo_nonrecip} shows that $\max({A}_X^{(t-1)},(A_X^T)^{(t-1)})=\max({A}_X^{(n-1)},(A_X^T)^{(n-1)})=u^{\NR}_X$. However, since $u^{\NR}_X$ is an ultrametric it is idempotent in the dioid algebra [cf. \eqref{def_diod_algebra_ultrametric_result}] and the outermost dioid power in \eqref{eqn_algo_semi_reciprocal_2} is moot. This recovery is consistent with the observations in \eqref{eqn_inter_reciprocal_clustering_1} and \eqref{eqn_inter_reciprocal_clustering_2} that reciprocal and nonreciprocal clustering are particular cases of semi-reciprocal clustering $\ccalH^{\SR(t)}$ in that for $t=2$ we have $u^{\SR(2)}_X(x,x')=u^{\R}_X(x,x')$ and for $t\geq n$ it holds that $u^{\SR(t)}_X(x,x')=u^{\NR}_X(x,x')$ for arbitrary points $x,x'$ of arbitrary network $N=(X,A_X)$. The results in Theorem \ref{theo_algo_recip_nonrecip} and Proposition \ref{prop_general_algo} emphasize the extremal nature of the reciprocal and nonreciprocal methods and characterize the semi-reciprocal ultrametrics as natural intermediate clustering methods in an algorithmic sense. 

\subsection{Algorithmic intermediate clustering methods}
This algorithmic perspective allows for a generalization in which the powers of the matrices $A_X$ and $A_X^T$ are different. To be precise consider positive integers $t,t'>0$ and define the algorithmic intermediate clustering method $\ccalH^{t,t'}$ with parameters $t,t'$ as the one that maps the given network $N=(X, A_X)$ to the ultrametric space $(X,u^{t,t'}_X) = \ccalH^{t,t'}(N)$ given by
\begin{equation}\label{eqn_algo_algorithmic_intermediate}
   u^{t,t'}_X := \left(\max\left({A}_X^{(t)},\left(A_X^T\right)^{(t')}\right)\right)^{(n-1)}. 
\end{equation}
The ultrametric \eqref{eqn_algo_algorithmic_intermediate} can be interpreted as a semi-reciprocal ultrametric where the allowed length of secondary chains varies with the direction. Forward secondary chains may have at most $t+1$ nodes whereas backward secondary chains may have at most $t'+1$ nodes. The algorithmic intermediate family $\ccalH^{t,t'}$ encapsulates the semi-reciprocal family since $\ccalH^{t,t}\equiv\ccalH^{\SR(t+1)}$ as well as the reciprocal method since $\ccalH^{\R}\equiv\ccalH^{1,1}$ as it follows from comparison of \eqref{eqn_algo_algorithmic_intermediate} with \eqref{eqn_algo_semi_reciprocal_2} and \eqref{eqn_algo_recip}, respectively. We also have that $\ccalH^{\NR}(N) = \ccalH^{n-1, n-1}(N)$ for all networks $N=(X, A_X)$ such that $|X| \leq n$. This follows from the comparison of \eqref{eqn_algo_algorithmic_intermediate} with \eqref{eqn_algo_nonrecip} and the idempotency of $u^{\NR}_X=\max({A}_X^{(n-1)},(A_X^T)^{(n-1)})$ with respect to the dioid algebra. The intermediate algorithmic methods $\ccalH^{t,t'}$ are admissible as we claim in the following proposition.

%
\begin{proposition}\label{prop_algorithmic_intermediate_ultrametric} The hierarchical clustering method $\ccalH^{t,t'}$ is valid and admissible. I.e., $u_X^{t,t'}$ defined by \eqref{eqn_algo_algorithmic_intermediate} is an ultrametric for all networks $N=(X, A_X)$ and $\ccalH^{t,t'}$ satisfies axioms (A1)-(A2).
\end{proposition}

%
\begin{myproofnoname} See Appendix \ref{appendix_sec_algorithms}. \end{myproofnoname}

%
\subsection{Algorithms for the grafting and convex combination families of methods}
Algorithms to compute ultrametrics associated with the grafting families in Section \ref{sec_grafting} entail simple combinations of matrices $u^{\R}_X$ and $u^{\NR}_X$. E.g., the ultrametrics in \eqref{def_mu_beta_1} corresponding to the grafting method $\ccalH^{\R/\NR}(\beta)$ can be computed as
\begin{equation}\label{eqn_algo_algorithmic_grafting}
   u^{\R/\NR}_X(\beta) =   u_X^{\NR}\circ\ind{u_X^{\R}\leq\beta}  
                       + u_X^{\R} \circ\ind{u_X^{\R}>\beta},
\end{equation}
where $A\circ B$ denotes the Hadamard product of matrices $A$ and $B$ and $\ind{\cdot}$ is an element-wise indicator function which outputs a matrix with a 1 in the positions of the elements that satisfy the condition to which its applied and a 0 otherwise.

In symmetric networks Corollary \ref{cor_single_linkage} states that any admissible algorithm must output an ultrametric equal to the single linkage ultrametric $u^{\SL}_X$ as defined in \eqref{eqn_single_linkage_ultrametric}. Thus, all algorithms in this section yield the same output when restricted to symmetric matrices $A_X$ and this output is $u^{\SL}_X$. Considering, e.g., the algorithm for the reciprocal ultrametric in \eqref{eqn_algo_recip} and noting that for a symmetric network $A_X=\max(A_X, A^T_X)$ we conclude that single linkage can be computed as
\begin{equation}\label{eqn_algorithm_single_linkage}
u^{\SL}_X = A_X^{(n-1)}.
\end{equation}
Algorithms for the convex combination family in Section \ref{sec_convex_comb} involve computing dioid algebra powers of a convex combination of ultrametric matrices. Given two admissible methods $\ccalH^1$ and $\ccalH^2$ with outputs $(X, u^1_X)=\ccalH^1(N)$ and $(X, u^2_X)=\ccalH^2(N)$, and  $\theta\in[0,1]$, the ultrametric in \eqref{eqn_def_u_1_2_bar} corresponding to the method $\ccalH^{12}_\theta$ can be computed as
\begin{equation}\label{eqn_algo_convex_comb}
u^{12}_X(\theta)= \Big(\theta\, u^1_X + (1-\theta) \, u^2_X\Big)^{(n-1)}.
\end{equation}
The operation $\theta\, u^1_X + (1-\theta) \, u^2_X$ is just the regular convex combination in 
\eqref{eqn_def_sym_net_conv_comb} and the dioid power in \eqref{eqn_algo_convex_comb} implements the single linkage operation in \eqref{eqn_def_u_1_2_bar} as it follows from \eqref{eqn_algorithm_single_linkage}.

\begin{remark}\normalfont 
It follows from \eqref{eqn_algo_recip}, \eqref{eqn_algo_nonrecip}, \eqref{eqn_algo_semi_reciprocal_2}, \eqref{eqn_algo_algorithmic_intermediate}, and \eqref{eqn_algorithm_single_linkage} that all methods presented in this paper can be computed in a number of operations of order $\text{O}(n^4)$ which coincides with the time it takes to compute $n$ matrix products of matrices of size $n\times n$. This complexity can be reduced to $\text{O}(n^3\log n)$ by noting that the dioid matrix power $A^n$ can be computed with the sequence $A, A^2, A^4, \ldots$ which requires $\text{O}(\log n)$ matrix products at a cost of $\text{O}(n^3)$ each. Complexity can be further reduced using the sub cubic dioid matrix multiplication algorithms in \cite{VassilevskaEtal09, DuanPettie09} that have complexity $\text{O}(n^{2.688})$ for a total complexity of $\text{O}(n^{2.688}\log n)$ to compute the $n$th matrix power. There are also related methods with even lower complexity. For the case of reciprocal clustering, complexity of order $\text{O}(n^2)$ can be achieved by leveraging an equivalence between single linkage and a minimum spanning tree problem \cite{Hu61,daniel}. For the case of nonreciprocal clustering, Tarjan's method \cite{tarjan-improved} can be implemented to reduce complexity to $\text{O}(n^2\log n)$. 
\end{remark}


%
\section{Quasi-Clustering methods}\label{sec_full_characterization_asymmetric}

A partition $P=\{B_1,\ldots, B_J\}$ of a set $X$ represents a clustering of $X$ into groups of nodes $B_1, \ldots, B_J \in P$ such that nodes within each group can influence each other more than they can influence or be influenced by the nodes in other groups. A partition can be interpreted as a reduction in data complexity in which variations between elements of a group are neglected in favor of the larger dissimilarities between elements of different groups. This is natural when clustering datasets endowed with symmetric dissimilarities because the concepts of a node $x\in X$ being similar to another node $x'\in X$ and $x'$ being similar to $x$ are equivalent. In an asymmetric network these concepts are different and this difference motivates the definition of structures more general than partitions.

Recalling that a partition $P=\{B_1,\ldots, B_J\}$ of $X$ induces and is induced by an equivalence relation $\sim_P$ on $X$ we search for the analogous of an asymmetric partition by removing the symmetry property in the definition of an equivalence relation. Thus, we define a \emph{quasi-equivalence} $\leadsto$ as a binary relation that satisfies the reflexivity and transitivity properties but is not necessarily symmetric as stated next. 

\begin{definition}\label{def_quasi_equivalence}
A binary relation $\leadsto$ between elements of a set $X$ is a quasi-equivalence if and only if the following properties hold true for all $x, x', x'' \in X$:
\begin{mylist}
\item[{\it (i) Reflexivity.}] Points are quasi-equivalent to themselves, $x \leadsto x$.
\item[{\it (ii) Transitivity.}] If $x\leadsto x'$ and $x'\leadsto x''$ then $x\leadsto x''$.
\end{mylist} \end{definition}

Quasi-equivalence relations are more often termed preorders or quasi-orders in the literature \cite{Harzheim05}. We choose the term quasi-equivalence to emphasize that they are a modified version of an equivalence relation.

We further define a \emph{quasi-partition} of the set $X$ as a directed unweighted graph as stated next.

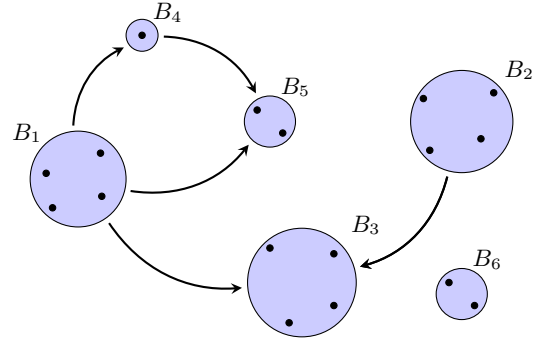
\begin{figure}
\centering
\def \thisplotscale {0.85}
\def \unit {\thisplotscale cm}

{\footnotesize
\begin{tikzpicture}[-stealth, shorten >=2,  shorten <=2, x = \unit, y=0.9*\unit]

    \node [blue vertex, minimum size = 1.5*\unit] at (0,0.1) (b1) {};
    \path (b1) ++ (-0.8,0.8) node {\small $B_1$};
    \node [point, minimum height = 0.1*\unit, minimum width  = 0.1*\unit, fill=black] at ( 0.37,-0.20) (1) {}; 
    \node [point, minimum height = 0.1*\unit, minimum width  = 0.1*\unit, fill=black] at ( 0.35, 0.55) (2) {};
    \node [point, minimum height = 0.1*\unit, minimum width  = 0.1*\unit, fill=black] at (-0.5, 0.2) (3) {}; 
    \node [point, minimum height = 0.1*\unit, minimum width  = 0.1*\unit, fill=black] at (-0.4,-0.4) (4) {}; 

    \path (b1) ++ (6.0,1.0) node [blue vertex, minimum size = 1.6*\unit] (b2) {}; 
    \path (b2) ++ (0.9,0.9) node {\small $B_2$};    
    \path (b2) ++ (-0.6,0.4) node [point, minimum height = 0.1*\unit, minimum width  = 0.1*\unit, fill=black] (1p) {};  \path (b2) ++ (0.3,-0.3) node [point, minimum height = 0.1*\unit, minimum width  = 0.1*\unit, fill=black] (2p) {};
    \path (b2) ++ (0.5,0.5) node [point, minimum height = 0.1*\unit, minimum width  = 0.1*\unit, fill=black] (3p) {};  \path (b2) ++ (-0.5,-0.5) node [point, minimum height = 0.1*\unit, minimum width  = 0.1*\unit, fill=black] (4p) {};
                            (4p) edge (1p) (2p) edge (1p) (4p) edge (3p) ;
     
    \path (b1) ++ (3.5,-1.8) node [blue vertex, minimum size = 1.7*\unit] (b3) {}; 
    \path (b3) ++ (1,1) node {\small $B_3$};    
    \path (b3) ++ (-0.5,0.6) node [point, minimum height = 0.1*\unit, minimum width  = 0.1*\unit, fill=black] (1pp) {};  \path (b3) ++ (0.5,-0.4) node [point, minimum height = 0.1*\unit, minimum width  = 0.1*\unit, fill=black] (2pp) {};
    \path (b3) ++ (0.5,0.5) node [point, minimum height = 0.1*\unit, minimum width  = 0.1*\unit, fill=black] (3pp) {};  \path (b3) ++ (-0.2,-0.7) node [point, minimum height = 0.1*\unit, minimum width  = 0.1*\unit, fill=black] (4pp) {};
   
    \path (b1) ++ (1.0,2.5) node [blue vertex, minimum size = 0.5*\unit] (b4) {}; 
    \path (b4) ++ (+0.4,0.4) node {\small $B_4$};    
    \path (b4) ++ (0,0) node [point, minimum height = 0.1*\unit, minimum width  = 0.1*\unit, fill=black] (1ppp) {}; 

    \path (b1) ++ (3,1) node [blue vertex, minimum size = 0.8*\unit] (b5) {}; 
    \path (b5) ++ (+0.4,+0.6) node {\small $B_5$};    
    \path (b5) ++ (-0.2,0.2) node [point, minimum height = 0.1*\unit, minimum width  = 0.1*\unit, fill=black] (1pppp) {};  
    \path (b5) ++ (0.2,-0.2) node [point, minimum height = 0.1*\unit, minimum width  = 0.1*\unit, fill=black] (2pppp) {};

    \path (b1) ++ (6,-2) node [blue vertex, minimum size = 0.8*\unit] (b6) {}; 
    \path (b6) ++ (+0.4,+0.6) node {\small $B_6$};    
    \path (b6) ++ (-0.2,0.2) node [point, minimum height = 0.1*\unit, minimum width  = 0.1*\unit, fill=black] (1pppp) {};  
    \path (b6) ++ (0.2,-0.2) node [point, minimum height = 0.1*\unit, minimum width  = 0.1*\unit, fill=black] (2pppp) {};

    \path (b1) edge [thick, bend right, above]  node {} (b3);
    \path (b1) edge [thick, bend left, above]  node {} (b4);    
    \path (b1) edge [thick, bend right, above] node {} (b5);    
    \path (b2) edge [thick, bend left, pos=0.52, right] node {} (b3);    
    \path (b2) edge [thick, bend left, pos=0.52, right] node {} (b3);            
    \path (b4) edge [thick, bend left, pos=0.52, right] node {} (b5);   
\end{tikzpicture}}
\caption{A quasi-partition $\tdP=(P,E)$ on a set of nodes. The vertex set $P$ of the quasi-partition is given by a partition of the nodes $P=\{B_1, B_2, \ldots, B_6\}$. Nodes within the same block of the partition $P$ can influence each other. The edges of the directed graph $\tdP=(P,E)$ represent unidirectional influence between the blocks of the partition. In this case, block $B_1$ can influence $B_3$, $B_4$ and $B_5$ while block $B_2$ and $B_4$ can only influence $B_3$ and $B_5$, respectively.}
\vspace{-0.1in}
\label{fig_quasi_partition_example}
\end{figure}

\begin{definition}\label{def_quasi_partition}
A quasi-partition of a given set $X$ is a directed unweighted graph $\tdP=(P,E)$ where the vertex set $P$ is a partition $P=\{B_1,\ldots, B_J\}$ of the space $X$ and the edge set $E \subset P \times P$ is such that it contains no self-loops and the following properties are satisfied (see Fig. \ref{fig_quasi_partition_example}):

\myindentedparagraph{(QP1) Unidirectionality} For any given pair of distinct blocks $B_i$ and $B_j \in P$ we have, at most, one edge between them. Thus, if for some $i \neq j$ we have $(B_i,B_j)\in E$ then forcibly $(B_j,B_i)\notin E$.

\myindentedparagraph{(QP2) Transitivity} If there are edges between blocks $B_i$ and $B_j$ and between blocks $B_j$ and $B_k$, then there is an edge between blocks $B_i$ and $B_k$. 

\end{definition}

%
\medskip\noindent The vertex set $P$ of a quasi-partition $\tdP=(P,E)$ is meant to capture sets of nodes that can influence each other, whereas the edges in $E$ intend to capture the notion of directed influence from one group to the next. In the example in Fig. \ref{fig_quasi_partition_example}, nodes which are drawn close to each other have low dissimilarities between them in both directions. Thus, the nodes inside each block $B_i$ are close to each other but dissimilarities between nodes of different blocks are large in at least one direction. E.g., the dissimilarity from $B_1$ to $B_4$ is small but the dissimilarity from $B_4$ to $B_1$ is large. This latter fact motivates keeping $B_1$ and $B_4$ as separate blocks in the partition whereas the former motivates addition of the directed influence edge $(B_1,B_4)$. Likewise, dissimilarities from $B_1$ to $B_3$, from $B_2$ to $B_3$ and from $B_4$ to $B_5$ are small whereas those on opposite directions are not. Dissimilarities from the nodes in $B_1$ to the nodes in $B_5$ need not be small, but $B_1$ can influence $B_5$ through $B_4$, hence the edge from $B_1$ to $B_5$, in accordance with (QP2). All other dissimilarities are large justifying the lack of connections between the other blocks. Further observe that there are no bidirectional edges as required by (QP1).

Requirements (QP1) and (QP2) in the definition of quasi-partition represent the relational structure that emerges from quasi-equivalence relations as we state in the following proposition.

\begin{proposition}\label{prop_quasi_equiv_quasi_part}
Given a node set $X$ and a quasi-equivalence relation $\leadsto$ on $X$ [cf. Definition \ref{def_quasi_equivalence}] define the relation $\leftrightarrow$ on $X$ as
\begin{equation}\label{eqn_quasi_equiv_equiv}
x \leftrightarrow x' \quad \iff \quad x \leadsto x' \,\,\, \text{\normalfont and} \,\,\, x' \leadsto x,
\end{equation}
for all $x, x' \in X$. Then, $\leftrightarrow$ is an equivalence relation. Let $P = \{B_1, \ldots , B_J\}$ be the partition of $X$ induced by $\leftrightarrow$. Define $E \subseteq P \times P$ such that for all distinct $B_i, B_j \in P$
\begin{equation}\label{eqn_quasi_equiv_edges_quasi_partition}
(B_i, B_j) \in E \quad \iff \quad x_i \leadsto x_j,
\end{equation}
for some $x_i \in B_i$ and $x_j \in B_j$. Then, $\tdP=(P,E)$ is a quasi-partition of $X$. Conversely, given a quasi-partition $\tdP=(P,E)$ of $X$, define the binary relation $\leadsto$ on $X$ so that for all $x, x' \in X$
\begin{equation}
x \leadsto x' \iff [x] = [x'] \,\,\, \text{or} \,\,\, ([x], [x']) \in E,
\end{equation}
where $[x] \in P$ is the block of the partition $P$ that contains the node $x$ and similarly for $[x']$. Then, $\leadsto$ is a quasi-equivalence on $X$.
\end{proposition}
\begin{myproofnoname}
See Theorem 4.9, Ch. 1.4 in \cite{Harzheim05}.
\end{myproofnoname}

In the same way that an equivalence relation induces and is induced by a partition on a given node set $X$, Proposition \ref{prop_quasi_equiv_quasi_part} shows that a quasi-equivalence relation induces and is induced by a quasi-partition on $X$. We can then adopt the construction of quasi-partitions as the natural generalization of clustering problems when given asymmetric data. Further, observe that if the edge set $E$ contains no edges, then $\tdP=(P,E)$ is such that $P$ is a standard partition of $X$. In this sense, partitions can be regarded as particular cases of quasi-partitions having the generic form $\tdP=(P,\emptyset)$. To allow generalizations of hierarchical clustering methods with asymmetric outputs we introduce the notion of \emph{quasi-dendrogram} in the following section.

\subsection{Quasi-dendrograms}\label{sec_quasi_dendrograms}

Recalling that a dendrogram is defined as a nested set of partitions, we define a \emph{quasi-dendrogram} $\tilde{D}_X$ of the set $X$ as a collection of nested quasi-partitions $\tilde{D}_X(\delta)=(D_X(\delta), E_X(\delta))$ indexed by a resolution parameter $\delta \geq 0$. Recall the definition of $[x]_\delta$ from Section \ref{sec_preliminaries}. Formally, for $\tilde{D}_X$ to be a quasi-dendrogram we require the following conditions:

\begin{indentedparagraph}{(\~D1) Boundary conditions} At resolution $\delta=0$ all nodes are in separate clusters with no influences between them and for some $\delta_0$ sufficiently large all elements of $X$ are in a single cluster,
\begin{align}\label{eqn_quasi_dendrogram_boundary_conditions}
   & \tdD_X(0)  = \Big ( \big\{ \{x\}, \, x\in X\big\},\ \emptyset \Big), \nonumber\\
   & \tdD_X(\delta_0) = \Big ( \{ X\}, \emptyset \Big) \quad \forsome\ \delta_0 > 0.
\end{align}\end{indentedparagraph}

\begin{indentedparagraph}{(\~D2) Equivalence hierarchy} For any pair of points $x,x'$ for which $x\sim_{D_X(\delta_1)} x'$ at resolution $\delta_1$ we must have $x\sim_{D_X(\delta_2)} x'$ for all resolutions $\delta_2 \geq \delta_1$. \end{indentedparagraph} 

\begin{indentedparagraph}{(\~D3) Influence hierarchy} If there is an influence edge $([x]_{\delta_1}, [x']_{\delta_1}) \in E_X(\delta_1)$ between the equivalence classes $[x]_{\delta_1}$ and $[x']_{\delta_1}$ of nodes $x$ and $x'$ at resolution $\delta_1$, at any resolution $\delta_2\geq \delta_1$ we either have $([x]_{\delta_2}, [x']_{\delta_2}) \in E_X(\delta_2)$ or $[x]_{\delta_2}=[x']_{\delta_2}$. \end{indentedparagraph}

\begin{indentedparagraph}{(\~D4) Right continuity} For all $\delta \geq 0$ there exists $\epsilon > 0$ such that $\tdD_X(\delta)=\tdD_X(\delta')$ for all $\delta'\in [\delta, \delta+\epsilon]$. \end{indentedparagraph}

\noindent Requirements (\~D1), (\~D2), and (\~D4) are counterparts to requirements (D1), (D2), and (D3) in the definition of dendrograms. The minor variation in (\~D1) is to specify that the edge sets at the extreme values of $\delta$ are empty. For $\delta=0$ this is because there are no influences at that resolution and for $\delta=\delta_0$ because there is a single cluster and we declared in Definition \ref{def_quasi_partition} that blocks do not have self-loops. Condition (\~D3) states for the edge set the analogous requirement that condition (D2), or (\~D2) for that matter, states for the node set. If there is an edge present at a given resolution $\delta_1$ that edge should persist at coarser resolutions $\delta_2>\delta_1$ except if two blocks linked by the edge merge into a single cluster.

Respective comparison of (\~D1), (\~D2), and (\~D4) to properties (D1), (D2), and (D3) in Section \ref{sec_preliminaries} implies that given a quasi-dendrogram $\tilde{D}_X=(D_X, E_X)$ on a node set $X$, the component $D_X$ is a dendrogram on $X$. I.e, the vertex sets $D_X(\delta)$ of the quasi-partitions $(D_X(\delta), E_X(\delta))$ for varying $\delta$ form a nested set of partitions. Hence, if the edge set $E_X(\delta) = \emptyset$ for every resolution parameter $\delta \geq 0$, $\tilde{D}_X$ recovers the structure of the dendrogram $D_X$. Thus, quasi-dendrograms are a generalization of dendrograms, or, equivalently, dendrograms are particular cases of quasi-dendrograms with empty edge sets. Redefining dendrograms $D_X$ so that they represent quasi-dendrograms $(D_X,\emptyset)$ with empty edge sets and reinterpreting $\ccalD$ as the set of quasi-dendrograms with empty edge sets we have that $\ccalD \subset \tilde{\ccalD}$, where $\tilde{\ccalD}$ is the space of quasi-dendrograms.

A hierarchical clustering method $\ccalH: \ccalN \to \ccalD$ is defined as a map from the space of networks $\ccalN$ to the space of dendrograms $\ccalD$ [cf. \eqref{eqn_clust_from_networks_to_dendrograms}]. Likewise, we define a hierarchical \emph{quasi-clustering} method as a map
\begin{equation}\label{eqn:def_hierarchical_quasi_clustering_methods}
   \tilde{\ccalH}: \ccalN \to \tilde{\ccalD},
\end{equation}
from the space of networks to the space of quasi-dendrograms such that the underlying space $X$ is preserved. Since $\ccalD \subset \tilde{\ccalD}$ we have that every clustering method is a quasi-clustering method but not vice versa. Our goal here is to study quasi-clustering methods satisfying suitably modified versions of the axioms of value and transformation introduced in Section \ref{sec_axioms}. Before that, we introduce quasi-ultrametrics as asymmetric versions of ultrametrics and show their equivalence to quasi-dendrograms in the following section after two pertinent remarks.

\begin{remark} \normalfont
If we are given a quasi-equivalence relation and its induced quasi-partition on a node set $X$, \eqref{eqn_quasi_equiv_equiv} implies that all nodes inside the same block of the quasi-partition are quasi-equivalent to each other. If we combine this with the transitivity property in Definition \ref{def_quasi_equivalence}, we have that if $x_i \leadsto x_j$ for some $x_i \in B_i$ and $x_j \in B_j$ or, equivalently, $(B_i, B_j) \in E$ then $x'_i \leadsto x'_j$ for all $x'_i\in B_i$ and all $x'_j\in B_j$.
\end{remark}

\begin{remark} \normalfont
Unidirectionality (QP1) ensures that no cycles containing exactly two blocks can exist in any quasi-partition $\tdP=(P,E)$. If there were longer cycles, transitivity (QP2) would imply that every two distinct blocks in a longer cycle would have to form a two-block cycle, contradicting (QP1). Thus, conditions (QP1) and (QP2) imply that every quasi-partition $\tdP=(P,E)$ is directed acyclic graph (DAG). The fact that a DAG represents a partial order shows that our construction of a quasi-partition from a quasi-equivalence relation is consistent with the known set theoretic construction of a partial order on a partition of a set given a preorder on the set \cite{Harzheim05}. 
\end{remark}

\subsection{Quasi-ultrametrics}\label{sec_quasi_ultrametrics}
Given a node set $X$, a \emph{quasi-ultrametric} $\tdu_X$ on $X$ is a function $\tdu_X: X \times X \to \reals_+$ satisfying the identity property and the strong triangle inequality in \eqref{eqn_strong_triangle_inequality} as we formally define next. 

%
\begin{definition}\label{def_quasi_ultrametric}
 Given a node set $X$ a quasi-ultrametric $\tdu_X$ is a nonnegative function $\tdu_X: X \times X \to \reals_+$ satisfying the following properties for all $x,x',x''\in X$:
\begin{mylist}
\item[{\it (i) Identity.}] $\tdu_X(x, x')=0$ if and only if $x=x'$.
\item[{\it (ii) Strong triangle inequality.}] $\tdu_X$ satisfies \eqref{eqn_strong_triangle_inequality}.
\end{mylist} \vspace{-10pt} \end{definition}

%
Comparison of definitions \ref{def_ultrametric} and \ref{def_quasi_ultrametric} shows that a quasi-ultrametric may be regarded as a relaxation of the notion of an ultrametric in that the symmetry property is not imposed. In particular, the space $\tilde{\ccalU}$ of quasi-ultrametric networks, i.e. networks with quasi-ultrametrics as dissimilarity functions, is a superset of the space of ultrametric networks $\ccalU\subset\tilde{\ccalU}$. See \cite{gurvich} for a study of some structural properties of quasi-ultrametrics.

Analogously to the claim in Theorem \ref{theo_dendrograms_as_ultrametrics} that provides a structure preserving bijection between dendrograms and ultrametrics, the following constructions and theorem establish a structure preserving equivalence between quasi-dendrograms and quasi-ultrametrics.

Consider the map $\tilde{\Psi}:\tilde{\mathcal{D}}\rightarrow\tilde{\mathcal{U}}$ defined as follows:
for a given quasi-dendrogram $\tilde{D}_X=(D_X, E_X)$ over the set $X$ write $\tilde{\Psi}(\tilde{D}_X) = (X,\tdu_X)$, where we define $\tdu_X(x,x')$ for each $x, x' \in X$ as the smallest resolution $\delta$ at which either both nodes belong to the same equivalence class $[x]_\delta=[x']_\delta$, i.e. $x \sim_{D_X(\delta)} x'$, or there exists an edge in $E_X(\delta)$ from the equivalence class $[x]_\delta$ to the equivalence class $[x']_\delta$,
\begin{align}\label{eqn_theo_dendrograms_as_quasi_ultrametrics_10}
   \tdu_X(x,x')& := \min \Big\{ \delta\geq 0 \, \Big| \\
   &  [x]_\delta = [x']_\delta \quad \text{\normalfont or} \quad ([x]_\delta, [x']_\delta) \in E_X(\delta) \Big\}. \nonumber
\end{align}

We also consider the map $\tilde{\Upsilon}:\tilde{\mathcal{U}}\rightarrow\tilde{\mathcal{D}}$ constructed as follows: for a given quasi-ultrametric $\tdu_X$ on the set $X$ and each $\delta \geq 0$ define the relation $\sim_{\tdu_X(\delta)}$ on $X$ as
\begin{equation}\label{eqn_theo_dendrograms_as_quasi_ultrametrics_20}
   x \sim_{\tdu_X(\delta)} x' \iff \max \big( \tdu_X(x,x'), \tdu_X(x',x) \big) \leq \delta.
\end{equation}
Define further $D_X(\delta) :=\big\{X \mod \sim_{\tdu_X(\delta)}\big\}$ and the edge set $E_X(\delta)$ for every $\delta \geq 0$ as follows: $B_1 \neq B_2 \in D_X(\delta)$ are such that 
\begin{equation}\label{eqn_theo_dendrograms_as_quasi_ultrametrics_30}
   (B_1, B_2) \in E_X(\delta) \iff \min_{\substack{x_1 \in B_1\\x_2 \in B_2}} \tdu_X(x_1, x_2) \leq \delta.
\end{equation}
\mbox{Finally, $\tilde{\Upsilon}(X, \tdu_X):= \tilde{D}_X$, where $\tilde{D}_X:=(D_X, E_X)$.}

\begin{theorem}\label{theo_equivalence_quasi_dendrogram_quasi_ultrametric}
The maps $\tilde{\Psi}:\tilde{\mathcal{D}}\rightarrow\tilde{\mathcal{U}}$ and $\tilde{\Upsilon}:\tilde{\mathcal{U}}\rightarrow\tilde{\mathcal{D}}$ are both well defined. Furthermore, $\tilde{\Psi}\circ\tilde{\Upsilon}$ is the identity on $\tilde{\mathcal{U}}$ and $\tilde{\Upsilon}\circ\tilde{\Psi}$ is the identity on $\tilde{\mathcal{D}}$.
\end{theorem}

\begin{myproofnoname} See Appendix \ref{appendix_sec_full_characterization_asymmetric}.
\end{myproofnoname}

\begin{remark}\label{rem_equivalence_quasi_ultram_quasi_dendro}\normalfont
Theorem \ref{theo_equivalence_quasi_dendrogram_quasi_ultrametric} implies that every quasi-dendrogram $\tilde{D}_X$ has an equivalent representation as a quasi-ultrametric network defined on the same underlying node set $X$ given by $\tilde{\Psi} ( \tilde{D}_X) $. Analogously, every quasi-ultrametric network $\tilde{U} = (X, \tilde{u}_X)$ has an equivalent quasi-dendrogram given by $\tilde{\Upsilon} (\tilde{U})$.
\end{remark}

The equivalence between quasi-dendrograms and quasi-ultrametric networks described in Remark \ref{rem_equivalence_quasi_ultram_quasi_dendro} allows us to reinterpret hierarchical quasi-clustering methods [cf. \eqref{eqn:def_hierarchical_quasi_clustering_methods}] as maps 
\begin{equation}
\tilde{\ccalH}:\ccalN \to \tilde{\ccalU},
\end{equation}
from the space of networks to the space of quasi-ultrametric networks. Apart from the 
theoretical importance of Theorem \ref{theo_equivalence_quasi_dendrogram_quasi_ultrametric}, this equivalence result is of practical importance since quasi-ultrametrics are mathematically more convenient to handle than quasi-dendrograms -- in the same sense in which regular ultrametrics are easier to handle than regular dendrograms. Quasi-dendrograms are still preferable for data representation as we discuss in the numerical examples in Section \ref{sec_numerical_experiments}.  

Given a quasi-dendrogram $\tdD_X=(D_X, E_X)$, the value $\tdu_X(x, x')$ of the associated quasi-ultrametric for $x, x' \in X$ is given by the minimum resolution $\delta$ at which $x$ can influence $x'$. This may occur when $x$ and $x'$ belong to the same block of $D_X(\delta)$ or when they belong to different blocks $B, B' \in D_X(\delta)$, but there is an edge from the block containing $x$ to the block containing $x'$, i.e. $(B, B') \in E_X(\delta)$. Conversely, given a quasi-ultrametric network $(X, \tdu_X)$, for a given resolution $\delta$ the graph $\tdD_X(\delta)$ has as a vertex set the classes of nodes whose quasi-ultrametric is less than $\delta$ in both directions. Furthermore, $\tdD_X(\delta)$ contains a directed edge between two distinct equivalence classes if the quasi-ultrametric from some node in the first class to some node in the second is not greater than $\delta$. 

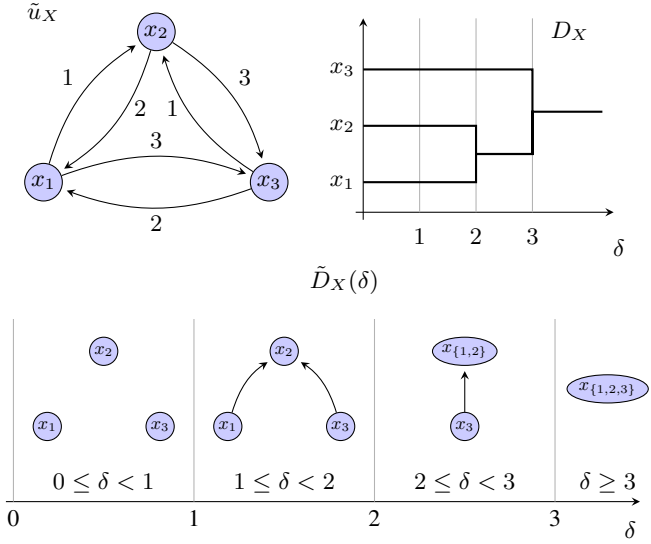
\begin{figure}
\def \thisplotscale {0.5}
\def \unit {\thisplotscale cm}

{\small
\begin{tikzpicture}[-stealth, shorten >=2, scale = \thisplotscale]

    \node [blue vertex] at (1,3) (1) {$x_2$};
    \node [blue vertex] at (4,-1) (2) {$x_3$};    
    \node [blue vertex] at (-2,-1) (3) {$x_1$};

    \path (1) edge [bend left=20, above right] node {$3$} (2);	
    \path (2) edge [bend left=20, below] node {$2$} (3);
    \path (3) edge [bend left=20, above left] node {$1$} (1);    	

    \path (2) edge [bend left=20,left, pos=0.6] node {$1$} (1);	
    \path (3) edge [bend left=20, above] node {$3$} (2);
    \path (1) edge [bend left=20, right, , pos=0.4]  node {$2$} (3);    	
    

    \node [below] at (6,-3) {$\tilde{D}_X(\delta)$};

       \node [blue vertex, scale=0.75] at (-0.4,-5.5) (4) {$x_2$};
    \node [blue vertex, scale=0.75] at (1.1,-7.5) (5) {$x_3$};    
    \node [blue vertex, scale=0.75] at (-1.9,-7.5) (6) {$x_1$};
    
    
    \node [below] at (-0.4,-8.5) {$0 \leq \delta < 1$};

           \node [blue vertex, scale=0.75] at (4.4,-5.5) (4p) {$x_2$};
    \node [blue vertex, scale=0.75] at (5.9,-7.5) (5p) {$x_3$};    
    \node [blue vertex, scale=0.75] at (2.9,-7.5) (6p) {$x_1$};
    

       \path (6p) edge [bend left=20, above] node {} (4p);
    \path (5p) edge [bend right=20, right, , pos=0.4]  node {} (4p);    	
    
    \node [below] at (4.4,-8.5) {$1 \leq \delta < 2$};

     \node [blue vertex, scale=0.75] at (9.2,-5.5) (7p) {$x_{\{1,2\}}$};
      \node [blue vertex, scale=0.75] at (9.2,-7.5) (8p) {$x_3$};
      
      
         \path (8p) edge node {} (7p);

      \node [below] at (9.2,-8.5) {$2 \leq \delta < 3$};

         \node [blue vertex, scale=0.75] at (13,-6.5) (9p) {$x_{\{1,2, 3\}}$};
         
           \node [below] at (13,-8.5) {$\delta \geq 3$};




   \small
   \draw [-stealth] (6.5,-2) -- (6.5,3.5);
    \draw [-stealth] (6.3,-1.8) -- (13.3,-1.8);
    \draw [-, draw=black!30] (8,-1.8) -- (8,3.5);
      \draw [-, draw=black!30] (9.5,-1.8) -- (9.5,3.5);
       \draw [-, draw=black!30] (11,-1.8) -- (11,3.5);
    
   \draw[thick, -] (6.5, -1) -- ++(3,0) -- ++(0,1.5) -- ++(-3,0) ++(0,1.5) -- ++(4.5,0) -- ++(0,-2.25) -- ++(-1.5,0)-- ++(1.5,0)-- ++(0,1.125)-- ++(2,0);

    \node [below] at (13.3,-2.3) {$\delta$};
    \node [below] at (8,-2) {$1$};
    \node [below] at (9.5,-2) {$2$};
    \node [below] at (11,-2) {$3$};
    \node [left] at (6.5,-1) {$x_1$};
    \node [left] at (6.5,0.5) {$x_2$};
    \node [left] at (6.5,2) {$x_3$};
    
    \node [below] at (12,3.5) {$D_X$};
    
    \node [below] at (-2,4) {$\tdu_X$};
    
      \draw [-stealth] (-3,-9.5) -- (14,-9.5);
       \draw [-, draw=black!30] (-2.8,-9.5) -- (-2.8,-4.5);
       \node [below] at (-2.8, -9.5) {0};
        \draw [-, draw=black!30] (2,-9.5) -- (2,-4.5);
        \node [below] at (2,-9.5) {1};
         \draw [-, draw=black!30] (6.8,-9.5) -- (6.8,-4.5);
         \node [below] at (6.8,-9.5) {2};
          \draw [-, draw=black!30] (11.6,-9.5) -- (11.6,-4.5);
    \node [below] at (11.6,-9.5) {3};
       \node [below] at (13.6,-9.8) {$\delta$};
    
    
    

\end{tikzpicture}
}
\vspace{-0.1in}
\caption{Equivalence between quasi-dendrograms and quasi-ultrametrics. A quasi-ultrametric $\tdu_X$ is defined on three nodes $\{x_1, x_2, x_3\}$ and the equivalent quasi-dendrogram $\tilde{D}_X=(D_X, E_X)$ is presented by depicting $D_X$ and graphs $\tilde{D}_X(\delta)$ for every resolution $\delta$.}
\vspace{-0.1in}
\label{fig_quasi_dendrogram_example}
\end{figure}

In Fig. \ref{fig_quasi_dendrogram_example} we present an example of the equivalence between quasi-dendrograms and quasi-ultrametric networks stated by Theorem \ref{theo_equivalence_quasi_dendrogram_quasi_ultrametric}. At the top left of the figure, we present a quasi-ultrametric $\tdu_X$ defined on a three-node set $X=\{x_1, x_2, x_3\}$. At the top right, we depict the dendrogram component $D_X$ of the quasi-dendrogram $\tilde{D}_X=(D_X, E_X)$ equivalent to $(X, \tdu_X)$ as given by Theorem \ref{theo_equivalence_quasi_dendrogram_quasi_ultrametric}. At the bottom of the figure, we present graphs $\tilde{D}_X(\delta)$ for a  range of resolutions $\delta \geq 0$.

To obtain $\tilde{D}_X$ from $\tdu_X$, we first obtain the dendrogram component $D_X$ by symmetrizing $\tdu_X$ to the maximum [cf. \eqref{eqn_theo_dendrograms_as_quasi_ultrametrics_20}], nodes $x_1$ and $x_2$ merge at resolution 2 and $x_3$ merges with $\{x_1, x_2\}$ at resolution 3. To see how the edges in $\tdD_X$ are obtained, at resolutions $0 \leq \delta < 1$, there are no edges since there is no quasi-ultrametric value between distinct nodes in this range [cf. \eqref{eqn_theo_dendrograms_as_quasi_ultrametrics_30}]. At resolution $\delta=1$, we reach the first nonzero values of $\tdu_X$ and hence the corresponding edges appear in $\tdD_X(1)$. At resolution $\delta=2$, nodes $x_1$ and $x_2$ merge and become the same vertex in graph $\tdD_X(2)$. Finally, at resolution $\delta=3$ all the nodes belong to the same equivalence class and hence $\tdD_X(3)$ contains only one vertex. Conversely, to obtain $\tdu_X$ from $\tilde{D}_X$ as depicted in the figure, note that at resolution $\delta=1$ two edges $([x_1]_1, [x_2]_1)$ and $([x_3]_1, [x_2]_1)$ appear in $\tdD_X(1)$, thus the corresponding values of the quasi-ultrametric are fixed to be $\tdu_X(x_1, x_2)=\tdu(x_3, x_2)=1$. At resolution $\delta=2$, when $x_1$ and $x_2$ merge into the same vertex in $\tdD_X(2)$, an edge is generated from $[x_3]_2$ to $[x_1]_2$ the equivalence class of $x_1$ at resolution $\delta=2$ which did not exist before, implying that $\tdu_X(x_3, x_1)=2$. Moreover, we have that $[x_2]_2 = [x_1]_2$, hence $\tdu_X(x_2, x_1)=2$. Finally, at $\tdD_X(3)$ there is only one equivalence class, thus the values of $\tdu_X$ that have not been defined so far must equal 3.

%

\subsection{Axioms for hierarchical quasi-clustering methods}\label{sec_quasi_clustering_axioms}

Mimicking the development in Section \ref{sec_axioms}, we encode desirable properties of quasi-clustering methods into axioms which we use as a criterion for admissibility. The axioms considered are the directed versions of the axioms of value (A1) and transformation (A2) introduced in Section \ref{sec_axioms}. The Directed Axiom of Value (\~A1) and the Directed Axiom of Transformation (\~A2) winnow the space of quasi-clustering methods by imposing conditions on their output quasi-dendrograms.


%
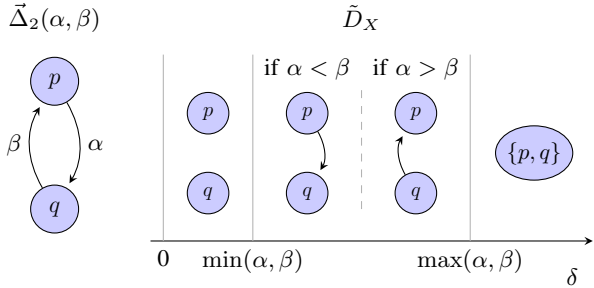
\begin{figure}
\centering
\centerline{\def \thisplotscale {0.85}
\def \unit {\thisplotscale cm}
\def \xdendogram{{1, 2}}
\def \ydendogram{{1, 2}}

{\small
\begin{tikzpicture}[shorten >=2, scale = \thisplotscale]

    \node [blue vertex, scale = 0.75] at (-4.5,1.5) (p) {\large{$p$}};
    \node [blue vertex, scale = 0.75] at (-4.5,-0.5) (q) {\large{$q$}};
    \path [-stealth](p) edge [bend left, right] node {$\alpha$} (q);	
    \path [-stealth] (q) edge [bend left, left] node {$\beta$}  (p);	
    
    \node at (-4.5,2.5) {$\vec{\Delta}_2(\alpha, \beta)$};

    \node at (0.3,2.5) {$\tdD_X$};
    
      \draw [-stealth] (-3,-1) -- (4,-1);
       \draw [-, draw=black!30] (-2.8,-1) -- (-2.8,2);
       \node [below] at (-2.8, -1) {0};
        \draw [-, draw=black!30] (-1.4,-1) -- (-1.4,2);
        \node [below] at (-1.4,-1) {$\min(\alpha, \beta)$};
         \draw [-, draw=black!30] (2,-1) -- (2,2);
         \node [below] at (2,-1) {$\max(\alpha, \beta)$};
       \node [below] at (3.6,-1.3) {$\delta$};
       
       \draw [dashed, draw=black!30] (0.3,-0.5) -- (0.3,1.5);
       
    \node [blue vertex, scale = 0.65] at (-2.1,1) (p1) {\large{$p$}};
    \node [blue vertex, scale = 0.65] at (-2.1,-0.25) (q1) {\large{$q$}};
    
        \node [blue vertex, scale = 0.65] at (-0.55,1) (p2) {\large{$p$}};
    \node [blue vertex, scale = 0.65] at (-0.55,-0.25) (q2) {\large{$q$}};
    \path [-stealth] (p2) edge [bend left, right] node {$$} (q2);	
           \node [above] at (-0.55,1.4) {if $\alpha < \beta$};

        \node [blue vertex, scale = 0.65] at (1.15,1) (p3) {\large{$p$}};
    \node [blue vertex, scale = 0.65] at (1.15,-0.25) (q3) {\large{$q$}};
    \path [-stealth] (q3) edge [bend left, left] node {$$}  (p3);	
               \node [above] at (1.15,1.4) {if $\alpha > \beta$};

        \node [blue vertex, scale = 0.85] at (3,0.3775) (p4) {\normalsize{$\{p, q\}$}};

\end{tikzpicture}
}}
\vspace{-0.1in}
\caption{Directed Axiom of Value. Nodes in a two-node network merge into one block at the minimum resolution at which both can influence each other. For smaller resolutions, the quasi-dendrogram captures unidirectional influence.}
\vspace{-0.1in}
\label{fig_axiom_directed_value}
\end{figure}

\begin{indentedparagraph}{(\~A1) Directed Axiom of Value} For each $\alpha,\beta\geq 0$, the quasi-dendrogram $\tdD_X=(D_X, E_X)=\tilde{\ccalH}(\vec{\Delta}_2(\alpha, \beta))$ produced by $\tilde{\ccalH}$ on the arbitrary two-node network $\vec{\Delta}_2(\alpha, \beta)$ is such that $D_{X}(\delta)=\big\{\{p\},\{q\}\big\}$ for $\delta<\max(\alpha,\beta)$ and $D_{X}(\delta)=\big\{\{p,q\}\big\}$ for $\delta \geq\max(\alpha,\beta)$. When $\alpha \neq \beta$, the edge sets $E_X(\delta)$ are non-empty for resolutions $\min(\alpha, \beta) \leq \delta < \max(\alpha, \beta)$ where $(q, p) \in E_X(\delta)$ if $\alpha > \beta$ and $(p, q) \in E_X(\delta)$ if $\alpha < \beta$; see Fig. \ref{fig_axiom_directed_value}.
\end{indentedparagraph}

\begin{indentedparagraph}{(\~A2) Directed Axiom of Transformation} 
Consider two networks $N_X=(X,A_X)$ and $N_Y=(Y,A_Y)$ and a dissimilarity-reducing map $\phi:X \to Y$. Then, the output quasi-dendrograms $\tdD_X=(D_X, E_X)=\ccalH(N_X)$ and $\tdD_Y=(D_Y, E_Y)=\ccalH(N_Y)$ are such that for all $\delta \geq 0$, if $[x]_\delta = [x']_\delta$ then $[\phi(x)]_\delta = [\phi(x')]_\delta$ and if $([x]_\delta, [x']_\delta) \in E_X(\delta)$ then $([\phi(x)]_\delta, [\phi(x')]_\delta) \in E_Y(\delta)$ or $[\phi(x)]_\delta = [\phi(x')]_\delta$ for all $x, x' \in X$.
\end{indentedparagraph}

\noindent Theorem \ref{theo_equivalence_quasi_dendrogram_quasi_ultrametric} allows us to rewrite axioms (\~A1) and (\~A2) in terms of quasi-ultrametric networks. As it was the case for ultrametrics and dendrograms, quasi-ultrametrics are mathematically more convenient to handle than quasi-dendrograms. The first indication of this fact is the simpler reformulation of axioms (\~A1) and (\~A2) in terms of quasi-ultrametrics:

\begin{indentedparagraph}{(\~A1) Directed Axiom of Value} $\tilde{\ccalH}(\vec{\Delta}_2(\alpha, \beta))= \vec{\Delta}_2(\alpha, \beta)$ for every two-node network $\vec{\Delta}_2(\alpha, \beta)$.\end{indentedparagraph}

\begin{indentedparagraph}{(\~A2) Directed Axiom of Transformation} Consider two networks $N_X=(X,A_X)$ and $N_Y=(Y,A_Y)$ and a dissimilarity-reducing map $\phi:X\to Y$, i.e. a map $\phi$ such that for all $x,x' \in X$ it holds that $A_X(x,x')\geq A_Y(\phi(x),\phi(x'))$. Then, for all $x, x' \in X$, the outputs $(X,\tdu_X)=\tilde{\ccalH}(X,A_X)$ and $(Y,\tdu_Y)=\tilde{\ccalH}(Y,A_Y)$ satisfy 
\begin{equation}\label{eqn_dissimilarity_reducing_quasi_ultrametric}
    \tdu_X(x,x') \geq \tdu_Y(\phi(x),\phi(x')).
\end{equation} \end{indentedparagraph}

The Directed Axiom of Transformation (\~A2) is just a restatement of the (regular) Axiom of Transformation (A2) where the ultrametrics $u_X$ and $u_Y$ in \eqref{eqn_dissimilarity_reducing_ultrametric} are replaced by the quasi-ultrametrics $\tdu_X$ and $\tdu_Y$ in \eqref{eqn_dissimilarity_reducing_quasi_ultrametric}. The axioms are otherwise conceptual analogues. In terms of quasi-dendrograms, (\~A2) states that no influence relation can be weakened by a dissimilarity reducing transformation. The Directed Axiom of Value (\~A1) simply recognizes that in any two-node network, the dissimilarity function is itself a quasi-ultrametric and that there is no valid justification to output a different quasi-ultrametric. In this sense, (\~A1) is similar to the Symmetric Axiom of Value (B1) that also requires two-node networks to be fixed points of (symmetric) hierarchical clustering methods. In terms of quasi-dendrograms, (\~A1) requires the quasi-clustering method to output the quasi-dendrogram equivalent according to Theorem \ref{theo_equivalence_quasi_dendrogram_quasi_ultrametric} to the dissimilarity function of the two-node network.

%
\subsection{Existence and uniqueness of admissible quasi-clustering methods: directed single linkage}\label{sec_existance_uniqueness_quasi_clustering}

We call a quasi-clustering method $\tilde{\ccalH}$ \emph{admissible} if it satisfies axioms (\~A1) and (\~A2) and, emulating the development in Section \ref{sec_reicprocal_and_nonreciprocal}, we want to find methods that are admissible with respect to these axioms. This is can be done in the following way. Recall the definition of the directed minimum chain cost $\tdu^*_X$ in \eqref{eqn_nonreciprocal_chains} and define the \emph{directed single linkage} quasi-clustering method $\tilde{\ccalH}^*$ as the one with output quasi-ultrametrics $(X, \tdu_X^*)=\tilde{\ccalH}^*(X, A_X)$ given by the directed minimum chain cost function $\tdu^*_X$. The directed single linkage method $\tilde{\ccalH}^*$ is valid and admissible as we show in the following proposition.

%
\begin{proposition}\label{prop_directed_axioms}
The hierarchical quasi-clustering method $\tilde{\ccalH}^*$ is valid and admissible. I.e., $\tdu^*_X$ defined by  \eqref{eqn_nonreciprocal_chains} is a quasi-ultrametric and $\tilde{\ccalH}^*$ satisfies axioms (\~A1)-(\~A2).
\end{proposition}

%
\begin{myproofnoname} 
In order to show that $\tdu^*_X$ is a valid quasi-ultrametric we may apply an argument based on concatenated chains as the one preceding Proposition \ref{prop_reciprocal_axioms}.

To show fulfillment of Axiom (\~A1), pick an arbitrary two-node network $\vec{\Delta}_2(\alpha, \beta)$ as defined in Section \ref{sec_preliminaries} and denote by $(\{p,q\}, \tdu^*_{p,q})=\tilde{\ccalH}^*(\vec{\Delta}_2(\alpha, \beta))$. Then, we have $\tdu^*_{p,q}(p,q)=\alpha$ and $\tdu^*_{p,q}(q,p)=\beta$ because there is only one possible chain selection in each direction [cf. \eqref{eqn_nonreciprocal_chains}]. Satisfaction of the Directed Axiom of Transformation (\~A2) is the intermediate result \eqref{eqn_theo_nonreciprocal_axioms_pf_50} in the proof of Proposition \ref{prop_nonreciprocal_axioms} in Appendix \ref{appendix_sec_reicprocal_and_nonreciprocal}.\end{myproofnoname}

From Proposition \ref{prop_directed_axioms} we know that $\tdu_X^*$ is a quasi-ultrametric. Its equivalent quasi-dendrogram according to Theorem \ref{theo_equivalence_quasi_dendrogram_quasi_ultrametric} [cf. Remark \ref{rem_equivalence_quasi_ultram_quasi_dendro}] is related to the nonreciprocal clustering method $\ccalH^{\NR}$ as we show next.

%


\begin{proposition}\label{prop_quasi_ultrametric_non_reciprocal}
For every network $N=(X, A_X)$, let $\tdD^*_X=(D^*_X, E^*_X)$ denote the quasi-dendrogram $\ccalH^*(N)$. Then, $D^*_X = D^{\NR}_X$ where $D^{\NR}_X=\ccalH^{\NR}(N)$ is the output dendrogram of applying nonreciprocal clustering as defined in \eqref{eqn_nonreciprocal_clustering} to the network $N$.
\end{proposition}

%
\begin{myproofnoname} 
Compare \eqref{eqn_nonreciprocal_clustering} with \eqref{eqn_theo_dendrograms_as_quasi_ultrametrics_20} and conclude that
\begin{equation}\label{eqn_quasi_ultrametric_non_reciprocal}
x \sim_{\tdu^*_X(\delta)} x' \iff u^{\NR}_X(x, x') \leq \delta,
\end{equation}
for all $x, x' \in X$.
The equivalence relation $\sim_{\tdu^*_X(\delta)}$ defines $D^*_X$ and by \eqref{eqn_theo_dendrograms_as_ultrametrics_20} in Theorem \ref{theo_dendrograms_as_ultrametrics} we obtain that the equivalence relation $\sim_{U^{\NR}_X(\delta)}$ defining  $D^{\NR}_X$ is such that
\begin{equation}\label{eqn_quasi_ultrametric_non_reciprocal_2}
x \sim_{U^{\NR}_X(\delta)} x' \iff u^{\NR}_X(x, x') \leq \delta.
\end{equation}
Comparing \eqref{eqn_quasi_ultrametric_non_reciprocal} and \eqref{eqn_quasi_ultrametric_non_reciprocal_2}, the result follows.
\end{myproofnoname}

Furthermore, from \eqref{eqn_single_linkage_ultrametric} and \eqref{eqn_algorithm_single_linkage} it follows that for every network $(X, A_X)$ with $|X|=n$, the quasi-ultrametric $\tdu_X^*$ can be computed as
\begin{equation}\label{eqn_algo_quasi_ultrametric}
\tdu_X^*=A_X^{(n-1)},
\end{equation}
where the operation $(\cdot)^{(n-1)}$ denotes the $(n-1)$st matrix power in the dioid algebra $(\reals^+\cup\{+\infty\},\min,\max)$ with matrix product as defined in \eqref{def_star}.

Mimicking the developments in sections \ref{sec_reicprocal_and_nonreciprocal} and \ref{sec_extremal_ultrametrics}, we next ask which other methods satisfy (\~A1)-(\~A2) and what special properties directed single linkage has. As it turns out, directed single linkage is the \emph{unique} quasi-clustering method that is admissible with respect to (\~A1)-(\~A2) as we assert in the following theorem.

%
\begin{theorem}\label{theo_uniqueness_quasi_clustering}
Let $\tilde{\ccalH}$ be a valid hierarchical quasi-clustering method satisfying axioms (\~A1) and (\~A2). Then, $\tilde{\ccalH} \equiv \tilde{\ccalH}^*$ where $\tilde{\ccalH}^*$ is the directed single linkage method with output quasi-ultrametrics as in \eqref{eqn_nonreciprocal_chains}.
\end{theorem}

%
\begin{myproofnoname} The proof is similar to the proof of Theorem \ref{theo_extremal_ultrametrics}. Given an arbitrary network $N=(X, A_X)$ denote by $(X,\tdu_X)=\tilde{\ccalH}(X,A_X)$ the output quasi-ultrametric resulting from application of an arbitrary admissible quasi-clustering method $\tilde{\ccalH}$. We will show that for all $x, x' \in X$
\begin{equation}\label{eqn_inequality_unicity_directed}
  \tdu_X^*(x, x') \leq\tdu_X(x, x') \leq\tdu_X^*(x, x').
\end{equation}
To prove the rightmost inequality in \eqref{eqn_inequality_unicity_directed} we begin by showing that the dissimilarity function $A_X$ acts as an upper bound on all admissible quasi-ultrametrics $\tdu_X$, i.e.
\begin{equation}\label{eqn_dissimilarity_upper_bound_directed}
\tilde{u}_X(x, x') \leq A_X(x, x'),
\end{equation}
for all $x, x' \in X$. To see this, suppose $A_X(x, x')=\alpha$ and $A_X(x', x)=\beta$. Define the two-node network $\vec{\Delta}_2(\alpha,\beta)=(\{p,q\}, A_{p,q})$ where $A_{p,q}(p,q)=\alpha$ and $A_{p,q}(q,p)=\beta$ and denote by $(\{p, q\}, \tilde{u}_{p,q})=\tilde{\ccalH}(\vec{\Delta}_2(\alpha,\beta))$ the output of applying the method $\tilde{\ccalH}$ to the network $\vec{\Delta}_2(\alpha,\beta)$. From axiom (\~A1), we have $\tilde{\ccalH}(\vec{\Delta}_2(\alpha,\beta))=\vec{\Delta}_2(\alpha,\beta)$, in particular
\begin{equation}\label{eqn_dissimilarity_two_node_network}
\tilde{u}_{p,q}(p, q) = A_{p,q}(p, q)= A_{X}(x, x').
\end{equation}
Moreover, notice that the map $\phi:\{p,q\} \to X$, where $\phi(p)=x$ and $\phi(q)=x'$ is a dissimilarity reducing map, i.e. it does not increase any dissimilarity, from $\vec{\Delta}_2(\alpha,\beta)$ to $N$. Hence, from axiom (\~A2), we must have
\begin{equation}\label{eqn_quasi-ultra_two_node_network}
\tdu_{p,q}(p,q) \geq \tdu_{X}(\phi(p),\phi(q)) = \tdu_{X}(x,x').
\end{equation}
Substituting \eqref{eqn_dissimilarity_two_node_network} in \eqref{eqn_quasi-ultra_two_node_network}, we obtain \eqref{eqn_dissimilarity_upper_bound_directed}.

Consider now an arbitrary chain $C(x, x')=[x=x_0, x_1, \ldots , x_l=x']$ linking nodes $x$ and $x'$. Since $\tilde{u}_X$ is a valid quasi-ultrametric, it satisfies the strong triangle inequality \eqref{eqn_strong_triangle_inequality}. Thus, we have that
\begin{equation}\label{eqn_stron_triangle_directed}
\tilde{u}_{X}(x, x') \! \leq \!\!\! \max_{i|x_i \in C(x, x')} \!\! \tdu_{X}(x_i, x_{i+1}) \leq \!\! \max_{i|x_i \in C(x, x')} \!\!\! A_X(x_i, x_{i+1}),
\end{equation}
where the last inequality is implied by \eqref{eqn_dissimilarity_upper_bound_directed}. Since by definition $C(x, x')$ is an arbitrary chain linking $x$ to $x'$, we can minimize \eqref{eqn_stron_triangle_directed} over all such chains maintaining the validity of the inequality,
\begin{equation}\label{eqn_stron_triangle_directed_minimizing}
\tilde{u}_{X}(x, x') \leq \min_{C(x, x')} \,\, \max_{i|x_i \in C(x, x')} A_X(x_i, x_{i+1}) =\tdu^*_X(x, x'),
\end{equation}
where the last equality is given by the definition of the directed minimum chain cost \eqref{eqn_nonreciprocal_chains}. Thus, the rightmost inequality in \eqref{eqn_inequality_unicity_directed} is proved.

To prove the leftmost inequality in \eqref{eqn_inequality_unicity_directed}, consider an arbitrary pair of nodes $x, x' \in X$ and fix $\delta = \tilde{u}^*_X(x, x')$. Then, by Lemma \ref{lemma_axiom_redundancy}, there exists a partition $P_\delta(x,x')=\{B_\delta(x), B_\delta(x')\}$ of the node space $X$ into blocks $B_\delta(x)$ and $B_\delta(x')$ with $x \in B_\delta(x)$ and $x' \in B_\delta(x')$ such that for all points $b \in B_\delta(x)$ and $b' \in B_\delta(x')$ we have
\begin{equation}\label{eqn_partition_proof_quasi_ultrametrics}
A_X(b, b') \geq \delta.
\end{equation}
Focus on a two-node network $\vec{\Delta}_2(\delta,s)=(\{u,v\}, A_{u,v})$ with $A_{u,v}(u,v)=\delta$ and $A_{u,v}(v,u)=s$ where $s=\sep(X, A_X)$ as defined in \eqref{eqn_def_separation_network}. Denote by $(\{u, v\}, \tdu_{u,v})=\tilde{\ccalH}(\vec{\Delta}_2(\delta,s))$ the output of applying the method $\tilde{\ccalH}$ to the network $\vec{\Delta}_2(\delta,s)$. Notice that the map $\phi: X \to \{u,v\}$ such that $\phi(b)=u$ for all $b \in B_\delta(x)$ and $\phi(b')=v$ for all $b' \in B_\delta(x')$ is dissimilarity reducing because, from \eqref{eqn_partition_proof_quasi_ultrametrics}, dissimilarities mapped to dissimilarities equal to $\delta$ in $\vec{\Delta}_2(\delta,s)$ were originally larger. Moreover, dissimilarities mapped into $s$ cannot have increased due to the definition of separation of a network \eqref{eqn_def_separation_network}. From axiom (\~A1),
\begin{equation}\label{eqn_two_node_delta_directed}
\tilde{u}_{u,v}(u,v)= A_{u,v}(u,v) = \delta,
\end{equation}
since $\vec{\Delta}_2(\delta,s)$ is a two-node network. Moreover, since $\phi$ is dissimilarity reducing, from (\~A2) we may assert that
\begin{equation}\label{eqn_transformation_directed_delta}
\tilde{u}_{X}(x,x') \geq \tdu_{u,v}(\phi(x), \phi(x')) = \delta,
\end{equation}
where we used \eqref{eqn_two_node_delta_directed} for the last equality. Recalling that $\tilde{u}^*_X(x, x') = \delta$ and substituting in \eqref{eqn_transformation_directed_delta} concludes the proof of the leftmost inequality in \eqref{eqn_inequality_unicity_directed}.

Since both inequalities in \eqref{eqn_inequality_unicity_directed} hold, we must have $\tdu_X^*(x, x') = \tdu_X(x, x')$ for all $x,x' \in X$. Since this is true for any arbitrary network $N=(X,A_X)$, it follows that the admissible quasi-clustering method must be $\tilde{\ccalH}\equiv\tilde{\ccalH}^*$.
\end{myproofnoname}

%
As it follows from Section \ref{sec_intermediate_ultrametrics},  there are exist many (actually, infinitely many) different admissible hierarchical clustering algorithms for asymmetric networks. In the case of symmetric networks, \cite{clust-um} establishes that there is a unique admissible method. Theorem \ref{theo_uniqueness_quasi_clustering} suggests that what prevents uniqueness in asymmetric networks is the insistence that the hierarchical clustering method should have a symmetric ultrametric output. If we remove the symmetry requirement there is also a unique admissible hierarchical quasi-clustering method. Furthermore, this unique method is an asymmetric version of single linkage. 

\begin{remark} \normalfont
The definition of directed single linkage as a natural extension of single linkage hierarchical clustering to asymmetric networks dates back to \cite{boyd-asymmetric}. Our contribution is to develop a framework to study hierarchical quasi-clustering that starts from quasi-equivalence relations, builds towards quasi-partitions and quasi-dendrograms, shows the equivalence of the latter to quasi-ultrametrics, and culminates with the proof that directed single linkage is the \emph{unique} admissible method to hierarchically quasi-cluster asymmetric networks. Furthermore, stability of directed single linkage is established in Section \ref{sec_stability}.
\end{remark}


%
\section{Alternative axiomatic constructions}\label{sec_alternative_axioms}
The axiomatic framework that we adopted allows alternative constructions by modifying the underlying set of axioms. Among the axioms in Section \ref{sec_axioms}, the Axiom of Value (A1) is perhaps the most open to interpretation. Although we required the two-node network in Fig. \ref{fig_axioms_value_influence} to first cluster into one single block at resolution $\max(\alpha,\beta)$ corresponding to the largest dissimilarity and argued that this was reasonable in most situations, it is also reasonable to accept that in some situations the two nodes should be clustered together as long as one of them is able to influence the other. To account for this possibility we replace the Axiom of Value by the following alternative.

\begin{indentedparagraph}{(A1'') Alternative Axiom of Value} The ultrametric output $(\{p,q\},u_{p,q}):=\ccalH(\vec{\Delta}_2(\alpha, \beta))$ produced by $\ccalH$ applied to the two-node network $\vec{\Delta}_2(\alpha, \beta)$ satisfies 
\begin{equation}\label{eqn_two_node_network_ultrametric_alternative}
   u_{p,q}(p,q) = \min(\alpha,\beta).
\end{equation}\end{indentedparagraph}

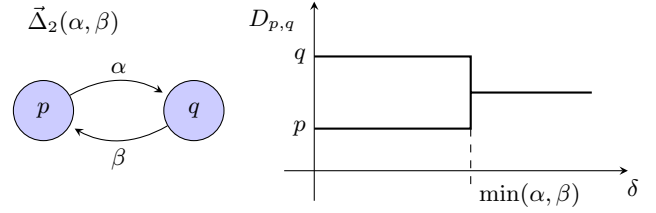
\begin{figure}
\def \thisplotscale {0.8}
\def \unit {\thisplotscale cm}
\def \xdendogram{{1, 2}}
\def \ydendogram{{1, 2}}

{\small
\begin{tikzpicture}[shorten >=2, scale = \thisplotscale]

    \node [blue vertex] at (-4.5,1) (p) {$p$};
    \node [blue vertex] at (-2,1) (q) {$q$};
    \path [-stealth](p) edge [bend left, above] node {$\alpha$} (q);	
    \path [-stealth] (q) edge [bend left, below] node {$\beta$}  (p);	
    
    \draw [-stealth] (-0.5,0) -- (5.3,0) node [below, at end] {$\delta$};
    \draw [-stealth] (0,-0.5) -- (0,2.9);
    
    \draw[thick] (0,0.7) -- ++(2.6,0) -- ++(0,1.2) -- +(-2.6,0) ++(0,-0.6) -- +(2.1,0);
    \draw[dashed](2.6,0.7) -- ++(0,-1.1) node [right, at end] {$\min(\alpha,\beta)$};
    \node [left] at (0,0.7) {$p$};
    \node [left] at (0,1.9) {$q$};
    
        \node at (-4,2.5) {$\vec{\Delta}_2(\alpha, \beta)$};
    \node at (-0.7,2.5) {$D_{p,q}$};
 
\end{tikzpicture}
}


    
\vspace{-0.1in}
\caption{Alternative Axiom of Value. For a two-node network, nodes are clustered together at the minimum resolution at which one of them can influence the other.}
\vspace{-0.1in}
\label{fig_two_node_network_dendrogram_modif}
\end{figure}

\noindent Axiom (A1'') replaces the requirement of bidirectional influence in Axiom (A1) to unidirectional influence; see Fig. \ref{fig_two_node_network_dendrogram_modif}. We say that a clustering method $\ccalH$ is admissible with respect to the alternative axioms if it satisfies axioms (A1'') and (A2).

The property of influence (P1), which is a keystone in the proof of Theorem \ref{theo_extremal_ultrametrics}, is not compatible with the Alternative Axiom of Value (A1''). Indeed, just observe that the minimum loop cost of the two-node network in Fig. \ref{fig_two_node_network_dendrogram_modif} is $\mlc(\vec{\Delta}_2(\alpha, \beta))=\max(\alpha, \beta)$ whereas in \eqref{eqn_two_node_network_ultrametric_alternative} we are requiring the output ultrametric to be $u_X(p,q)=\min(\alpha,\beta)$. We therefore have that Axiom (A1'') itself implies $u_{p,q}(p,q) = \min(\alpha, \beta) < \max(\alpha, \beta) = \mlc(\vec{\Delta}_2(\alpha, \beta))$ for the cases when $\alpha \neq \beta$. Thus, we reformulate (P1) into the Alternative Property of Influence (P1') that we define next.

\begin{indentedparagraph}{(P1') Alternative Property of Influence} For any network $N_X=(X,A_X)$ the output ultrametric $(X, u_X)=\ccalH(X,A_X)$ corresponding to the application of a hierarchical clustering method $\ccalH$ is such that the ultrametric value $u_X(x,x')$ between any two distinct points $x$ and $x'$ cannot be smaller than the separation [cf. \eqref{eqn_def_separation_network}] of the network
\begin{equation}\label{eqn_mlc_lowerbounds_ultrametric_alternative}
    u_X(x,x') \geq \sep(X,A_X) \quad \quad \forall \quad x \neq x'.
\end{equation} \end{indentedparagraph}

\noindent Observe that the Alternative Property of Influence (P1') coincides with the Symmetric Property of Influence (Q1) defined in Section \ref{secsymmetric_networks}. This is not surprising because for symmetric networks the Axiom of Value (A1) and the Alternative Axiom of Value (A1'') impose identical restrictions. Moreover, since the separation of a network cannot be larger than its minimum loop cost, the Alternative Property of Influence (P1') is implied by the (regular) Property of Influence (P1), but not vice versa.

The Alternative Property of Influence (P1') states that no clusters are formed at resolutions at which there are no unidirectional influences between any pair of nodes and is consistent with the Alternative Axiom of Value (A1''). Moreover, in studying methods admissible with respect to (A1'') and (A2), (P1') plays a role akin to the one played by (P1) when studying methods that are admissible with respect to (A1) and (A2). In particular, as (P1) is implied by (A1) and (A2), (P1') is true if (A1'') and (A2) hold as we assert in the following theorem.

%
\begin{theorem}\label{theo_alternative_influence}
If a clustering method $\ccalH$ satisfies the Alternative Axiom of Value (A1'') and the Axiom of Transformation (A2) then it also satisfies the Alternative Property of Influence (P1').
\end{theorem}

\begin{myproofnoname}
See Appendix \ref{appendix_sec_alternative_axioms}.
\end{myproofnoname}

Theorem \ref{theo_alternative_influence} admits the following interpretation. In (A1'') we require two-node networks to cluster at the resolution where unidirectional influence occurs. When we consider (A1'') in conjunction with (A2) we can translate this requirement into a statement about  clustering in arbitrary networks. Such requirement is the Alternative Property of Influence (P1') which prevents nodes to cluster at resolutions at which each node in the network is disconnected from the rest.

%

%
\subsection{Unilateral clustering}\label{sec_unilateral_clustering}

Mimicking the developments in sections \ref{sec_axioms}-\ref{sec_extremal_ultrametrics}, we move on to identify and define methods that satisfy axioms (A1'')-(A2) and then bound the range of admissible methods respect to these axioms. To do so let $N=(X, A_X)$ be a given network and consider the symmetric dissimilarity function 
\begin{equation}\label{eqn_unilateral_clustering_001} 
    \hat{A}_X(x, x') : = \min(A_X(x, x'),A_X(x', x)),
\end{equation} 
for all $x, x' \in X$. Notice that as opposed to the definition of $\bar{A}_X$, where the symmetrization is done by means of a $\max$ operation, $\hat{A}$ is defined by using a $\min$ operation.

We define the \emph{unilateral} clustering method $\ccalH^\U$ with output ultrametric $(X, u^\U_X)= \ccalH^\U(N)$, where $u^\U_X$ is defined as
\begin{equation}\label{eqn_unilateral_clustering_2} 
    u^{\U}_X(x,x') :=
       \min_{C(x,x')} \, \max_{i | x_i\in C(x,x')} \hat{A}_X(x_i,x_{i+1}),
\end{equation} 
for all $x, x' \in X$.
To show that $\ccalH^\U$ is a properly defined clustering method, we need to establish that $u_X^\U$ as defined in \eqref{eqn_unilateral_clustering_2} is a valid ultrametric. However, comparing \eqref{eqn_unilateral_clustering_2} and \eqref{eqn_single_linkage_ultrametric} we see that 
\begin{equation}\label{eqn_unilateral_as_single_linkage_hat}
\ccalH^\U(X, A_X) \equiv \ccalH^{\SL}(X, \hat{A}_X),
\end{equation}
i.e. applying the unilateral clustering method to an asymmetric network $(X, A_X)$ is equivalent to applying single linkage clustering method to the symmetrized network $(X, \hat{A}_X)$. Since we know that single linkage produces a valid ultrametric when applied to any symmetric network such as $(X, \hat{A}_X)$, \eqref{eqn_unilateral_clustering_2} is a properly defined ultrametric. Moreover, as an elaboration of the results in Section \ref{sec_algorithms}, from \eqref{eqn_unilateral_as_single_linkage_hat}, \eqref{eqn_algorithm_single_linkage}, and \eqref{eqn_unilateral_clustering_001} we obtain an algorithmic way of computing the unilateral ultrametric output for any network,
\begin{align}
   u^{\U}_X = \Big(\min\left( {A}_X, A_X^T \right)\Big)^{(n-1)}, \label{eqn_algo_unilateral} 
\end{align}
where the operation $(\cdot)^{(n-1)}$ denotes the $(n-1)$st matrix power in the dioid algebra $(\reals^+\cup\{+\infty\},\min,\max)$ with matrix product as defined in \eqref{def_star}. Furthermore, it can be shown that $\ccalH^\U$ satisfies axioms (A1'') and (A2).

%
\begin{proposition}\label{prop_unilateral_axioms}
The unilateral clustering method $\ccalH^\U$ with output ultrametrics defined in \eqref{eqn_unilateral_clustering_2} satisfies axioms (A1'') and (A2). 
\end{proposition}

%
\begin{myproofnoname}
See Appendix \ref{appendix_sec_alternative_axioms}.
\end{myproofnoname}

In the case of admissibility with respect to (A1) and (A2), in Section \ref{sec_intermediate_ultrametrics} we constructed an infinite number of clustering methods whose outcomes are uniformly bounded between those of nonreciprocal and reciprocal clustering as predicted by Theorem \ref{theo_extremal_ultrametrics}. In contrast, in the case of admissibility with respect to (A1'') and (A2), unilateral clustering is the \emph{unique} admissible method as stated in the following theorem.

%
\begin{theorem}\label{theo_unilateral_unicity}
Let $\ccalH$ be a hierarchical clustering method satisfying axioms (A1'') and (A2). Then, $\ccalH \equiv \ccalH^\U$ where $\ccalH^\U$ is the unilateral clustering method with output ultrametrics as in \eqref{eqn_unilateral_clustering_2}.
\end{theorem}

%
\begin{myproofnoname} 
See Appendix \ref{appendix_sec_alternative_axioms}.
\end{myproofnoname}

\vspace{-0.1in}
\begin{remark}\normalfont
By Theorem \ref{theo_unilateral_unicity}, the space of methods that satisfy the Alternative Axiom of Value (A1'') and the Axiom of Transformation (A2) is inherently simpler than the space of methods that satisfy the (regular) Axiom of value (A1) and the Axiom of Transformation (A2). 
\end{remark}

Further note that in the case of symmetric networks, for all $x, x' \in X$ we have $\hat{A}_X(x, x')=A_X(x, x')=A_X(x', x)$ [cf. \eqref{eqn_unilateral_clustering_001}] and as a consequence unilateral clustering is equivalent to single linkage as it follows from comparison of \eqref{eqn_single_linkage_ultrametric} and 
\eqref{eqn_unilateral_clustering_2}. Thus, the result in Theorem \ref{theo_unilateral_unicity} reduces to the statement in Corollary \ref{cor_single_linkage}, which was derived upon observing that in symmetric networks reciprocal and nonreciprocal clustering yield identical outcomes. The fact that reciprocal, nonreciprocal, and unilateral clustering all coalesce into single linkage when restricted to symmetric networks is consistent with the fact that the Axiom of Value (A1) and the Alternative Axiom of Value (A1'') are both equivalent to the Symmetric Axiom of Value (B1) when restricted to symmetric dissimilarities.

\vspace{-0.1in}
%
\subsection{Agnostic Axiom of Value}\label{sec_agnostic_axiom_of_value}

Axiom (A1) stipulates that every two-node network  $\vec{\Delta}_2(\alpha, \beta)$ is clustered into a single block at resolution $\max(\alpha,\beta)$, whereas Axiom (A1'') stipulates that they should be clustered at $\min(\alpha,\beta)$. One can also be agnostic with respect to this issue and say that both of these situations are admissible. An agnostic version of axioms (A1) and (A1'') is given next.

\begin{indentedparagraph}{(A1\emph{'''}) Agnostic Axiom of Value} The ultrametric output $(X,u_{p,q})=\ccalH(\vec{\Delta}_2(\alpha, \beta))$ produced by $\ccalH$ applied to the two-node network $\vec{\Delta}_2(\alpha, \beta)$ satisfies 
\begin{equation}\label{eqn_two_node_network_ultrametric_agnostic}
   \min(\alpha,\beta) \leq u_X(p,q) \leq \max(\alpha,\beta).
\end{equation}\end{indentedparagraph}

\noindent Since fulfillment of (A1) or (A1'') implies fulfillment of (A1'''), any admissible clustering method with respect to the original axioms (A1)-(A2) or with respect to the alternative axioms (A1'')-(A2) must be admissible with respect to the agnostic axioms (A1''')-(A2). In this sense, (A1''')-(A2) is the most general combination of axioms described in this paper. For methods that are admissible with respect to (A1''') and (A2) we can bound the range of  outcome ultrametrics as explained in the following theorem.

%
\begin{theorem}\label{theo_extremal_ultrametrics_2}
Consider a clustering method $\ccalH$ satisfying axioms (A1\emph{'''}) and (A2). For an arbitrary given network $N=(X,A_X)$ denote by $(X, u_X)=\ccalH(X,A_X) $ the outcome of $\ccalH$ applied to $N$. Then, for all pairs of nodes $x,x' \in X$
\begin{equation}\label{eqn_theo_extremal_ultrametrics_2} 
    u^{\U}_X(x,x') \leq  u_X(x,x') \leq  u^{\R}_X(x,x'),
\end{equation} 
where $u^{\U}_X(x,x')$ and $u^{\R}_X(x,x')$ denote the unilateral and reciprocal ultrametrics as defined by \eqref{eqn_unilateral_clustering_2} and \eqref{eqn_reciprocal_clustering}, respectively.
\end{theorem}

%
\begin{myproofnoname}
See Appendix \ref{appendix_sec_alternative_axioms}.
\end{myproofnoname}

By Theorem \ref{theo_extremal_ultrametrics_2}, given an asymmetric network $(X,A_X)$, any hierarchical clustering method abiding by axioms (A1''') and (A2) produces outputs contained between those corresponding to  two methods. The first method, unilateral clustering, symmetrizes $A_X$ by calculating $\hat{A}_X(x, x')=\min(A_X(x, x'), A_X(x',x))$ for all $x, x' \in X$ and computes single linkage on $(X,\hat{A}_X)$. The other method, reciprocal clustering, symmetrizes $A_X$ by calculating $\bbarA_X(x, x')=\max(A_X(x, x'), A_X(x', x))$ for all $x, x' \in X$ and computes single linkage on $(X,\bbarA_X)$. 


%
\section{Stability}\label{sec_stability}

The collection of all compact metric spaces modulo isometry becomes a metric space of its own when endowed with the Gromov-Hausdorff distance \cite[Chapter 7.3]{burago-book}. This distance has been proven very useful in studying the stability of different methods of data analysis \cite{metric-structures,dgh-focm,clust-um} and here we generalize it to the space of networks $\ccalN$ modulo a properly defined notion of isomorphism. For a given hierarchical clustering method $\ccalH$ we can then ask the question of whether networks that are close to each other result in dendrograms that are also close to each other. In analogy to the symmetric case \cite{clust-um}, the answer to this question is affirmative for semi-reciprocal methods -- of which reciprocal and nonreciprocal methods are particular cases --, and most other constructions introduced earlier, as we discuss in the following sections.

%
\vspace{-0.1in}
\subsection{Gromov-Hausdorff distance for asymmetric networks}\label{sub_sec_network_distance}

Relabeling the nodes of a given network $N_X=(X,A_X)$ results in a network $N_Y=(Y,A_Y)$ that is identical from the perspective of the dissimilarity relationships between nodes. To capture this notion formally, we say that $N_X$ and  $N_Y$ are \emph{isomorphic} whenever there exists a bijective map $\phi:X\to Y$ such that for all points $x,x'\in X$ we have
\begin{equation}\label{eqn_network_isomorphism}
   A_X(x,x')=A_Y(\phi(x),\phi(x')).
\end{equation}
When networks $N_X$ and $N_Y$ are isomorphic we write $N_X \cong N_Y$. The space of networks where all isomorphic networks are represented by a single point is called the space of networks modulo isomorphism and denoted as $\ccalN\mod\cong$.

To motivate the definition of a distance on the space $\ccalN\mod\cong$ of networks modulo isomorphism, we start by considering networks $N_X$ and $N_Y$ with the same number of nodes and assume that a bijective transformation $\phi:X\to Y$ is given. It is then natural to define the \emph{distortion} $\dis(\phi)$ of the map $\phi$ as 
\begin{equation}\label{eqn_infinity_norm_distance}
   \dis(\phi):= \max_{(x,x')}\big| A_X(x,x')-A_Y(\phi(x),\phi(x') \big|.
\end{equation}
Since different maps $\phi:X\to Y$ are possible, we further focus on those maps $\phi$ that makes the networks $N_X$ and $N_Y$ as similar as possible and define the distance $d_{\infty}$ between networks $N_X$ and $N_Y$ with the same cardinality as
\begin{align}\label{eqn_permuted_infinity_norm_distance}
   d_{\infty}(N_X,N_Y)
      := \frac{1}{2} \min_{\phi} \dis(\phi),
\end{align}
where the factor $1/2$ is added for consistency with the definition of the Gromov-Hausdorff distance for metric spaces \cite[Chapter 7.3]{burago-book}.
To generalize \eqref{eqn_permuted_infinity_norm_distance} to networks that may have different numbers of nodes we consider the notion of correspondence between node sets to take the role of the bijective transformation $\phi$ in \eqref{eqn_infinity_norm_distance} and \eqref{eqn_permuted_infinity_norm_distance}. More specifically, for node sets $X$ and $Y$ consider subsets $R\subseteq X\times Y$ of the Cartesian product space $X\times Y$ with elements $(x,y)\in R$. The set $R$ is a \emph{correspondence} between $X$ and $Y$ if for all $x_0\in X$ we have at least one element $(x_0,y)\in R$ whose first component is $x_0$, and for all $y_0\in Y$ we have at least one element $(x,y_0)\in R$ whose second component is $y_0$. The distortion of the correspondence $R$ is defined as
\begin{equation}\label{eqn_correspondence_distance}
   \dis(R)
       :=\max_{(x,y),(x',y')\in R}
             \big|A_X(x,x')-A_Y(y,y')\big|.
\end{equation}
In a correspondence $R$ all the elements of $X$ are paired with some point in $Y$ and, conversely, all the elements of $Y$ are paired with some point in $X$. We can then think of $R$ as a mechanism to superimpose the node spaces on top of each other so that no points are orphaned in either $X$ or $Y$. As we did in going from \eqref{eqn_infinity_norm_distance} to \eqref{eqn_permuted_infinity_norm_distance} we now define the distance between networks $N_X$ and $N_Y$ as the distortion associated with the correspondence $R$ that makes $N_X$ and $N_Y$ as close as possible,
\begin{align}\label{eqn_gh_distance}
   d_\ccalN(N_X,N_Y)
       := &\frac{1}{2} \min_{R} \dis(R)\\\nonumber
        = &\frac{1}{2}\min_{R}\!\max_{(x,y),(x',y')\in R}
               \big|A_X(x,x')-A_Y(y,y')\big|.
\end{align}
Notice that \eqref{eqn_gh_distance} does not necessarily reduce to \eqref{eqn_permuted_infinity_norm_distance} when the networks have the same number of nodes. Since for networks $N_X, N_Y$ with $|X| = |Y|$, correspondences are more general than bijective maps there may be a correspondence $R$ that results in a distance $d_\ccalN(N_X,N_Y)$ smaller than the distance $d_{\infty}(N_X,N_Y)$. 

The definition in \eqref{eqn_gh_distance} is a verbatim generalization of the Gromov-Hausdorff distance in \cite[Theorem 7.3.25]{burago-book} except that the dissimilarity functions $A_X$ and $A_Y$ are not restricted to be metrics. It is legitimate to ask whether the relaxation of this condition renders $d_\ccalN(N_X,N_Y)$ in \eqref{eqn_gh_distance} an invalid metric. We prove in the following theorem that this is not the case since $d_\ccalN(N_X,N_Y)$ becomes a legitimate metric in the space $\ccalN\mod\cong$ of networks modulo isomorphism.

%
\begin{theorem}\label{theo_gromov_hausdorff}
The function $d_\ccalN: \ccalN \times \ccalN \to \reals_+$ defined in \eqref{eqn_gh_distance} is a metric on the space $\ccalN\mod\cong$ of networks modulo isomorphism. I.e., for all networks $N_X,N_Y,N_Z\in\ccalN$, $d_\ccalN$ satisfies the following properties:
\begin{mylist}
\item[Nonnegativity:] $d_\ccalN(N_X,N_Y)\geq0$.
\item[Symmetry:] $d_\ccalN(N_X,N_Y)= d_\ccalN(N_Y,N_X)$.
\item[Identity:]  $d_\ccalN(N_X,N_Y)=0$ if and only if $N_X \cong N_Y$.
\item[Triangle ineq.:] $\! d_\ccalN(\!N_X,N_Y\!)\!\leq \! d_\ccalN(\!N_X,N_Z\!)+d_\ccalN(\!N_Z,N_Y\!)$.
\end{mylist} \end{theorem}
\vspace{-0.1in}
\begin{myproofnoname}
See Appendix \ref{appendix_sec_stability}.
\end{myproofnoname}

%
The guarantee offered by Theorem \ref{theo_gromov_hausdorff} entails that the space $\ccalN\mod\cong$ of networks modulo isomorphism endowed with the distance defined in \eqref{eqn_gh_distance} is a metric space. Restriction of \eqref{eqn_gh_distance} to symmetric networks shows that the space $\ccalM\mod\cong$ of symmetric networks [cf. Section \ref{secsymmetric_networks}] modulo isomorphism is also a metric space. Further restriction to metric spaces shows that the space of finite metric spaces modulo isomorphism is properly metric \cite[Chapter 7.3]{burago-book}. A final restriction of \eqref{eqn_gh_distance} to finite ultrametric spaces shows that the space $\ccalU\mod\cong$ of ultrametrics modulo isomorphism is a metric space. As implemented in \cite{clust-um}, having a properly defined metric to measure distances between networks $\ccalN$ and therefore also between ultrametrics $\ccalU \subset \ccalN$ permits the study of stability of hierarchical clustering methods for asymmetric networks that we undertake in the following section.

%
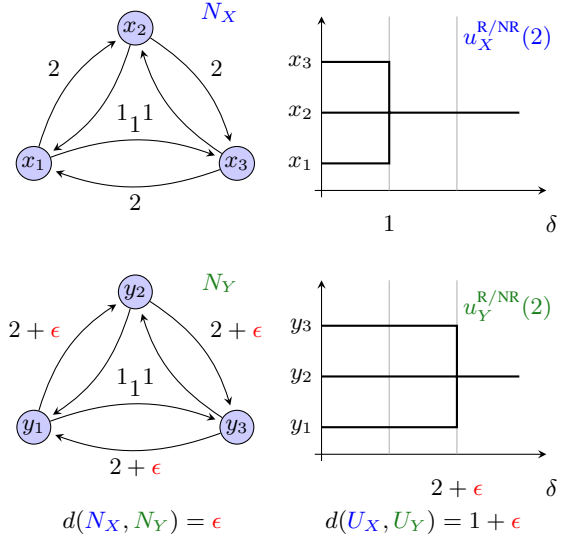
\begin{figure}
\centering
\def \thisplotscale {0.45}
\def \unit {\thisplotscale cm}

{\small
\begin{tikzpicture}[-stealth, shorten >=2, scale = \thisplotscale]

    \node [blue vertex] at (1,3) (1) {$x_2$};
    \node [blue vertex] at (4,-1) (2) {$x_3$};    
    \node [blue vertex] at (-2,-1) (3) {$x_1$};

    \path (1) edge [bend left=20, above right] node {$2$} (2);	
    \path (2) edge [bend left=20, below] node {$2$} (3);
    \path (3) edge [bend left=20, above left] node {$2$} (1);    	

    \path (2) edge [bend left=20, below left, pos=0.6] node {$1$} (1);	
    \path (3) edge [bend left=20, above] node {$1$} (2);
    \path (1) edge [bend left=20, below right, , pos=0.4]  node {$1$} (3);    	
    
    
        \node [blue vertex] at (1,-4.8) (4) {$y_2$};
    \node [blue vertex] at (4,-8.8) (5) {$y_3$};    
    \node [blue vertex] at (-2,-8.8) (6) {$y_1$};

    \path (4) edge [bend left=20, above right] node {$2+\red{\epsilon}$} (5);	
    \path (5) edge [bend left=20, below] node {$2+\red{\epsilon}$} (6);
    \path (6) edge [bend left=20, above left] node {$2+\red{\epsilon}$} (4);    	

    \path (5) edge [bend left=20, below left, pos=0.6] node {$1$} (4);	
    \path (6) edge [bend left=20, above] node {$1$} (5);
    \path (4) edge [bend left=20, below right, , pos=0.4]  node {$1$} (6);

   
   \draw [-stealth] (6.5,-2) -- (6.5,3.5);
    \draw [-stealth] (6.3,-1.8) -- (13.3,-1.8);
    \draw [-, draw=black!30] (8.5,-1.8) -- (8.5,3.5);
      \draw [-, draw=black!30] (10.5,-1.8) -- (10.5,3.5);
    
   \draw[thick, -] (6.5, -1) -- ++(2,0) -- ++(0,1.5) -- ++(-2,0) ++(0,1.5) -- ++(2,0) -- ++(0,-1.5) -- ++(4,0);

    \node [below] at (13.3,-2.3) {$\delta$};
    \node [below] at (8.5,-2.3) {$1$};
    \node [left] at (6.5,-1) {$x_1$};
    \node [left] at (6.5,0.5) {$x_2$};
    \node [left] at (6.5,2) {$x_3$};
    
    \node [below] at (12,3.5) {$\blue{u_X^{\R/\NR}(2)}$};
    
    \node [below] at (3.5,4) {$\blue{N_X}$};
    

       \draw [-stealth] (6.5,-9.8) -- (6.5,-4.3);
    \draw [-stealth] (6.3,-9.6) -- (13.3,-9.6);
    \draw [-, draw=black!30] (8.5,-9.6) -- (8.5,-4.3);
      \draw [-, draw=black!30] (10.5,-9.6) -- (10.5,-4.3);
    
   \draw[thick, -] (6.5, -8.8) -- ++(4,0) -- ++(0,1.5) -- ++(-4,0) ++(0,1.5) -- ++(4,0) -- ++(0,-1.5) -- ++(2,0);

    \node [below] at (13.3,-10.1) {$\delta$};
    \node [below] at (10.5,-10.1) {$2+\red{\epsilon}$};
    \node [left] at (6.5,-8.8) {$y_1$};
    \node [left] at (6.5,-7.3) {$y_2$};
    \node [left] at (6.5,-5.8) {$y_3$};
    
    \node [below] at (12,-4.5) {$\green{u_Y^{\R/\NR}(2)}$};
    
    \node [below] at (3.5,-4) {$\green{N_Y}$};
    
    \node [below] at (1.2,-11) {$d(\blue{N_X}, \green{N_Y})=\red{\epsilon}$};
     \node [below] at (9.5,-11) {$d(\blue{U_X}, \green{U_Y})=1+\red{\epsilon}$};
    
    

\end{tikzpicture}
}
\vspace{-0.1in}
\caption{Instability of the method $\ccalH^{\R/\NR}(2)$. Some dissimilarities in the network $N_X$ are perturbed by an arbitrarily small $\epsilon$ to obtain $N_Y$ such that the distance between both networks is $\epsilon$. However, the distance between the output ultrametrics cannot be bounded by a multiple of $\epsilon$, violating the definition of stability \eqref{eqn_stability_definition}.}
\vspace{-0.1in}
\label{fig_stability_counter_example}
\end{figure}

%
\subsection{Stability of clustering methods}\label{sec_ultrametric_stability}

Intuitively, a hierarchical clustering method $\ccalH$ is stable if its application to networks that have small distance between each other results in dendrograms that are close to each other. Formally, we require the distance between output ultrametrics to be bounded by the distance between the original networks as we define next.

\begin{indentedparagraph}{(P2) Stability} We say that the clustering method $\ccalH: \ccalN \to \ccalU$ is \emph{stable} if 
\begin{equation}\label{eqn_stability_definition}
   d_\ccalN(\ccalH(N_X), \ccalH(N_Y)) \leq d_\ccalN(N_X, N_Y),
\end{equation}
for all $N_X, N_Y \in \ccalN$. \end{indentedparagraph}

\begin{remark}\normalfont
Note that our definition of a stable hierarchical clustering method $\ccalH$ coincides with the property of $\ccalH: (\ccalN, d_\ccalN) \to (\ccalU, d_{\ccalN}| _ { \ccalU \times \ccalU})$ being a 1-Lipschitz map between the metric spaces $(\ccalN, d_\ccalN)$ and $(\ccalU, d_{\ccalN}| _ { \ccalU \times \ccalU})$.
\end{remark}

\noindent Recalling that the space of ultrametrics $\ccalU$ is included in the space of networks $\ccalN$, the distance $d_\ccalN(\ccalH(N_X), \ccalH(N_Y))$ in \eqref{eqn_stability_definition} is well defined and endows $\ccalU$ with a metric by Theorem \ref{theo_gromov_hausdorff}. The relationship in \eqref{eqn_stability_definition} means that a stable hierarchical clustering method is a non-expansive map from the metric space of networks endowed with the distance defined in \eqref{eqn_gh_distance} into itself. A particular consequence of \eqref{eqn_stability_definition} is that if networks $N_X$ and $N_Y$ are at small distance $d_\ccalN(N_X, N_Y)\leq\epsilon$ of each other, the output ultrametrics of the stable method $\ccalH$ are also at small distance of each other $ d_\ccalN(\ccalH(N_X), \ccalH(N_Y)) \leq d_\ccalN(N_X, N_Y) \leq \epsilon$. This latter observation formalizes the idea that nearby networks yield nearby dendrograms when processed with a stable hierarchical clustering method.

Notice that the stability definition in (P2) extends to the hierarchical quasi-clustering methods introduced in Section \ref{sec_full_characterization_asymmetric}, since the space of quasi-ultrametric networks, just like the space of ultrametric networks, is a subset of the space of asymmetric networks. Thus, we begin by showing the stability of the directed single linkage quasi-clustering method $\tilde{\ccalH}^*$. The reason to start the analysis with $\tilde{\ccalH}^*$ is that the proof of the following theorem can be used to simplify the proof of stability of other clustering methods. Theorem \ref{theo_stab_directed_single_linkage} below is a generalization of \cite[Proposition 26]{clust-um}.

%
\begin{theorem}\label{theo_stab_directed_single_linkage}
The directed single linkage quasi-clustering method $\tilde{\ccalH}^*$ with outcome quasi-ultrametrics as defined in \eqref{eqn_nonreciprocal_chains} is stable in the sense of property (P2).
\end{theorem}
\begin{myproofnoname}
Given two arbitrary networks $N_X=(X, A_X)$ and $N_Y=(Y, A_Y)$, assume $\eta=d_{\mathcal{N}}(N_X,N_Y)$ and let $R$ be a correspondence between $X$ and $Y$ such that $\mathrm{dis}(R)=2\eta$. Write $(X,\tilde{u}_X)= \tilde{\ccalH}^*(N_X)$ and  $(Y,\tilde{u}_Y)= \tilde{\ccalH}^*(N_Y)$. Fix $(x,y)$ and $(x',y')$ in $R$. Pick any $x=x_0,x_1,\ldots,x_n=x'$ in $X$ such that $\max_{i}A_X(x_i,x_{i+1})=\tilde{u}_X(x,x').$ Choose $y_0,y_1,\ldots,y_n\in Y$ so that $(x_i,y_i)\in R$ for all $i=0,1,\ldots,n.$ Then, by definition of $\tilde{u}_Y(y,y')$ in \eqref{eqn_nonreciprocal_chains} and the definition of $\eta$ in \eqref{eqn_gh_distance}:
\begin{align}\label{eqn:proof_theo_stab_directed_single_linkage_010}
\tilde{u}_Y(y,y')&\leq \max_iA_Y(y_i,y_{i+1})\nonumber \\
&\leq \max_i A_X(x_i,x_{i+1})+2\eta =\tilde{u}_X(x,x')+ 2\eta.
\end{align}
By symmetry, one also obtains $\tilde{u}_X(x,x')\leq \tilde{u}_Y(y,y')+2\eta$, which combined with \eqref{eqn:proof_theo_stab_directed_single_linkage_010} implies that
\begin{align}\label{eqn:proof_theo_stab_directed_single_linkage_020}
|\tilde{u}_X(x,x') - \tilde{u}_Y(y,y')| \leq 2\eta.
\end{align}
Since this is true for arbitrary pairs $(x,y)$ and $(x',y') \in R$, it must also be true for the maximum as well. Moreover, $R$ need not be the minimizing correspondence for the distance between the networks $(X, \tilde{u}_X)$ and $(Y, \tilde{u}_Y)$. However, it suffices to obtain an upper bound implying that
\begin{align}\label{eqn:proof_theo_stab_directed_single_linkage_030}
d_\ccalN((X, \tilde{u}_X), (Y, \tilde{u}_Y))  & \leq \frac{1}{2} \max_{(x,y),(x',y')\in R}|\tilde{u}_X(x,x')-\tilde{u}_Y(y,y')| \nonumber \\
 & \leq \eta = d_\ccalN(N_X, N_Y), 
\end{align}
concluding the proof.
\end{myproofnoname}

Moving into the realm of clustering methods, we show that semi-reciprocal methods $\ccalH^{\SR(t)}$ are stable in the sense of property (P2) in the following theorem.

%
\begin{theorem}\label{theo_stab_rec}
The semi-reciprocal clustering method $\ccalH^{\SR(t)}$ with outcome ultrametrics as defined in \eqref{eqn_inter_reciprocal_clustering} is stable in the sense of property (P2) for every integer $t \geq 2$.
\end{theorem}

The following lemma is used to prove Theorem \ref{theo_stab_rec}.

\begin{lemma}\label{lem:bounded_maximum}
Given $a, \bar{a}, b, \bar{b}, c \in \reals_+$ such that $|a-b| \leq c$ and $|\bar{a}-\bar{b}| \leq c$, then $|\max(a, \bar{a}) - \max(b, \bar{b})| \leq c$.
\end{lemma}
\begin{myproofnoname}
Begin by noticing that 
\begin{equation}\label{eqn:proof_lemma_bounded_maximum_010}
a = |a - b +b| \leq |a-b|+|b| = |a-b|+b,
\end{equation}
and similarly for $\bar{a}$ and $\bar{b}$. Thus, we may write
\begin{equation}\label{eqn:proof_lemma_bounded_maximum_020}
\max(a, \bar{a}) \leq \max(|a-b|+b, |\bar{a}-\bar{b}|+\bar{b}).
\end{equation}
By using the bounds assumed in the statement of the lemma, we obtain
\begin{equation}\label{eqn:proof_lemma_bounded_maximum_030}
\max(a, \bar{a}) \leq \max(c+b, c+\bar{b}) = c+ \max(b, \bar{b}).
\end{equation}
By applying the same reasoning but starting with $\max(b, \bar{b})$, we obtain that
\begin{equation}\label{eqn:proof_lemma_bounded_maximum_040}
\max(b, \bar{b}) \leq c+ \max(a, \bar{a}).
\end{equation}
Finally, by combining \eqref{eqn:proof_lemma_bounded_maximum_030} and \eqref{eqn:proof_lemma_bounded_maximum_040} we obtain the result stated in the lemma.
\end{myproofnoname}

%
\begin{myproof}[of Theorem \ref{theo_stab_rec}]
Here we present the proof for $t=2$ in order to illustrate the main conceptual steps, which are similar to those in the proof of Proposition 26 of \cite{clust-um}. The general proof for any $t \geq 2$ can be found in Appendix \ref{appendix_sec_stability}. Recall that from \eqref{eqn_inter_reciprocal_clustering_1}, we know that $\ccalH^{\SR(2)} \equiv \ccalH^{\R}$.
Given two networks $N_X=(X, A_X)$ and $N_Y=(Y, A_Y)$ denote by $(X, u^\R_X)=\ccalH^\R(N_X)$ and $(X, u^\R_Y)=\ccalH^\R(N_Y)$ the outputs of applying the reciprocal clustering method to such networks. Let $\eta = d_\ccalN(N_X, N_Y)$ be the distance between $N_X$ and $N_Y$ as defined by \eqref{eqn_gh_distance} and $R$ be an associated minimizing correspondence such that
\begin{equation}\label{abs_dif}
|A_X(x,x')-A_Y(y,y')| \leq 2\eta,
\end{equation}
for all $(x,y)$,$(x',y') \in R$. By reversing the order of $(x, y)$ and $(x', y')$ we obtain that
\begin{equation}\label{abs_dif_2}
|A_X(x',x)-A_Y(y',y)| \leq 2\eta.
\end{equation}
From \eqref{abs_dif}, \eqref{abs_dif_2}, and the definition $\bar{A}_X(x, x') = \max(A_X(x, x'),A_X(x', x))$ for all $x, x' \in X$, we obtain from Lemma \ref{lem:bounded_maximum} that 
\begin{equation}\label{abs_dif_3}
|\bar{A}_X(x,x')-\bar{A}_Y(y,y')| \leq 2\eta,
\end{equation}
for all $(x,y)$,$(x',y') \in R$. By using the same argument applied in the proof of Theorem \ref{theo_stab_directed_single_linkage} to go from \eqref{eqn:proof_theo_stab_directed_single_linkage_020} to \eqref{eqn:proof_theo_stab_directed_single_linkage_030}, we obtain that
\begin{align}\label{eq:theo_stab_rec_010}
d_\ccalN&((X, \bar{A}_X), (Y, \bar{A}_Y))  \leq d_\ccalN(N_X, N_Y). 
\end{align}
By comparing \eqref{eqn_nonreciprocal_chains} with \eqref{eqn_reciprocal_clustering} (or equivalently in terms of algorithms by comparing \eqref{eqn_algo_quasi_ultrametric} with \eqref{eqn_algo_recip}) it follows that 
\begin{align}\label{eq:theo_stab_rec_020}
(X, u^\R_X) = \tilde{\ccalH}^*(X, \bar{A}_X),
\end{align}
and similarly for $(Y, u^\R_Y)$. However, since $\tilde{\ccalH}^*$ is stable from Theorem \ref{theo_stab_directed_single_linkage}, we obtain that
\begin{align}\label{eq:theo_stab_rec_030}
d_\ccalN((X, u^\R_X),(Y, u^\R_Y)) \leq  d_\ccalN((X, \bar{A}_X), (Y, \bar{A}_Y)),
\end{align}
which combined with \eqref{eq:theo_stab_rec_010} completes the proof.
\end{myproof}


Reciprocal clustering is a particular case of semi-reciprocal clustering for $t=2$. Moreover, given any network, nonreciprocal clustering behaves as a semi-reciprocal clustering for some big enough $t$ which, by Theorem \ref{theo_stab_rec} is a stable method. It thus follows that these two methods are stable. This result is of sufficient merit so as to be stated separately in the following corollary.

%
\begin{corollary}\label{prop_stab_rnr}
The reciprocal and nonreciprocal clustering methods $\ccalH^{\R}$ and $\ccalH^{\NR}$ with output ultrametrics given as in \eqref{eqn_reciprocal_clustering} and \eqref{eqn_nonreciprocal_clustering}, respectively, are stable in the sense of property (P2).
\end{corollary}

%

%
By \eqref{eqn_stability_definition}, Theorem \ref{theo_stab_rec} shows that semi-reciprocal clustering methods -- subsuming the particular cases of reciprocal and nonreciprocal clustering -- do not expand distances between pairs of input and their corresponding output networks. In particular, for any method of the above, nearby networks yield nearby dendrograms. This is important when we consider noisy dissimilarity data. Property (P2) ensures that noise has limited effect on output dendrograms. 
\begin{remark} \normalfont Theorem \ref{theo_stab_rec} notwithstanding, not all methods that are admissible with respect to axioms (A1) and (A2) are stable. For example, the admissible grafting method $\ccalH^{\R/\NR}(\beta)$ introduced in Section \ref{sec_grafting} does not abide by (P2). To see this fix $\beta=2$ and turn attention to the networks $N_X$ and $N_Y$ shown in Fig. \ref{fig_stability_counter_example}, where $\epsilon>0$. For network $N_X$ we have $u^{\NR}_X(x, x')=1$ and $u^{\R}_X(x, x')=2$ for all pairs $x, x'$. Since $u^{\R}_X(x, x') = \beta=2$ for all $x, x'$, the top condition in definition \eqref{def_mu_beta_1} is active and we have $u^{\R/\NR}_X(x, x';2)=u^{\NR}_X(x, x')=1$ leading to the top dendrogram in Fig. \ref{fig_stability_counter_example}. For the network $N_Y$ we have that $u^{\R}_Y(y, y')=2+\epsilon > 2 = \beta$ for all $y, y'$. Thus, the bottom condition in definition \eqref{def_mu_beta_1} is active and we have $u^{\R/\NR}_Y(y,y';2)=u^{\R}_Y(y, y')=2+\epsilon$ for all $y,y'$. Given the symmetry in the original network and the output ultrametrics, the correspondence $R$ with $(x_i, y_i) \in R$ for $i=1,2,3$ is an optimal correspondence in the definition in \eqref{eqn_gh_distance}. It then follows that
\begin{align}\label{eqn_grafting_not_stable}
   d_\ccalN(\ccalH^{\R/\NR}(N_X;2), &\ccalH^{\R/\NR}(N_Y;2)) \nonumber \\
   &= 1+\epsilon >  d_\ccalN(N_X, N_Y)= \epsilon.
\end{align}
Comparing \eqref{eqn_grafting_not_stable} with \eqref{eqn_stability_definition} we conclude that methods in the grafting family $\ccalH^{\R/\NR}(\beta)$ are in general not stable in the sense of property (P2). This observations concurs with our intuition on instability. A small perturbation in the original data results in a large variation in the output ultrametrics. The discontinuity in the grafting method $\ccalH^{\R/\NR}(\beta)$ arises due to the switching between reciprocal and nonreciprocal ultrametrics implied by \eqref{def_mu_beta_1}. 
\end{remark}
%
\begin{remark}\normalfont
The same tools used in the proofs of theorems \ref{theo_stab_directed_single_linkage} and \ref{theo_stab_rec} can be used to show that the unilateral clustering method $\ccalH^U$ introduced in Section \ref{sec_unilateral_clustering} is stable. Moreover, convex combination methods introduced in Section \ref{sec_convex_comb} need not be stable in general, even when the methods combined are stable. Nevertheless, it can be shown that the combination of any two of the stable methods described in this section is also stable. However, the respective proofs are omitted to avoid repetition.
\end{remark}


%
\section{Applications}\label{sec_numerical_experiments}

We apply the hierarchical clustering and quasi-clustering methods developed throughout the paper to determine dendrograms and quasi-dendrograms for two asymmetric network datasets. In Section \ref{sec_state_to_state_migration} we analyze the internal migration network between states of the United States (U.S.) for the year 2011. In Section \ref{sec_economic_sectors} we analyze the network of interactions between sectors of the U.S. economy for the same year.

\subsection{Internal migration between states of the United States}\label{sec_state_to_state_migration}

The number of migrants from state to state, including the District of Columbia (DC) as a separate entity, is published yearly by the geographical mobility section of the U.S. census bureau \cite{USmigration}. We denote by $S$, with cardinality $|S|=51$, the set containing every state plus DC and as $M: S \times S \to \reals_+ \cup \{+\infty\}$ the migration flow similarity function given by the U.S. census bureau in which $M(s, s')$ is the number of individuals that migrated from state $s$ to state $s'$ and $M(s, s)=+\infty$ for all $s, s' \in S$. We then construct the asymmetric network $N_S=(S, A_S)$ with node set $S$ and dissimilarities $A_S$ such that $A_S(s, s)=0$ for all $s \in S$ and
\begin{equation}\label{eqn_def_migration_dissimilarity}
   A_S(s, s') := f \left( \frac{M(s, s')}{\sum_t M(t, s')}\right),
\end{equation}
for all $s \neq s' \in S$ where $f: [0, 1) \to \reals_{++}$ is a given decreasing function (to be specified below). The normalization $M(s, s')/\sum_t M(t, s')$ in \eqref{eqn_def_migration_dissimilarity} can be interpreted as the probability that an immigrant to state $s'$ comes from state $s$. The role of the decreasing function $f$ is to transform the similarities $M(s, s')/\sum_t M(t, s')$ into corresponding dissimilarities. For the experiments here we use $f(x) = 1-x$. Dissimilarities $A_S(s, s')$ focus attention on the composition of migration flows rather than on their magnitude. A small dissimilarity from state $s$ to state $s'$ implies that from all the immigrants into $s'$ a high percentage comes from $s$. E.g., if 85\% of the immigration into $s'$ comes from $s$, then $A_S(s, s')=1-0.85=0.15$. The application of hierarchical clustering to migration data has been extensively investigated by Slater, see \cite{slater1976hierarchical,slater1984partial}.

%
\begin{figure*}
\centering
\input{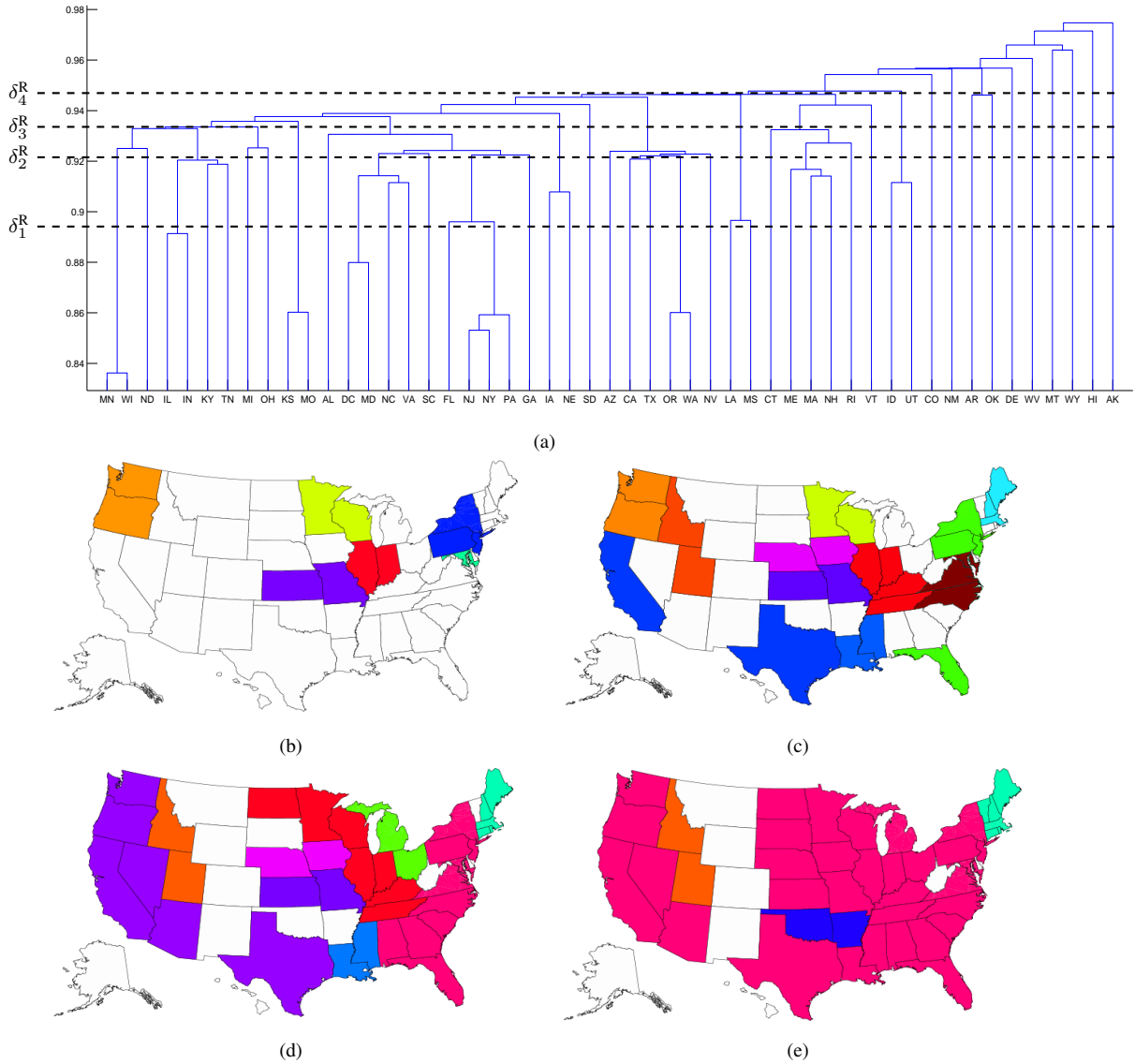}
\vspace{-0.1in}
\caption{(a) Reciprocal dendrogram. Output of clustering method $\ccalH^\R$ when applied to the migration network $N_S$. (b) Clusters at resolution $\delta^\R_1$. States that share urban metropolitan areas merge together first. States in white form singleton clusters at this resolution. (c) Clusters at resolution $\delta^\R_2$. Clusters are highly determined by geographical proximity except for Texas and Florida. (d) Clusters at resolution $\delta^\R_3$. The two coasts form separate clusters. (e) Clusters at resolution $\delta^\R_4$. Most of the nation forms a single cluster. Observe New England's relative isolation.}
\vspace{-0.1in}
\label{fig_reciprocal_example_us}
\end{figure*}

\subsubsection{Reciprocal clustering $\ccalH^\R$}\label{sec_reciprocal_migration}

The outcome of applying the reciprocal clustering method $\ccalH^\R$ defined in \eqref{eqn_reciprocal_clustering} to the migration network $N_S$ was computed with the algorithmic formula in \eqref{eqn_algo_recip}. The resulting output dendrogram is shown in Fig. \ref{fig_reciprocal_example_us}-(a). Figs. \ref{fig_reciprocal_example_us}-(b) through \ref{fig_reciprocal_example_us}-(e) illustrate the partitions that are obtained at four representative resolutions $\delta^\R_1=0.895$, $\delta^\R_2=0.921$, $\delta^\R_3=0.933$, and $\delta^\R_4=0.947$. States marked with the same color other than white are co-clustered at the given resolution whereas states in white are singleton clusters. For a given $\delta$, states that are clustered together in partitions produced by $\ccalH^\R$ are those connected by a chain of intense bidirectional migration flows in the sense dictated by the resolution under consideration. 

The most definite pattern arising from Fig. \ref{fig_reciprocal_example_us} is that migration is highly correlated with geographical proximity. With the exceptions of California, Florida, and Texas that we discuss below, all states merge into clusters with other neighboring states. In particular, the first non singleton clusters to form are pairs of neighboring states that join together at resolutions smaller than $\delta^\R_1$ with the exception of one cluster formed by three states -- New York, New Jersey, and Pennsylvania -- as shown in Fig. \ref{fig_reciprocal_example_us}-(b). In ascending order of resolutions at which they are formed, these pairs are Minnesota and Wisconsin (green, at resolution $\delta=0.836$), Oregon and Washington (orange, at resolution $\delta=0.860$), Kansas and Missouri (purple, at resolution $\delta=0.860$), District of Columbia and Maryland (turquoise, at resolution $\delta=0.880$), as well as Illinois and Indiana (red, at resolution $\delta=0.891$). In the group of three states composed of New Jersey, New York, and Pennsylvania, we observe that New York and New Jersey form a cluster (blue) at a smaller resolution ($\delta=0.853$) than the one at which they merge with Pennsylvania ($\delta=0.859$). The formation of these clusters can be explained by the fact that these states share respective metropolitan areas. These areas are Minneapolis and Duluth for Minnesota and Wisconsin, Portland for Oregon and Washington, Kansas City for Kansas and Missouri, Washington for the District of Columbia and Maryland, Chicago for Illinois and Indiana, New York City for New York State and New Jersey, as well as Philadelphia for Pennsylvania and New Jersey. Even while crossing state lines, migration within shared metropolitan areas corresponds to people moving to different neighborhoods or suburbs and occurs frequently enough to suggest it is the reason behind the clusters formed at low resolutions in the reciprocal dendrogram.

As we continue to increase the resolution, clusters formed by pairs of neighboring states continue to appear and a few clusters with multiple states emerge. At resolution $\delta^\R_2$, shown in Fig. \ref{fig_reciprocal_example_us}-(c), clusters with two adjacent states include Louisiana and Mississippi, Iowa and Nebraska, and Idaho and Utah. Kentucky and Tennessee join Illinois and Indiana to form a midwestern cluster while Maine, Massachusetts, and New Hampshire form a cluster of New England states. The only two exceptions to geographic proximity appear at this resolution.  These exceptions are the merging of Florida into the northeastern cluster formed by New Jersey, Pennsylvania, and New York, due to its closeness with the latter, and the formation of a cluster made of California and Texas. This anomaly occurs among the four states with the most intense outgoing and incoming migration in the country during 2011. The data analyzed shows that people move from all over the United States to New York, California, Texas, and Florida. For instance, Texas has the lowest standard deviation in the proportion of immigrants from each other state indicating a homogenous migration flow from the whole country. Hence, the proportion of incoming migration from neighboring states is not as significant as for other states. E.g., only 19\% of the migration into California comes from its three neighboring states whereas for North Dakota, which also has three neighboring states, these provide 45\% of its immigration. Based on the data, we observe that New York, California, Texas, and Florida have a strong influence on the immigration into their neighboring states but, given the mechanics of $\ccalH^\R$, the lack of influence in the opposite direction is the reason why Texas joins California and Florida joins New York before forming a cluster with their neighbors. If we require only unidirectional influence as in Section \ref{sec_unilateral_migration}, then these four states first join their neighboring states as observed in Fig. \ref{fig_unilateral_example_us}.

Higher resolutions see the appearance of three regional clusters in the Atlantic Coast, Midwest, and New England, as well as a cluster composed of the West Coast states plus Texas. This is illustrated in Fig. \ref{fig_reciprocal_example_us}-(d) for resolution $\delta^\R_3$. This points towards the fact that people living in a coastal state have a preference to move within the same coast, that people in the midwest tend to stay in the midwest, and that New Englanders tend to stay in New England.

At larger resolutions states start collapsing into a single cluster. At resolution $\delta^\R_4$, shown in Fig. \ref{fig_reciprocal_example_us}-(e), all states except those in New England and the Mountain West, along with Alaska, Arkansas, Delaware, West Virginia, Hawaii, and Oklahoma are part of a single cluster. The New England cluster includes all six New England states which shows a remarkable degree of migrational isolation with respect to the rest of the country. This indicates that people living in New England tend to move within the region, that people outside New England rarely move into the area, or both. The same observation can be made of the pairs Arkansas-Oklahoma and Idaho-Utah. The latter could be partially attributed to the fact that Idaho and Utah are the two states with the highest percentage of mormon population in the country \cite{USreligion}. Four states in the Mountain West, New Mexico, Colorado, Wyoming, and Montana as well as Delaware, West Virginia, Hawaii and Alaska stay as singleton clusters. Hawaii and Alaska are respectively the next to last, and last state to merge with the rest of the nation further adding evidence to the correlation between geographical proximity and migration clustering.

%
\begin{figure*}
\hspace{-0.6in}
\includegraphics[width=1.15 \linewidth, height=0.33 \linewidth]
                {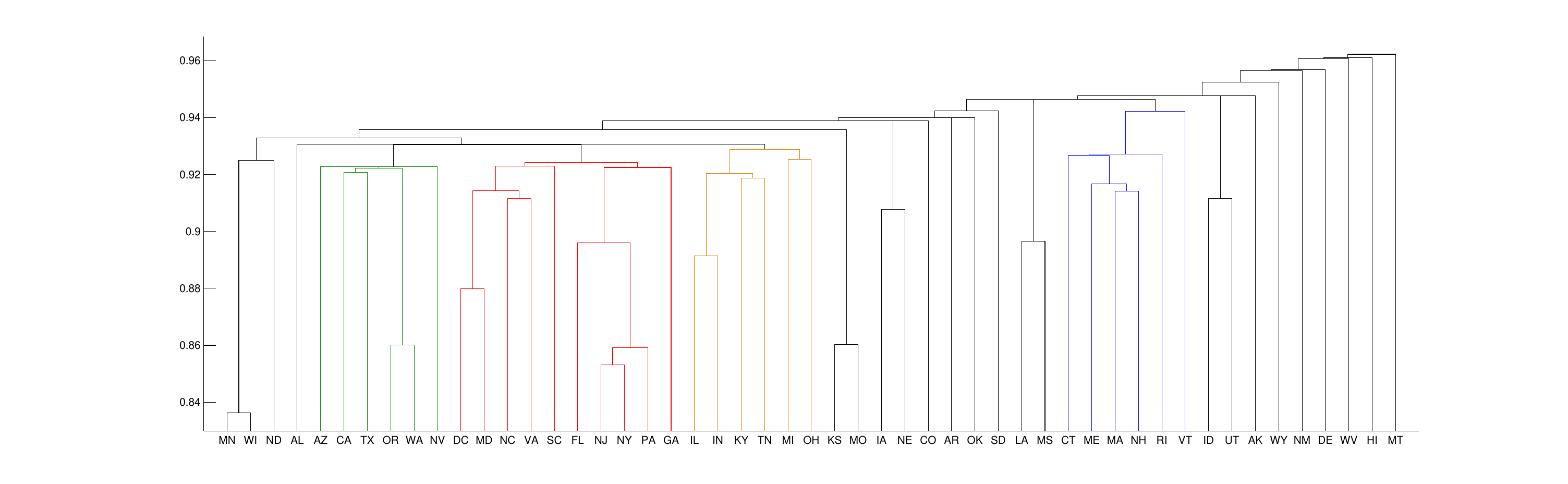}
                \vspace{-0.2in}
\caption{Nonreciprocal dendrogram. Dendrogram obtained when applying the nonreciprocal method $\ccalH^{\NR}$ to the state-to-state migration network $N_S$. The resemblance with the dendrogram in Fig. \ref{fig_reciprocal_example_us}(a) indicates that migration cycles are not ubiquitous.}
\vspace{-0.1in}
\label{fig_nonreciprocal_us_dendrogram}
\end{figure*}

\subsubsection{Nonreciprocal clustering $\ccalH^{\NR}$}\label{sec_nonreciprocal_migration}

The outcome of applying the nonreciprocal clustering method $\ccalH^\NR$ defined in \eqref{eqn_nonreciprocal_clustering} to the migration network $N_S$ is computed with the algorithmic formula in \eqref{eqn_algo_nonrecip}. The resulting output dendrogram is shown in Fig. \ref{fig_nonreciprocal_us_dendrogram}. Comparing the reciprocal and nonreciprocal dendrograms in figs. \ref{fig_reciprocal_example_us}-(a) and \ref{fig_nonreciprocal_us_dendrogram} shows that the nonreciprocal clustering method merges any pair of states into a common cluster at a resolution not higher than the resolution at which they are co-clustered by reciprocal clustering. This is as it should be because the uniform dominance of nonreciprocal ultrametrics by reciprocal  ultrametrics holds for all networks [cf. \eqref{eqn_nonreciprocal_smaller_than_reciprocal}]. E.g., for the reciprocal method, Colorado and Florida become part of the same cluster at resolution $\delta=0.954$ whereas for the nonreciprocal case they become part of the same cluster at resolution $\delta=0.939$. The nonreciprocal resolution need not be strictly smaller, for example, Illinois and Tennessee are merged by both clustering methods at a resolution $\delta=0.920$. 

Further observe that there are many striking similarities between the reciprocal and nonreciprocal dendrograms in figs. \ref{fig_reciprocal_example_us}-(a) and \ref{fig_nonreciprocal_us_dendrogram}. In both dendrograms, the first three clusters to emerge are the pair Minnesota and Wisconsin (at resolution $\delta=0.836$), followed by the pair New York and New Jersey (at resolution $\delta=0.853$) which are in turn co-clustered with Pennsylvania at resolution $\delta=0.859$. We then see the emergence of the four pairs: Oregon and Washington (at resolution $\delta=0.860$), Kansas and Missouri (at resolution $\delta=0.860$), District of Columbia and Maryland (at resolution $\delta=0.880$), and Illinois and Indiana (at resolution $\delta=0.891$). These are the same seven groupings and resolutions at which clusters form in the reciprocal dendrogram that we attributed to the existence of shared metropolitan areas spanning more than one state [cf. Fig. \ref{fig_reciprocal_example_us}-(b)].

Recall that the difference between the reciprocal and nonreciprocal clustering methods $\ccalH^\R$ and $\ccalH^\NR$ is that the latter allows influence to propagate through cycles whereas the former requires direct bidirectional influence for the formation of a cluster. In the particular case of the migration network $N_S$ this means that nonreciprocal clustering may be able to detect migration cycles of arbitrary length that are overlooked by reciprocal clustering. E.g., if people in state $A$ tend to move predominantly to $B$, people in $B$ to move predominantly to $C$, and people in $C$ move predominantly to $A$, nonreciprocal clustering merges these three states according to this migration cycle but reciprocal clustering does not. The overall similarity of the reciprocal and nonreciprocal dendrograms in figs. \ref{fig_reciprocal_example_us}-(a) and \ref{fig_nonreciprocal_us_dendrogram} suggests that migration cycles are rare in the United States. In particular, the formation of the seven clusters due to shared metropolitan areas indicates that the bidirectional migration flow between these pairs of states is higher than any migration cycle in the country. Notice that highly symmetric data would also correspond to similar reciprocal and nonreciprocal dendrograms. Nevertheless, another consequence of highly symmetric data would be to obtain a unilateral dendrogram similar to the reciprocal and the nonreciprocal ones. This is not the case, as can be seen in Section \ref{sec_unilateral_migration}, thus, symmetry cannot be the reason for the similarity observed between the reciprocal and nonreciprocal dendrograms.

However similar, the reciprocal and nonreciprocal dendrograms in figs. \ref{fig_reciprocal_example_us}-(a) and \ref{fig_nonreciprocal_us_dendrogram} are not identical. E.g., the last state to merge with the rest of the country in the reciprocal dendrogram is Alaska at resolution $\delta=0.975$ whereas the last state to merge in the
nonreciprocal dendrogram is Montana at resolution $\delta=0.962$ with Alaska joining the rest of the country at resolution $\delta=0.948$. Given the mechanics of $\ccalH^{\NR}$, this must occur due to the existence of a cycle of migration involving Alaska which is stronger than the bidirectional exchange between Alaska and any other state, and direct analysis of the data confirms this fact.

As we have argued, the areas of the country that cluster together when applying the nonreciprocal method are similar to the ones depicted in Fig. \ref{fig_reciprocal_example_us}-(d) for the reciprocal clustering method. When we cut the nonreciprocal dendrogram in Fig. \ref{fig_nonreciprocal_us_dendrogram} at resolution $\delta=0.930$, three major clusters arise -- highlighted in green, red, and orange in the dendrogram in Fig. \ref{fig_nonreciprocal_us_dendrogram}. The green cluster corresponds to the exact same block containing the West Coast plus Texas that arises in the reciprocal dendrogram and is depicted in purple in Fig. \ref{fig_reciprocal_example_us}-(d). The red cluster in the dendrogram corresponds to the East Coast cluster found with the reciprocal method with the exception that Alabama is not included. However, Alabama joins this block at a slightly higher resolution of $\delta=0.931$, coinciding with the merging of the green, red and orange clusters. The orange cluster in the nonreciprocal dendrogram corresponds to the Midwest cluster found in \ref{fig_reciprocal_example_us}-(d). However, in contrast with the reciprocal case, Michigan and Ohio join the Midwest cluster before Minnesota, Wisconsin and North Dakota. For the nonreciprocal case, these last three states join the main cluster at resolution $\delta=0.933$, after the East Coast, West Coast and Midwest become a single block. 

The migrational isolation of New England with respect to the rest of the country, which we observed in reciprocal clustering, also arises in the nonreciprocal case. The New England cluster is depicted in blue in the nonreciprocal dendrogram in Fig. \ref{fig_nonreciprocal_us_dendrogram} and joins the main cluster at a resolution of $\delta=0.946$, which coincides with the merging resolution for the reciprocal case. However, the order in which states become part of the New England cluster varies. In the nonreciprocal case, Connecticut merges with the cluster of Maine-Massachusetts-New Hampshire at resolution $\delta=0.926$ before Rhode Island which merges at resolution $\delta=0.927$. However, for the reciprocal case, Rhode Island still merges at the same resolution but Connecticut merges after this at a resolution $\delta=0.933$. The reason for this is that in the reciprocal case, the states of Connecticut and Rhode Island merge with the cluster Maine-Massachusetts-New Hampshire at the resolution where there exist bidirectional flows with the state of Massachusetts. In the nonreciprocal case, this same situation applies for Rhode Island, but from the data it can be inferred that Connecticut joins the mentioned cluster at a lower resolution due to a migration cycle composed of the chain $[$Connecticut, Maine, New Hampshire, Massachusetts, Connecticut$]$.

Up to this point we see that all the conclusions that we have extracted when applying $\ccalH^{\NR}$ are qualitatively similar to those obtained when applying $\ccalH^\R$. This is not surprising because the differences between the reciprocal and nonreciprocal dendrograms either occur at coarse resolutions or are relatively small. In fact, one should expect any conclusion stemming from the application of $\ccalH^\R$ and $\ccalH^\NR$ to the migration network $N_S$ to be qualitatively similar.

\subsubsection{Intermediate methods}\label{sec_intermediate_migration}

From Theorem \ref{theo_extremal_ultrametrics} we know that any clustering method satisfying the axioms of value and transformation applied to the migration network $N_S$ yields an outcome dendrogram such that the resolution at which any pair of states merge in a common cluster is bounded by the resolutions at which the same pair of states is co-clustered in the dendrograms resulting from application of the nonreciprocal and reciprocal clustering methods. Given the similar conclusions obtained upon analysis of the reciprocal and nonreciprocal clustering outputs we can assert that any other hierarchical clustering method satisfying the axioms of value and transformation would lead to similar conclusions. In particular, this is true for the intermediate methods described in Section \ref{sec_intermediate_ultrametrics} and the algorithmic intermediate of Section \ref{sec_algorithms}.

\subsubsection{Unilateral clustering $\ccalH^\U$}\label{sec_unilateral_migration}

%
\begin{figure*}
\centering
\centerline{\input{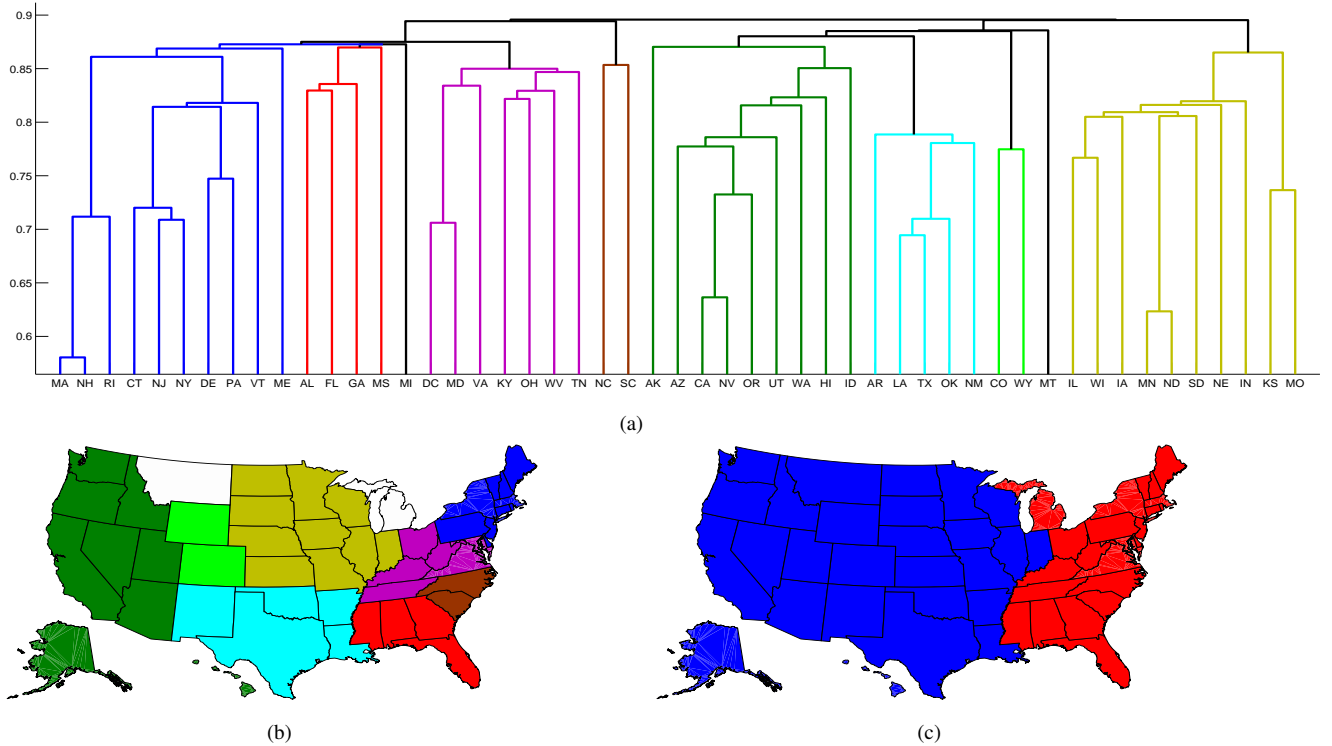} }
\vspace{-0.2in}
\caption{Unilateral clustering of state-to-state migration network. (a) Dendrogram output of applying the unilateral clustering method $\ccalH^\U$ to the network of state-to-state migration $N_S$. Clusters at resolution $\delta^\U_1=0.872$ are highlighted in color. (b) Highlighted clusters are identified in a map. Clusters tend to form around high populated states. (c) Map colored according to the partition at resolution $\delta^\U_2=0.896$. Two clear clusters, east and west, arise.}
\vspace{-0.2in}
\label{fig_unilateral_example_us}
\end{figure*}

The outcome of applying the unilateral clustering method $\ccalH^\U$ defined in \eqref{eqn_unilateral_clustering_2} to the migration network $N_S$ is computed with the algorithmic formula in \eqref{eqn_algo_unilateral}. The resulting output dendrogram is shown in Fig. \ref{fig_unilateral_example_us}-(a). The colors in the dendrogram correspond to the clusters formed at resolution $\delta^\U_1=0.872$ which are also shown in the map in Fig. \ref{fig_unilateral_example_us}-(b) with the same color code. States shown in black in Fig. \ref{fig_unilateral_example_us}-(a) and white in Fig. \ref{fig_unilateral_example_us}-(b) are singleton clusters at this resolution. In Fig. \ref{fig_unilateral_example_us}-(c) we show the two clusters that appear when the unilateral dendrogram is cut at resolution $\delta^\U_2=0.896$. States that are clustered together in unilateral partitions are those connected by a chain of intense unidirectional migration flows in the sense dictated by the resolution under consideration. 

In unilateral clustering, the relation between geographical proximity and tendency to form clusters is even more determinant than in reciprocal and nonreciprocal clustering [cf. sections \ref{sec_reciprocal_migration} and \ref{sec_nonreciprocal_migration}] since the exceptions of Texas, California, and Florida do not occur in this case. Indeed, California first merges with Nevada at resolution $\delta=0.637$, Texas with Louisiana at $\delta=0.694$, and Florida with Alabama at $\delta=0.830$, the three pairs of states being neighbors. Moreover, from Fig. \ref{fig_unilateral_example_us}-(b) it is immediate that at resolution $\delta^\U_1$ every non singleton cluster is formed by a set of neighboring states.  

Recall that unilateral clustering $\ccalH^\U$ abides by the alternative axioms of value and transformation (A1'')-(A2) in contrast to the (regular) axioms of value and transformation satisfied by reciprocal $\ccalH^\R$ and nonreciprocal $\ccalH^{\NR}$ clustering. Consequently, unidirectional influence is enough for the formation of a cluster. In the particular case of the migration network $N_S$ this means that unilateral clustering may detect one-way migration flows that are overlooked by reciprocal and nonreciprocal clustering. E.g., if people in state $A$ tend to move to $B$ but people in $B$ rarely move to $A$ either directly or through intermediate states, unilateral clustering merges these two states according to the one-way intense flow from $A$ to $B$ but reciprocal and nonreciprocal clustering do not. The differences between the unilateral dendrogram in Fig. \ref{fig_unilateral_example_us}-(a) with the reciprocal and nonreciprocal dendrograms in figs. \ref{fig_reciprocal_example_us}-(a) and \ref{fig_nonreciprocal_us_dendrogram} indicate that migration flows which are intense in one way but not in the other are common. E.g., the first two states to merge in the unilateral dendrogram in Fig. \ref{fig_unilateral_example_us}-(a) are Massachusetts and New Hampshire at resolution $\delta=0.580$ because from all the people that moved into New Hampshire, 42\% came from Massachusetts, this being the highest value in all the country. The flow in the direction from New Hampshire to Massachusetts is lower, only 9\% of the immigrants entering the latter come from the former. This is the reason why these two states are not the first to merge in the reciprocal and nonreciprocal dendrograms. In these previous cases, Minnesota and Wisconsin were the first to merge because the relative flow in both directions is 16\% and 19\%.

Unilateral clusters tend to form around populous states. In Fig. \ref{fig_unilateral_example_us}-(b), the six clusters with more than two states  contain the seven states with largest population -- California, Texas, New York, Florida, Illinois, Pennsylvania, and Ohio \cite{USmigration} -- one in each cluster except for the blue one that contains New York and Pennsylvania. The data suggests that the reason for this is that populous states have a strong influence on the immigration into neighboring states. Indeed, if we focus on the cyan cluster formed around Texas, the proportional immigration into Louisiana, New Mexico, Oklahoma, and Arkansas coming from Texas is 31\%, 22\%, 29\%, and 21\% respectively. The opposite is not true, since the immigration into Texas from the four aforementioned neighboring state is of 5\%, 3\%, 4\%, and 3\%, respectively. However, this flow in the opposite direction is not required for unilateral clustering to merge the states into one cluster. Between two states with large population, the immigration is more balanced in both directions, thus merging at high resolutions in the unilateral dendrogram. E.g., 11\% of the immigration into Texas comes from California and 8\% in the opposite direction.

Unilateral clustering detects an east-west division of migration flows in the United States. The last merging in the unilateral dendrogram occurs at resolution $\delta=0.8958$ and just below the merging resolution, e.g. at resolution $\delta^\U_2$, there are two clusters -- east and west -- corresponding to the ones depicted in Fig. \ref{fig_unilateral_example_us}-(c). The cut at $\delta^\U_2$ corresponds to a migrational flow of 10.45\%. This implies that for any two different states within the same cluster we can find a unilateral chain where every flow is at least 10.45\%. More interestingly, there is no pair of states, one from the east and one form the west, with a flow of 10.45\% or more in any direction.

\subsubsection{Directed single linkage quasi-clustering $\tilde{\ccalH}^*$}\label{sec_directed_migration}

%
\begin{figure}
\centering
\input{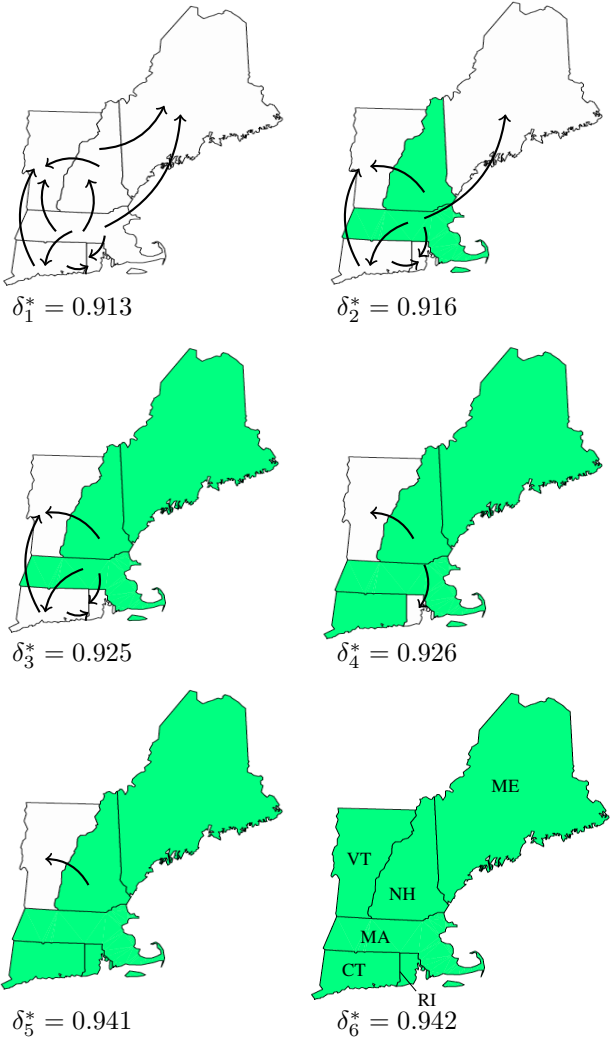}
\vspace{-0.1in}
\caption{Directed single linkage quasi-clustering method applied to New England's migration flow. Quasi-partitions shown for resolutions before every merging and after the last. Massachusetts migrational influence over the region is represented by the outgoing edges in the quasi-partitions.}
\vspace{-0.2in}
\label{fig_quasi-clustering_new_england}
\end{figure}

\begin{figure}
\centering
\input{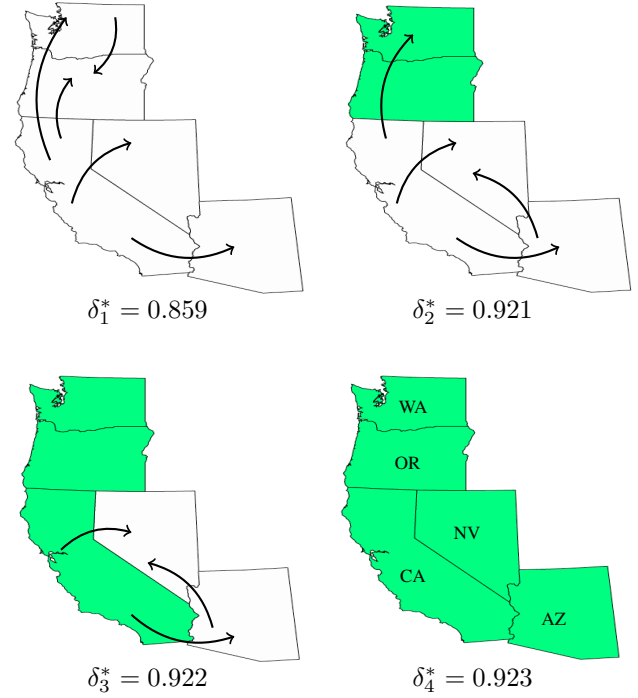}
\vspace{-0.1in}
\caption{Directed single linkage quasi-clustering method applied to the extended West Coast migration flow. Quasi-partitions shown for resolutions before every merging and after the last. California acts as an agglutination agent in the region.}
\vspace{-0.2in}
\label{fig_quasi-clustering_west_coast}
\end{figure}

The outcome of applying the directed single linkage quasi-clustering method $\tilde{\ccalH}^*$ with output quasi-ultrametrics defined in \eqref{eqn_nonreciprocal_chains} to the migration network $N_S$ is computed with the algorithmic formula in \eqref{eqn_algo_quasi_ultrametric}. In figs. \ref{fig_quasi-clustering_new_england} and \ref{fig_quasi-clustering_west_coast} we show some quasi-partitions of the output quasi-dendrogram $\tdD^*_S=(D^*_S, E^*_S)$ focusing on New England and an extended West Coast including Arizona and Nevada. States represented with the same color are part of the same cluster at the given resolution and states in white form singleton clusters. Arrows between clusters for a given resolution $\delta$ represent the edge set $E^*_S(\delta)$ for resolution $\delta$. The resolutions $\delta$ at which quasi-partitions are shown in figs. \ref{fig_quasi-clustering_new_england} and \ref{fig_quasi-clustering_west_coast} correspond to those 0.001 smaller than those in which mergings in the dendrogram component $D^*_S$ of the output quasi-dendrogram $\tdD^*_S$ occur or, in the case of the last map in each figure, correspond to the resolution of the last merging in the region shown. E.g., in Fig. \ref{fig_quasi-clustering_west_coast} Oregon and Washington merge at resolution $\delta=0.860$, thus, in the first map we look at the quasi-partition at resolution $\delta^*_1=0.859$. 

The directed single linkage quasi-clustering method $\tilde{\ccalH}^*$ captures not only the formation of clusters but also the asymmetric influence between them. E.g. the quasi-partition in Fig. \ref{fig_quasi-clustering_new_england} for resolution $\delta^*_1=0.913$ is of little interest since every state forms a singleton cluster. The influence structure, however, reveals a highly asymmetric migration pattern. At this resolution Massachusetts has migrational influence over every other state in the region as depicted by the five arrows leaving Massachusetts and entering each of the other five states. No state has influence over Massachusetts at this resolution since this would imply the formation of a non singleton cluster by the mechanics of $\ccalH^*$. This influence could be explained by the fact that Massachusetts contains Boston, the largest urban area of the region. Hence, Boston attracts immigrants from all over the country reducing the proportional immigration into Massachusetts from its neighbors and generating the asymmetric influence structure observed. This is consistent with the conclusions regarding clustering around populous states that we reached by analyzing the unilateral clusters in Fig. \ref{fig_unilateral_example_us}-(b). However, in the quasi-partition analysis, as opposed to the unilateral clustering analysis, the influence of Massachusetts over the other states can be seen clearly as it is formally captured in the edge set $E_S^*(0.913)$. The rest of the influence pattern at this resolution sees Connecticut influencing Rhode Island and Vermont and New Hampshire influencing Maine and Vermont. 

At resolution $\delta^*_2=0.916$, we see that Massachusetts has merged with New Hampshire and this main cluster exerts influence over the rest of the region. Similarly, at resolution $\delta^*_3=0.925$, Maine has joined the cluster formed by Massachusetts and New Hampshire and together they exert influence over the singleton clusters of Connecticut, Rhode Island, and Vermont. The influence arcs from Connecticut to Rhode Island and Vermont persist in these two diagrams. We know that this has to be the case due to the influence hierarchy property of the the edge sets $E_S^*$ stated in condition (\~D3) in the definition of quasi-dendrogram in Section \ref{sec_quasi_dendrograms}. At resolution $\delta^*_4=0.926$ Connecticut joins the main cluster while Rhode Island joins at resolution $\delta^*=0.927$, thus we depict the corresponding maps at resolutions 0.001 smaller than these merging resolutions. The whole region becomes one cluster at resolution $\delta^*_6=0.942$ -- which marks the joining of Vermont into the cluster.

%

For the case of the West Coast in Fig. \ref{fig_quasi-clustering_west_coast}, California is the most influential state as expected from its large population. The quasi-partition at resolution $\delta^*_1=0.859$ is such that all states are singleton clusters with California exerting influence onto all other West Coast states and Washington exerting influence on Oregon. The first cluster to form does not involve California but Washington and Oregon merging at resolution $\delta=0.860$ and the cluster can be observed from the map at resolution $\delta^*_2=0.921$. However, California has influence over this two-state cluster as shown by the arrow going from California to the green cluster in the corresponding figure. The influence over the two other states, Nevada and Arizona, remains. This is as it should be because of the persistence property of the edge set $E_S^*$. At this resolution we also see an influence arc appearing from Arizona to Nevada. At resolution $\delta^*_3=0.922$ California joins the Washington-Oregon cluster that exerts influence over Arizona and Nevada. The whole region merges in a common cluster at resolution $\delta^*_4=0.923$.

An important property of quasi-dendrograms is that the quasi-partitions at any given resolution define a partial order between the clusters. Recall that slicing a {\it dendrogram} at certain resolution yields a partition of the node set where there is no defined order between the blocks of the partition. Slicing a {\it quasi-dendrogram} yields also an edge set $E_S^*(\delta)$ that defines a partial order among the clusters at such resolution. This partial order is useful because it allows us to ascertain the relative importance of different clusters. E.g., in the case of the extended West Coast in Fig. \ref{fig_quasi-clustering_west_coast} one would expect California to be the dominant migration force in the region. The quasi-partition at resolution $\delta^*_1=0.859$ permits asserting this fact formally because the partial order at this resolution has California ranked as more important than any other state. We also see the not unreasonable dominance of Washington over Oregon, while the remaining pairs of the ordering are not defined. 

At larger resolutions we can ascertain relative importance of clusters. At resolution $\delta^*_2=0.921$ we can say that California is more important than the cluster formed by Oregon and Washington as well as more important than Arizona and Nevada. We can also see that Arizona precedes Nevada in the migration ordering at this resolution while the remaining pairs of the ordering are undefined. At resolution $\delta^*_3=0.922$ there is an interesting pattern as we can see the cluster formed by the three West Coast states preceding Arizona and Nevada in the partial order. At this resolution the partial order also happens to be a complete order as Arizona is seen to precede Nevada. This is not true in general as we have already seen.

In New England and the West Coast, the respective importance of Massachusetts and California over nearby states acts as an agglutination force towards regional clustering. Indeed, if we delete any of these two states and cluster the remaining states in the corresponding region, the resolution at which the whole region becomes one cluster is increased, showing a decreasing tendency to cluster. E.g., for the case of New England, if we delete Massachusetts and cluster the remaining five states, they become one regional cluster at a resolution of $\delta^*=0.979$ whereas if we delete, e.g. Maine or Rhode Island, the remaining five states merge into one single cluster at resolution $\delta^*_6=0.942$ as in the original case [cf. Fig. \ref{fig_quasi-clustering_new_england}].


\renewcommand\arraystretch{1.2} 
\begin{table*}
\small
\centering
\caption{Code and description of industrial sectors}
\vspace{-0.1in}
\begin{tabular}{ l  l p{1cm}  l  l }
  \hline    
Code & Industrial Sector                                    &&
Code & Industrial Sector                                    \\  \hline                    
AC & Accommodation                                          &&
AG & Amusements, gambling, and recreation industries        \\ 
AH & Ambulatory health care services                        &&
AP & Apparel and leather and allied products                \\ 
AT & Air transportation                                     &&
AS & Administrative and support services                    \\ 
BT & Broadcasting and telecommunications                    &&
CE & Computer and electronic products                       \\ 
CH & Chemical products                                      &&
CO & Construction                                           \\ 
CS & Computer systems design and related services           &&
ED & Educational services                                   \\ 
EL & Electrical equipment, appliances, and components       &&
FA & Farms                                                  \\
\hline

\end{tabular}
\label{table_industrial_sectors}
\vspace{-0.1in}
\end{table*}

Further observe that if we limit our attention to the dendrogram component of the quasi-dendrogram depicted in Fig. \ref{fig_quasi-clustering_new_england}, i.e., if we ignore the edge sets $E_S^*(\delta)$, we recover the information in the nonreciprocal dendrogram in Fig. \ref{fig_nonreciprocal_us_dendrogram}. In the case of New England the dendrogram part $D^*_S$ of the quasi-dendrogram $\tdD_S^*$ has the mergings occurring at resolutions 0.001 larger than the resolutions used to depict the quasi-partitions, i.e. Massachusetts first merges with New Hampshire ($\delta=0.914$), then Maine joins this cluster ($\delta=0.917$), followed by Connecticut ($\delta=0.926$), Rhode Island ($\delta=0.927$) and finally Vermont ($\delta=0.942$). The order and resolutions in which states join the main cluster coincides with the blue part of the nonreciprocal dendrogram in Fig. \ref{fig_nonreciprocal_us_dendrogram}. In the case of the extended West Coast in Fig. \ref{fig_quasi-clustering_west_coast} we have Oregon joining Washington ($\delta=0.860$), which are then joined by California ($\delta=0.922$), which are then joined by Arizona and Nevada at resolution $\delta=0.923$. Observe that Arizona and Nevada do not form a separate cluster before joining California, Oregon, and Washington. They both join the rest of the states at the exact same resolution. This is the same order and the same resolutions corresponding to the green part of the nonreciprocal dendrogram in Fig. \ref{fig_nonreciprocal_us_dendrogram}. Notice that while Texas appears in the nonreciprocal dendrogram it does not appear in the quasi-partitions. This is only because we decided to show a partial view of the extended West Coast without including Texas. The fact that when we limit our attention to the dendrogram component of the quasi-dendrogram we recover the nonreciprocal dendrogram is not a coincidence. We know from Proposition \ref{prop_quasi_ultrametric_non_reciprocal} that the dendrogram component of the quasi-partitions generated by directed single linkage is equivalent to the dendrograms generated by nonreciprocal clustering.

\subsection{Interactions between sectors of the U.S. economy}\label{sec_economic_sectors}

The Bureau of Economic Analysis of the U.S. Department of Commerce publishes a yearly table of input and outputs organized by economic sectors \cite{USinputoutput}. This table records how economic sectors interact to generate gross domestic product. We focus on a particular section of this table, called \emph{uses}, corresponds to the inputs to production for year 2011 . More precisely, we are given a set $I$ of 61 industrial sectors as defined by the North American Industry Classification System (NAICS) -- see Table \ref{table_industrial_sectors} -- and a similarity function $U: I \times I \to \reals_+$ where $U(i, i')$ represents how much of the production of sector $i$, expressed in dollars, is used as an input of sector $i'$. Notice that it is common for part of the output of some sector $i \in I$ to be used as input in the same sector, i.e. $U(i,i)$ can be strictly positive. We define the network $N_I=(I, A_I)$ where the dissimilarity function $A_I$ satisfies $A_I(i,i)=0$ for all $i\in I$ and, for $i \neq i' \in I$, is given by
\begin{equation}\label{eqn_def_io_dissimilarity}
A_I(i, i') := f \left( \frac{U(i, i')}{\sum_j U(j, i')}\right),
\end{equation}
where $f: [0, 1) \to \reals_{++}$ is a given decreasing function. For the experiments here we use $f(x) = 1-x$. The normalization $U(i, i')/\sum_j U(j, i')$ in \eqref{eqn_def_io_dissimilarity} can be interpreted as the proportion of the input in dollars to productive sector $i'$ that comes from sector $i$. In this way, we focus on the combination of inputs of a sector rather than the size of the economic sector itself. That is, a small dissimilarity from sector $i$ to sector $i'$ implies that sector $i'$ highly relies on the use of sector $i$ output as an input for its own production. E.g., if 40\% of the input into sector $i'$ comes from sector $i$, we say that sector $i$ has an influence of 40\% over $i'$ and the dissimilarity $A_I(i, i') = 1-0.40 = 0.60$. Given that part of the output of some sector can be used as input in the same sector, if we sum the input proportion from every other sector, we obtain a number less than 1. The role of the decreasing function $f$ is to transform the similarities into corresponding dissimilarities. 

\subsubsection{Reciprocal clustering $\ccalH^\R$}\label{sec_reciprocal_io}

\begin{figure*}
\centering
\centerline{\input{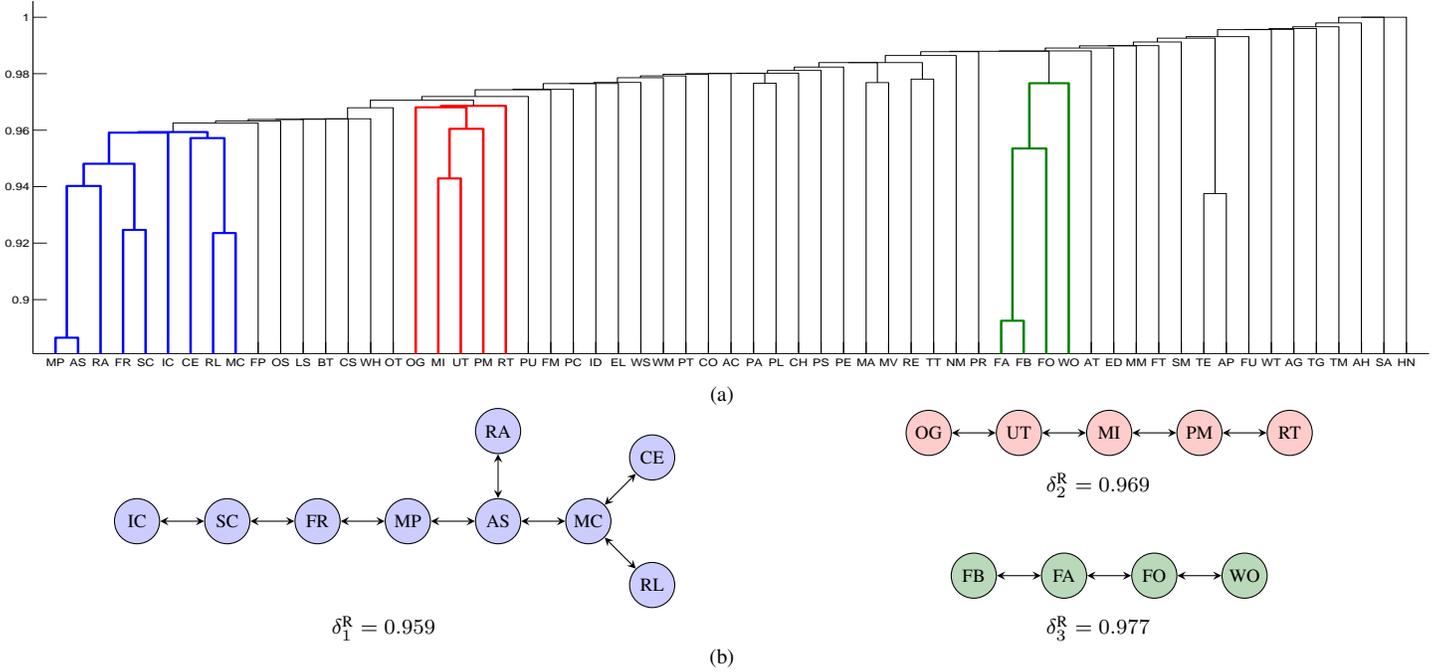} }
\vspace{-0.1in}
\caption{(a) Reciprocal dendrogram. Output of the reciprocal clustering method $\ccalH^\R$ when applied to the network $N_I$. Three clusters formed at resolutions $\delta^\R_1=0.959$, $\delta^\R_2=0.969$, and $\delta^\R_3=0.977$ are highlighted in blue, red and green, respectively. (b) Highlighted clusters. Edges between sectors represent bidirectional influence between them at the corresponding resolution.}
\vspace{-0.2in}
\label{fig_reciprocal_example_io}
\end{figure*}

The outcome of applying the reciprocal clustering method $\ccalH^\R$ defined in \eqref{eqn_reciprocal_clustering} to the network $N_I$ is computed with the algorithmic formula in \eqref{eqn_algo_recip}. The resulting output dendrogram is shown in Fig. \ref{fig_reciprocal_example_io}-(a) where three clusters are highlighted in blue, red and green. These clusters appear at resolutions $\delta^\R_1=0.959$, $\delta^\R_2=0.969$, and $\delta^\R_3=0.977$, respectively. In Fig. \ref{fig_reciprocal_example_io}-(b) we present the three highlighted clusters with edges representing bidirectional influence between industrial sectors at the corresponding resolution. That is, a double arrow is drawn between two nodes if and only if the dissimilarity between these nodes in both directions is less than or equal to the resolution at which the corresponding cluster appears. In particular, it shows the bidirectional chains of minimum cost between two nodes. E.g., for the blue cluster ($\delta^\R_1=0.959$) the bidirectional chain of minimum cost from the sector `Rental and leasing services of intangible assets' (RL) to `Computer and electronic products' (CE) goes through `Management of companies and enterprises' (MC). 

According to our analysis, the reciprocal clustering method $\ccalH^\R$ tends to cluster sectors that satisfy one of two possible typologies. The first type of clustering occurs among sectors of balanced influence in both directions. E.g., the first two sectors to be merged by $\ccalH^\R$ are `Administrative and support services' (AS) and `Miscellaneous professional, scientific and technical services' (MP) at a resolution of $\delta=0.887$. This occurs because 13.2\% of the input of AS comes from MP -- corresponding to $A_I(\text{MP}, \text{AS})=0.868$ -- and 11.3\% of MP's input comes from AS -- implied by $A_I(\text{AS}, \text{MP})=0.887$ -- both influences being similar in magnitude. It is reasonable that these two sectors hire services from each other in order to better perform their own service. This balanced behavior is more frequently observed among service sectors than between raw material extraction (primary) or manufacturing (secondary) sectors. Notice that for two manufacturing sectors A and B to have balanced bidirectional influence we need the outputs of A to be inputs of B in the same proportion as the outputs of B are inputs of A. This situation is rarer. Further examples of this clustering typology where the influence in both directions is balanced can be found between pairs of service sectors with bidirectional edges in the blue cluster formed at resolution $\delta^\R_1=0.959$. E.g., the participation of RL in the input to MC is of 7.6\% -- since $A_I(\text{RL}, \text{MC})=0.924$ -- whereas the influence in the opposite direction is 8.5\%, given by $A_I(\text{MC},\text{RL})=0.915$. Similarly, 6.5\% of the input to the `Real estate' (RA) sector comes from AS and 6.0\% vice versa. This implies that the RA sector hires external administrative and support services and the AS sector depends on the real estate services to, e.g., rent a location for their operation. The second type of clustering occurs between sectors with one natural direction of influence but where the influence in the opposite direction is meaningful. E.g., the second merging in the reciprocal dendrogram in Fig. \ref{fig_reciprocal_example_io}-(a) occurs at resolution $\delta=0.893$ between the `Farm' (FA) sector and the `Food, beverage and tobacco products' (FB) sector. In this case, one expects a big portion of FB's input to come from FA -- 35.2\% to be precise -- as raw materials for processed food products but there is also a dependency on the opposite direction of 10.7\% from, e.g., food supplementation for livestock not entirely fed with grass. This second clustering typology generally occurs between consecutive sectors in the production chain of a particular industry, with the strong influence in the natural direction of the material movement and the non negligible influence in the opposite direction which is particular of each industry. E.g., for the food industry, the primary FA sector precedes in the production process the secondary FB sector. Thus, the influence of FA over FB is clear. However, there is an influence of FB over FA that could be explained by the provision of food supplementation for livestock. Further examples of this interaction between sectors can be found in the textile and metal industries. Representing the textile industry, at resolution $\delta=0.938$ the sectors `Textile mills and textile product mills' (TE) and `Apparel and leather and allied products' (AP) merge. In the garment production process, there is a natural direction of influence from TE that generates fabric from a basic fiber to AP that cuts and sews the fabric to generate garments. Indeed, the influence in this direction is of 17.8\% represented by $A_I(\text{TE}, \text{AP})=0.822$. However, there is an influence of 6.2\% -- corresponding to $A_I(\text{AP}, \text{TE})=0.938$ -- in the opposite direction. This influence can be partially attributed to companies in the TE sector which also manufacture garments and buy intermediate products from companies in the AP sector. For example, a textile mill that produces wool fabric and also manufactures wool garments with some details in leather. This leather comes from a company in the AP sector and represents a movement from AP back to TE. In the metal industry, at resolution $\delta=0.960$ `Mining, except oil and gas' (MI) merges with `Primary metals' (PM). The bidirectional influence between these two sectors can be observed in the red cluster formed at resolution $\delta^\R_2=0.969$ in Fig. \ref{fig_reciprocal_example_io}-(b). As before, the natural influence is in the direction of the production process, i.e. from MI to PM. Indeed, 9.3\% of PM's input comes from MI mainly as ores for metal manufacturing. Moreover, there is an influence of 4.0\% in the opposite direction from PM to MI due to, e.g., structural metals for mining infrastructure.

%

The cluster in Fig. \ref{fig_reciprocal_example_io} that forms at resolution $\delta^\R_1=0.959$ (blue) is mainly composed of services. The first two mergings, described in the previous paragraph, occur between MP-AS and RL-MC representing professional, support, rental and management services, respectively. At resolution $\delta=0.925$, the sectors `Federal Reserve banks, credit intermediation, and related activities' (FR) and `Securities, commodity contracts, and investments' (SC) merge. This is an exception to the described balanced mergings  between service sectors. Indeed, 24.1\% of FR's input comes from SC whereas only 7.5\% of SC's input comes from FR. This is expected since credit intermediation entities in FR have as input investments done in the SC sector. At resolution $\delta=0.940$, RA joins the MP-AS cluster due to the bidirectional influence between RA and AS described in the previous paragraph. The MP-AS-RA cluster merges with the FR-SC cluster at resolution $\delta=0.948$ due to the relation between MP and FR. More precisely, MP provides 11.3\% of FR input -- corresponding to $A_I(\text{MP}, \text{FR})=0.887$ -- and 5.2\% of MP's input comes from FR, given by $A_I(\text{FR}, \text{MP})=0.948$. At resolution $\delta=0.957$, CE joins the RL-MC cluster due to its bidirectional influence relation with MC. The sector of electronic products CE is the only sector in the blue cluster formed at resolution $\delta^\R_1=0.959$ that does not represent a service. The `Insurance carriers and related activities' (IC) sector joins the MP-AS-RA-FR-SC cluster at resolution $\delta=0.959$ because of its relation with SC. In fact, 4.5\% of IC's input comes from SC in the form of securities and investments and 4.1\% of SC's input comes from IC in the form of insurance policies for investments. Finally, at resolution $\delta^\R_1=0.959$, the clusters MP-AS-RA-FR-SC-IC and CE-RL-MC merge due to the relation between the supporting services AS and the management services MC.

The cluster in Fig. \ref{fig_reciprocal_example_io} that forms at resolution $\delta^\R_2=0.969$ (red) mixes the three levels of the economy: raw material extraction or primary, manufacturing or secondary and services or tertiary. The `Mining, except oil and gas' sector (MI), which is a primary activity of extraction, merges at resolution $\delta=0.943$ with the `Utilities' (UT) sector which extends vertically into the secondary and tertiary industrial sectors since it generates and distributes energy. This merging occurs because 5.7\% of UT's input comes from MI and 8.8\% vice versa. This pair then merges at resolution $\delta=0.961$ with the manufacturing sector of `Primary metals' (PM). PM joins this cluster due to its bidirectional relation with MI previously described. At resolution $\delta=0.968$, the primary sector of `Oil and gas extraction' (OG) joins the MI-UT-PM cluster because 3.2\% of OG's input comes from UT, mainly as electric power supply, and 57.3\% of UT's input comes from OG as natural gas for combustion and distribution. Finally, at resolution $\delta^\R_2=0.969$ the service sector of `Rail transportation' (RT) merges with the rest of the cluster due to its influence relation with PM. Indeed, PM provides 7.0\% of the input of RT for the construction of railroads -- corresponding to $A_I(\text{PM}, \text{RT})=0.930$ -- and RT provides 3.1\% of PM's input -- given by $A_I(\text{RT}, \text{PM})=0.969$ -- as transportation services for final metal products. 

The cluster in Fig. \ref{fig_reciprocal_example_io} that forms at resolution $\delta^\R_3=0.977$ (green) is composed of food and wood generation and processing. It starts with the aforementioned merging between FA and FB at $\delta=0.893$. At resolution $\delta=0.956$, `Forestry, fishing, and related activities' (FO) joins the FA-FB cluster due to its relation with FA. The farming sector FA depends 9.2\% on FO due to, e.g., deforestation for crop growth. The dependence in the opposite direction is of 4.7\%. Finally, at $\delta^\R_3=0.977$, `Wood products' (WO) joins the cluster. Its relation with FO is highly asymmetric and corresponds to the second clustering typology described at the beginning of this section. There is a natural influence in the direction of  the material movement from FO to WO. Indeed, 26.2\% of WO's input comes from FO whereas the influence is of 2.3\% in the opposite direction. 

Requiring direct bidirectional influence for clustering generates some cluster which are counter-intuitive. E.g., in the reciprocal dendrogram in Fig. \ref{fig_reciprocal_example_io}-(a), at resolution $\delta=0.971$ when the blue and red clusters merge together we have that the oil and gas sector OG in the red cluster joins the insurance sector IC in the blue cluster. However, OG does not merge with `Petroleum and coal products' (PC), a sector that one would expect to be more closely related, until resolution $\delta=0.975$. In order to avoid this situation, we may allow nonreciprocal influence as we do in the following section.

\subsubsection{Nonreciprocal clustering $\ccalH^{\NR}$}\label{sec_nonreciprocal_io}

\begin{figure*}
\centering
\centerline{\input{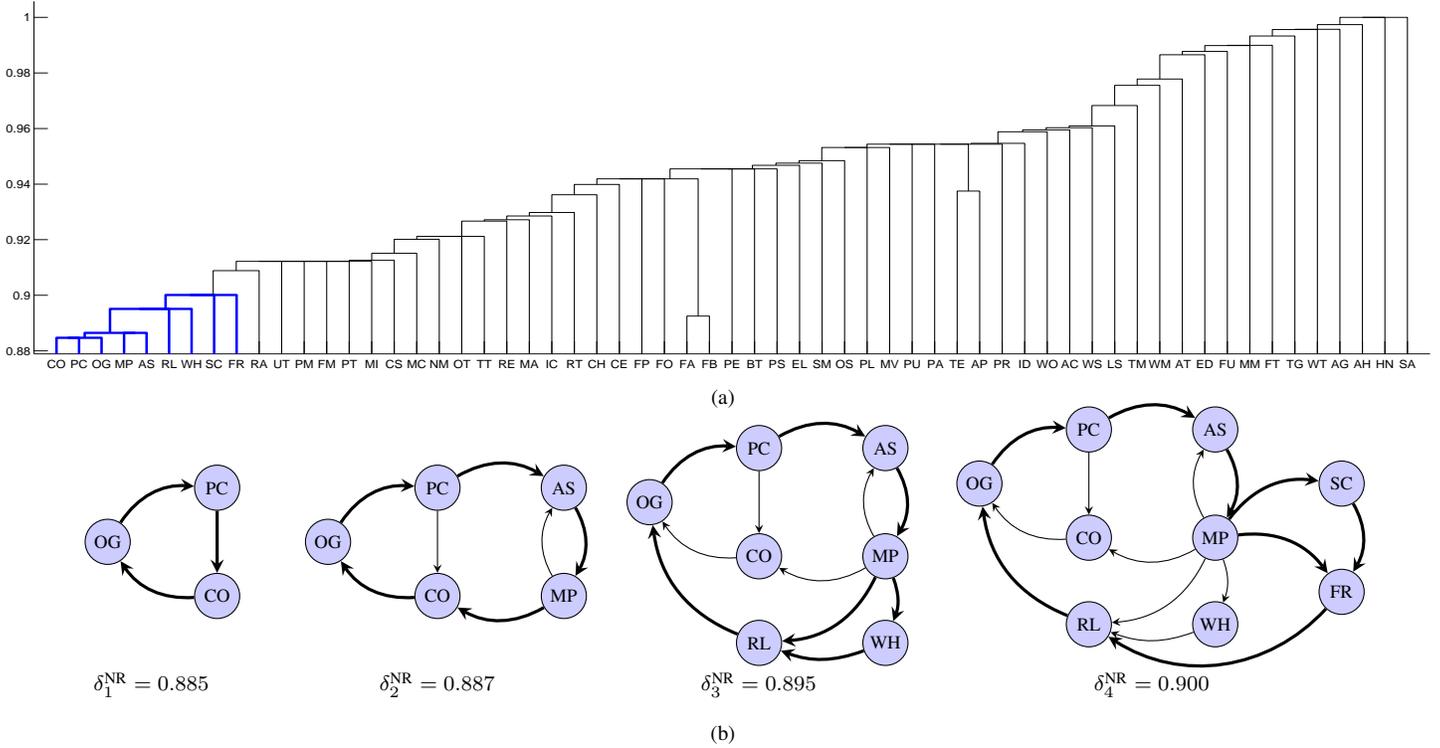} }
\vspace{-0.1in}
\caption{(a) Nonreciprocal dendrogram. Output of the nonreciprocal clustering method $\ccalH^{\NR}$ when applied to the network $N_I$. One cluster, formed at resolution $\delta^{\NR}_4=0.900$, is highlighted in blue. (b) Generation of highlighted cluster. Sequential mergings of sectors at resolutions $\delta^{\NR}_1=0.885$, $\delta^{\NR}_2=0.887$, $\delta^{\NR}_3=0.895$, and $\delta^{\NR}_4=0.900$ are shown. Directed edges between sectors imply unidirectional influence between them at the corresponding resolution. Notice the cyclic influences between the sectors, e.g., OG $\to$ PC $\to$ CO $\to$ OG in the leftmost diagram.}
\vspace{-0.15in}
\label{fig_nonreciprocal_example_io}
\end{figure*}

The outcome of applying the nonreciprocal clustering method $\ccalH^\NR$ defined in \eqref{eqn_nonreciprocal_clustering} to the network $N_I$ is computed with formula \eqref{eqn_algo_nonrecip}. The resulting output dendrogram is shown in Fig. \ref{fig_nonreciprocal_example_io}-(a). Let us first observe, as we did for the case of the migration matrix in Section \ref{sec_nonreciprocal_io}, that the nonreciprocal ultrametric distances in Fig. \ref{fig_nonreciprocal_example_io}-(a) are not larger than the reciprocal ultrametric distances in Fig. \ref{fig_reciprocal_example_io}-(a) as it should be the case given the inequality in \eqref{eqn_nonreciprocal_smaller_than_reciprocal}. As a test case we have that the mining sector MI and the `Pipeline transportation' (PT) sectors become part of the same cluster in the reciprocal dendrogram at a resolution $\delta= 0.979$ whereas they merge in the nonreciprocal dendrogram at resolution $\delta'=0.912<0.979$.

A more interesting observation is that, in contrast with the case of the migration matrix of Section \ref{sec_state_to_state_migration}, the nonreciprocal dendrogram is qualitatively very different from the reciprocal dendrogram. In the reciprocal dendrogram we tended to see the formation of definite clusters that then merged into larger clusters at coarser resolutions. The cluster formed at resolution $\delta^\R_1=0.959$ (blue) shown in Fig. \ref{fig_reciprocal_example_io}-(b) grows by merging with singleton clusters (FP, OS, LS, BT, CS, WH, and OT in progressive order of resolution) until it merges at resolution $\delta=0.971$ with a cluster of five nodes which emerges at resolution $\delta^\R_2=0.969$. This whole cluster then grows by adding single nodes and pairs of nodes until it merges at resolution $\delta=0.988$ with a cluster of four nodes that forms at resolution $\delta^\R_3=0.977$. In the nonreciprocal dendrogram, in contrast, we see the progressive agglutination of economic sectors into a central cluster.

Indeed, the first non singleton cluster to arise is formed at resolution $\delta^{\NR}_1=0.885$ by the sectors of oil and gas extraction OG, petroleum and coal products PC, and `Construction' (CO). For reference, observe that this happens before the first reciprocal merging between AS and MP, which occurs at resolution $\delta=0.887$ [cf. Fig. \ref{fig_reciprocal_example_io}-(a)]. The cluster formed by OG, PC, and MP is shown in the leftmost graph in Fig. \ref{fig_nonreciprocal_example_io}-(b) where the directed edges represent all the dissimilarities $A_I(i, i') \leq \delta^{\NR}_1=0.885$ between these three nodes. We see that this cluster forms due to the influence cycle $[$OG, PC, CO, OG$]$. Of all the economic input to PC, $82.6\%$ comes from the OG sector -- which is represented by the dissimilarity $A_I(\text{OG}, \text{PC})=0.174$ -- in the form of raw material for its productive processes of which the dominant process is oil refining. In the input to CO a total of $11.5\%$ comes from PC  -- tantamount to dissimilarity $A_I(\text{PC}, \text{CO})=0.885$ -- as fuel and lubricating oil for heavy machinery as well as asphalt coating, and $12.3\%$ of OG's input comes from CO -- corresponding to dissimilarity $A_I(\text{CO}, \text{OG})=0.877$ -- mainly from engineering projects to enable extraction such as perforation and the construction of pipelines and their maintenance.

At resolution $\delta^{\NR}_2=0.887$ this cluster grows by the simultaneous incorporation of the support service sector AS and the professional service sector MP. These sectors join due to the loop $[$AS, MP, CO, OG, PC, AS$]$. The three new edges in this loop that involve the new sectors are the ones from PC to AS, from AS to MP and from MP to CO. Of all the economic input to AS, $13.4\%$ comes from the PC sector -- which is represented by the dissimilarity $A_I(\text{PC}, \text{AS})=0.866$ -- in the form of, e.g., fuel for the transportation of manpower. Of MP's input, 11.3\% comes from AS -- given by $A_I(\text{AS}, \text{MP})=0.887$ -- corresponding to administrative and support services hired by the MP sector for the correct delivery of MP's professional services and in the input to CO a total of 12.8\% comes from MP -- corresponding to $A_I(\text{MP}, \text{CO})=0.872$ -- from, e.g., architecture and consulting services for the construction. 

%

We then see the incorporation of the rental service sector RL and `Wholesale trade' (WH) to the five-node cluster at resolution $\delta^{\NR}_3=0.895$ given by the loop $[$WH, RL, OG, PC, AS, MP, WH$]$. To be more precise, the sector RL joins the main cluster by the aforementioned loop and by another one excluding WH, i.e. $[$RL, OG, PC, AS, MP, RL$]$. The formation of both loops is simultaneous since the last edge to appear is the one going from RL to OG at resolution $A_I(\text{RL}, \text{OG})=\delta^{\NR}_3=0.895$. This implies that from OG's inputs, 10.5\% comes from RL from, e.g., rental and leasing of generators, pumps, welding equipment and other machinery for extraction. The other edges depicted in the cluster at resolution $\delta^{\NR}_3$ that complete the two mentioned loops are the ones from MP to RL, from MP to WH, and from WH to RL. These edges are associated with the corresponding dissimilarities $A_I(\text{MP}, \text{RL})=0.886$, $A_I(\text{MP}, \text{WH})=0.836$, and $A_I(\text{WH}, \text{RL})=0.894$, all of them less than $\delta^{\NR}_3$.

At resolution $\delta^{\NR}_4=0.900$ the financial sectors SC and FR join this cluster due to the chain $[$SC, FR, RL, OG, PC, AS, SC$]$. Analogous to RL's merging at resolution $\delta^{\NR}_3$, the sector FR merges the main cluster by the aforementioned loop and by the one excluding SC, i.e., $[$FR, RL, OG, PC, AS, FR$]$. Both chains are formed simultaneously since the last edge to appear is the one from FR to RL at resolution $A_I(\text{FR}, \text{RL})=\delta^{\NR}_4=0.900$. This means that from RL's inputs, 10\% comes from FR. The remaining edges depicted in the cluster at resolution $\delta^{\NR}_4$ that complete the two mentioned loops are the ones from MP to SC, from MP to FR, and from SC to FR. These edges are associated with the corresponding dissimilarities $A_I(\text{MP}, \text{SC})=0.837$, $A_I(\text{MP}, \text{FR})=0.887$, and $A_I(\text{SC}, \text{FR})=0.759$, all of them less than $\delta^{\NR}_4$. 

The sole exceptions to this pattern of progressive agglutination are the pairings of the farms FA and the food products FB sectors at resolution $\delta=0.893$ and the textile mills TE and apparel products AP sectors at resolution $\delta=0.938$.

The nonreciprocal clustering method $\ccalH^{\NR}$ detects cyclic influences which, in general, lead to clusters that are more reasonable than those requiring the bidirectional influence that defines the reciprocal method $\ccalH^{\R}$. E.g., $\ccalH^{\NR}$ merges OG with PC at resolution $\delta=0.885$ before they merge with the insurance sector IC at resolution $\delta=0.923$. As we had already noted in the last paragraph of the preceding section, $\ccalH^{\R}$ merges OG with IC before their common joining with PC. However, the preponderance of cyclic influences in the network of economic interactions $N_I$ leads to the formation of clusters that look more like artifacts than fundamental features. E.g., the cluster that forms at resolution $\delta^{\NR}_2=0.887$ has AS and MP joining the three-node cluster CO-PC-OG because of an influence cycle of five nodes composed of $[$AS, MP, CO, OG, PC, AS$]$. From our discussion above, it is thus apparent that allowing clusters to be formed by arbitrarily long cycles overlooks important bidirectional influences between co-clustered nodes. If we wanted a clustering method which at resolution $\delta^{\NR}_2=0.887$ would cluster the nodes PC, CO, and OG into one cluster and AS and MP into another cluster, we should allow influence to propagate through cycles of at most three or four nodes. A family of methods that permits this degree of flexibility is the family of semi-reciprocal methods $\ccalH^{\SR(t)}$ that we discussed in Section \ref{sec_inter_reciprocal} and whose application we exemplify in the following section.

\subsubsection{Semi-reciprocal clustering $\ccalH^{\SR(3)}$}\label{sec_semi_reciprocal_io}

%
\begin{figure*}
\centering
\centerline{\input{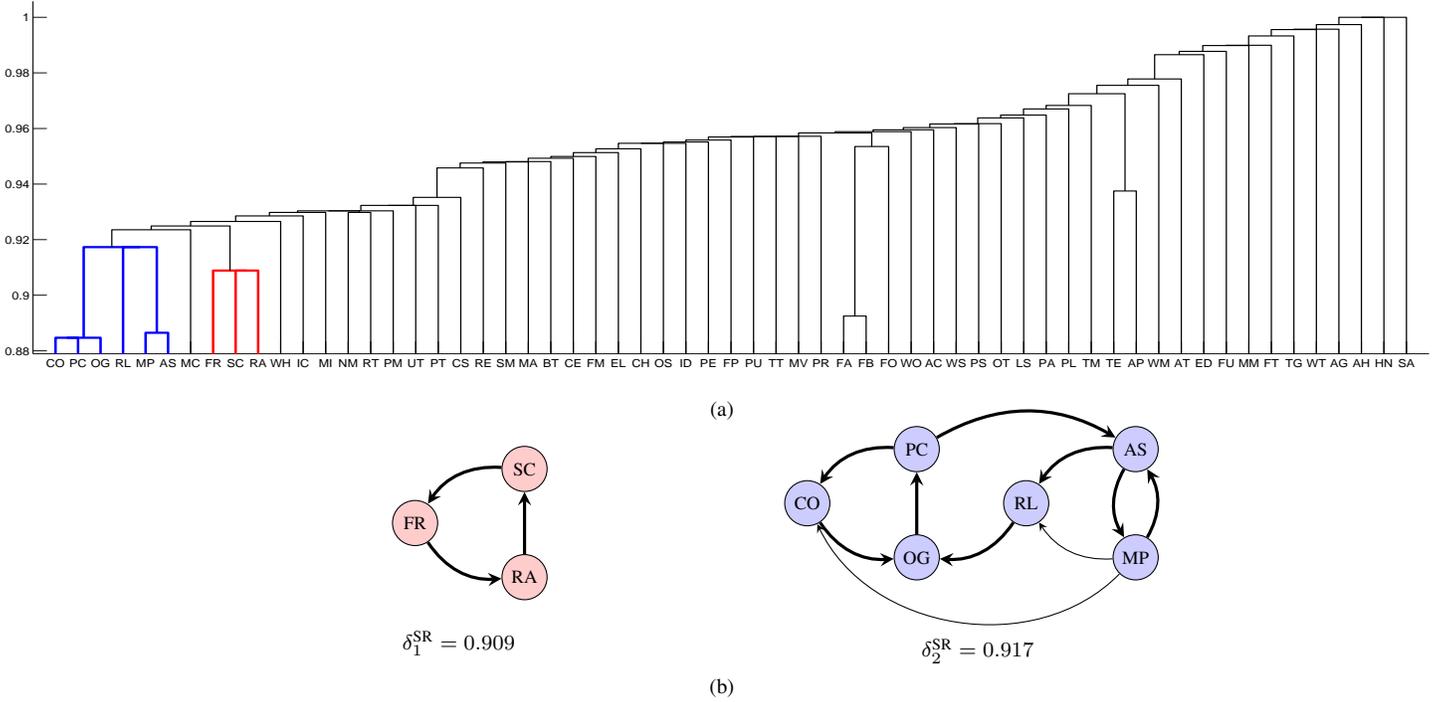} }
\vspace{-0.1in}
\caption{(a) Semi-reciprocal dendrogram. Output of the semi-reciprocal clustering method $\ccalH^{\SR(3)}$ when applied to the network $N_I$. Two clusters formed at resolution $\delta^\SR_1=0.909$ and $\delta^\SR_2=0.917$ are highlighted in red and blue, respectively. (b) Highlighted clusters. Directed edges between sectors imply unidirectional influence between them at the corresponding resolution. Cyclic influences can be observed.}
\vspace{-0.1in}
\label{fig_semi_reciprocal_example_io}
\end{figure*}

The outcome of applying the semi-reciprocal clustering method $\ccalH^{\SR(3)}$ defined in Section \ref{sec_inter_reciprocal} to the network $N_I$ is computed with the formula in \eqref{eqn_algo_semi_reciprocal_2}. The resulting output dendrogram is shown in Fig. \ref{fig_semi_reciprocal_example_io}-(a). 
Two clusters generated at resolutions $\delta^\SR_1=0.909$ and $\delta^\SR_2=0.917$ are highlighted in red and blue, respectively. These clusters are depicted in Fig. \ref{fig_semi_reciprocal_example_io}-(b) with directed edges between the nodes representing dissimilarities less than or equal to the corresponding resolution. E.g., for the cluster generated at resolution $\delta^\SR_1=0.909$ (red), we draw an edge from sector $i$ to sector $i'$ if and only if $A_I(i, i') \leq \delta^\SR_1$. 
Comparing the semi-reciprocal dendrogram in Fig. \ref{fig_semi_reciprocal_example_io}-(a) with the reciprocal and nonreciprocal dendrograms in figs. \ref{fig_reciprocal_example_io}-(a) and \ref{fig_nonreciprocal_example_io}-(a), we observe that semi-reciprocal clustering merges any pair of sectors  into a cluster at a resolution not higher than the resolution at which they are co-clustered by reciprocal clustering and not lower than the one at which they are co-clustered by nonreciprocal clustering. E.g., the sectors of construction CO and `Fabricated metal products'  (FM) become part of the same cluster at resolution $\delta_\R=0.980$ in the reciprocal dendrogram, at resolution $\delta_{\SR}=0.950$ in the semi-reciprocal dendrogram and at resolution $\delta_{\NR}=0.912$ in the nonreciprocal dendrogram, satisfying $\delta_{\NR} \leq \delta_{\SR} \leq \delta_\R$. The inequalities described among the merging resolutions need not be strict as in the previous example, e.g., the farms (FA) sector merges with the food products FB sector at resolution $\delta=0.893$ for the reciprocal, nonreciprocal and semi-reciprocal clustering methods. This ordering of the merging resolutions is as it should be since the reciprocal and nonreciprocal ultrametrics uniformly bound the output ultrametric of any clustering method satisfying the axioms of value and transformation such as the semi-reciprocal clustering method [cf. \eqref{eqn_theo_extremal_ultrametrics}].

The semi-reciprocal clustering method $\ccalH^{\SR(3)}$ allows reasonable cyclic influences and is insensitive to intricate influences described by long cycles. As we pointed out in the two preceding subsections, $\ccalH^\R$ does not recognize the obvious relation between the sectors oil and gas extraction OG and the petroleum products PC sectors because it requires direct bidirectional influence whereas $\ccalH^{\NR}$ merges OG and PC at a low resolution but also considers other counter-intuitive cyclic influence structures represented by long loops such as the merging of the service sectors AS and MP with the cluster OG-PC-CO before forming a cluster by themselves [cf. Fig. \ref{fig_nonreciprocal_example_io}]. The semi-reciprocal method $\ccalH^{\SR(3)}$ combines the desirable features of the reciprocal and nonreciprocal methods. Indeed, as can be seen from the semi-reciprocal dendrogram in Fig. \ref{fig_semi_reciprocal_example_io}-(a), $\ccalH^{\SR(3)}$ recognizes the heavy industry cluster OG-PC-CO since these three sectors are the first to merge at resolution $\delta=0.885$. However, the service sectors MP and AS form a cluster of their own before merging with the heavy industry cluster. To be more precise, MP and AS merge at resolution $\delta=0.887$ due to the bidirectional influence between them. This resolution coincides with the first merging in the reciprocal dendrogram [cf. Fig. \ref{fig_reciprocal_example_io}-(a)]. 
When we increase the resolution, at $\delta^{\SR}_2=0.917$ the `Rental and leasing services' (RL) sector acts as an intermediary merging the OG-PC-CO cluster with the MP-AS cluster forming the blue cluster in Fig. \ref{fig_semi_reciprocal_example_io}-(b). The cycle containing RL with secondary chains of length at most 3 nodes is $[$RL, OG, PC, AS, RL$]$. The sector RL uses administrative and support services from AS to provide their own leasing services and leasing is a common practice in the OG sector. Thus, the influences depicted in the blue cluster.
At resolution $\delta^{\SR}_1=0.909$ the credit intermediation sector FR, the investment sector SC and the real estate sector RA form a three-node cluster given by the influence cycle $[$RA, SC, FR, RA$]$ and depicted in red in Fig. \ref{fig_semi_reciprocal_example_io}-(b). Of all the economic input to SC, $9.1\%$ comes from the RA sector -- which is represented by the dissimilarity $A_I(\text{RA}, \text{SC})=0.909$ -- in the form of, e.g., leasing services related to real estate investment trusts. The sector SC provides 24.1\% of FR's input -- corresponding to $A_I(\text{SC}, \text{FR})=0.759$ -- whereas FR represents 35.1\% of RA's input -- corresponding to $A_I(\text{FR}, \text{RA})=0.649$. We interpret the relation among these three sectors as follows: the credit intermediation sector FR acts as a vehicle to connect the investments sector SC with the sector that attracts investments RA. Notice that in the nonreciprocal dendrogram in Fig. \ref{fig_nonreciprocal_example_io}-(a), these three sectors join the main blue cluster separately due to the formation of intricate influence loops. The semi-reciprocal method, by not allowing the formation of long loops, distinguishes the more reasonable cluster formed by FR-RA-SC.

\subsubsection{Unilateral clustering $\ccalH^\U$}\label{sec_unilateral_io}

%
\begin{figure*}
\centering
\centerline{\input{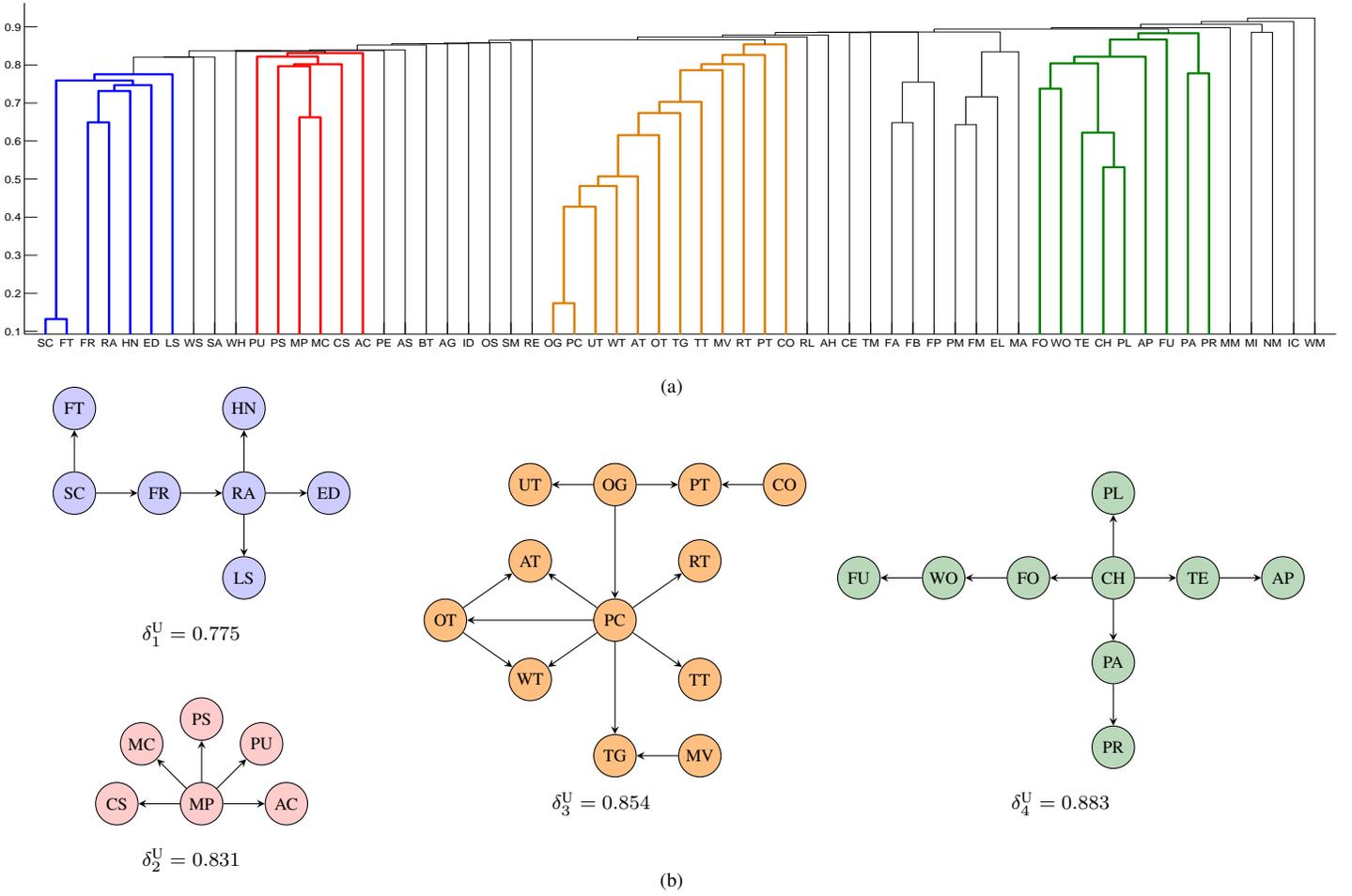} }
\vspace{-0.05in}
\caption{(a) Unilateral dendrogram. Output of the unilateral clustering method $\ccalH^{\U}$ when applied to the network $N_I$. Four clusters formed at resolutions $\delta^\U_1=0.775$, $\delta^\U_2=0.831$, $\delta^\U_3=0.854$, and $\delta^\U_4=0.883$ are highlighted in blue, red, orange, and green, respectively. (b) Highlighted clusters. Directed edges between sectors imply unidirectional influence between them at the corresponding resolution. Cycles are not required for the formation of clusters due to the definition of unilateral clustering $\ccalH^\U$.}
\vspace{-0.2in}
\label{fig_unilateral_example_io}
\end{figure*}

The outcome of applying the unilateral clustering method $\ccalH^\U$ defined in \eqref{eqn_unilateral_clustering_2} to the network $N_I$ is computed with the algorithmic formula in \eqref{eqn_algo_unilateral}. The resulting output dendrogram is shown in Fig. \ref{fig_unilateral_example_io}-(a). Four clusters appearing at resolutions $\delta^\U_1=0.775$, $\delta^\U_2=0.831$, $\delta^\U_3=0.854$, and $\delta^\U_4=0.883$ are highlighted in blue, red, orange, and green, respectively. In Fig. \ref{fig_unilateral_example_io}-(b) we explicit the highlighted clusters and draw a directed edge between two nodes if and only if the dissimilarity between them is less than or equal to the corresponding resolution at which the clusters are formed. E.g., for the cluster generated at resolution $\delta^\U_1=0.775$ (blue), we draw an edge from sector $i$ to sector $i'$ if and only if $A_I(i, i') \leq \delta^\U_1$. Unidirectional influence is enough for clusters to form when applying the unilateral clustering method. 

The asymmetry of the original network $N_I$ is put in evidence by the difference between the unilateral dendrogram in Fig. \ref{fig_unilateral_example_io}-(a) and the reciprocal dendrogram in Fig. \ref{fig_reciprocal_example_io}-(a). The last merging in the unilateral dendrogram, i.e. when `Waste management and remediation services' (WM) joins the main cluster, occurs at $\delta=0.923$. If, in turn, we cut the reciprocal dendrogram at this resolution, we observe 57 singleton clusters and two pairs of nodes merged together. Recall that if the original network is symmetric, the unilateral and the reciprocal dendrograms must coincide [cf. \eqref{eqn_nonreciprocal_sl_reciprocal} and \eqref{eqn_unilateral_as_single_linkage_hat}] and so must every other method satisfying the agnostic set of axioms in Section \ref{sec_agnostic_axiom_of_value} [cf. \eqref{eqn_theo_extremal_ultrametrics_2}]. Thus, the observed difference between the dendrograms is a manifestation of asymmetries in the network $N_I$.

Unilateral clustering detects intense one-way influences between sectors. The first two sectors to be merged into a single cluster by the unilateral clustering method $\ccalH^\U$ are the financial sectors SC and `Funds, trusts, and other financial vehicles' (FT) at a resolution $\delta=0.132$. This occurs because 86.8\% of FT's input comes from SC, corresponding to $A_I(\text{SC}, \text{FT})=0.132$ the smallest positive dissimilarity in the network $N_I$. The strong influence of SC over FT is expected since FT is comprised of entities organized to pool securities coming from the SC sector. The next merging when increasing the resolution occurs at $\delta=0.174$ between oil and gas extraction OG and petroleum and coal products PC since 82.6\% of PC's input comes from OG -- tantamount to $A_I(\text{OG}, \text{PC})=0.174$ -- mainly as crude oil for refining. The following three mergings correspond to sequential additions to the OG-PC cluster of the utilities UT, `Water transportation' (WT), and `Air transportation' (AT) sectors at resolution $\delta=0.428$, $\delta=0.482$, and $\delta=0.507$, respectively. These mergings occur because 57.2\% of UT's input comes from OG in the form of natural gas for both distribution and fuel for the generation of electricity and for the transportation sectors WT and AT, 51.8\% and 49.3\% of the respective inputs come from PC as the provision of liquid fuel.

%
\begin{figure*}
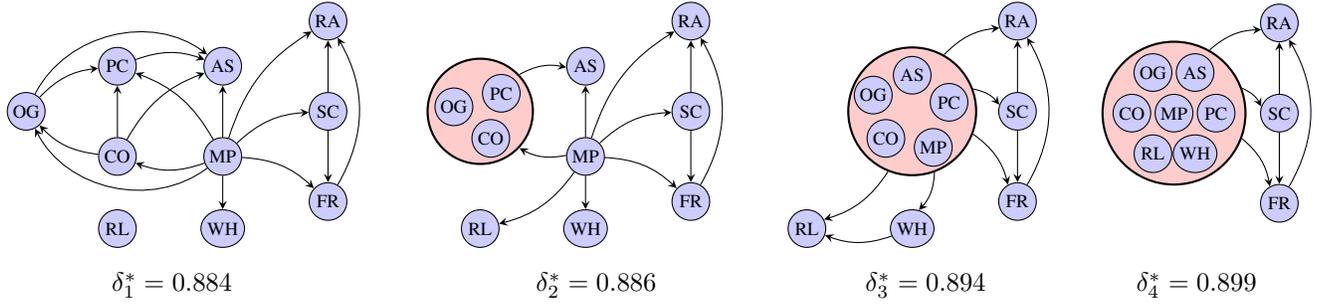

\centering
\centerline{\def \thisplotscale {0.5}
\def \unit {\thisplotscale cm}
\tikzstyle {blue vertex here} = [blue vertex, 
                                 minimum width = 0.7*\unit, 
                                 minimum height = 0.7*\unit, 
                                 anchor=center]
{\begin{tikzpicture}[thick, x = 1.2*\unit, y = 0.96*\unit]

       \node[anchor=south west,inner sep=0] at (-8,0) {\centering\input{quasi-clustering_io_network_1_2.tex}};
        \node[anchor=south west,inner sep=0] at (1.3,0) {\centering\input{quasi-clustering_io_network_2_2.tex}};
         \node[anchor=south west,inner sep=0] at (9.3,0) {\centering\input{quasi-clustering_io_network_3_2.tex}};
          \node[anchor=south west,inner sep=0] at (15.5,0.7) {\centering\input{quasi-clustering_io_network_4_2.tex}};

\node at (-3.7,-1) {$\delta^*_1=0.884$};
\node at (5.7, -1) {$\delta^*_2=0.886$};
\node at (13, -1) {$\delta^*_3=0.894$};
\node at (19, -1) {$\delta^*_4=0.899$};

\end{tikzpicture}} }
\vspace{-0.1in}
\caption{Directed single linkage quasi-clustering method applied to a portion of the sectors of the economy. Quasi-partitions shown for resolutions 0.001 smaller than the first four merging resolutions in the dendrogram component $D^*_I$ of the quasi-dendrogram $\tdD^*_I$. The edges define a partial order among the blocks of every quasi-partition.}
\vspace{-0.1in}
\label{fig_quasi-dendrogram_example_io_2}
\end{figure*}

Unilateral clusters tend to form around sectors of intense output. This observation is analogous to the formation of clusters around populous states, hence with intense population movement, that we observed in Section \ref{sec_unilateral_migration}. Indeed, if for each sector we evaluate the commodity intermediate value in dollars, i.e. the total output not destined to final uses, the professional service MP sector achieves the maximum followed by, in decreasing order, the sectors RA, OG, FR, AS and `Chemical products' (CH). These top sectors are composed of massively demanded services like professional, support, real estate and financial services plus the core activities of two important industries, namely oil \& gas and chemical products. Of these top six sectors, five are contained in the four clusters highlighted in Fig. \ref{fig_unilateral_example_io}-(b), with every cluster containing at least one of these sectors and the cluster formed at resolution $\delta^\U_1=0.775$ (blue) containing two, FR and RA. These clusters of intense output have influence, either directly or indirectly, over most of the sectors in their same cluster. E.g., in the cluster formed at resolution $\delta^\U_2=0.831$ (red) in Fig. \ref{fig_unilateral_example_io}-(b) there is a directed edge from MP to every other sector in the cluster. This occurs because MP provides professional and technical services that represent, in decreasing order, 33.8\%, 20.3\%, 19.8\%, 17.8\%, and 16.9\% of the input to the sectors of management of companies MC, `Motion picture and sound recording industries' (PS), `Computer systems design and related services' (CS), `Publishing industries' (PU), and `Accommodation' (AC), respectively. Consequently, in the unilateral clustering we can observe the MP sector merging with MC at resolution $\delta=0.662$ followed by a sequential merging of the remaining singleton clusters, i.e. PS at $\delta=0.797$, CS at $\delta=0.802$, PU at $\delta=0.822$ and finally AC joins at resolution $\delta^\U_2=0.831$. As another example consider the cluster formed at resolution $\delta^\U_4=0.833$ (green) containing the influential sector CH. Its influence over four different industries, namely plastics, apparel, paper and wood, is represented by the four directed branches leaving from CH in Fig. \ref{fig_unilateral_example_io}-(b). The sector CH first merges with `Plastics and rubber products' (PL) at resolution $\delta=0.531$ because 46.9\% of PL's input comes from CH as materials needed for the handling and manufacturing of plastics. The textile mills TE sector then merges at resolution $\delta=0.622$ because 37.8\% of TE's input comes from CH as dyes and other chemical products for the fabric manufacturing. At resolution $\delta=0.804$ the previously formed cluster composed of the forestry FO and wood products WO sectors join the CH-PL-TE cluster due to the dependence of FO on CH for the provision of chemicals for soil treatment and pest control. At resolution $\delta=0.822$, the apparel sector AP joins the main cluster due to its natural dependence on the fabrics generated by TE. Indeed, 17.8\% of AP's input comes from TE. In a similar way, at resolution $\delta=0.867$, `Furniture and related products' (FU) joins the cluster due to the influence from the WO sector. Finally, at resolution $\delta^\U_4=0.833$, the previously clustered paper industry comprised of the sectors `Paper products' (PA) and `Printing and related support activities' (PR) joins the main cluster due to the intense utilization of chemical products in the paper manufacturing process.

\subsubsection{Directed single linkage quasi-clustering $\tilde{\ccalH}^*$}\label{sec_directed_io}

The outcome of applying the directed single linkage quasi-clustering method $\tilde{\ccalH}^*$ with output quasi-ultrametrics defined in \eqref{eqn_nonreciprocal_chains} to the network $N_I$ is computed with the algorithmic formula in \eqref{eqn_algo_quasi_ultrametric}. In Fig. \ref{fig_quasi-dendrogram_example_io_2} we present four quasi-partitions of the output quasi-dendrogram $\tdD^*_I=(D^*_I, E^*_I)$ focusing on ten economic sectors. We limit the view of the quasi-partitions -- which were computed for the whole network -- to ten sectors to facilitate the interpretation. These ten sectors are the first to cluster in the dendrogram component $D^*_I$ of the quasi-dendrogram $\tdD^*_I$. To see this, recall that from Proposition \ref{prop_quasi_ultrametric_non_reciprocal} we have that $D^*_I=\ccalH^{\NR}(N_I)$, i.e. the dendrogram component $D^*_I$ coincides with the output dendrogram of applying the nonreciprocal clustering method to the network $N_I$. Hence, the ten sectors depicted in the quasi-partitions in Fig. \ref{fig_quasi-dendrogram_example_io_2} coincide with the ten leftmost sectors in the dendrogram in Fig. \ref{fig_nonreciprocal_example_io}-(a). We present quasi-partitions $\tdD^*_I(\delta)$ for four different resolutions $\delta^*_1=0.884$, $\delta^*_2=0.886$, $\delta^*_3=0.894$, and $\delta^*_4=0.899$. These resolutions are 0.001 smaller than the first four merging resolutions in the dendrogram component $D^*_I$ or, equivalently, in the nonreciprocal dendrogram [cf. Fig. \ref{fig_nonreciprocal_example_io}-(b)]. 

The edge component $E^*_I$ of the quasi-dendrogram $\tdD^*_I$ captures the asymmetric influence between clusters. E.g. in the quasi-partition in Fig. \ref{fig_quasi-dendrogram_example_io_2} for resolution $\delta^*_1=0.884$ every cluster is a singleton since the resolution is smaller than that of the first merging. However, the influence structure reveals an asymmetry in the dependence between the economic sectors. At this resolution the professional service sector MP has influence over every other sector except for the rental services RL as depicted by the eight arrows leaving the MP sector. No sector has influence over MP at this resolution since this would imply, except for RL, the formation of a non singleton cluster. The influence of MP reaches primary sectors as OG, secondary sectors as PC and tertiary sectors as AS or SC. The versatility of MP's influence can be explained by the diversity of services condensed in this economic sector, e.g. civil engineering and architectural services are demanded by CO, production engineering by PC and financial consulting by SC. For the rest of the influence pattern, we can observe an influence of CO over OG mainly due to the construction and maintenance of pipelines, which in turn influences PC due to the provision of crude oil for refining. Thus, from the transitivity (QP2) property of quasi-partitions introduced in Section \ref{sec_full_characterization_asymmetric} we have an influence edge from CO to PC. The sectors CO, PC and OG influence the support service sector AS. Moreover, the service sectors RA, SC and FR have a totally hierarchical influence structure where SC has influence over the other two and FR has influence over RA. Since these three nodes remain as singleton clusters for the resolutions studied, the influence structure described is preserved for higher resolutions as it should be from the influence hierarchy property of the the edge set $E_S^*(\delta)$ stated in condition (\~D3) in the definition of quasi-dendrogram in Section \ref{sec_quasi_dendrograms}.

At resolution $\delta^*_2=0.886$, we see that the sectors OG-PC-CO have formed a three-node cluster depicted in red that influences AS. At this resolution, the influence edge from MP to RL appears and, thus, MP gains influence over every other cluster in the quasi-partition including the three-node cluster. At resolution $\delta=0.887$ the service sectors AS and MP join the cluster OG-PC-CO and for $\delta^*_3=0.894$ we have this five-node cluster influencing the other five singleton clusters plus the mentioned hierarchical structure among SC, FR, and RA and an influence edge from WH to RL. When we increase the resolution to $\delta^*_4=0.899$ we see that RL and WH have joined the main cluster that influences the other three singleton clusters. If we keep increasing the resolution, we would see at resolution $\delta=0.900$ the sectors SC and FR joining the main cluster which would have influence over RA the only other cluster in the quasi-partition. Finally, at resolution $\delta=0.909$ RA joins the main cluster and the quasi-partition contains only one block.

The influence structure between clusters at any given resolution defines a partial order. More precisely, for every resolution $\delta$, the edge set $E_I^*(\delta)$ defines a partial order between the blocks given by the partition $D^*_I(\delta)$. We can use this partial order to evaluate the relative importance of different clusters by stating that more important sectors have influence over less important ones. E.g., at resolution $\delta^*_1=0.884$ we have that MP is more important than every other sector except for RL, which is incomparable at this resolution. There are three totally ordered chains that have MP as the most important sector at this resolution. The first one contains five sectors which are, in decreasing order of importance, MP, CO, OG, PC, and AS.  The second one is comprised of MP, SC, FR, and RA and the last one only contains MP and WH. At resolution $\delta^*_2=0.886$ we observe that the three-node cluster OG-PC-CO, although it contains more nodes than any other cluster, it is not the most important of the quasi-partition. Instead, the singleton cluster MP has influence over the three-node cluster and, on top of that, is comparable with every other cluster in the quasi-partition. From resolution $\delta^*_3=0.894$ onwards, after MP joins the red cluster, the cluster with the largest number of nodes coincides with the most important of the quasi-partition. At resolution $\delta^*_4=0.899$ we have a total ordering among the four clusters of the quasi-partition. This is not true for the other three depicted quasi-partitions.

%
\begin{figure*}
\centering
\centerline{\input{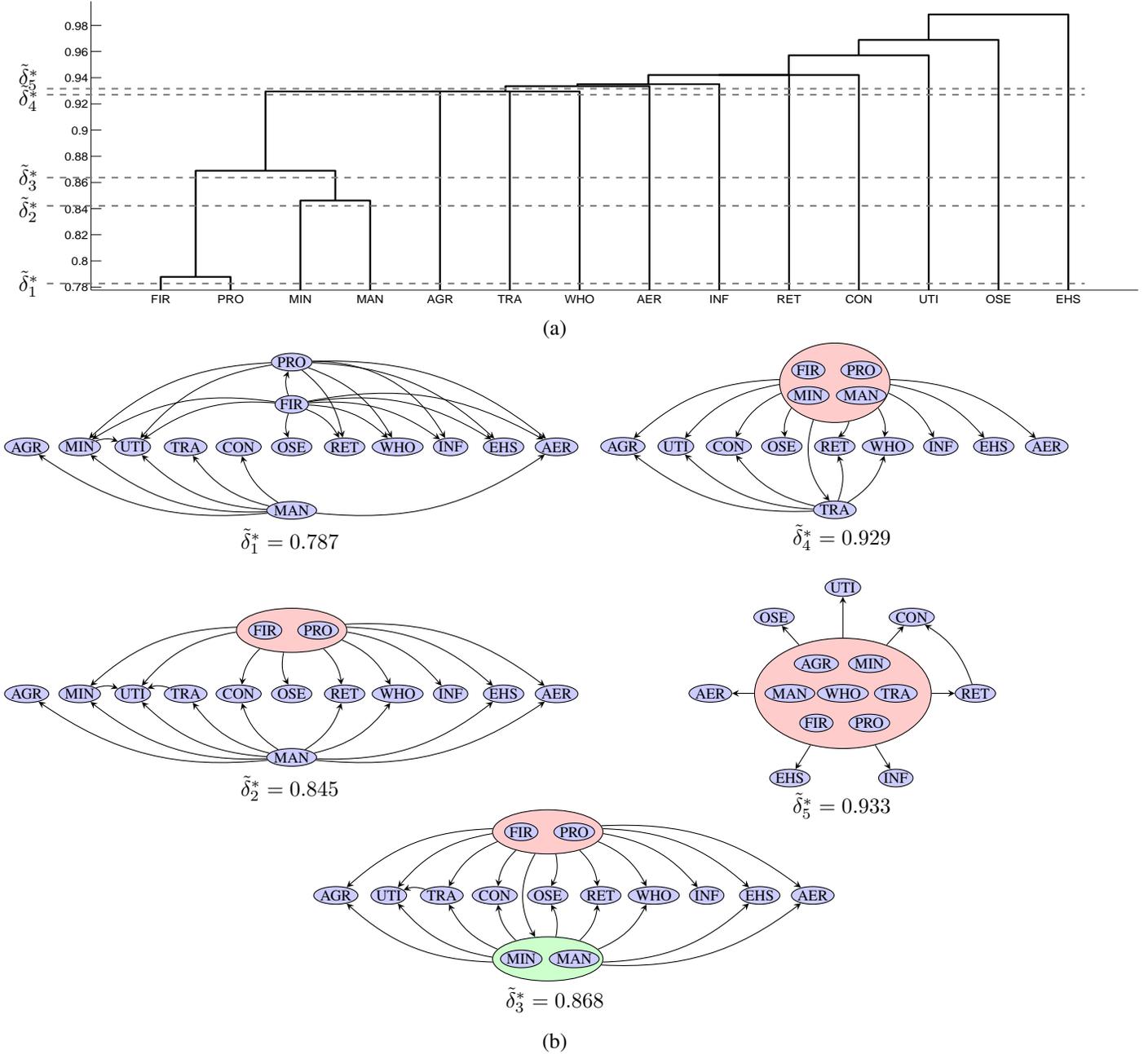} }
\vspace{-0.1in}
\caption{(a) Dendrogram component $D_C^*$ of the quasi-dendrogram $\tdD_C^*=(D_C^*, E_C^*)$. Output of the directed single linkage quasi-clustering method $\tilde{\ccalH}^*$ when applied to the network $N_{C}$. (b) Quasi-partitions. Given by the specification of the quasi-dendrogram $\tdD_C^*$ at a particular resolution $\tdD_C^*(\tilde{\delta}^*_k)$ for $k=1, \ldots, 5$.}
\label{fig_quasi-clustering_example_io}
\vspace{-0.1in}
\end{figure*}

As a further illustration of the quasi-clustering method $\tilde{\ccalH}^*$, we apply this method to the network $N_{C}=(C, A_C)$ of consolidated industrial sectors \cite{USinputoutput} of year 2011 where $|C|=14$ -- see Table \ref{table_consolidated_industrial_sectors} -- instead of the original 61 sectors. To generate the dissimilarity function $A_C$ from the similarity data available in \cite{USinputoutput} we use \eqref{eqn_def_io_dissimilarity}. The outcome of applying the directed single linkage quasi-clustering method $\tilde{\ccalH}^*$ with output quasi-ultrametrics defined in \eqref{eqn_nonreciprocal_chains} to the network $N_C$ is computed with the algorithmic formula in \eqref{eqn_algo_quasi_ultrametric}. Of the output quasi-dendrogram $\tdD^*_C=(D^*_C, E^*_C)$, in Fig. \ref{fig_quasi-clustering_example_io}-(a) we show the dendrogram component $D^*_C$ and in Fig. \ref{fig_quasi-clustering_example_io}-(b) we depict the quasi-partitions $\tdD^*_C(\tilde{\delta}^*_i)$ for $\tilde{\delta}^*_1=0.787$, $\tilde{\delta}^*_2=0.845$, $\tilde{\delta}^*_3=0.868$, $\tilde{\delta}^*_4=0.929$, and $\tilde{\delta}^*_5=0.933$, corresponding to resolutions 0.001 smaller than mergings in the dendrogram $D^*_C$. The reason why we use the consolidated network $N_C$ is to facilitate the visualization of quasi-partitions that capture every sector of the economy instead of only ten particular sectors as in the previous application.


\renewcommand\arraystretch{1.2} 

\begin{table*}
\small
\centering
\caption{Code and description of consolidated industrial sectors}
\vspace{-0.1in}
\begin{tabular}{ l  l p{1cm}  l  l }
  \hline    
Code & Consolidated Industrial Sector                                    &&
Code & Consolidated Industrial Sector                                    \\  \hline   
AER & Arts, entertain., recreation, accomm., and food serv. && 
AGR & Agriculture, forestry, fishing, and hunting \\ 
CON & Construction && 
EHS & Educational services, health care, and social assistance \\ 
FIR & Finance, insurance, real estate, rental, and leasing && 
INF & Information \\ 
MAN & Manufacturing && 
MIN & Mining \\ 
OSE & Other services, except government && 
PRO & Professional and business services \\ 
RET & Retail trade && 
TRA & Transportation and warehousing \\ 
UTI & Utilities && 
WHO & Wholesale trade \\ \hline
\end{tabular}
\label{table_consolidated_industrial_sectors}
\vspace{-0.1in}
\end{table*}

The quasi-dendrogram $\tdD^*_C$ captures the asymmetric influences between clusters of industrial sectors at every resolution. E.g., at resolution $\tilde{\delta}^*_1=0.787$ the dendrogram $D^*_C$ in Fig. \ref{fig_quasi-clustering_example_io}-(a) indicates that every industrial sector forms its own singleton cluster. However, this simplistic representation, characteristic of clustering methods, ignores the asymmetric relations between clusters at resolution $\tilde{\delta}^*_1$. These influence relations are formalized in the quasi-dendrogram $\tdD^*_C$ with the introduction of the edge set $E^*_C(\delta)$ for every resolution $\delta$. In particular, for $\tilde{\delta}^*_1$ we see in Fig. \ref{fig_quasi-clustering_example_io}-(b) that the sectors of `Finance, insurance, real estate, rental, and leasing' (FIR) and `Manufacturing' (MAN) combined have influence over the remaining 12 sectors. More precisely, the influence of FIR is concentrated on the service and commercialization sectors of the economy whereas the influence of MAN is concentrated on primary sectors, transportation, and construction. Furthermore, note that due to the transitivity (QP2) property of quasi-partitions defined in Section \ref{sec_full_characterization_asymmetric}, the influence of FIR over `Professional and business services' (PRO) implies influence of FIR over every sector influenced by PRO. The influence among the remaining 11 sectors, i.e. excluding MAN, FIR and PRO, is minimal, with the `Mining' (MIN) sector influencing the `Utilities' (UTI) sector. This influence is promoted by the influence of the `Oil and gas extraction' (OG) subsector of MIN over the utilities sector as observed in the cluster formed at resolution $\delta^\U_3=0.854$ (orange) by the unilateral clustering method [cf. Fig. \ref{fig_unilateral_example_io}-(b)]. At resolution $\tilde{\delta}^*_2=0.845$, FIR and PRO form one cluster, depicted in red, and they add an influence to the `Construction' (CON) sector apart from the previously formed influences that must persist due to the influence hierarchy property of the the edge set $E_C^*(\delta)$ stated in condition (\~D3) in the definition of quasi-dendrogram in Section \ref{sec_quasi_dendrograms}. The manufacturing sector also intensifies its influences by reaching the commercialization sectors `Retail trade' (RET) and `Wholesale trade' (WHO) and the service sector `Educational services, health care, and social assistance' (EHS). The influence among the rest of the sectors is still scarce with the only addition of the influence of `Transportation and warehousing' (TRA) over UTI. At resolution $\tilde{\delta}^*_3=0.868$ we see that mining MIN and manufacturing MAN form their own cluster, depicted in green. The previously formed red cluster has influence over every other cluster in the quasi-partition, including the green one. At resolution $\tilde{\delta}^*_4=0.929$, the red and green clusters become one, composed of four original sectors. Also, the influence of the transportation TRA sector over the rest is intensified with the appearance of edges to the primary sector `Agriculture, forestry, fishing, and hunting' (AGR), the construction CON sector and the commercialization sectors RET and WHO. Finally, at resolution $\tilde{\delta}^*_5=0.933$ there is one clear main cluster depicted in red and composed of seven sectors spanning the primary, secondary, and tertiary sectors of the economy. This main cluster influences every other singleton cluster. The only other influence in the quasi-partition $\tdD^*_C(0.933)$ is the one of RET over CON. For increasing resolutions, the singleton clusters join the main red cluster until at resolution $\delta=0.988$ the 14 sectors form one single cluster.

The influence structure at every resolution induces a partial order in the blocks of the corresponding quasi-partition. As done in previous examples, we can interpret this partial order as an ordering of relative importance of the elements within each block. E.g., we can say that at resolution $\tilde{\delta}^*_1=0.787$, MAN is more important that MIN which in turn is more important than UTI which is less important that PRO. However, PRO and MAN are not comparable at this resolution. At resolution $\tilde{\delta}^*_4=0.929$, after the red and green clusters have merged together at resolution $\delta=0.869$, we depict the combined cluster as red. This representation is not arbitrary, the red color of the combined cluster is inherited from the most important of the two component cluster. The fact that the red cluster is more important than the green one can be seen from the edge from the former to the latter in the quasi-partition at resolution $\tilde{\delta}^*_3$. In this sense, the edge component $E^*_C$ of the quasi-dendrogram formally provides a hierarchical structure between clusters at a fixed resolution apart from the hierarchical structure across resolutions given by the dendrogram component $D^*_C$ of the quasi-dendrogram. E.g., if we focus only on the dendrogram $D^*_C$ in Fig. \ref{fig_quasi-clustering_example_io}-(a), the nodes MIN and MAN seem to play the same role. However, when looking at the quasi-partitions at resolutions $\tilde{\delta}^*_1$ and $\tilde{\delta}^*_2$, it follows that MAN has influence over a larger set of nodes than MIN and hence plays a more important role in the clustering for increasing resolutions. Indeed, if we delete the three nodes with the strongest influence structure, namely PRO, FIR, and MAN, and apply the quasi-clustering method $\tilde{\ccalH}^*$ on the remaining 11 nodes, the first merging occurs between the mining MIN and utilities UTI sectors at $\delta=0.960$. At this same resolution, in the original dendrogram component in Fig. \ref{fig_quasi-clustering_example_io}-(a), a main cluster composed of 12 nodes only excluding `Other services, except government' (OSE) and EHS is formed. This indicates that by removing influential sectors of the economy, the tendency to co-cluster of the remaining sectors is decreased.


%
\section{Conclusion}\label{sec_conclusion}

Continuing the line of work in \cite{clust-um,CarlssonMemoli10,multi-param}, we have developed a theory for hierarchically clustering asymmetric -- weighted and directed -- networks. Starting from the observation that generalizing methods used to cluster metric data to asymmetric networks is not always intuitive, we identified simple reasonable properties and proceeded to characterize the space of methods that are admissible with respect to them. The properties that we have considered are the following: 

\begin{mylist}
\item[{\it (A1) Axiom of Value.}] In a network with two nodes, the output dendrogram consists of two singleton clusters for resolutions smaller than the maximum of the two intervening dissimilarities and a single two-node cluster for larger resolutions.
\item[{\it (A1') Extended Axiom of Value.}] Define a canonical asymmetric network of $n$ nodes in which the two directed dissimilarities mediating between any given pair of points are the same for any pair of nodes. These two dissimilarity values are allowed to be different.The output dendrogram consists of $n$ singleton clusters for resolutions smaller than the maximum of the two intervening dissimilarities and, consists of a single $n$-node cluster for larger resolutions.
\item[{\it (A1'') Alternative Axiom of Value.}] In a network with two nodes, the output dendrogram consists of two singleton clusters for resolutions smaller than the minimum of the two intervening dissimilarities, and consists of a single two-node cluster for larger resolutions.
\item[{\it (A1\emph{'''}) Agnostic Axiom of Value.}] In a network with two nodes, the output dendrogram consists of two singleton clusters for resolutions smaller than the minimum of the two intervening dissimilarities, and consists of a single two-node cluster for  resolutions larger than their maximum.
\item[{\it (A2) Axiom of Transformation.}] Consider two given networks $N$ and $M$ and a dissimilarity reducing map from the nodes of $N$ to the nodes of $M$, i.e. a map such that dissimilarities between the image nodes in $M$ are smaller than or equal to the corresponding dissimilarities of the pre-image nodes in $N$. Then, the resolution at which any two nodes merge into a common cluster in the network $M$ is smaller than or equal to the resolution at which their pre-images merge in the network $N$.
\item[{\it (P1) Property of Influence.}] For any network with $n$ nodes, the output dendrogram consists of $n$ singleton clusters for resolutions smaller than the minimum loop cost of the network -- the loop cost is the maximum directed dissimilarity when traversing the loop in a given direction, and the minimum loop cost is the cost of the loop of smallest cost.
\item[{\it (P1') Alternative Property of Influence.}] For any network with $n$ nodes, the output dendrogram consists of $n$ singleton clusters for resolutions smaller than the separation of the network -- defined as the smallest positive dissimilarity across all pairs of nodes.
\item[{\it (P2) Stability.}] For any two networks $N$ and $M$, the generalized Gromov-Hausdorff distance between the corresponding output dendrograms is uniformly bounded by the generalized Gromov-Hausdorff distance between the networks.
\end{mylist}

Throughout the paper we identified and described clustering methods satisfying different subsets of the above properties. Several methods were based on finding directed chains of minimum cost, where the chain cost was defined as the maximum dissimilarity encountered when traversing the given chain. The set of clustering methods that we have considered in the present paper is comprised of the following: 

\begin{mylist}
\item[{\it Reciprocal.}] Nodes $x$ and $x'$ are clustered together at a given resolution $\delta$ if there exists a chain linking $x$ to $x'$ such that the directed chain costs are not larger than $\delta$ in either direction.
\item[{\it Nonreciprocal.}] Nodes $x$ and $x'$ are clustered together at a given resolution $\delta$ if there exist two chains, one linking $x$ to $x'$ and the other linking $x'$ to $x$, such that both directed chain costs are not larger than $\delta$ in either direction. In contrast to the reciprocal method, the chains linking $x$ to $x'$ and $x'$ to $x$ may be different.
\item[{\it Grafting.}] Grafting methods are defined by exchanging branches between the reciprocal and nonreciprocal dendrograms as dictated by an exogenous parameter $\beta$. Two grafting methods were studied. In both methods, the reciprocal dendrogram is sliced at resolution $\beta$. In the first method, the branches of resolution smaller than $\beta$ are replaced by the corresponding branches of the nonreciprocal dendrogram. In the second method, the branches of resolution smaller than $\beta$ are preserved and these branches merge either at resolution $\beta$ or at the resolution given by the nonreciprocal dendrogram, whichever is larger.
\item[{\it Convex combinations.}] Given a network $N$ and two clustering methods $\ccalH_1$ and $\ccalH_2$, denote by $D_1$ and $D_2$ the corresponding output dendrograms. Construct a symmetric network $M$ so that the dissimilarities between any pair $(x,x')$ is given by the convex combination of the minimum resolutions at which $x$ and $x'$ are clustered together in $D_1$ and $D_2$. Cluster the network $M$ with the single linkage method to define a valid dendrogram.
\item[{\it Semi-reciprocal.}] A semi-reciprocal chain of index $t \geq 2$ between two nodes $x$ and $x'$ is formed by concatenating directed chains of length at most $t$, called secondary chains, from $x$ to $x'$ and back. The nodes at which secondary chains in both directions concatenate must coincide, although the chains themselves might differ. Nodes $x$ and $x'$ are clustered together at a given resolution $\delta$ if they can be linked by a semi-reciprocal chain of cost not larger than $\delta$.
\item[{\it Algorithmic intermediate.}] Generalizes the semi-reciprocal clustering methods by allowing the maximum length $t$ of secondary chains to be different in both directions.
\item[{\it Unilateral.}] Consider the cost of an undirected chain as one where the edge cost between two consecutive nodes is given by the minimum directed cost in both directions. Nodes $x$ and $x'$ are clustered together at a given resolution $\delta$ if there exists an undirected chain linking $x$ and $x'$ of cost not larger than $\delta$.
\end{mylist}

%
\renewcommand\arraystretch{1.4} 
\begin{table*}\footnotesize\centering
\caption{Summary of methods and properties}
\vspace{-0.1in}
\begin{tabular}{ l l p{0.5mm} c  c  c  c  
                  c  c  c }\hline    
   &&&{\bf Reciprocal}    & {\bf Nonreciprocal}   & Grafting            
   & Convex              & Semi-        & Algorithmic         
         & {\bf Unilateral}   \\[-0.7ex] 
   &&&                        &                &                         
   & combs.              & reciprocal   & intermediate        
   &                         \\ \hline
   \!{\bf(A1)} & \!\!\!\!\!\!\!\! {\bf Axiom of Value}
   && x & x & x & x & x & x &   \\ 
   \!(A1') & \!\!\!\!\!\!\!\! Extended Axiom of Value 
   && x & x & x & x & x & x &   \\ 
   \!{\bf(A1'')} & \!\!\!\!\!\!\!\! {\bf Alt. Axiom of Value}
   &&   &   &   &   &   &   & x \\ 
   \!(A1''') & \!\!\!\!\!\!\!\! Agnostic Axiom of Value
   && x & x & x & x & x & x & x \\ 
   \!{\bf(A2)} & \!\!\!\!\!\!\!\! {\bf Axiom of Transformation}
   && x & x & x & x & x & x & x \\ 
   \!(P1) & \!\!\!\!\!\!\!\! Property of Influence
   && x & x & x & x & x & x &   \\ 
   \!(P1') & \!\!\!\!\!\!\!\! Alt. Property of Influence
   && x  &  x &  x &  x &  x &  x & x \\ 
   \!(P2) & \!\!\!\!\!\!\!\! Stability\!
   && x & x &   &   & x & x & x \\ \hline
\end{tabular}
\label{table_summary_methods_properties}
\vspace{-0.1in}
\end{table*}

%
We can build a taxonomy of this paper from the perspective of axioms and properties and an intertwined taxonomy from the perspective of clustering methods as we summarize in Table \ref{table_summary_methods_properties} and elaborate in the following sections. 

%
\subsection{Taxonomy of axioms and properties}\label{sec_conclusions_axiom_taxonomy}

The taxonomy from the perspective of axioms and properties is encoded in the rows in Table \ref{table_summary_methods_properties}. For most of the paper, the axioms of value (A1) and transformation (A2) were requirement for admissibility. All of the methods enumerated above satisfy the Axiom of Transformation whereas all methods, except for unilateral clustering, satisfy the Axiom of Value. Although seemingly weak, (A1) and (A2) are a stringent source of structure. E.g., we showed   that admissibility with respect to (A1) and (A2) is equivalent to admissibility with respect to the apparently stricter conditions given by the Extended Axiom of Value (A1') combined with (A2). Likewise, we showed that the Property of Influence (P1) is implied by (A1) and (A2). This latter fact can be interpreted as stating that the requirement of bidirectional influence in two-node networks combined with the Axiom of Transformation implies a requirement for loops of influence in all networks. Given that (A1') and (P1) are implied by (A1) and (A2) and that all methods except for unilateral clustering satisfy (A1) and (A2) it follows that all methods other than unilateral clustering satisfy  (A1') and (P1) as well.

The Alternative Axiom of Value (A1'') is satisfied by unilateral clustering only, which is also the unique method listed above that satisfies the Alternative Property of Influence (P1') but does not satisfy the (regular) Property of Influence. We have also proved that (P1') is implied by (A1'') and (A2) in the same manner that (P1) is implied by (A1) and (A2). Since the Agnostic Axiom of Value (A1''') encompasses (A1) and (A1'') all of the methods listed above satisfy (A1''').

To study stability, we adopted the Gromov-Hausdorff distance. This adopted distance was shown to be properly defined which therefore allows the quantification of differences between networks. Since output dendrograms are equivalent to finite ultrametric spaces which in turn are particular cases of networks, this distance can be used to compare both the given networks and their corresponding output ultrametrics. The notion of stability of a given method that we adopted is that the distance between two outputs produced by the given hierarchical clustering method is bounded by the distance between the original networks. This means that clustering methods are non-expansive maps in the space of networks, i.e. they do not increase the distance between the given networks. An intuitive interpretation of the stability property is that similar networks yield similar dendrograms. The Stability Property (P2) is satisfied by reciprocal, nonreciprocal, semi-reciprocal, algorithmic intermediates, and unilateral clustering methods. The grafting and convex combination families are not stable in this sense.

%
\subsection{Taxonomy of methods}

A classification from the perspective of methods follows from reading the columns in Table \ref{table_summary_methods_properties}. This taxonomy is more interesting than the one in Section \ref{sec_conclusions_axiom_taxonomy} because the reciprocal, nonreciprocal, and unilateral methods not only satisfy desirable properties but have also been proved to be either extremal or unique among those methods that are admissible with respect to some subset of properties. 

Indeed, reciprocal $\ccalH^\R$ and nonreciprocal $\ccalH^{\NR}$ clustering were shown to be extremes of the range of methods that satisfy (A1)-(A2) in that the clustering outputs of these two methods provide uniform upper and lower bounds, respectively, for the output of every other method under this axiomatic framework. These two methods also satisfy all the other desirable properties that are compatible with (A1). I.e., they satisfy the extended and agnostic axioms of value, the Property of Influence, and, implied by it, the Alternative Property of Influence. They are also stable in terms of the generalized Gromov-Hausdorff distance.

Unilateral clustering $\ccalH^\U$ is the unique method that abides by the alternative set of axioms (A1'')-(A2). In that sense it plays the dual role of reciprocal and nonreciprocal clustering when we replace the Axiom of Value (A1) with the Alternative Axiom of Value (A1''). Unilateral clustering also satisfies all the desirable properties that are compatible with (A1''). It satisfies the Agnostic Axiom of Value, the Alternative Property of Influence, and Stability.

In this paper, unilateral clustering  $\ccalH^\U$ and reciprocal $\ccalH^\R$ were shown to be extremal among methods that are admissible with respect to (A1''')-(A2). Unilateral clustering yields uniformly minimal ultrametric distances, while reciprocal clustering yields uniformly maximal ultrametric distances. 

We also considered families of intermediate methods that yield ultrametrics that lie between the outputs of reciprocal and nonreciprocal clustering. The first such family considered is that of grafting methods $\ccalH^{\R/\NR}(\beta)$ and $\ccalH^{\R/\R_{\max}}(\beta)$. They satisfy the axioms and properties that can be derived from (A1)-(A2), i.e. the Extended Axiom of Value, the Agnostic Axiom of Value, the Property of Influence and the Alternative Property of Influence. Their dependance on a cutting parameter $\beta$ is the reason why they fail to fulfill Stability, hence, impairing practicality of these methods.
Convex combination clustering methods $\ccalH^{12}_\theta$ constitute another family of intermediate methods considered in this paper. Their admissibility is based on the result that the convex combination of two admissible methods is itself an admissible clustering method. However, although not proved here, one can show that the convex combination operation does not preserve Stability in general.
Semi-reciprocal clustering methods $\ccalH^{\SR(t)}$ allow the formation of cyclic influences in a more restrictive way than nonreciprocal clustering but more permissive than reciprocal clustering, controlled by the integer parameter $t$. Semi-reciprocal clustering methods were shown to be stable. 
Algorithmic intermediate clustering methods $\ccalH^{t, t'}$ are a generalization of semi-reciprocal methods and share their same properties.

%
\renewcommand{\arraystretch}{2.55}
\begin{table*}\footnotesize\centering
\caption{Summary of Algorithms}
\vspace{-0.1in}
\begin{tabular}{ l  l  l  l }\hline 
   
   Method & Observations & Notation & Formula \\  \hline
   {\bf Reciprocal} 
       &
       & $u^{\R}_X $&$ \Big(\max\left( {A}_X, A_X^T \right)\Big)^{(n-1)}$ 
       \\ 
   {\bf Nonreciprocal}             
       & 
       & $u^{\NR}_X $&$\max\left({A}_X^{(n-1)},\left(A_X^T\right)^{(n-1)}\right)$   
       \\ 
   Grafting 
       & Reciprocal/nonreciprocal                        
       & $u^{\R/\NR}_X(\beta) $&$   u_X^{\NR}\circ\ind{u_X^{\R}\leq\beta}  
                       + u_X^{\R} \circ\ind{u_X^{\R}>\beta}$   \\
   Convex Combinations             
       & Given $\ccalH^1$ and $\ccalH^2$ 
       &  $u^{12}_X(\theta)$&$ \Big(\theta\, u^1_X + (1-\theta) \, u^2_X\Big)^{(n-1)}$   
       \\
   Semi-reciprocal ($t$)             
       & Secondary chains of length $t$
       &  $u_X^{\SR(t)} $&$ \left(\max\left({A}_X^{(t-1)},
                 \left(A_X^T\right)^{(t-1)}\right)\right)^{(n-1)}$  
       \\
   Algorithmic intermediate 
       & Given parameters $t$ and $t'$       
       & $u^{t,t'}_X $&$ \left(\max\left({A}_X^{(t)},
               \left(A_X^T\right)^{(t')}\right)\right)^{(n-1)} $   
       \\
   {\bf Unilateral}    
       &            
       & $u^{\U}_X $&$ \Big(\min\left( {A}_X, A_X^T \right)\Big)^{(n-1)}$   \\ \hline
   {\bf Directed single linkage}
       & Quasi-clustering
       & $\tdu_X^*$&$A_X^{(n-1)}$   \\\hline
   Single linkage
       & Symmetric networks
       & $u^{\SL}_X $&$ A_X^{(n-1)}$   \\\hline
\end{tabular}
\label{table_summary_algorithms}
\vspace{-0.1in}
\end{table*}

\subsection{Algorithms and applications to real datasets}

Algorithms for the application of the methods described throughout the paper were developed using the min-max dioid algebra $\mathfrak{A}$ on the extended nonnegative reals. In this algebra, the regular sum is replaced by the minimization operator and the regular product by maximization. In this algebra, the $k$-th power of the dissimilarity matrix was shown to contain in position $i,j$ the minimum chain cost corresponding to going from node $i$ to node $j$ in at most $k$ hops. Since chain costs played a major role in the definition of clustering methods, dioid matrix powers were presented as a natural framework for algorithmic development.

The reciprocal ultrametric was computed by first symmetrizing directed dissimilarities to their maximum and then computing increasing powers of the symmetrized dissimilarity matrix until stabilization. For the nonreciprocal case, the opposite was shown to be true, i.e., we first take successive powers of the asymmetric dissimilarity matrix until stabilization and then symmetrize the result via a maximum operation. The opposite nature of both algorithms illustrated the extremal properties of reciprocal and nonreciprocal clustering in the algorithmic domain. In a similar fashion, algorithms for the remaining clustering methods presented throughout the paper were developed in terms of finite matrix powers, thus exhibiting computational tractability of our clustering constructions. A summary of all the algorithms presented in this paper is available in Table \ref{table_summary_algorithms}.

Clustering algorithms were applied to two real-world networks. We gained insight about migrational preferences of individuals within United States by clustering a network of internal migration. In addition, we applied the developed theory to a network containing information about how sectors of the U.S. economy interact to generate gross domestic product. In this way, we learned about economic sectors exhibiting pronounced interdependence and reasoned their relation with the rest of the economy.

The clusters appearing in the reciprocal dendrogram of the migration network revealed that population movements are dominated by geographical proximity. In particular, the reciprocal dendrogram showed that the strongest bidirectional migration flows correspond to pairs of states sharing urban areas. E.g., Minnesota and Wisconsin formed a tight cluster due to the spillover of Minneapolis and Duluth's suburbs into Wisconsin, and Illinois joined with Indiana because of the southern reaches of Chicago. As we looked for clusters at coarser resolutions, a separation between larger geographical regions such as East, West, Midwest, and New England could be observed. The two exceptions to geographical proximity were Texas that clustered with the West Coast states and Florida that clustered with the Northeast states. The relative isolation of New England and the state pairs Arkansas-Oklahoma and Idaho-Utah was observed in their persistence as clusters for very high resolutions. 

For this particular dataset the outputs of the reciprocal and nonreciprocal dendrograms were very similar, being indicative of the rarity of migrational cycles. Combining this observation with the fact that reciprocal and nonreciprocal clustering are uniform lower and upper bounds on all methods that satisfy (A1) and (A2), it further follows that all the methods that satisfy these axioms yield similar clustering outputs. Unilateral clustering is the only hierarchical clustering method included here that does {\it not} satisfy these axioms. Its application to the migration network revealed regional separations more marked than the ones that appeared in the reciprocal and nonreciprocal dendrograms. For coarse resolutions, we observed a clear East-West separation along the west borders of Michigan, Ohio, Kentucky, Tennessee, and Missouri. For finer resolutions we observed clustering around the most populous states. The West clustered around California, the South around Texas, The Southeast around Florida, the Northeast and New England around New York, Appalachia around Virginia, and the Midwest around Illinois. This latter pattern is indicative of the ability of unidirectional clustering to capture the unidirectional influence of the populous states on the smaller ones -- as opposed to the methods that satisfy (A1)-(A2), which capture bidirectional influence. To study the influence between states, we applied the directed single linkage quasi-clustering method, revealing the dominant roles of California and Massachusetts in the population influxes into the West Coast and New England, respectively.

For the network of interactions between sectors of the U.S. economy, in contrast to the migration network, the reciprocal and nonreciprocal dendrograms uncovered different clustering structures. The reciprocal dendrogram generated distinctive clusters of sectors that have significant interactions. These include a cluster of service sectors such as financial, professional, insurance, and support services; a cluster of extractive industries such as mining, primary metals, and oil and gas extraction; and a cluster formed by farms, forestry, food, and wood processing. As is required for the formation of clusters when reciprocal clustering is applied, sectors in these clusters use as inputs large fractions of each other's outputs. The nonreciprocal dendrogram did not output distinctive separate clusters but rather a single cluster around which sectors coalesced as the resolution coarsened. This cluster started with the sectors oil and gas, petroleum and coal products, and construction as the tightest coupled triplet to which support services were then added, with financial services then joining the group and so on. This pattern indicates that considering cycles of influence yields a different understanding of interactions between sectors of the U.S. economy than what can be weaned from the direct mutual influence required by reciprocal clustering. We further observed that allowing cycles of arbitrary length generates clusters based on rather convoluted influence structures. An intermediate picture that allows cycles of restricted length was obtained by use of the semi-reciprocal method with parameter 3. This method recognizes the importance of cycles by allowing cyclic influences involving at most three sectors in each direction but discards intricate influences created by longer cycles. Unilateral clustering yielded clusters that group around large sectors of the economy. This is akin to the agglomeration around populous states observed in the case of the migration network. We finally considered the use of the directed single linkage quasi-clustering method to understand influences between economic sectors. This analysis revealed the dominant influence of energy, manufacturing, and financial and professional services over the rest of the economy.

\subsection{Symmetric networks and asymmetric quasi-ultrametrics}\label{sec_conclusion_other_results}

In hierarchical clustering of asymmetric networks we output a symmetric ultrametric to summarize information about the original asymmetric structure. As a particular case, we considered the construction of symmetric ultrametrics when the original network is symmetric. As a generalization, we studied the problem of defining and constructing asymmetric ultrametrics associated with asymmetric networks.

By restricting our general results to the particular case of symmetric networks, we strengthened the uniqueness result from \cite{clust-um,CarlssonMemoli10} which showed that single linkage is the unique admissible clustering method on finite metric spaces under a framework determined by three axioms. In the current paper, we showed that single linkage is the unique admissible method for symmetric networks -- a superset of metric spaces -- in a framework determined only by two axioms, i.e. the Symmetric Axiom of Value (B1) and the Axiom of Transformation (A2), out of the three axioms considered in \cite{clust-um,CarlssonMemoli10}.

Hierarchical clustering methods output dendrograms, which are symmetric data structures. When clustering asymmetric networks, requiring the output to be symmetric might be undesirable. In this context we defined quasi-dendrograms, a generalization of dendrograms that admits asymmetric relations, and developed a theory for quasi-clustering methods, i.e. methods that output quasi-dendrograms when applied to asymmetric networks. In this context, we revised the notion of admissibility by introducing the Directed Axiom of Value (\~A1) and the Directed Axiom of Transformation (\~A2). Under this framework, we showed that directed single linkage -- an asymmetric version of the single linkage clustering method -- is the unique admissible method. Furthermore, we proved an equivalence between quasi-dendrograms and quasi-ultrametrics that generalizes the known equivalence between dendrograms and ultrametrics. Algorithmically, the quasi-ultrametric produced by directed single linkage can be computed by applying iterated min-max matrix power operations to the dissimilarity matrix of the network until stabilization.

Directed single linkage can be used to understand relationships that cannot be understood when performing (standard) hierarchical clustering. In particular, directed influences between clusters of a given resolution define a partial order between clusters which permits making observations about the relative importances of different clusters. This was corroborated through the application of directed single linkage to the United Stated internal migration network. Regular hierarchical clustering uncovers the grouping of California with other West Coast states and the grouping of Massachusetts with other New England States. Directed single linkage shows that California is the dominant state in the West Coast whereas Massachusetts appears as the dominant state in New England. When applied to the network of interactions between sectors of the United States economy, directed single linkage revealed the prominent influence of manufacturing, finance and professional services over the rest of the economy.

\subsection{Future developments}

In order to winnow the admissible space of methods that satisfy the axioms of value (A1) and transformation (A2), one can require additional properties to be fulfilled by these methods. The property of stability, discussed in this paper, is a first step in this direction. Further desirable properties will be considered in future work including scale invariance, representability, and excisiveness.

Scale invariance is defined by the requirement that the formation of clusters does not depend on the scale used to measure dissimilarities. Representability, a concept introduced in \cite{CarlssonMemoli10}, is an attempt to characterize methods that are described through the specification of their effect over particular exemplar networks thus giving rise to generative models for clustering methods. Excisiveness \cite{CarlssonMemoli10} encodes the property that clustering a previously clustered network does not generate new clusters. 
By further restricting the space of methods when imposing these additional properties, we aim at achieving a full characterization of the space of hierarchical clustering methods.

\begin{appendices}


\section{Proofs in Section \ref{sec_reicprocal_and_nonreciprocal}}\label{appendix_sec_reicprocal_and_nonreciprocal}

\begin{myproof}[of Proposition \ref{prop_nonreciprocal_axioms}] That $\ccalH^{\NR}$ outputs valid ultrametrics was already argued prior to the statement of Proposition \ref{prop_nonreciprocal_axioms}. The proof of admissibility is analogous to the proof of Proposition \ref{prop_reciprocal_axioms} and presented for completeness. For Axiom (A1) notice that for the two-node network $\vec{\Delta}_2(\alpha,\beta)$ we have $\tdu^*_{p, q}(p,q)=\alpha$ and $\tdu^*_{p, q}(q,p)=\beta$ because there is only one possible chain selection. According to \eqref{eqn_nonreciprocal_clustering} we then have
\begin{equation}\label{eqn_theo_nonreciprocal_axioms_pf_10}
    u^{\NR}_{p, q}(p,q) = \max \Big( \tdu^*_{p, q}(p,q), \tdu^*_{p, q}(q,p)\Big)
                 = \max(\alpha,\beta).
\end{equation} 
To prove that Axiom (A2) is satisfied consider arbitrary points $x,x' \in X$ and denote by $C^*(x, x')$ one chain achieving the minimum chain cost in \eqref{eqn_nonreciprocal_chains},
\begin{align}\label{eqn_theo_nonreciprocal_axioms_pf_30} 
   \tdu^*_X(x, x') = \max_{i | x_i\in C^*(x,x')} A(x_i,x_{i+1}).
\end{align} 

Consider the transformed chain $C_Y(\phi(x),\phi(x'))=[\phi(x)=\phi(x_0),\ldots, \phi(x_l)=\phi(x')]$ in the space $Y$. Since the map $\phi:X\to Y$ reduces dissimilarities we have that for all links in this chain $A_Y(\phi(x_i),\phi(x_{i+1}))\leq A_X(x_i,x_{i+1})$. Consequently,
\begin{align}\label{eqn_theo_nonreciprocal_axioms_pf_40}
    &\max_{i | x_i\in C_Y(\phi(x),\phi(x'))} A_Y(\phi(x_i),\phi(x_{i+1})) 
                       \\\nonumber &\hspace{40mm}
           \leq \max_{i | x_i\in C^*(x,x')} A_X(x_i,x_{i+1}).
\end{align}
Further note that the minimum chain cost $\tdu^*_Y(\phi(x), \phi(x'))$ among all chains linking $\phi(x)$ to $\phi(x')$ cannot exceed the cost in the given chain $C_Y(\phi(x),\phi(x'))$. Combining this observation with the inequality in \eqref{eqn_theo_nonreciprocal_axioms_pf_40} it follows that
\begin{align}\label{eqn_theo_nonreciprocal_axioms_pf_50} 
   \tdu^*_Y(\phi(x),\phi(x')) \leq \max_{i | x_i\in C^*(x,x')} A_X(x_i,x_{i+1})
                                 = \tdu^*_X(x, x'),
\end{align} 
where we also used \eqref{eqn_theo_nonreciprocal_axioms_pf_30} to write the equality.

The bound in \eqref{eqn_theo_nonreciprocal_axioms_pf_50} is true for arbitrary ordered pair $(x,x')$. In particular, it is true if we reverse the order to consider the pair $(x',x)$. Consequently, we can write
\begin{align}\label{eqn_theo_nonreciprocal_axioms_pf_60}
   \max \Big(\tdu^*_Y(\phi(x),&\phi(x')),\ \tdu^*_Y(\phi(x'),\phi(x))\Big) \nonumber \\
     &  \leq \max \Big(\tdu^*_X(x, x'),\ \tdu^*_X(x', x)\Big),
\end{align} 
because both maximands in the left are smaller than their corresponding maximand in the right. To complete the proof just notice that the expressions in \eqref{eqn_theo_nonreciprocal_axioms_pf_60} correspond to the nonreciprocal ultrametric distances $u^{\NR}_Y(\phi(x),\phi(x'))$ and $u^{\NR}_X(x, x')$ [cf. \eqref{eqn_nonreciprocal_clustering}]. Thus we have that for a dissimilarity reducing map $\phi:X\to Y$ the nonreciprocal ultrametric distances satisfy $u^{\NR}_Y(\phi(x),\phi(x'))\leq u^{\NR}_X(x, x')$ as required by Axiom (A2) [cf. \eqref{eqn_dissimilarity_reducing_ultrametric}].
\end{myproof}

\section{Proofs in Section \ref{sec_intermediate_ultrametrics}}\label{appendix_sec_intermediate_ultrametrics}

\begin{myproof}[of Proposition \ref{prop_beta_1}]
The function $u^{\R/\NR}_X(\beta)$ fulfills the symmetry $u^{\R/\NR}_X(x,x'; \beta)=u^{\R/\NR}_X(x',x;\beta)$, non negativity and identity $u^{\R/\NR}_X(x,x';\beta)=0 \Leftrightarrow x=x'$ properties because $u^{\NR}_X$ and $u^{\R}_X$ fulfill them separately. Hence, to show that $u^{\R/\NR}_X(\beta)$ is a properly defined ultrametric, we need to show that it satisfies the strong triangle inequality (\ref{eqn_strong_triangle_inequality}).
%
 %
To show this, we split the proof into two cases: $u^{\R}_X(x,x') \leq \beta$ and $u^{\R}_X(x,x') > \beta$. Note that, by definition,
\begin{equation}\label{inter_beta_1}
u^{\NR}_X(x,x') \leq u^{\R/\NR}_X(x,x';\beta) \leq u^{\R}_X(x,x').
\end{equation}
Starting with the case where $u^{\R}_X(x,x') \leq \beta$, since $u^{\NR}_X$ satisfies (\ref{eqn_strong_triangle_inequality}) we can state that,
\begin{align}\label{beta_1_1}
u^{\R/\NR}_X(x,x';\beta)=&u^{\NR}_X(x,x') \nonumber\\
 \leq & \max \Big( u^{\NR}_X(x,x'') \, , \, u^{\NR}_X(x'',x')\Big).
\end{align}
Using the lower bound inequality in (\ref{inter_beta_1}) we can write 
\begin{align}\label{beta_1_2}
\max \Big( u^{\NR}_X&(x,x'') \, , \, u^{\NR}_X(x'',x')\Big) \nonumber \\
 \leq &\max \Big( u^{\R/\NR}_X(x,x'';\beta) \, , \, u^{\R/\NR}_X(x'',x';\beta)\Big).
\end{align}
Combining (\ref{beta_1_1}) and (\ref{beta_1_2}), we obtain 
\begin{equation}\label{beta_1_2_bis}
u^{\R/\NR}_X(x,x';\beta) \leq \max \Big( u^{\R/\NR}_X(x,x'';\beta) \, , \, u^{\R/\NR}_X(x'',x';\beta)\Big),
\end{equation}
which implies that $u^{\R/\NR}_X(\beta)$ fulfills the strong triangle inequality in this case.
 
In the second case, suppose that $u^{\R}_X(x,x') > \beta$, from the validity of the strong triangle inequality (\ref{eqn_strong_triangle_inequality}) for $u^{\R}_X$, we can write 
\begin{align}\label{beta_1_3}
\beta<u^{\R/\NR}_X(x,x';\beta)&=u^{\R}_X(x,x') \nonumber \\
&\leq \max \Big( u^{\R}_X(x,x'') \, , \, u^{\R}_X(x'',x')\Big).
\end{align}
This implies that at least one of $u^{\R}_X(x,x'')$ and $u^{\R}_X(x'',x')$ is greater than $\beta$. When this occurs, $u^{\R/\NR}_X(\beta)=u^{\R}_X$. Hence, 
\begin{align}\label{beta_1_4}
\max \Big( u^{\R}_X&(x,x'') \, , \, u^{\R}_X(x'',x')\Big) \nonumber \\
= \max &\Big( u^{\R/\NR}_X(x,x'';\beta) \, , \, u^{\R/\NR}_X(x'',x';\beta) \Big).
\end{align}
By substituting (\ref{beta_1_4}) into (\ref{beta_1_3}), we can justify the same inequality as in (\ref{beta_1_2_bis}) for this second case. Since the two cases studied include all possible situations, we can conclude that $u^{\R/\NR}_X(\beta)$ always satisfies the strong triangle inequality.

To show that $\ccalH^{\R/\NR}(\beta)$ satisfies Axiom (A1) it suffices to see that in a two-node network $u^{\NR}_X$ and $u^{\R}_X$ coincide, meaning that we must have $u^{\R/\NR}_X(\beta) = u^{\NR}_X = u^{\R}_X$. Since $\ccalH^{\R}$ and $\ccalH^{\NR}_X$ fulfill (A1), the clustering method $\ccalH^{\R/\NR}(\beta)$ must satisfy (A1) as well. 

To prove (A2) consider a dissimilarity reducing map $\phi:X\to Y$ as defined in Section \ref{sec_axioms} and split consideration with regards to whether the reciprocal ultrametric is $u^{\R}_X(x,x') \leq \beta$ or $u^{\R}_X(x,x') > \beta$. When $u^{\R}_X(x,x') \leq \beta$ we must have $u^{\R}_Y(\phi(x),\phi(x')) \leq \beta$ because $\ccalH^{\R}$ satisfies (A2) and $\phi$ is a dissimilarity reducing map. Hence, according to the definition in (\ref{def_mu_beta_1}) we must have that both $u^{\R/\NR}_X(x,x';\beta)$ and $u^{\R/\NR}_Y(\phi(x),\phi(x');\beta)$ coincide with the nonreciprocal ultrametric 
\begin{align}\label{beta_1_6}
   u^{\R/\NR}_X(x,x';\beta)&=u^{\NR}_X(x,x'), \nonumber\\
   u^{\R/\NR}_Y(\phi(x),\phi(x');\beta)&=u^{\NR}_Y(\phi(x),\phi(x')). 
\end{align}
Since $\ccalH^{\NR}$ satisfies (A2) it is an immediate consequence of the equalities in \eqref{beta_1_6} that
\begin{equation}\label{beta_1_6_bis}
u^{\R/\NR}_X(x,x';\beta) \geq u^{\R/\NR}_Y(\phi(x),\phi(x');\beta).
\end{equation}
This means that $\ccalH^{\R/\NR}(\beta)$ satisfies Axiom (A2) when $u^{\R}_X(x,x') \leq \beta$. 

In the second case, when $u^{\R}_X(x,x') > \beta$, the validity of (A2) for the reciprocal ultrametric $u^{\R}_X$ allows us to write
\begin{equation}\label{beta_1_7}
u^{\R/\NR}_X(x,x';\beta)=u^{\R}_X(x,x') \geq u^{\R}_Y(\phi(x),\phi(x')).
\end{equation}
Applying the fact that $u^{\R}_Y$ is an upper bound on $u^{\R/\NR}_Y(\beta)$ (\ref{inter_beta_1}), we have
\begin{equation}\label{beta_1_8}
u^{\R}_Y(\phi(x),\phi(x')) \geq u^{\R/\NR}_Y(\phi(x),\phi(x');\beta).
\end{equation}
By combining (\ref{beta_1_7}) and (\ref{beta_1_8}), we can obtain an equation analogous to (\ref{beta_1_6_bis}) for the second case. This proves the fulfillment of (A2) by $\ccalH^{\R/\NR}(\beta)$ in the general case.
\end{myproof}

\begin{myproof}[of Proposition \ref{prop_beta_5}]
As in Proposition \ref{prop_beta_1}, to prove that $u^{\R/\R_{\max}}_X(\beta)$ is properly defined, it suffices to show the strong triangle inequality (\ref{eqn_strong_triangle_inequality}). To show this, we divide the proof into two cases: $u^{\R}_X(x,x') \leq \beta$ and $u^{\R}_X(x,x') > \beta$. Note that, by definition,
\begin{equation}\label{inter_beta_5}
u^{\NR}_X(x,x') \leq u^{\R/\R_{\max}}_X(x,x';\beta) \leq u^{\R}_X(x,x').
\end{equation}

In the case where $u^{\R}_X(x,x') \leq \beta$, recalling the strong triangle inequality (\ref{eqn_strong_triangle_inequality}) validity on $u^{\R}_X$, we can assert that 
\begin{align}\label{beta_5_1}
u^{\R/\R_{\max}}_X&(x,x';\beta) = u^{\R}_X(x,x') \nonumber \\
&\leq \min \Big(\beta , \max \big(u^{\R}_X(x,x'') \, , \, u^{\R}_X(x'',x') \big)\Big).
\end{align}
Using the definition (\ref{def_mu_beta_5}), one can say 
\begin{align}\label{beta_5_2}
\min &\Big( \beta , \max \big(u^{\R}_X(x,x'') \, , \, u^{\R}_X(x'',x') \big)\Big) \nonumber \\
&\leq \max \Big( u^{\R/\R_{\max}}_X(x,x'';\beta) \, , \, u^{\R/\R_{\max}}_X(x'',x';\beta)\Big).
\end{align}
The combination of (\ref{beta_5_1}) and (\ref{beta_5_2}) leads to 
\begin{align}\label{beta_5_2_bis}
u^{\R/\R_{\max}}&_X(x,x';\beta) \nonumber \\
\leq \max & \Big( u^{\R/\R_{\max}}_X(x,x'';\beta) \, , \, u^{\R/\R_{\max}}_X(x'',x';\beta)\Big),
\end{align}
which shows the strong triangle inequality in this first case.

In the case where $u^{\R}_X(x,x') > \beta$, using the definition (\ref{def_mu_beta_5}) and the strong triangle inequality applied to $u^{\NR}_X$, we get
\begin{align}\label{beta_5_3}
u^{\R/\R_{\max}}_X&(x,x';\beta) = \max \big( \beta , u^{\NR}_X(x,x') \big) \nonumber  \\
&\leq \max \Big( \beta , \max \big(u^{\NR}_X(x,x'') , u^{\NR}_X(x'',x') \big) \Big).
\end{align}
However, since $u^{\R}_X(x,x') > \beta$ from the strong triangle inequality applied to $u^{\R}_X$ we know that either $u^{\R}_X(x,x'') > \beta$ or $u^{\R}_X(x'',x') > \beta$. This implies that,
\begin{align}\label{beta_5_4}
\max &\Big( \beta , \max \big(u^{\NR}_X(x,x'') , u^{\NR}_X(x'',x') \big) \Big)  \nonumber \\
&= \max \Big(u^{\R/\R_{\max}}_X(x,x'';\beta) , u^{\R/\R_{\max}}_X(x'',x';\beta) \Big).
\end{align}
By substituting (\ref{beta_5_4}) into (\ref{beta_5_3}) we obtain a result analogous to (\ref{beta_5_2_bis}) for this second case. This proves that $u^{\R/\R_{\max}}_X(\beta)$ fulfills the strong triangle inequality.


The proof that $\ccalH^{\R/\R_{\max}}(\beta)$ satisfies (A1) is identical to the proof in Proposition \ref{prop_beta_1} and is based on (\ref{inter_beta_5}), so we omit it.

Finally, we divide the proof that $\ccalH^{\R/\R_{\max}}(\beta)$ satisfies (A2) into the cases where $u^{\R}_X(x,x') \leq \beta$ and $u^{\R}_X(x,x') > \beta$. Consider a dissimilarity reducing map $\phi:X\to Y$ as defined in Section \ref{sec_axioms}. In the first case, if $u^{\R}_X(x,x') \leq \beta$ then $u^{\R}_Y(\phi(x),\phi(x')) \leq \beta$ since $\ccalH^{\R}$ satisfies (A2). Hence, 
\begin{equation}\label{beta_5_5}
u^{\R/\R_{\max}}_X(x,x';\beta)=u^{\R}_X(x,x'), \nonumber\\
\end{equation}
\begin{equation}\label{beta_5_6}
u^{\R/\R_{\max}}_Y(\phi(x),\phi(x');\beta)=u^{\R}_Y(\phi(x),\phi(x')). 
\end{equation}
Since $\ccalH^{\R}$ satisfies (A2), we can conclude that,
\begin{equation}\label{beta_5_7}
u^{R/R_{\max}}_X(x,x';\beta) \geq u^{R/R_{\max}}_Y(\phi(x),\phi(x');\beta),
\end{equation}
showing the fulfillment of Axiom (A2) in this first case. 

In the case where $u^{\R}_X(x,x') > \beta$, we apply definition (\ref{def_mu_beta_5}) and the fact that $\ccalH^{\NR}$ satisfies (A2) to get
\begin{align}\label{beta_5_8}
u^{\R/\R_{\max}}_X(x,x';\beta) &= \max \big(\beta\, ,\, u^{\NR}_X(x,x')\big) \nonumber \\
&\geq \max \big(\beta\, ,\, u^{\NR}_Y(\phi(x),\phi(x'))\big).
\end{align}
However, the piecewise definition (\ref{def_mu_beta_5}), implies that, 
\begin{equation}\label{beta_5_9}
\max \big(\beta\, ,\, u^{\NR}_Y(\phi(x),\phi(x'))\big) \geq u^{\R/\R_{\max}}_Y(\phi(x),\phi(x');\beta).
\end{equation}
By substitution of (\ref{beta_5_9}) into (\ref{beta_5_8}), an analogous result to (\ref{beta_5_7}) can be shown for the second case. This proves the fulfillment of (A2) in the general case.
\end{myproof}


\begin{myproof}[of Proposition \ref{prop_convex_combination}]
We need to show that (1) $u^{12}_X(\theta)$ is a valid ultrametric and (2) that the method $\ccalH^{12}_\theta$ satisfies (A1) and (A2). As discussed in the paragraph preceding the statement of this proposition, $u^{12}_X(\theta)$ is the output of applying single linkage clustering method to the symmetric network $N^{12}_\theta$. Hence, $u^{12}_X(\theta)$ is well defined.

To see that axiom (A1) is fulfilled, pick an arbitrary two-node network $\vec{\Delta}_2(\alpha,\beta)=(\{p,q\}, A_{p, q})$ with $A_{p, q}(p, q)=\alpha$ and $A_{p, q}(q, p)=\beta$. Since methods $\ccalH^1$ and $\ccalH^2$ are admissible, in particular they satisfy (A1), hence $u^1_{p, q}(p,q)=u^2_{p, q}(p,q)=\max(\alpha, \beta)$. It then follows from \eqref{eqn_def_sym_net_conv_comb} that $A^{12}_{p, q}(p,q;\theta)=A^{12}_{p, q}(q,p;\theta)=\max(\alpha, \beta)$ for all possible values of $\theta$. Moreover, since in \eqref{eqn_def_u_1_2_bar} all possible chains joining $p$ and $q$ must contain these two nodes as consecutive elements, we have that
\begin{equation}
u^{12}_{p, q}(p,q;\theta)=A^{12}_{p, q}(p,q;\theta)=\max(\alpha, \beta),
\end{equation}
for all $\theta$, satisfying axiom (A1).

Fulfillment of axiom (A2) also follows from admissibility of methods $\ccalH^1$ and $\ccalH^2$. Suppose there are two networks $N_X=(X,A_X)$ and $N_Y=(Y, A_Y)$ and a dissimilarity reducing map $\phi: X \to Y$. From the fact that $\ccalH^1$ and $\ccalH^2$ satisfy (A2) we have
\begin{align}
u^1_X(x, x') \geq u^1_Y(\phi(x), \phi(x')) \label{eqn_ultram_1_convex_comb}, \\
u^2_X(x, x') \geq u^2_Y(\phi(x), \phi(x')) \label{eqn_ultram_2_convex_comb}.
\end{align}
By multiplying the inequality \eqref{eqn_ultram_1_convex_comb} by $\theta$ and \eqref{eqn_ultram_2_convex_comb} by $(1-\theta)$, and adding both inequalities we obtain [cf. \eqref{eqn_def_sym_net_conv_comb}]
\begin{align}
A^{12}_X(x, x';\theta) \geq A^{12}_Y(\phi(x), \phi(x');\theta),
\end{align}
for all $0 \leq \theta \leq 1$. This implies that the map $\phi$ is also dissimilarity reducing between the networks $(X, A^{12}_X(\theta))$ and $(Y, A^{12}_Y(\theta))$. Recall that $(X, u^{12}_X(\theta))=\ccalH(X, A^{12}_X(\theta))$ and $(Y, u^{12}_Y(\theta))=\ccalH(Y, A^{12}_Y(\theta))$ for the admissible method $\ccalH$ since the networks are symmetric. Moreover, we know that $\phi$ is a dissimilarity reducing map between these two symmetric networks. Hence, from admissibility of the method $\ccalH$ it follows that
\begin{align}
u^{12}_X(x, x';\theta) \geq u^{12}_Y(\phi(x), \phi(x');\theta),
\end{align}
for all $\theta$, showing that axiom (A2) is satisfied by the convex combination method.
\end{myproof}


\begin{myproof}[of Proposition \ref{inter_reciprocal_axioms}]
We begin the proof by showing that \eqref{eqn_inter_reciprocal_clustering} outputs a valid ultrametric. That $u^{\SR(t)}_X(x,x')=0 \Leftrightarrow x=x'$ and $u^{\SR(t)}_X(x,x')=u^{\SR(t)}_X(x',x)$ are immediate for all $t \geq 2$ from the definition \eqref{eqn_inter_reciprocal_clustering}. Hence, we need to show fulfillment of the strong triangle inequality \eqref{eqn_strong_triangle_inequality}. For a fixed $t$, pick an arbitrary pair of nodes $x$ and $x'$ and an arbitrary intermediate node $x''$. Let us denote by $C^*(x,x'')$ and $C^*(x'',x')$ a pair of main chains that satisfy definition \eqref{eqn_inter_reciprocal_clustering} for $u^{\SR(t)}_X(x,x'')$ and $u^{\SR(t)}_X(x'',x')$ respectively. Construct $C(x,x') =C^*(x,x'') \uplus C^*(x'',x')$ by concatenating the aforementioned minimizing chains. However, $C(x,x')$ is a particular chain for computing $u^{\SR(t)}_X(x,x')$ and need not be the minimizing one. This implies that
\begin{equation}\label{eqn_triangle_inter_reciprocal}
u^{\SR(t)}_X(x,x') \leq \max \Big(u^{\SR(t)}_X(x,x''), u^{\SR(t)}_X(x'',x')\Big),
\end{equation}
proving the strong triangle inequality.

To show fulfillment of (A1), consider the network $\vec{\Delta}_2(\alpha,\beta)=(\{p,q\}, A_{p, q})$ with $A_{p, q}(p,q)=\alpha$ and $A_{p, q}(q,p)=\beta$. Note that in this situation, $A_{p, q}^{\SR(t)}(p, q)=\alpha$ and $A_{p, q}^{\SR(t)}(q, p)=\beta$ for all $t$ [cf. \eqref{eqn_def_sym_net_conv_comb}], since there is only one possible chain between them and contains only two nodes. Hence, from definition \eqref{eqn_inter_reciprocal_clustering},
\begin{equation}
u^{\SR(t)}_{p, q}(p,q) = \max (\alpha, \beta),
\end{equation}
for all $t$. Consequently, axiom (A1) is satisfied.

To show fulfillment of axiom (A2), consider two arbitrary networks $(X, A_X)$ and $(Y, A_Y)$ and a dissimilarity reducing map $\phi: X \to Y$ between them. Further, denote by $C^*_X(x,x')=[x=x_0,\ldots, x_l =x']$ a main chain that achieves the minimum semi-reciprocal cost in \eqref{eqn_inter_reciprocal_clustering}. Then, for a fixed $t$, we can write
\begin{equation}\label{eqn_inter_reciprocal_axiom_3}
    u^{\SR(t)}_X(x,x') = \max_{i | x_i\in C^*_X(x,x')} \overline{A^{\SR(t)}_X}(x_i,x_{i+1}).
\end{equation}
Consider now a secondary chain $C^X_t(x_i, x_{i+1})=[x_i=x^{(0)},\ldots, x^{(l')}=x_{i+1}]$ between two consecutive nodes $x_i$ and $x_{i+1}$ of the minimizing chain $C^*_X(x,x')$. Further, focus on the image of this secondary chain under the map $\phi$, that is $C^Y_t(\phi(x_i),\phi(x_{i+1})):=\phi\big(C^X_t(x_i, x_{i+1})\big) = [\phi(x_i)=\phi(x^{(0)}),\ldots, \phi(x^{(l')})=\phi(x_{i+1})]$ in the space $Y$. 

Since the map $\phi:X\to Y$ is dissimilarity reducing, $A_Y(\phi(x^{(i)}),\phi(x^{(i+1)}))\leq A_X(x^{(i)},x^{(i+1)})$ for all links in this chain. Analogously, we can bound the dissimilarities in secondary chains $C^X_t(x_{i+1}, x_{i})$ from $x_{i+1}$ back to $x_i$. Thus, from \eqref{eqn_inter_cost} we can state that,
\begin{align}\label{eqn_inter_reciprocal_axiom_3_2}
    A^{\SR(t)}_X(x_i,x_{i+1}) \geq A^{\SR(t)}_Y(\phi(x_i),\phi(x_{i+1})), \nonumber \\
    A^{\SR(t)}_X(x_{i+1},x_{i}) \geq A^{\SR(t)}_Y(\phi(x_{i+1}),\phi(x_{i})).
\end{align}
Denote by $C_Y(\phi(x),\phi(x'))$ the image of the main chain $C^*_X(x,x')$ under the map $\phi$. Notice that $C_Y(\phi(x),\phi(x'))$ is a particular chain joining $\phi(x)$ and $\phi(x')$, whereas the semi-reciprocal ultrametric computes the minimum across all main chains. Therefore,
\begin{equation}\label{eqn_inter_reciprocal_axiom_3_3}
u^{\SR(t)}_Y\!(\phi(x),\phi(x'))  \leq  \max_{i}  \overline{A^{\SR(t)}_Y} (\phi(x_i),\phi(x_{i+1})).
\end{equation}
By bounding the right-hand side of \eqref{eqn_inter_reciprocal_axiom_3_3} using \eqref{eqn_inter_reciprocal_axiom_3_2} we can write
\begin{equation}\label{eqn_inter_reciprocal_axiom_3_4}
u^{\SR(t)}_Y(\phi(x),\phi(x')) \leq  \max_{i} \overline{A^{\SR(t)}_X}(x_i,x_{i+1}). 
\end{equation}
From the combination of \eqref{eqn_inter_reciprocal_axiom_3_4} and  \eqref{eqn_inter_reciprocal_axiom_3}, it follows that $u^{\SR(t)}_Y(\phi(x),\phi(x'))\leq u^{\SR(t)}_X(x,x')$. This proves that (A2) is satisfied. 
\end{myproof}

\section{Proofs in Section \ref{sec_algorithms}}\label{appendix_sec_algorithms}

\begin{myproof}[of Proposition \ref{prop_general_algo}]
By comparison with \eqref{eqn_algo_recip}, in \eqref{eqn_algo_semi_reciprocal_2} we in fact compute reciprocal clustering on the network $(X, A^{(t-1)}_X)$. Furthermore, from the definition of matrix multiplication \eqref{def_star} in the dioid algebra $(\reals^+\cup\{+\infty\},\min,\max)$, the $(l-1)$th dioid power $A_X^{(t-1)}$ is such that its $i,j$ entry $[A_X^{(t-1)}]_{ij}$ represents the minimum infinity norm cost of a chain containing at most $t$ nodes, i.e.
\begin{equation}\label{eqn_dioid_semi_recip}
[A_X^{(t-1)}]_{ij} = \min_{C_t(x_i, x_j)} \,\,\,  \max_{k | x_k\in C_t(x_i, x_j)} A_X(x_k, x_{k+1}).
\end{equation} 
It is just a matter of notation, when comparing \eqref{eqn_dioid_semi_recip} and \eqref{eqn_inter_cost} to see that
\begin{equation}\label{eqn_dioid_semi_recip_2}
A_X^{(t-1)} = A^{\SR(t)}_X
\end{equation} 
Hence, \eqref{eqn_algo_semi_reciprocal_2} can be reinterpreted as computing reciprocal clustering on the network $(X, A^{\SR(t)}_X)$, which is the definition of semi-reciprocal clustering [cf. \eqref{eqn_inter_reciprocal_clustering} and \eqref{eqn_reciprocal_clustering}].
\end{myproof}

\begin{myproof}[of Proposition \ref{prop_algorithmic_intermediate_ultrametric}]
Since method $\ccalH^{t,t'}$ is a generalization of $\ccalH^{\SR(t)}$ that allows different length for forward and backward secondary chains, the proof is almost identical to the one of Proposition \ref{inter_reciprocal_axioms}. The only major difference is that showing the symmetry of $u^{t,t'}_X$, i.e. $u^{t,t'}_X(x, x')=u^{t,t'}_X(x', x)$ for all $x, x' \in X$, is not immediate as in the case of $u_X^{\SR(t)}$.
In a fashion similar to \eqref{eqn_inter_reciprocal_clustering}, we rewrite the definition of $u^{t,t'}_X$ given an arbitrary network $(X, A_X)$ in terms of minimizing chains,
\begin{equation}\label{eqn_proof_algo_general_1}
u^{t,t'}_X(x,x')= \min_{C(x,x')} \,\,\, \max_{i | x_i\in C(x,x')} A^{t,t'}_X(x_i, x_{i+1})
\end{equation}
where the function $A^{t,t'}_X$ is defined as
\begin{equation}\label{eqn_proof_algo_general_2}
A^{t,t'}_X\!(x,x')\!=\! \max\! \left(\!A^{\SR(t+1)}_X\!(x,x'), A^{\SR(t'+1)}_X\!(x',x)\!\right),
\end{equation}
for all $x, x' \in X$.
The functions $A^{\SR(\cdot)}_X$ in \eqref{eqn_proof_algo_general_2} are defined as in \eqref{eqn_inter_cost}. 
Notice that $A^{t,t'}_X$ is not symmetric in general, hence symmetry of $u^{t,t'}_X$ has to be explicitly verified. In order to do so, we use the result in the following claim.

\begin{claim}\label{claim_algorithmic_intermediate}
Given an arbitrary network $(X, A_X)$ and a pair of nodes $x, x' \in X$ such that $u^{t,t'}_X(x, x') = \delta$, then $u^{t,t'}_X(x', x) \leq \delta$.
\end{claim}
\begin{myproofnoname}
To show Claim \ref{claim_algorithmic_intermediate}, we must show that there exists a chain $\hat{C}(x', x)$ from $x'$ back to $x$ with the same cost $\delta$ given by \eqref{eqn_proof_algo_general_1}. Suppose $u^{t,t'}_X(x, x')=\delta$ and let $C(x, x')=[x=x_0, x_1, ... , x_l=x']$ be a minimizing chain achieving the cost $\delta$ in \eqref{eqn_proof_algo_general_1}. From definition \eqref{eqn_proof_algo_general_2}, there must exist secondary chains in both directions between every pair of consecutive nodes $x_i, x_{i+1}$ in $C(x, x')$ with cost no greater than $\delta$. These secondary chains $C_{t+1}(x_i, x_{i+1})$ and $C_{t'+1}(x_{i+1}, x_{i})$ can have at most $t+1$ nodes in the forward direction and at most $t'+1$ nodes in the opposite direction. Moreover, without loss of generality we may consider the secondary chains as having exactly $t+1$ nodes in one direction and $t'+1$ in the other if we do not require consecutive nodes to be distinct. In this way, if a minimizing secondary chain has, e.g., $t-1$ nodes, we can think of it as having $t+1$ nodes where the last two links are self loops with null cost. 

Focus on a pair of consecutive nodes $x_i, x_{i+1}$ of the main chain $C(x, x')$. If we can construct a chain from $x_{i+1}$ back to $x_i$ with cost not greater than $\delta$, then we can concatenate these chains for pairs $x_{i+1}, x_i$ for all $i$ and obtain a chain $\hat{C}(x', x)$ from $x'$ back to $x$ of cost not higher than $\delta$, concluding the proof of Claim \ref{claim_algorithmic_intermediate}. 

Notice that the secondary chains $C_{t'+1}(x_{i+1}, x_{i})$ and $C_{t+1}(x_i, x_{i+1})$ can be concatenated to form a loop, i.e. a chain starting and ending at the same node, $L(x_{i+1},x_{i+1})=C_{t'+1}(x_{i+1}, x_{i}) \uplus C_{t+1}(x_i, x_{i+1})$ of $t'+t+1$ nodes and cost not larger than $\delta$. We rename the nodes in $L(x_{i+1},x_{i+1})=[x_{i+1}=x^{0}, x^{1}, ... , x^{t'}=x_i, ..., x^{t'+t-1}, x^{t'+t}=x_{i+1}]$ starting at $x_{i+1}$ and following the direction of the loop.

Now we are going to construct a main chain $C(x_{i+1}, x_i)$ from $x_{i+1}$ to $x_i$. We may reinterpret the loop $L(x_{i+1},x_{i+1})$ as the concatenation of two secondary chains $[x^{0}, x^1, \ldots, x^{t}]$ and $[x^t, x^{t+1}, \ldots, x^{t+t'}=x^0]$ each of them having cost not greater than $\delta$. Thus, we may pick $x^{0}=x_{i+1}$ and $x^{t}$ as the first two nodes of the main chain $C(x_{i+1}, x_i)$. With the same reasoning, we may link $x^{t}$ with $x^{\,2t \!\!\! \mod \!(t+t')}$ through the secondary chains $[x^{t}, x^{t+1}, \ldots, x^{\,2t \!\!\! \mod \!(t+t')}]$ and $[x^{\,2t \!\!\! \mod \!(t+t')}, \ldots, x^{\,2t+t' \!\!\! \mod \!(t+t')} = x^t]$  with cost not exceeding $\delta$, and we may link $x^{\,2t \!\!\! \mod \!(t+t')}$ with $x^{\,3t \!\!\! \mod \!(t+t')}$ with cost not exceeding $\delta$, and so on. Hence, we construct the main chain
\begin{align}\label{eqn_proof_algo_general_3}
C(x_{i+1}, x_i)=[x^0, x^t, x^{2t \!\!\! \mod \!(t+t')}, \ldots , x^{(t+t'-1)t \!\!\! \mod \!(t+t')}], 
\end{align}
which, by construction, has cost not exceeding $\delta$.

In order to finish the proof, we need to verify that the last node in the chain in \eqref{eqn_proof_algo_general_3} is in fact $x^{t'}$. To do so, we have to show that
\begin{equation}\label{eqn_proof_algo_general_4}
(t+t'-1) \, t \equiv t' \mod (t+t').
\end{equation}
This equality is immediate when rearranging the terms in the left hand side
\begin{equation}\label{eqn_proof_algo_general_5}
(t+t') (t-1) + t' \equiv t' \mod (t+t').
\end{equation}
Consequently, using the chain in \eqref{eqn_proof_algo_general_3} we can go back from $x_{i+1}$ to $x_i$ with cost not exceeding $\delta$. Since this pair was picked arbitrarily, we may concatenate chains like the one in \eqref{eqn_proof_algo_general_3} for every value of $i$ and generate the chain $\hat{C}(x', x)$ coming back from $x'$ to $x$ with cost less than or equal to $\delta$, completing the proof of Claim \ref{claim_algorithmic_intermediate}.
\end{myproofnoname}

From Claim \ref{claim_algorithmic_intermediate}, we know that if $u^{t,t'}_X(x, x') = \delta$ then $u^{t,t'}_X(x', x) \leq \delta$. However, suppose that $u^{t,t'}_X(x', x) = \delta' < \delta$, then, by applying Claim \ref{claim_algorithmic_intermediate} for the pair $x', x \in X$, it must be that $u^{t,t'}_X(x, x') \leq \delta' < \delta$, which is a contradiction since $u^{t,t'}_X(x, x') = \delta$. Thus, it cannot be that $u^{t,t'}_X(x', x) < \delta$ and, since $u^{t,t'}_X(x', x) \leq \delta$, we have that that $u^{t,t'}_X(x', x) = \delta$, showing symmetry of $u^{t,t'}_X$ as wanted.
\end{myproof}

\section{Proofs in Section \ref{sec_full_characterization_asymmetric}}\label{appendix_sec_full_characterization_asymmetric}

\begin{myproof}[of Theorem \ref{theo_equivalence_quasi_dendrogram_quasi_ultrametric}]
In order to show that $\Psi$ is a well-defined map, we must show that $\Psi(\tdD_X)$ is a quasi-ultrametric network for every quasi-dendrogram $\tdD_X$. Given an arbitrary quasi-dendrogram $\tdD_X=(D_X, E_X)$, for a particular $\delta' \geq 0$ consider the quasi-partition $\tdD_X(\delta')$. Consider the range of resolutions $\delta$ associated with such quasi-partition. I.e.,
\begin{equation}\label{eqn_theo_pf_dendrograms_as_quasi_ultrametrics_00}
\{\delta \geq 0 \given \tdD_X(\delta)=\tdD_X(\delta')\}.
\end{equation}
Right continuity (\~D4) of $\tdD_X$ ensures that the minimum of the set in \eqref{eqn_theo_pf_dendrograms_as_quasi_ultrametrics_00} is well-defined and hence definition \eqref{eqn_theo_dendrograms_as_quasi_ultrametrics_10} is valid. To prove that $\tdu_X$ in \eqref{eqn_theo_dendrograms_as_quasi_ultrametrics_10} is a quasi-ultrametric we need to show that it attains non-negative values as well as the identity and strong triangle inequality properties. That $\tdu_X$ attains non-negative values is clear from the definition \eqref{eqn_theo_dendrograms_as_quasi_ultrametrics_10}. The identity property is implied by the first boundary condition in (\~D1). Since $[x]_0=[x]_0$ for all $x \in X$, we must have $\tdu_X(x, x)=0$. Conversely, since for all $x \neq x' \in X$, $([x]_0, [x']_0) \not\in E_X(0)$ and $[x]_0 \neq [x']_0$ we must have that $\tdu_X(x, x')>0$ for $x \neq x'$ and the identity property is satisfied.
To see that $\tdu_X$ satisfies the strong triangle inequality in \eqref{eqn_strong_triangle_inequality}, consider nodes $x$, $x'$, and $x''$ such that the lowest resolution for which $[x]_\delta = [x'']_\delta$ or $([x]_\delta, [x'']_\delta) \in E_X(\delta)$ is $\delta_1$ and the lowest resolution for which $[x'']_\delta = [x']_\delta$ or $([x'']_\delta, [x']_\delta) \in E_X(\delta)$ is $\delta_2$. Right continuity (\~D4) ensures that these lowest resolutions are well-defined. According to \eqref{eqn_theo_dendrograms_as_quasi_ultrametrics_10} we then have
\begin{alignat}{3}\label{eqn_theo_pf_dendrograms_as_quasi_ultrametrics_10}
   \tdu_X(x, x'') = \delta_1, \quad\quad\quad
   \tdu_X(x'',x') = \delta_2.
\end{alignat}
Denote by $\delta_0:=\max(\delta_1,\delta_2)$. From the equivalence hierarchy (\~D2) and influence hierarchy (\~D3) properties, it follows that $[x]_{\delta_0}=[x'']_{\delta_0}$ or $([x]_{\delta_0}, [x'']_{\delta_0}) \in E_X({\delta_0})$ and $[x'']_{\delta_0}=[x']_{\delta_0}$ or $([x'']_{\delta_0}, [x']_{\delta_0}) \in E_X({\delta_0})$. Furthermore, from transitivity (QP2) of the quasi-partition $\tdD_X(\delta_0)$, it follows that $[x]_{\delta_0}=[x']_{\delta_0}$ or $([x]_{\delta_0}, [x']_{\delta_0}) \in E_X({\delta_0})$. Using the definition in \eqref{eqn_theo_dendrograms_as_quasi_ultrametrics_10} for $x$, $x'$ we conclude that
\begin{equation}\label{eqn_theo_pf_dendrograms_as_quasi_ultrametrics_20}
   \tdu_X(x,x') \leq \delta_0.
\end{equation}
By definition $\delta_0:=\max(\delta_1,\delta_2)$, hence we substitute this expression in \eqref{eqn_theo_pf_dendrograms_as_quasi_ultrametrics_20} and compare with \eqref{eqn_theo_pf_dendrograms_as_quasi_ultrametrics_10} to obtain
\begin{equation}\label{eqn_theo_pf_dendrograms_as_ultrametrics_30}
   \tdu_X(x,x') \! \leq \! \max(\delta_1,\delta_2) \! = \! \max \Big(\tdu_X(x,x''), \tdu_X(x'',x')\Big).
\end{equation}
Consequently, $\tdu_X$ satisfies the strong triangle inequality and is therefore a quasi-ultrametric, proving that the map $\Psi$ is well-defined.

For the converse result, we need to show that $\Upsilon$ is a well-defined map. Given a quasi-ultrametric $\tdu_X$ on a node set $X$ and a resolution $\delta \geq 0$, we first define the relation
\begin{equation}\label{eqn_theo_pf_dendrograms_as_ultrametrics_31}
x \leadsto_{\tdu_X(\delta)} x' \quad \iff \quad \tdu_X(x, x') \leq \delta,
\end{equation}
for all $x, x' \in X$. Notice that $ \leadsto_{\tdu_X(\delta)}$ is a quasi-equivalence relation as defined in Definition \ref{def_quasi_equivalence} for all $\delta \geq 0$. The reflexivity property is implied by the identity property of the quasi-ultrametric $\tdu_X$ and transitivity is implied by the fact that $\tdu_X$ satisfies the strong triangle inequality. Furthermore, definitions \eqref{eqn_theo_dendrograms_as_quasi_ultrametrics_20} and \eqref{eqn_theo_dendrograms_as_quasi_ultrametrics_30} are just reformulations of \eqref{eqn_quasi_equiv_equiv} and \eqref{eqn_quasi_equiv_edges_quasi_partition} respectively, for the special case of the quasi-equivalence defined in \eqref{eqn_theo_pf_dendrograms_as_ultrametrics_31}. Hence, Proposition \ref{prop_quasi_equiv_quasi_part} guarantees that $\Upsilon(X, \tdu_X)=\tdD_X(\delta)=(D_X(\delta), E_X(\delta))$ is a quasi-partition for every resolution $\delta \geq 0$. In order to show that $\Upsilon$ is well-defined, we need to show that these quasi-partitions are nested, i.e. that $\tdD_X$ satisfies (\~D1)-(\~D4).

The first boundary condition in (\~D1) is implied by \eqref{eqn_theo_dendrograms_as_quasi_ultrametrics_20} and the identity property of $\tdu_X$. The second boundary condition in (\~D1) is implied by the fact that $\tdu_X$ takes finite real values on a finite domain since the node set $X$ is finite. Hence, any $\delta_0$ satisfying
\begin{equation}\label{eqn_theo_pf_dendrograms_as_ultrametrics_32}
\delta_0 \geq \max_{x, x' \in X} \tdu_X(x, x'),
\end{equation}
is a valid candidate to show fulfillment of (\~D1).

To see that $\tdD_X$ satisfies (\~D2) assume that for a resolution $\delta_1$ we have two nodes $x, x' \in X$ such that $x \sim_{\tdu_X(\delta_1)} x'$ as in \eqref{eqn_theo_dendrograms_as_quasi_ultrametrics_20}, then it follows that $\max \big( \tdu_X(x,x'), \tdu_X(x',x) \big) \leq \delta_1$. Thus, if we pick any $\delta_2 > \delta_1$ it is immediate that $\max \big( \tdu_X(x,x'), \tdu_X(x',x) \big) \leq \delta_2$ which by \eqref{eqn_theo_dendrograms_as_quasi_ultrametrics_20} implies that $x \sim_{\tdu_X(\delta_2)} x'$.

Fulfillment of (\~D3) can be shown in a similar way as fulfillment of (\~D2). Given a scalar $\delta_1 \geq 0$ and $x, x' \in X$ such that  $([x]_{\delta_1}, [x']_{\delta_1}) \in E_X(\delta_1)$ then by \eqref{eqn_theo_dendrograms_as_quasi_ultrametrics_30} we have that
\begin{equation}\label{eqn_theo_dendrograms_as_quasi_ultrametrics_33}
   \min_{x_1 \in [x]_{\delta_1},x_2 \in [x']_{\delta_1}} \tdu_X(x_1, x_2) \leq \delta_1.
\end{equation}
From property (\~D2), we know that for all $x \in X$, $[x]_{\delta_1} \subset [x]_{\delta_2}$ for all $\delta_2 > \delta_1$. Hence, two things might happen. Either $\max(\tdu_X(x, x'), \tdu_X(x', x)) \leq \delta_2$ in which case $[x]_{\delta_2}=[x']_{\delta_2}$ or it might be that $[x]_{\delta_2} \neq [x']_{\delta_2}$ but 
\begin{equation}\label{eqn_theo_dendrograms_as_quasi_ultrametrics_34}
   \min_{x_1 \in [x]_{\delta_2},x_2 \in [x']_{\delta_2}} \tdu_X(x_1, x_2) \leq \delta_1 < \delta_2,
\end{equation}
which implies that $([x]_{\delta_2}, [x']_{\delta_2}) \in E_X(\delta_2)$, satisfying (\~D3).

Finally, to see that $\tdD_X$ satisfies the right continuity condition (\~D4), for each $\delta \geq 0$ such that $\tdD_X(\delta) \neq ( \{ X\}, \emptyset )$ we may define $\epsilon(\delta)$ as any positive scalar satisfying
\begin{equation}\label{eqn_epsilon_d_3_ultrametric_quasi_dendrogram}
0 < \epsilon(\delta) < \displaystyle \min_{\substack{x, x' \in X \\ \text{s.t.} \,\, \tdu_X(x, x') > \delta}} \tdu_X(x, x')- \delta,
\end{equation} 
where the finiteness of $X$ ensures that $\epsilon(\delta)$ is well-defined.
Hence, \eqref{eqn_theo_dendrograms_as_quasi_ultrametrics_20} and \eqref{eqn_theo_dendrograms_as_quasi_ultrametrics_30} guarantee that $\tdD_X(\delta)=\tdD_X(\delta')$ for $\delta' \in [\delta, \delta + \epsilon(\delta)]$. For all other resolutions $\delta$ such that $\tdD_X(\delta) = ( \{ X\}, \emptyset)$, right continuity is trivially satisfied since the quasi-dendrogram remains unchanged for increasing resolutions. Consequently, $\Upsilon(X, \tdu_X)$ is a valid quasi-dendrogram for every quasi-ultrametric network $(X, \tdu_X)$, proving that $\Upsilon$ is well-defined.

In order to conclude the proof, we need to show that $\Psi \circ \Upsilon$ and $\Upsilon \circ \Psi$ are the identities on $\tilde{\mathcal{U}}$ and $\tilde{\mathcal{D}}$, respectively. To see why the former is true, pick any quasi-ultrametric network $(X, \tdu_X)$ and consider an arbitrary pair of nodes $x, x' \in X$ such that $\tdu_X(x, x')=\delta_0$. Also, consider the ultrametric network $\Psi \circ \Upsilon (X, \tdu_X):=(X, \tdu^*_X)$. From \eqref{eqn_theo_dendrograms_as_quasi_ultrametrics_20} and \eqref{eqn_theo_dendrograms_as_quasi_ultrametrics_30}, in the quasi-dendrogram $\Upsilon (X, \tdu_X)$, $x$ and $x'$ belong to different classes for resolutions $\delta < \delta_0$ and there is no edge from $[x]_\delta$ to $[x']_\delta$. Moreover, at resolution $\delta=\delta_0$ either an edge appears from $[x]_{\delta_0}$ to $[x']_{\delta_0}$, or both nodes merge into one single cluster. In any case, when we apply $\Psi$ to the resulting quasi-dendrogram, we obtain $\tdu^*_X(x, x')=\delta_0$. Since $x, x' \in X$ were chosen arbitrarily, we have that $\tdu_X = \tdu^*_X$, showing that $\Psi \circ \Upsilon$ is the identity on $\tilde{\mathcal{U}}$. A similar argument shows that $\Upsilon \circ \Psi$ is the identity on $\tilde{\mathcal{D}}$.
\end{myproof}

\section{Proofs in Section \ref{sec_alternative_axioms}}\label{appendix_sec_alternative_axioms}
\begin{myproof}[of Theorem \ref{theo_alternative_influence}]
Suppose there exists a clustering method $\ccalH$ that satisfies axioms (A1'') and (A2) but does not satisfy property (P1'). This means that there exists a network $N=(X, A_X)$ with output ultrametrics $(X, u_X)=\ccalH(N)$ for which
\begin{equation}\label{eqn_alternative_axioms_imply_influence}
   u_X(x_1, x_2) < \sep(X, A_X),
\end{equation}
for at least one pair of nodes $x_1 \neq x_2 \in X$. Focus on a symmetric two-node network $\vec{\Delta}_2(s,s)=(\{p,q\}, A_{p,q})$ with $A_{p,q}(p,q)=A_{p,q}(q,p)=s = \sep(X, A_X)$ and denote by $(X, u_{p,q})=\ccalH(\vec{\Delta}_2(s,s))$ the output of applying method $\ccalH$ to the two-node network $\vec{\Delta}_2(s,s)$. From axiom (A1''), we must have that
\begin{equation}\label{eqn_alternative_axioms_imply_influence_2}
u_{p,q}(p,q)=\min \Big(\sep(X, A_X),\sep(X, A_X)\Big)=\sep(X, A_X).
\end{equation}
Construct the map $\phi:X \to \{p,q\}$ from the network $N$ to $\vec{\Delta}_2(s,s)$ that takes node $x_1$ to $\phi(x_1)=p$ and every other node $x \neq x_1$ to $\phi(x)=q$. No dissimilarity can be increased when applying $\phi$ since every dissimilarity is mapped either to zero or to $\sep(X, A_X)$ which is by definition the minimum dissimilarity in the original network \eqref{eqn_def_separation_network}. Hence, $\phi$ is a dissimilarity reducing map and from Axiom (A2) it follows that
\begin{equation}\label{eqn_alternative_axioms_imply_influence_3}
u_X(x_1, x_2) \geq u_{p,q}(\phi(x_1), \phi(x_2)) = u_{p,q}(p,q).
\end{equation}
By substituting \eqref{eqn_alternative_axioms_imply_influence_2} in \eqref{eqn_alternative_axioms_imply_influence_3} we contradict \eqref{eqn_alternative_axioms_imply_influence} proving that such method $\ccalH$ cannot exist.
\end{myproof}

\begin{myproof}[of Proposition \ref{prop_unilateral_axioms}]
To show fulfillment of (A1''), consider the two-node network $\vec{\Delta}_2(\alpha, \beta)$ and denote by $(\{p,q\}, u^{\U}_{p, q})=\ccalH^\U(\vec{\Delta}_2(\alpha, \beta))$ the output of applying unilateral clustering to $\vec{\Delta}_2(\alpha, \beta)$. Since every chain connecting $p$ and $q$ must contain these two nodes as consecutive nodes, applying the definition in \eqref{eqn_unilateral_clustering_2} yields
\begin{equation}\label{eqn_theo_unilateral_axioms_pf_10}
    u^{\U}_{p, q}(p,q) = \min \big(A_{p,q}(p,q), A_{p,q}(q,p)\big) 
                 = \min(\alpha,\beta),
\end{equation} 
and axiom (A1'') is thereby satisfied. 

In order to show fulfillment of axiom (A2), the proof is analogous to the one developed in Proposition \ref{prop_reciprocal_axioms}. The proof only differs in the appearance of minimization operations instead of maximizations to account for the difference in the definitions of unilateral and reciprocal ultrametrics [cf.  \eqref{eqn_unilateral_clustering_2} and \eqref{eqn_reciprocal_clustering}].
\end{myproof}


\begin{myproof}[of Theorem \ref{theo_unilateral_unicity}]
Given an arbitrary network $(X, A_X)$, denote by $\ccalH$ a clustering method that fulfills axioms (A1'') and (A2) and define $\ccalH(X, A_X)=(X, u_X)$. Then, the output ultrametric $u_X$ must satisfy the inequality

\begin{equation}\label{eqn_theo_unilateral_unicity_pf_010}
    u^\U_X(x,x')\leq u_X(x,x')\leq u^\U_X(x,x'),
\end{equation} 
for every pair of nodes $x, x' \in X$. 

\begin{myproof}[of leftmost inequality in \eqref{eqn_theo_unilateral_unicity_pf_010}] Consider the unilateral clustering equivalence relation $\sim_{\U_X(\delta)}$ at resolution $\delta$ according to which $x \sim_{\U_X(\delta)} x'$ if and only if $x$ and $x'$ belong to the same unilateral cluster at resolution $\delta$. That is, 
\begin{equation}\label{eqn_theo_unilateral_unicity_pf_010_1}
    x \sim_{\U_X(\delta)} x'  \iff  u^{\U}_X(x,x')\leq\delta.
\end{equation} 
Further, as in the proof of Theorem \ref{theo_extremal_ultrametrics}, consider the space $Z$ of equivalence classes at resolution $\delta$. That is, $Z := X \mod \sim_{\U_X(\delta)}$. Also, consider the map $\phi_{\delta}:X\to Z$ that maps each point of $X$ to its equivalence class. Notice that $x$ and $x'$ are mapped to the same point $z$ if and only if they belong to the same block at resolution $\delta$, consequently
\begin{equation}\label{eqn_theo_unilateral_unicity_pf_011}
    \phi_\delta(x) = \phi_\delta(x') \iff  u^{\U}_X(x,x')\leq\delta.
\end{equation} 
We define the network $N_Z=(Z,A_Z)$ by endowing $Z$ with the dissimilarity matrix $A_Z$ derived from $A_X$ in the following way
\begin{equation}\label{eqn_theo_unilateral_unicity_pf_020}
    A_Z(z,z') = \min_{x\in\phi_\delta^{-1}(z), x'\in\phi_\delta^{-1}(z')} A_X(x,x').
\end{equation} 
For further details on this construction, review the corresponding proof in Theorem \ref{theo_extremal_ultrametrics} and see Fig. \ref{fig_proof_theo_extremal_ultrametrics}. Nonetheless, we stress the fact that the map $\phi_\delta$ is dissimilarity reducing for all $\delta$. I.e.,
\begin{equation}\label{eqn_theo_unilateral_unicity_pf_025}
    A_X(x,x') \geq A_Z(\phi_\delta(x),\phi_\delta(x')).
\end{equation} 
%

%
\begin{claim}
The separation as defined in \eqref{eqn_def_separation_network} of the equivalence class network $N_Z$ is 
\begin{equation}\label{eqn_theo_unilateral_unicity_pf_028}
    \sep(N_Z) > \delta.
\end{equation} \end{claim}
\begin{myproof} First, observe that by definition of unilateral clustering
\eqref{eqn_unilateral_clustering_2}, we know that,
\begin{equation}\label{eqn_theo_unilateral_unicity_pf_030}
    u^U_X(x,x') \leq \min(A_X(x,x'), A_X(x',x)),
\end{equation}
since a two node chain between nodes $x$ and $x'$ is a particular chain joining the two nodes whereas the ultrametric is calculated as the minimum over all chains. Now, assume that $\sep(N_Z) \leq \delta$. Therefore, by \eqref{eqn_theo_unilateral_unicity_pf_020}  there exists a pair of nodes $x$ and $x'$ that belong to different equivalence classes and have 
\begin{equation}\label{eqn_theo_unilateral_unicity_pf_031}
A_X(x,x')\leq \delta.
\end{equation}
However, if $x$ and $x'$ belong to different equivalence classes, they cannot be clustered at resolution $\delta$, hence,
\begin{equation}\label{eqn_theo_unilateral_unicity_pf_032}
u^U_X(x,x')>\delta.
\end{equation}
Inequalities \eqref{eqn_theo_unilateral_unicity_pf_031} and \eqref{eqn_theo_unilateral_unicity_pf_032} cannot hold simultaneously since they contradict \eqref{eqn_theo_unilateral_unicity_pf_030}. Thus, it must be the case that $\sep(N_Z) > \delta$.\end{myproof}

%
Denote by $\ccalH(Z,A_Z) = (Z,u_Z)$ the outcome of the clustering method $\ccalH$ applied to the equivalence class network $N_Z$. Since $\sep(N_Z)>\delta$, it follows from property (P1') that for all $z,z'$ such that $z \neq z'$ 
\begin{equation}\label{eqn_theo_unilateral_unicity_pf_070}
    u_Z(z,z') > \delta .
\end{equation}
Further, recalling that $\phi_\delta$ is a dissimilarity reducing map \eqref{eqn_theo_unilateral_unicity_pf_025}, from Axiom (A2) we must have $u_X(x,x') \geq u_Z(\phi_\delta(x), \phi_\delta(x')) = u_Z(z, z')$ for some $z, z' \in Z$. This fact, combined with \eqref{eqn_theo_unilateral_unicity_pf_070}, entails that when $\phi_\delta(x)$ and $\phi_\delta(x')$ belong to different equivalence classes
\begin{equation}\label{eqn_theo_unilateral_unicity_pf_080}
    u_X(x,x') \geq u_Z(\phi(x),\phi(x')) >\delta.
\end{equation}
Notice now that according to \eqref{eqn_theo_unilateral_unicity_pf_011}, $\phi_\delta(x)$ and $\phi_\delta(x')$ belonging to different equivalence classes is equivalent to $ u^{\U}_X(x,x')>\delta$. Hence, we can state that $u^{\U}_X(x,x')>\delta$ implies $u_X(x,x')>\delta$ for any arbitrary $\delta>0$. In set notation,
\begin{equation}\label{eqn_theo_unilateral_unicity_pf_090}
    \{(x,x') : u^{\U}_X(x,x')>\delta\} \subseteq \{(x,x') : u_X(x,x')>\delta\}.
\end{equation}
Since \eqref{eqn_theo_unilateral_unicity_pf_090} is true for arbitrary $\delta>0$, this implies that $u^{\U}_X(x,x') \leq  u_X(x,x')$, proving the left inequality in \eqref{eqn_theo_unilateral_unicity_pf_010}. \end{myproof}

\begin{myproof}[of rightmost inequality in \eqref{eqn_theo_unilateral_unicity_pf_010}] Consider two nodes $x$ and $x'$ with unilateral ultrametric value $u^{\U}_X(x,x') = \delta$. Let $C^*(x,x')=[x=x_0,\ldots, x_l=x']$ be a minimizing chain in the definition \eqref{eqn_unilateral_clustering_2} so that we can write
\begin{align}\label{eqn_theo_unilateral_unicity_pf_100}
   \delta & =  u^{\U}_X(x,x') \\
   &= \max_{i | x_i \in C^*(x,x')} \,  \min \Big(A_X(x_i,x_{i+1}),  A_X(x_{i+1},x_i)\Big). \nonumber
\end{align}
Consider the two-node network $\vec{\Delta}_2(\delta, M)=(\{p,q\}, A_{p,q})$ with $A_{p,q}(p,q)=\delta$ and $A_{p,q}(q,p)=M := \max_{x,x'} A_X(x,x') $. Denote by $(\{p, q\}, u_{p,q})=\ccalH(\{p,q\},A_{p,q})$ the output of the clustering method $\ccalH$ applied to network $\vec{\Delta}_2(\delta, M)$. Notice that according to Axiom (A1'') we have 
\begin{equation}\label{eqn_theo_unilateral_unicity_pf_104}
u_{p,q}(p,q) = u_{p,q}(q,p) = \min( \delta, M) = \delta,
\end{equation}
where the last equality is enforced by the definition of $M$.

Focus now on each link of the minimizing chain in \eqref{eqn_theo_unilateral_unicity_pf_100}. For every successive pair of nodes $x_i$ and $x_{i+1}$, we must have
\begin{align}
  \max \Big(A_X(x_i,x_{i+1}),  A_X(x_{i+1},x_i)\Big) \leq M, \label{eqn_theo_unilateral_unicity_pf_101} \\
  \min \Big(A_X(x_i,x_{i+1}),  A_X(x_{i+1},x_i)\Big) \leq \delta.  \label{eqn_theo_unilateral_unicity_pf_102}
\end{align}
Expression \eqref{eqn_theo_unilateral_unicity_pf_101} is true since $M$ is defined as the maximum dissimilarity in $A_X$. Inequality \eqref{eqn_theo_unilateral_unicity_pf_102}
is justified by \eqref{eqn_theo_unilateral_unicity_pf_100}, since $\delta$ is defined as the maximum among links of the minimum distance in both directions of the link.
This observation allows the construction of dissimilarity reducing maps $\phi_i:\{p,q\}\to X$,

\begin{equation}\label{eqn_theo_unilateral_unicity_pf_103}
\phi_i :=
\begin{cases}
\phi_i(p)=x_i, \phi_i(q)=x_{i+1}, \\ \quad\quad\quad\quad\quad\quad\quad\quad \text{if} \,\, \hat{A}_X(x_i,x_{i+1}) = A_X(x_i,x_{i+1}) \\
\phi_i(q)=x_i, \phi_i(p)=x_{i+1},  \quad \text{otherwise.}
\end{cases}
\end{equation}

In this way, we can map $p$ and $q$ to subsequent nodes in the chain $C(x,x')$ used in \eqref{eqn_theo_unilateral_unicity_pf_100}. Inequalities \eqref{eqn_theo_unilateral_unicity_pf_101} and \eqref{eqn_theo_unilateral_unicity_pf_102} combined with the map definition in \eqref{eqn_theo_unilateral_unicity_pf_103} guarantee that $\phi_i$ is a dissimilarity reducing map for every $i$. 
Since clustering method $\ccalH$ satisfies Axiom (A2), it follows that 
\begin{align}\label{eqn_theo_unilateral_unicity_pf_120}
   u_X(\phi_i(p),\phi_i(q)) \leq u_{p,q}(p,q) = \delta, \quad \forall\ i,
\end{align}
where we used \eqref{eqn_theo_unilateral_unicity_pf_104} for the last equality.
Substituting $\phi_i(p)$ and $\phi_i(q)$ in \eqref{eqn_theo_unilateral_unicity_pf_120} by the corresponding nodes given by the definition \eqref{eqn_theo_unilateral_unicity_pf_103}, we can write 
\begin{equation}\label{eqn_theo_unilateral_unicity_pf_131}
u_X(x_i, x_{i+1})= u_X(x_{i+1}, x_i) \leq \delta, \quad \forall\ i,
\end{equation}
where the symmetry property of ultrametrics was used.
To complete the proof we invoke the strong triangle inequality \eqref{eqn_strong_triangle_inequality} and apply it to $C(x,x')=[x=x_0,\ldots, x_l=x']$, the minimizing chain in \eqref{eqn_theo_unilateral_unicity_pf_100}. As a consequence,
\begin{align}\label{eqn_theo_unilateral_unicity_pf_140}
   u_X(x,x') \leq \max_i u_X(x_i,x_{i+1}) \leq \delta,
\end{align}
where \eqref{eqn_theo_unilateral_unicity_pf_131} was used in the second inequality. The proof of the right inequality in \eqref{eqn_theo_unilateral_unicity_pf_010} is completed by substituting $\delta =  u^{\U}_X(x,x')$ [cf. \eqref{eqn_theo_unilateral_unicity_pf_100}] into \eqref{eqn_theo_unilateral_unicity_pf_140}. \end{myproof}

Having proved both inequalities in \eqref{eqn_theo_unilateral_unicity_pf_010}, the conclusion that the unilateral clustering method is the only one that satisfies axioms (A1'') and (A2) is immediate, completing the global proof.
\end{myproof}


\begin{myproof}[of Theorem \ref{theo_extremal_ultrametrics_2}]
The leftmost inequality in \eqref{eqn_theo_extremal_ultrametrics_2} can be proved using the same method of proof used for the leftmost inequality in \eqref{eqn_theo_unilateral_unicity_pf_010} within the proof of Theorem \ref{theo_unilateral_unicity}. The proof of the rightmost inequality in \eqref{eqn_theo_extremal_ultrametrics_2} is equivalent to the proof of the rightmost inequality in Theorem \ref{theo_extremal_ultrametrics}.
\end{myproof}

\section{Proofs in Section \ref{sec_stability}}\label{appendix_sec_stability}

\begin{myproof}[of Theorem \ref{theo_gromov_hausdorff}]
%

\begin{myproof}[of nonnegativity and symmetry statements] That the distance $d_\ccalN(N_X, N_Y) $ is nonnegative follows from the absolute value in the definition of \eqref{eqn_gh_distance}. The symmetry $d_\ccalN(N_X,N_Y)=d_\ccalN(N_Y,N_X)$ follows because a correspondence $R\subseteq X\times Y$ with elements $r_i=(x_i,y_i)$ results in the same associations as the correspondence $S\subseteq Y\times X$ with elements $s_i=(y_i,x_i)$. This proves the first two statements. \end{myproof}

%
\begin{myproof}[of identity statement] In order to show the identity statement, assume that $N_X$ and $N_Y$ are isomorphic and let $\phi:X\rightarrow Y$ be a bijection realizing this isomorphism. Then, consider the particular correspondence $R_\phi=\{(x,\phi(x)),\,x\in X\}$. By construction, for all $x_0 \in X$ there is an element $r = (x_0, y) \in R_\phi$ and since $\phi$ is surjective -- indeed, bijective -- for all $y_0 \in Y$ there is an element $s=(x, y_0) \in R_\phi$. Thus, $R_\phi$ is a valid correspondence between $X$ and $Y$, which satisfies \eqref{eqn_network_isomorphism},
\begin{equation}\label{eqn_proof_gromov_hausdorf_isomo}
   A_Y(y,y') = A_Y(\phi(x),\phi(x')) = A_X(x,x')
\end{equation}
for all $(x,y),(x',y')\in R_\phi$. Since $R_\phi$ is a particular correspondence while in definition \eqref{eqn_gh_distance} we minimize over all possible correspondences it must be
\begin{equation}
   d_\ccalN(N_X,N_Y) \leq \frac{1}{2}\max_{(x,y),(x',y')\in R_\phi}|A_X(x,x')-A_Y(y,y')| 
              = 0,
\end{equation}
where the equality follows because $A_X(x,x')-A_Y(y,y')=0$ for all $(x,y),(x',y')\in R_\phi$ by \eqref{eqn_proof_gromov_hausdorf_isomo}. Since we already argued that $d_\ccalN(N_X,N_Y)\geq0$ it must be that $d_\ccalN(N_X,N_Y)=0$ when the networks $N_X \cong N_Y$ are isomorphic.

We now argue that the converse is also true, i.e., if the distance is $d_\ccalN(N_X,N_Y)=0$ it implies that $X$ and $Y$ are isomorphic. If $d_\ccalN(N_X,N_Y)=0$ there is a correspondence $R_0$ such that $A_X(x,x')=A_Y(y,y')$ for all $(x,y),(x',y')\in R_0$. Define then the function $\phi:X\to Y$ that associates to $x$ any value $y$ among those that form a pair with $x$ in the correspondence $R_0$,
\begin{equation}\label{eqn_prop_gh_distance_identity_pf_10}
   \phi(x) = y_0 \in \left\{y \given (x,y) \in R_0 \right\}.
\end{equation}
Since $R_0$ is a correspondence the set $\left\{y \given (x,y) \in R_0 \right\}$ is nonempty implying that \eqref{eqn_prop_gh_distance_identity_pf_10} is defined for all $x\in X$. Moreover, since we know that $(x,\phi(x))\in R_0$ we must have $A_X(x,x')=A_Y(\phi(x),\phi(x'))$ for all $x,x'$. From this observation it follows that the function $\phi$ must be injective. If it were not, there would be a pair of points $x\neq x'$ for which $\phi(x)=\phi(x')$. For this pair of points we can then write,
\begin{equation}\label{eqn_proof_gromov_hausdorf_isomo_3}
   A_X(x,x')=A_Y(\phi(x),\phi(x'))=0,
\end{equation}
where the first equality follows form the definition of $\phi$ and the second equality from the fact that $\phi(x)=\phi(x')$ and that dissimilarity functions are such that $A_Y(y,y)=0$. However, \eqref{eqn_proof_gromov_hausdorf_isomo_3} is inconsistent with $x\neq x'$ because the dissimilarity function is $A_X(x,x')=0$ if and only $x=x'$. Then, $\phi(x)=\phi(x')$ if and only if  $x=x'$, implying that $\phi$ is an injection.

Likewise, define the function $\psi:Y\to X$ that associates to $y$ any value $x$ among those that form a pair with $y$ in the correspondence $R_0$,
\begin{equation}\label{eqn_prop_gh_distance_identity_pf_20}
   \psi(y) = x_0 \in \left\{x \given (x,y) \in R_0 \right\}
\end{equation}
Since $R_0$ is a correspondence the set $\left\{x \given (x,y) \in R_0 \right\}\neq\emptyset$ is nonempty implying that \eqref{eqn_prop_gh_distance_identity_pf_20} is defined for all $y\in Y$ and since we know that $(\psi(y),y)\in R_0$ we must have $A_X(\psi(y),\psi(y'))=A_Y(y,y')$ for all $y,y'$ from where it follows that the function $\psi$ must be injective.

We have then constructed injections $\phi:X\to Y$ and $\psi:Y\to X$. The Cantor-Bernstein-Schroeder theorem \cite[Chapter 2.6]{Kolmogorov75} applies and guarantees that there exists a bijection between $X$ and $Y$. This forces $X$ and $Y$ to have the same cardinality and, as a consequence, it forces $\phi$ and $\psi$ to be bijections. Pick the bijection $\phi$ and recall that since $(x,\phi(x))\in R_0$ we must have $A_X(x,x')=A_Y(\phi(x),\phi(x'))$ for all $x,x'$ from where it follows that $N_X \cong N_Y$. Since we already showed $d_\ccalN(N_X,N_Y)=0$ when the networks $N_X \cong N_Y$ are isomorphic the identity statement follows. \end{myproof}

%
\begin{myproof}[of triangle inequality] To show the triangle inequality let correspondences $R^*$ between $X$ and $Z$ and $S^*$ between $Z$ and $Y$ be the minimizing correspondences in \eqref{eqn_gh_distance} so that we can write
\begin{alignat}{3}\label{eqn_prop_gh_distance_triangle_inequality_pf_10}
   &d_\ccalN(N_X,N_Z)
        \ =\ &&\frac{1}{2}\max_{(x,z),(x',z')\in R^*}
                 &&\big|A_X(x,x')-A_Z(z,z')\big|. \nonumber\\
   &d_\ccalN(N_Z,N_Y)
        \ =\ &&\frac{1}{2}\max_{(z,y),(z',y')\in S^*}
                 &&\big|A_Z(z,z')-A_Y(y,y')\big|.
\end{alignat}
Define now the correspondence $T$ between $X$ and $Y$ as the one induced by pairs $(x,z)$ and $(z,y)$ sharing a common point $z\in Z$,
\begin{equation}\label{eq_definition_correspondence_t}
T := \left\{(x,y) \given \exists\ z\in Z \text{ with } (x,z)\in R^*, 
  (z,y)\in S^*\right\}.
\end{equation}
To show that $T$ is a correspondence we have to prove that for every $x \in X$ there exists $y_0 \in Y$ such that $(x,y_0) \in T$ and that for every $y \in Y$ there exists $x_0 \in X$ such that $(x_0,y) \in T$. To see this pick arbitrary $x \in X$. Because $R$ is a correspondence there exists $z_0 \in Z$ such that $(x,z_0) \in R$. Since $S$ is also a correspondence, there exists $y_0\in Y$ such that $(z_0, y_0) \in S$. Hence, there exists $(x, y_0) \in T$ for every $x \in X$. Conversely, pick an arbitrary $y \in Y$. Since $S$ and $R$ are correspondences there exist $z_0 \in Z$ and $x_0\in X$ such that $(z_0,y) \in S$ and $(x_0, z_0) \in R$. Thus, there exists $(x_0, y) \in T$ for every $y \in Y$. Therefore, $T$ is a correspondence.

The correspondence $T$ need not be a minimizing correspondence for the distance $d_\ccalN(N_X,N_Y)$, but since it is a valid correspondence we can write [cf. \eqref{eqn_gh_distance}]
\begin{equation}\label{eqn_triangle_inequality_gromov_haus}
    d_\ccalN(N_X,N_Y)\leq \frac{1}{2} \max_{(x,y),(x',y')\in T}|A_X(x, x')-A_Y(y, y')|.
\end{equation}
According to the definition of $T$ in \eqref{eq_definition_correspondence_t} the requirement $(x,y),(x',y')\in T$ is equivalent to requiring $(x,z),(x',z')\in R^*$ and $(z,y),(z',y')\in S^*$.  Further adding and subtracting $A_Z(z, z')$ from the maximand and using the triangle inequality on the absolute value yields
\begin{align}\label{eqn_triangle_inequality_gromov_haus_15}
   d_\ccalN(N_X,N_Y) 
      \leq \frac{1}{2} \max_{ {(x,z),(x',z')\in R^*} \atop {(z,y),(z',y')\in S^*} } 
                &  |A_X(x, x')-A_Z(z, z')|  \\ \nonumber
                &\quad + |A_Z(z, z') -A_Y(y, y')|.
\end{align}
We can further bound \eqref{eqn_triangle_inequality_gromov_haus_15} by maximizing each summand independently so as to write
\begin{align}\label{eqn_triangle_inequality_gromov_haus_20}
   d_\ccalN(N_X,N_Y)  
      \leq \ &        \frac{1}{2} \max_{(x,z),(x',z')\in R^*} |A_X(x, x')-A_Z(z, z')| \nonumber\\
             &\quad + \frac{1}{2} \max_{(z,y),(z',y')\in S^*} |A_Z(z, z')-A_Y(y, y')|.
\end{align}
Substituting the equalities in \eqref{eqn_prop_gh_distance_triangle_inequality_pf_10} for the summands on the right hand side of \eqref{eqn_triangle_inequality_gromov_haus_20} yields the triangle inequality.
\end{myproof}

Having shown the four statements in Theorem \ref{theo_gromov_hausdorff}, the main proof concludes.
\end{myproof}

\begin{myproof}[of Theorem \ref{theo_stab_rec}]
In order to prove the statement for any $t \geq 2$, we first show that the difference between the costs of secondary chains is bounded as the following claim states.

\begin{claim}\label{lem_stability_interrec}
Given two networks $N_X=(X, A_X)$ and $N_Y=(Y, A_Y)$, let $\eta = d_\ccalN(N_X,N_Y)$ and $R$ be the associated minimizing correspondence. Given two pair of nodes $(x,y)$, $(x', y') \in R$ we have
\begin{equation}\label{abs_dif_2_interrec}
|A^{\SR(t)}_X(x,x')-A^{\SR(t)}_Y(y,y')| \leq 2\eta,
\end{equation}
where $A^{\SR(t)}_X$ and $A^{\SR(t)}_Y$ are defined as in \eqref{eqn_inter_cost}.
\end{claim}

\begin{myproofnoname}
Let $C^*(x, x')=[x=x_0, x_1, ... , x_l=x']$ be a minimizing chain in the definition \eqref{eqn_inter_cost}, implying that
\begin{equation}\label{ineq_1_interrec_0}
 A^{\SR(t)}_X(x,x') = \max_{i | x_i \in C^*(x, x')} A_X(x_i, x_{i+1}). 
 \end{equation}
Construct the chain $C(y, y')=[y=y_0, y_1, ... , y_l=y']$ in $N_Y$ from $y$ to $y'$ such that $(x_i, y_i) \in R$ for all $i$. This chain is guaranteed to exist from the definition of correspondence. Using the definition in \eqref{eqn_inter_cost} and the inequality stated in \eqref{abs_dif}, we write 
\begin{align}\label{ineq_1_interrec}
A^{\SR(t)}_Y(y,y') &\leq \max_{i | y_i \in C(y, y')} A_Y(y_i, y_{i+1}) \nonumber\\
&\leq \max_{i | x_i \in C^*(x, x')} A_X(x_i, x_{i+1}) + 2\eta.
\end{align}
Substituting \eqref{ineq_1_interrec_0} in \eqref{ineq_1_interrec} we obtain,
\begin{equation}\label{ineq_2_interrec}
A^{\SR(t)}_Y(y,y') \leq A^{\SR(t)}_X(x,x') + 2\eta.
\end{equation}
By following an analogous procedure starting with a minimizing chain in the network $N_Y$, we can show that,
\begin{equation}\label{ineq_3_interrec}
A^{\SR(t)}_X(x,x') \leq A^{\SR(t)}_Y(y,y') + 2\eta.
\end{equation}
From \eqref{ineq_2_interrec} and \eqref{ineq_3_interrec}, the desired result in \eqref{abs_dif_2_interrec} follows. \end{myproofnoname}

We use Lemma \ref{lem:bounded_maximum} to show that \eqref{abs_dif_2_interrec} implies
\begin{equation}\label{ineq_1_interrec_00_bis}
|\overline{A^{\SR(t)}_X}(x,x')-\overline{A^{\SR(t)}_Y}(y,y')| \leq 2\eta,
 \end{equation}
where $\overline{A^{\SR(t)}_X}$ and $\overline{A^{\SR(t)}_Y}$ are defined as in \eqref{eqn_inter_reciprocal_clustering_auxiliary}. We then compare \eqref{eqn_nonreciprocal_chains} and \eqref{eqn_inter_reciprocal_clustering} to see that
\begin{equation}\label{eqn:semi_reciprocal_in_terms_of_dsl}
(X, u^{\SR(t)}_X) = \tilde{\ccalH}^*(X, \overline{A^{\SR(t)}_X}),
\end{equation}
and similarly for $(Y, u^{\SR(t)}_Y)$. Finally, as done for the case $t=2$, by using stability of $\tilde{\ccalH}^*$ [cf. Theorem \ref{theo_stab_directed_single_linkage}], the result follows.
\end{myproof}

\end{appendices}


\bibliographystyle{unsrt}
\bibliography{clustering_biblio}

\end{document}